\documentclass[lettersize,journal]{IEEEtran}
\usepackage{amsmath,amsfonts}
\usepackage{array}
\usepackage{textcomp}
\usepackage{stfloats}
\usepackage{url}
\usepackage{verbatim}
\usepackage{graphicx}
\usepackage{booktabs}
\usepackage{algorithm}
\usepackage{algorithmicx}
\usepackage{algpseudocode}
\usepackage{xspace}
\usepackage{mathrsfs}
\usepackage{multirow}
\usepackage{enumitem}
\usepackage{dutchcal}
\usepackage[colorlinks, linkcolor=black, urlcolor=black, citecolor=black]{hyperref}
\usepackage{subcaption}
\usepackage{pifont}
\usepackage{svg}
\usepackage{threeparttable}
\usepackage{color}
\usepackage{diagbox}
\usepackage{amssymb}
\usepackage{bm}
\usepackage{caption}

\def\BibTeX{{\rm B\kern-.05em{\sc i\kern-.025em b}\kern-.08em
    T\kern-.1667em\lower.7ex\hbox{E}\kern-.125emX}}
\usepackage{balance}

\newcommand{\method}{{\textsc{TRC}}\xspace}

\renewcommand{\textcolor}[2]{#2}

\begin{document}
\title{Deep Tabular Representation Corrector}
% \title{Post-Learning for Deep Tabular Representations}
\author{Hangting Ye, Peng Wang, Wei Fan, Xiaozhuang Song, He Zhao, \textit{Member, IEEE}, Dandan Guo*, Yi Chang*, \textit{Senior Member, IEEE}
\thanks{
% Manuscript created October, 2020; This work was developed by the IEEE Publication Technology Department. This work is distributed under the \LaTeX \ Project Public License (LPPL) ( http://www.latex-project.org/ ) version 1.3. A copy of the LPPL, version 1.3, is included in the base \LaTeX \ documentation of all distributions of \LaTeX \ released 2003/12/01 or later. The opinions expressed here are entirely that of the author. No warranty is expressed or implied. User assumes all risk.
Hangting Ye, Peng Wang, Dandan Guo, and Yi Chang are with the School of Artificial
Intelligence, Jilin University, China. Yi Chang is also with the Engineering Research Center of Knowledge-Driven
Human-Machine Intelligence, Ministry of Education, China and the International Center of Future Science, Jilin University, China.  E-mail: yeht22@mails.jlu.edu.cn, pwang23@mails.jlu.edu.cn, guodandan@jlu.edu.cn, yichang@jlu.edu.cn.

Wei Fan is with the School of Computer Science, University of Auckland, New Zealand. E-mail: wei.fan@auckland.ac.nz.

Xiaozhuang Song is with the Chinese University of Hong Kong, Shenzhen, China. E-mail: xiaozhuangsong1@link.cuhk.edu.cn.

He Zhao is with CSIRO’s Data61 and Monash University, Australia. E-mail: he.zhao@data61.csiro.au.

*Dandan Guo and Yi Chang are the corresponding authors.

Our code is available at \url{https://github.com/HangtingYe/TRC}.
}}

\markboth{Journal of \LaTeX\ Class Files,~Vol.~18, No.~9, September~2020}%
{How to Use the IEEEtran \LaTeX \ Templates}

\maketitle

\begin{abstract}
Tabular data have been playing a mostly important role in diverse real-world fields, such as healthcare, engineering, finance, etc. The recent success of deep learning has fostered many deep networks (e.g., Transformer, ResNet) based tabular learning methods. Generally, existing deep tabular machine learning methods are along with the two paradigms, i.e., in-learning and pre-learning.
In-learning methods need to train networks from scratch or impose extra constraints to regulate the representations which nonetheless train multiple tasks simultaneously and make learning more difficult, while pre-learning methods design several pretext tasks for pre-training and then conduct task-specific fine-tuning, which however need much extra training effort with prior knowledge. In this paper, we introduce a novel deep Tabular Representation Corrector, TRC, to enhance any trained deep tabular model's representations without altering its parameters in a model-agnostic manner. Specifically, targeting the representation shift and representation redundancy that hinder prediction, we propose two tasks, i.e., (i) \textit{Tabular Representation Re-estimation}, that involves training a shift estimator to calculate the inherent shift of tabular representations to subsequently mitigate it, thereby re-estimating the representations and (ii) \textit{Tabular Space Mapping}, that transforms the above re-estimated representations into a light-embedding vector space via a coordinate estimator while preserves crucial predictive information to minimize redundancy. The two tasks jointly enhance the representations of deep tabular models without touching on the original models thus enjoying high efficiency.
Finally, we conduct extensive experiments on state-of-the-art deep tabular machine learning models coupled with TRC on various tabular benchmarks which have shown consistent superiority.
\end{abstract}


\section{Introduction}
\label{sec:introduction}
\IEEEPARstart{T}{abular} data, typically represented as tables with rows standing for the data samples and columns standing for the heterogeneous feature variables (e.g. categorical and numerical features), is a fundamental data type across diverse fields, including healthcare~\cite{hernandez2022synthetic}, finance~\cite{assefa2020generating}, engineering~\cite{ye2023uadb} and psychology~\cite{urban2021deep}.
Unlike perceptual data (e.g., image and text), tabular data lacks prior structural information~\cite{borisov2022deep}, making it essential to discover relationships within the data without relying on preconceived structures.
% Unlike perceptual data (e.g., image and text), there is no prior information on the structure of tabular data~\cite{borisov2022deep}.
% Hence, it is necessary to discover and exploit relations of tabular data without relying on prior structure information.
Over the course of several decades, tabular machine learning has garnered significant attention from researchers, transitioning from traditional machine learning techniques (e.g. linear regression~\cite{su2012linear}, logistic regression~\cite{wright1995logistic}) to more advanced tree-based approaches (e.g. XGBoost~\cite{chen2016xgboost}, CatBoost~\cite{prokhorenkova2018catboost}).
Recently, witnessing the great success of deep learning in many domains such as computer vision~\cite{he2016deep, Goodfellow-et-al-2016} and natural language processing~\cite{devlin2018bert}, researchers have observed there is still a large gap in the effective usage of deep learning methods in the tabular domain~\cite{borisov2022deep, gorishniy2021revisiting}.

Research on deep tabular machine learning currently encompasses two main directions: in-learning and pre-learning styles.
For the in-learning style, typically neural network backbones with different architectures are trained from scratch, or additional constraints are applied to regularize the representations. 
For example, AutoInt~\cite{song2019autoint} transforms features to embeddings and applies a series of attention-based transformations to the embeddings based on Transformer~\cite{vaswani2017attention}. 
FT-Transformer~\cite{gorishniy2021revisiting} further improves AutoInt through better token embeddings. 
SNN~\cite{klambauer2017self} proposes an MLP-like model with neuron activations aiming for zero mean and unit variance, fostering an effective regularization for learning complex representations.
\textit{Kadra et al.}~\cite{kadra2021well} further regularizes the MLP with multiple modern regularization techniques. 
Recently, PTaRL~\cite{ye2024ptarl} constructs a new projection space and projects data samples into this space, facilitating the learning of disentangled representations with constraints tailored for tabular data.
However, these techniques are limited to learning from scratch, or training multiple loss functions simultaneously, leading to difficulties in effectively optimizing the target loss function.
% However, these techniques are limited to learning from scratch, or training multiple loss functions simultaneously, leading to difficulties in effectively optimizing the target loss function, thereby possibly resulting in sub-optimal representation learning.

Another novel line of works explore the direction of pre-learning for the tabular domain. 
This line of works design a set of pretext tasks that are challenging but highly relevant to the objective tasks and optimize the parameters of the backbone models via pre-training on these pretext tasks.  
For example, TabNet~\cite{arik2021tabnet} and VIME~\cite{yoon2020vime} try to recover the corrupted inputs with auto-encoding loss. 
SCARF~\cite{bahriscarf} utilizes a contrastive loss similar to SimCLR~\cite{chen2020simple} between the sample and its corrupted counterpart.
SubTab~\cite{ucar2021subtab} incorporates a combination of both. 
% However, the designing of pretext tasks is often inspired by some prior knowledge of the objective tasks. 
However, these pretext tasks are often inspired by some prior knowledge and need much extra training effort.

Our paper introduces a novel deep Tabular Representation Corrector, TRC, to enhance the representations of any trained deep tabular backbone model without altering its parameters. 
Unlike existing approaches that directly intervene in the learning process of backbones (in-learning) or rely on prior knowledge to design pretext tasks for pre-training (pre-learning), TRC provides a cost-effective, parameter-efficient technique to enhance representations for deep tabular models to improve performance in a model-agnostic way.
Besides, with TRC, we aim to solve two inherent issues hindering prediction: (i) \textit{Representation Shift}, where inherent noise within the observations can lead the learned representations by deep tabular models to deviate from the ideal optimal representations, and (ii) \textit{Representation Redundancy}, where deep tabular models may incorporate redundant information in the representation space, thereby resulting in inaccurate prediction.

To overcome the aforementioned challenges, we propose two tasks: \textit{Tabular Representation Re-estimation} and \textit{Tabular Space Mapping}. 
Specifically, for \textit{Tabular Representation Re-estimation}, we fit an estimator to compute and eliminate the latent shift within each representation of the trained tabular model, thereby re-estimating the representations. 
For \textit{Tabular Space Mapping}, we transform the re-estimated representations into a newly defined space characterized by a set of embedding vectors to condense the information. 
Furthermore, additional strategies are proposed to preserve the critical knowledge for prediction.
The redundant information is removed by reducing the information capacity while preserving critical information necessary for accurate prediction.
The contributions of this paper include:

% 1) We propose a novel post-learning paradigm, \method for deep tabular machine learning, which could enhance learned representations of trained deep tabular backbone models without altering any of their parameters.
% 2) We propose two post-learning tasks, the \textit{Tabular Representation Re-estimation} and \textit{Tabular Space Mapping} to overcome the issues of \textit{Representation Shift} and \textit{Representation Redundancy}.
% 3) We conducted extensive experiments in state-of-the-art (SOTA) deep tabular models coupled with \method on various tabular benchmarks and the comprehensive results demonstrate our effectiveness.

\begin{itemize}[leftmargin=*]
\item We propose a novel tabular representation corrector, \method, for deep tabular machine learning, which enhances the learned representations of any trained deep tabular backbone without altering its parameters in a model-agnostic manner.
\item We propose two tasks in TRC, i.e., \textit{Tabular Representation Re-estimation} and \textit{Tabular Space Mapping} to overcome the issues of {Representation Shift} and {Representation Redundancy} respectively hindering tabular predictions.
\item We conduct extensive experiments on state-of-the-art (SOTA) deep tabular models coupled with \method on various tabular benchmarks. 
The comprehensive results along with analysis and visualizations demonstrate the effectiveness of \method, even in challenging scenarios involving missing values and reduced training samples.
\end{itemize}

% \begin{itemize}[leftmargin=*]
%     \item We investigated the learned patterns of deep tabular models and explore a novel direction of \textit{applying prototype learning for tabular machine learning} to address representation entanglement and localization.
%     \item We propose a model-agnostic prototype-based tabular representation learning framework, 
%     \textsc{PTaRL} for tabular prediction tasks, which transforms data into the prototype-based projection space and optimize representations via Optimal Transport.
%     \item We propose two different strategies, the Coordinates Diversifying Constraint and the Matrix Orthogonalization Constraint to make \textsc{PTaRL} learn disentangled representations.
%     \item We conducted extensive experiments in \textsc{PTaRL} coupled with state-of-the-art (SOTA) deep tabular ML models on various tabular benchmarks and the comprehensive results along with analysis and visualizations demonstrate our effectiveness.
%     %various tabular datasets along with comprehensive analysis and visualization. 
% % These results validate the effectiveness of the proposed framework and may provide valuable insights for further research on versatile tabular deep network boosters.
% \end{itemize}

\section{Related Work}
Inspired by the success of deep learning in CV~\cite{he2016deep} and NLP~\cite{devlin2018bert}, numerous deep learning methods have been proposed for tabular domain to accomplish prediction tasks.
Research on deep tabular machine learning currently encompasses two main directions: in-learning and pre-learning styles.

\subsection{In-learning for Tabular Data}
In the context of the \textit{in-learning} style, neural networks with various architectures are typically trained from scratch, or extra constraints are applied to better regulate learned representations~\cite{song2019autoint,gorishniy2021revisiting,wang2021dcn,klambauer2017self,kadra2021well,jeffares2022tangos,ye2024ptarl}.
Among these approaches, AutoInt~\cite{song2019autoint} transforms input features into embeddings, which are then processed through a series of attention-based transformations. 
Building on this, FT-Transformer~\cite{gorishniy2021revisiting} improves upon AutoInt by enhancing token embeddings, using element-wise multiplication for numerical features and element-wise lookup tables for categorical features. 
Similarly, ResNet for tabular data~\cite{gorishniy2021revisiting} demonstrates significant performance improvements. DCN2~\cite{wang2021dcn} introduces a feature-crossing module that combines linear layers and multiplication operations with an MLP-like structure, enhancing its ability to capture feature interactions. 
SNN~\cite{klambauer2017self} proposes neuron activations designed to maintain zero mean and unit variance, promoting better regularization. 
\textit{Kadra et al.}~\cite{kadra2021well} further regularizes the MLP by searching for the optimal combination of multiple regularization techniques
for each dataset using a joint optimization over the decision on which regularizers to apply.
TANGOS~\cite{jeffares2022tangos} proposes a new regularization technique for tabular data, encouraging neurons to focus on sparse, non-overlapping input features.
More recently, PTaRL~\cite{ye2024ptarl} constructs a new projection space consisting of prototypes and projects data samples into this space, facilitating the learning of disentangled representations with constraints tailored for tabular data.
However, these techniques are limited to learning from scratch, or training multiple loss functions simultaneously, leading to difficulties in effectively optimizing the target loss function.

\subsection{Pre-learning for Tabular Data}
\textit{Pre-learning} methods design several pretext tasks for pre-training and then conduct task-specific fine-tuning~\cite{yoon2020vime,bahriscarf,ucar2021subtab,somepalli2022saint,rubachev2022revisiting}.
For instance, VIME~\cite{yoon2020vime} introduces a novel pretext task that estimates mask vectors from corrupted tabular data alongside a reconstruction task for self-supervised learning. 
% For example, TabNet~\cite{arik2021tabnet} and VIME~\cite{yoon2020vime} try to recover the corrupted inputs with auto-encoding loss. 
SCARF~\cite{bahriscarf} employs a contrastive loss akin to SimCLR~\cite{chen2020simple}, contrasting samples with their corrupted versions by corrupting a random subset of features. 
SubTab~\cite{ucar2021subtab} divides the input features to multiple subsets,
and then incorporates a combination of both reconstruction and contrastive loss.
SAINT~\cite{somepalli2022saint} enhances this by integrating a Transformer with row-wise attention to capture inter-sample interactions and contrastive information during pre-training. 
In addition, \textit{Rubachev et al.}~\cite{rubachev2022revisiting} indicates that pre-training MLP with several self-supervised learning objectives could achieve promising results.
However, these pretext tasks are often inspired by some prior knowledge and need much extra training effort.
Recently, alternative approaches have leveraged external information outside target dataset during pre-training to enhance deep learning for tabular data. 
TransTab ~\cite{wang2022transtab} incorporates feature name information into Transformer to achieve cross table learning.
XTab ~\cite{zhu2023xtab} pre-trains Transformer on a variety of datasets and aims to enhance tabular deep learning on a previously unseen dataset. 
These methods need additional information outside target dataset and also increase the time complexity, which we do not consider as closely related baseline methods to ours as the primary goal, motivation, and methodology are different.

Different from existing approaches that directly intervene in the learning process of deep tabular models (in-learning), or rely on prior knowledge or additional information outside the target dataset to design pretext tasks for pre-training thus increasing time overhead (pre-learning), tabular representation corrector provides a cost-effective, parameter-efficient technique to enhance representations for deep tabular models.

% \section{Background}
\section{Problem Formulations}
\label{sec: background}
\textbf{Notation.} 
Denote the $i$-th sample as $(x_i, y_i)$,  where $x_i = (x_i^{(num)}, x_i^{(cat)}) \in \mathbb{X}$ represents numerical and categorical features respectively and $y_i \in \mathbb{Y}$ is the corresponding label.
% The total number of features is denoted as $N  \rr{ n?}$.
A real-world tabular dataset $\mathcal{D}=\{(x_i, y_i)\}_{i=1}^N$ is a collection of $N$ data samples.
We denote the training set as $\mathcal{D}_{train}$, the validation set for early stopping as $\mathcal{D}_{val}$, and the test set for the final evaluation as $\mathcal{D}_{test}$.
We consider deep learning for supervised tabular prediction tasks: regression $\mathbb{Y} = \mathbb{R}$, binary classification $\mathbb{Y} = \{0, 1\}$ and multiclass classification $\mathbb{Y} = \{1, . . . , C\}$. 
The goal is to obtain an accurate deep tabular model $F(\cdot;\theta): \mathbb{X} \to \mathbb{Y}$ trained on $\mathcal{D}_{train}$\textcolor{blue}{, that minimizes the expected loss $\mathbb{E}[\mathcal{L}(F(x;\theta), y)]$. Here, $\mathcal{L}$ denotes a task-specific loss function, typically the mean squared error for regression and the cross-entropy loss for classification.}

\textbf{In-learning and Pre-learning.}
\textcolor{blue}{Recall that the deep tabular model $F(\cdot;\theta)$ usually includes the backbone $G_f(\cdot;\theta_f)$ parameterized with $\theta_f$ and prediction head $G_h(\cdot;\theta_h)$ parameterized with $\theta_h$. 
Standard in-learning methods aim to minimize the following loss function:
\begin{equation}
\label{equ:in learning}
\begin{aligned}
    \min_{\theta_f, \theta_h} \frac{1}{N}\sum_{i=1}^N \mathcal{L}(G_h(z_i;\theta_h), y_i), \quad z_i = G_f(x_i;\theta_f),
\end{aligned}
\end{equation}
where $N$ is the size of $\mathcal{D}_{train}$, $z_i$ denotes the representation extracted by $G_f(x_i;\theta_f)$.}
% Let us decompose the deep tabular model $F(\cdot;\theta)$ as $G_h(G_f(\cdot;\theta_f);\theta_h)$, where $G_f(\cdot;\theta_f)$ is the backbone and $G_h(\cdot;\theta_h)$ is the projection head parameterized by $\theta_f$ and $\theta_h$ respectively. The representations are the output of the backbone.
% Standard in-learning methods aim to minimize the following loss function:
% \begin{equation}
% \label{equ:in learning}
% \begin{aligned}
%     &\min_{\theta_f, \theta_h} \frac{1}{N}\sum_{i=1}^N \mathcal{L}(G_h(z_i;\theta_h), y_i)\\
%     = \min_{\theta_f, \theta_h}& \frac{1}{N}\sum_{i=1}^N \mathcal{L}(G_h(G_f(x_i;\theta_f);\theta_h), y_i),
% \end{aligned}
% \end{equation}
% where $N$ is the size of $\mathcal{D}_{train}$, $z_i$ denotes the representation.
In some cases, additional constraints may be optimized simultaneously. 
The trained $G_h(G_f(\cdot;\theta_f);\theta_h)$ is then used for prediction.

For pre-learning methods, pretext tasks $\mathcal{D}_s = \{(x_i^s, y_i^s)\}_{i=1}^N$, such as reconstructing original features from corrupted ones, are designed to pre-train the backbone $G_f(\cdot;\theta_f)$:
\textcolor{blue}{
\begin{equation}
\label{equ:pre learning}
\begin{aligned}
    \min_{\theta_f, \theta_h^s} \frac{1}{N}\sum_{i=1}^N \mathcal{L}(G_h^s(z_i^s;\theta_h^s), y_i^s), \quad z_i^s = G_f(x_i^s;\theta_f),
\end{aligned}
\end{equation}
}
% \begin{equation}
% \label{equ:pre learning}
% \begin{aligned}
%     &\min_{\theta_f, \theta_h^s} \frac{1}{N}\sum_{i=1}^N \mathcal{L}(G_h^s(z_i^s;\theta_h^s), y_i^s)\\
%     = \min_{\theta_f, \theta_h^s}& \frac{1}{N}\sum_{i=1}^N \mathcal{L}(G_h^s(G_f(x_i^s;\theta_f);\theta_h^s), y_i^s),
% \end{aligned}
% \end{equation}
where $N$ is the size of $\mathcal{D}_{s}$ and $G_h^s$ is used to predict $y_i^s$ based on $z_i^s$.
After pre-training, the backbone $G_f(\cdot;\theta_f)$ can be used to extract better data representations.
A new head $G_h(\cdot;\theta_h)$ is then attached to the trained backbone $G_f(\cdot;\theta_f)$ to predict the true labels, and either $G_h(\cdot;\theta_h)$ or both $G_h(\cdot;\theta_h)$ and $G_f(\cdot;\theta_f)$ are further optimized (as in Eq.~\ref{equ:in learning}).

\textbf{Tabular Representation Corrector (TRC).}
\textcolor{blue}{In this paper, we consider the problem of enhancing the representation $z = G_f(x;\theta_f)$ produced by a trained deep tabular backbone, while fixing the whole parameters of $G_f(\cdot;\theta_f)$. 
To achieve this, TRC introduces a learnable correction module $G_p(\cdot;\theta_p)$, which takes the frozen representation $z$ as input and outputs an improved representation $G_p(z;\theta_p)$.
A new head $G_h(\cdot;\theta_h)$ is then attached to $G_p(\cdot;\theta_p)$ for label prediction.
During training, only $G_p(\cdot;\theta_p)$ and $G_h(\cdot;\theta_h)$ are updated.
TRC applies regardless of whether the backbone $G_f(\cdot;\theta_f)$ was trained in an in-learning or pre-learning manner.}
\section{Methodology}

% \subsection{Motivation and Preliminaries}
\subsection{Motivation}
\label{sec:motivation details}
In the context of deep tabular machine learning, the inherent heterogeneity of features poses significant challenges to achieving satisfactory performance using deep models. 
Existing approaches learn tabular representations by directly intervening in the learning process of deep tabular models (in-learning) or designing pretext tasks for pre-training based on prior knowledge or additional information outside the target dataset (pre-learning). 
We propose a novel deep Tabular Representation Corrector (TRC), that provides a cost-effective, parameter-efficient technique to enhance the representations of trained deep tabular models.
Specifically, the TRC could solve two inherent representation issues hindering predictions.

\begin{figure*}[t!]
    \centering
    \begin{subfigure}[b]{0.24\linewidth}
        \centering
        \includegraphics[width=\linewidth]{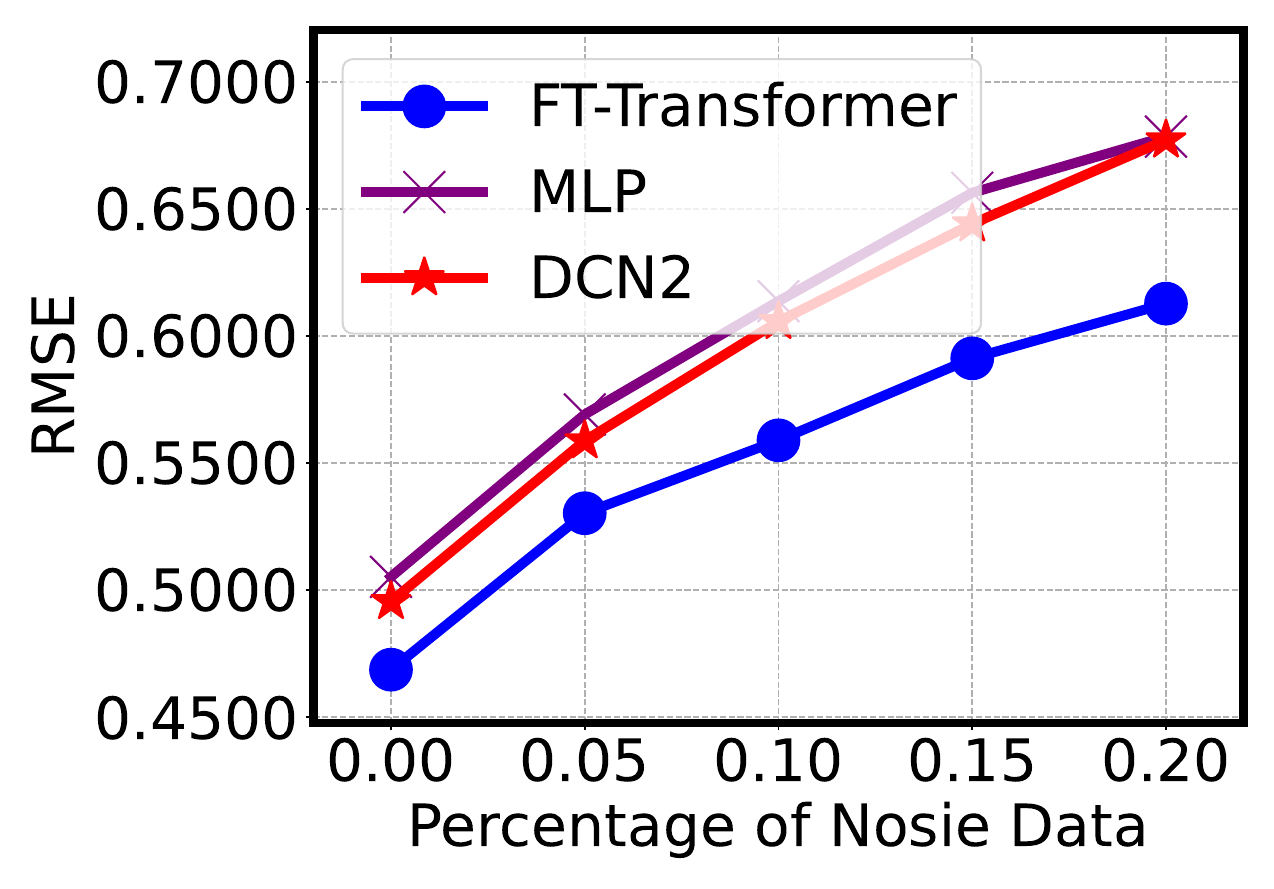}
        % \captionsetup{font=small}
        % \captionsetup{skip=0pt}
        \caption{CA dataset $\downarrow$}
    \end{subfigure}
    \begin{subfigure}[b]{0.24\linewidth}
        \centering
        \includegraphics[width=\linewidth]{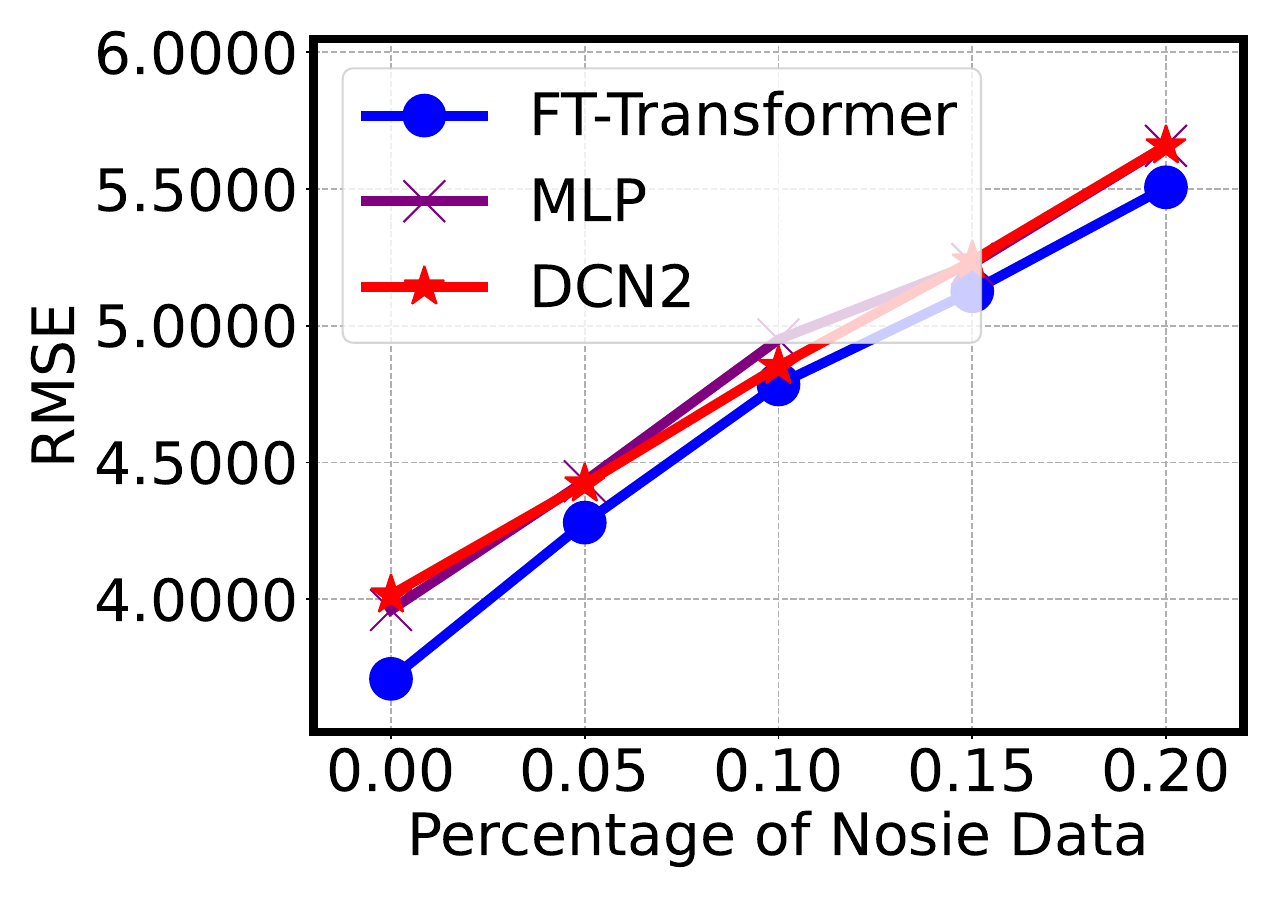}
        % \captionsetup{font=small}
        % \captionsetup{skip=0pt}
        \caption{CO dataset $\downarrow$}
    \end{subfigure}
    \begin{subfigure}[b]{0.24\linewidth}
        \centering
        \includegraphics[width=\linewidth]{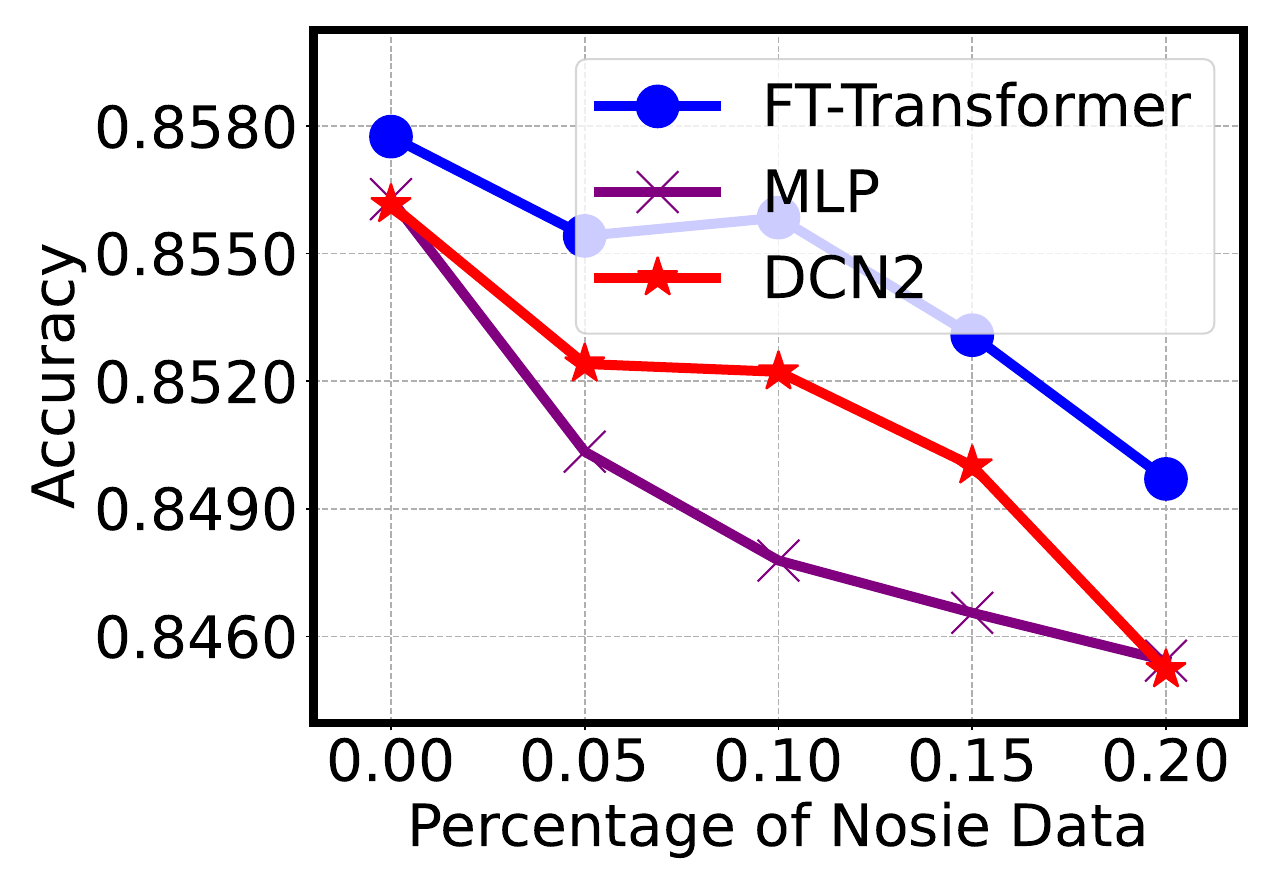}
        % \captionsetup{font=small}
        % \captionsetup{skip=0pt}
        \caption{AD dataset $\uparrow$}
    \end{subfigure}
    \begin{subfigure}[b]{0.24\linewidth}
        \centering
        \includegraphics[width=\linewidth]{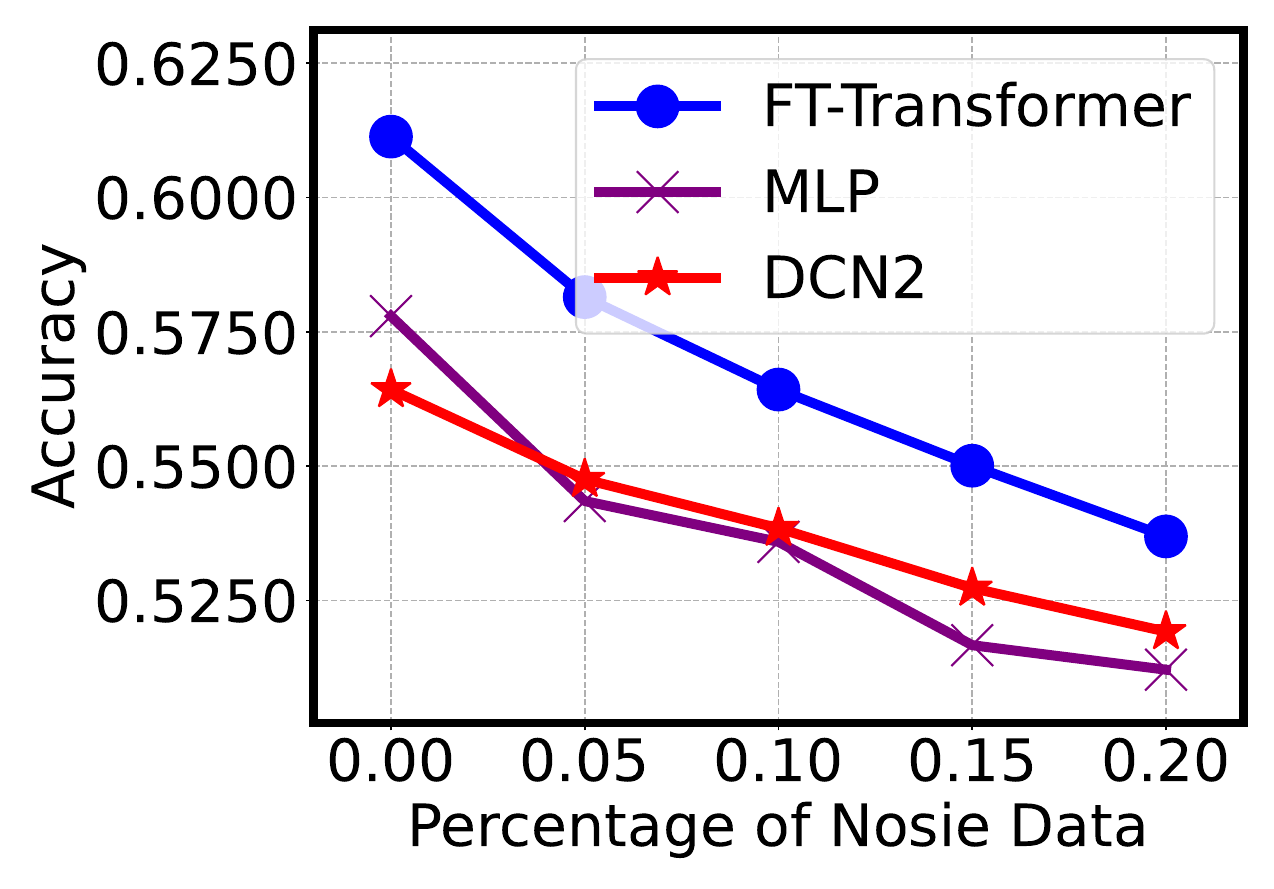}
        % \captionsetup{font=small}
        % \captionsetup{skip=0pt}
        \caption{GE dataset $\uparrow$}
    \end{subfigure}
    % \captionsetup{font=small}
    % \captionsetup{skip=4pt}
    \caption{The performance of deep tabular models with varying noise levels in observations. \textcolor{blue}{FT-Transformer}, MLP, and DCN2 indicate different deep tabular models. For regression tasks, lower RMSE is better, and for classification tasks, higher accuracy is better.}
    \label{fig:motivation1}
\end{figure*}

(i) \textit{Representation Shift}. As stated by~\cite{chang2023data, borisov2022deep}, there is a natural possibility of noise in the collection process due to the heterogeneous nature of tabular data. 
% Owing to feature heterogeneity, observations in the dataset are susceptible to varying degrees of noise. 
In this work, we claim that such inherent but unobserved noise within the original observations can lead the representations extracted by deep tabular models to deviate from the ideal optimal representations (called representation shift in this paper), thereby decreasing the model performance. 
To verify this point, we perturb the original features with noise as a form of data augmentation, leading to the deteriorated performance of existing deep tabular models as illustrated in Fig.~\ref{fig:motivation1}.
Moreover, to further substantiate the impact of data noise on deep tabular models, we systematically increase the noise ratio and observe a consistent decline in model performance.
% Our empirical observations underscore the realization that the presence of noise in the original observed tabular dataset is the inherent limitation that can hinder the deep tabular models from attaining optimal representations.  
Therefore, given the well-trained deep tabular models learned by the original observed tabular dataset, we assume that they might be limited in extracting optimal representations due to the existence of inherent noise in tabular data.

\begin{figure*}[!t]
    \centering
    \begin{subfigure}[b]{0.24\linewidth}
        \centering
        \includegraphics[width=\linewidth]{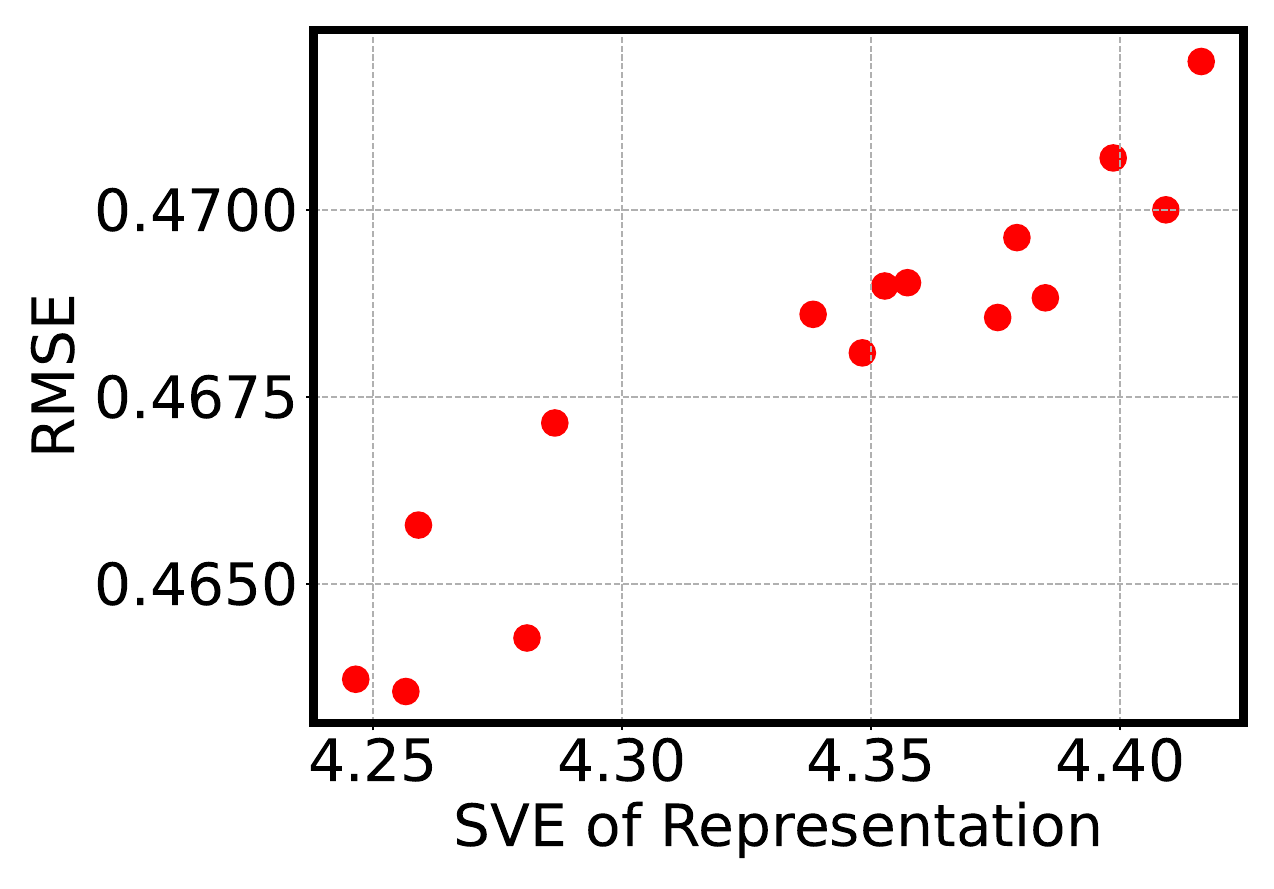}
        % \captionsetup{font=small}
        % \captionsetup{skip=0pt}
        \caption{\textcolor{blue}{FT-Transformer} on CA $\downarrow$}
    \end{subfigure}
    \begin{subfigure}[b]{0.24\linewidth}
        \centering
        \includegraphics[width=\linewidth]{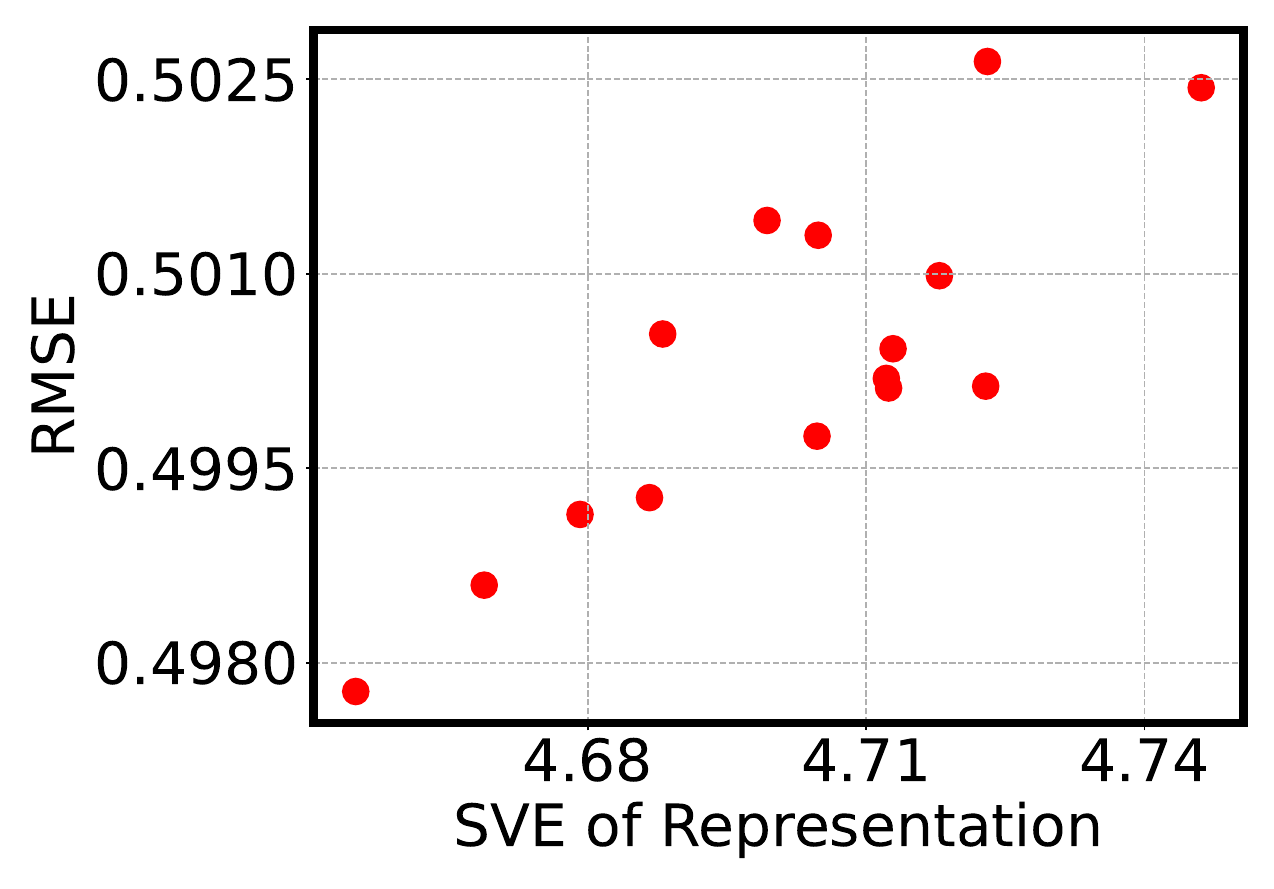}
        % \captionsetup{font=small}
        % \captionsetup{skip=0pt}
        \caption{MLP on CA $\downarrow$}
    \end{subfigure}
    \begin{subfigure}[b]{0.24\linewidth}
        \centering
        \includegraphics[width=\linewidth]{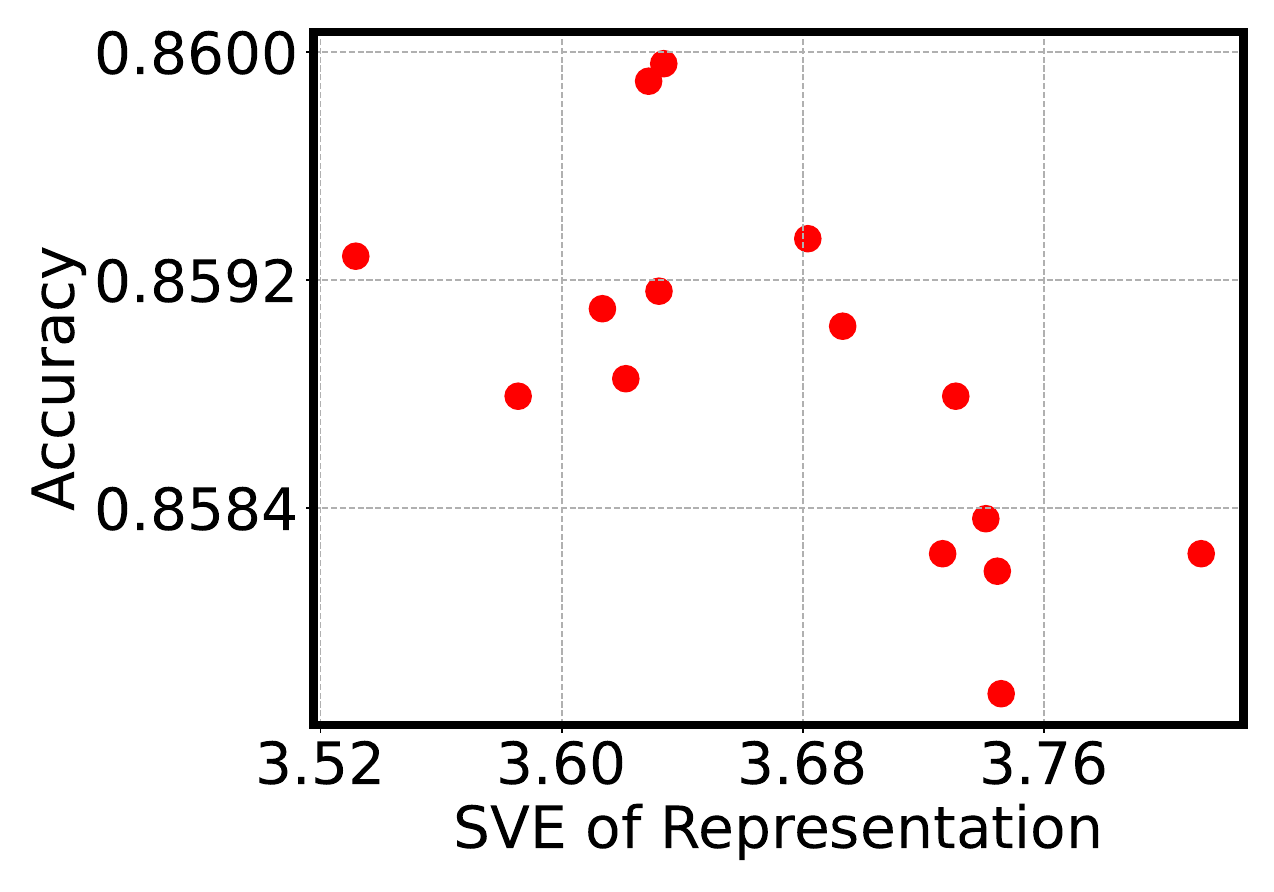}
        % \captionsetup{font=small}
        % \captionsetup{skip=0pt}
        \caption{\textcolor{blue}{FT-Transformer} on AD $\uparrow$}
    \end{subfigure}
    \begin{subfigure}[b]{0.24\linewidth}
        \centering
        \includegraphics[width=\linewidth]{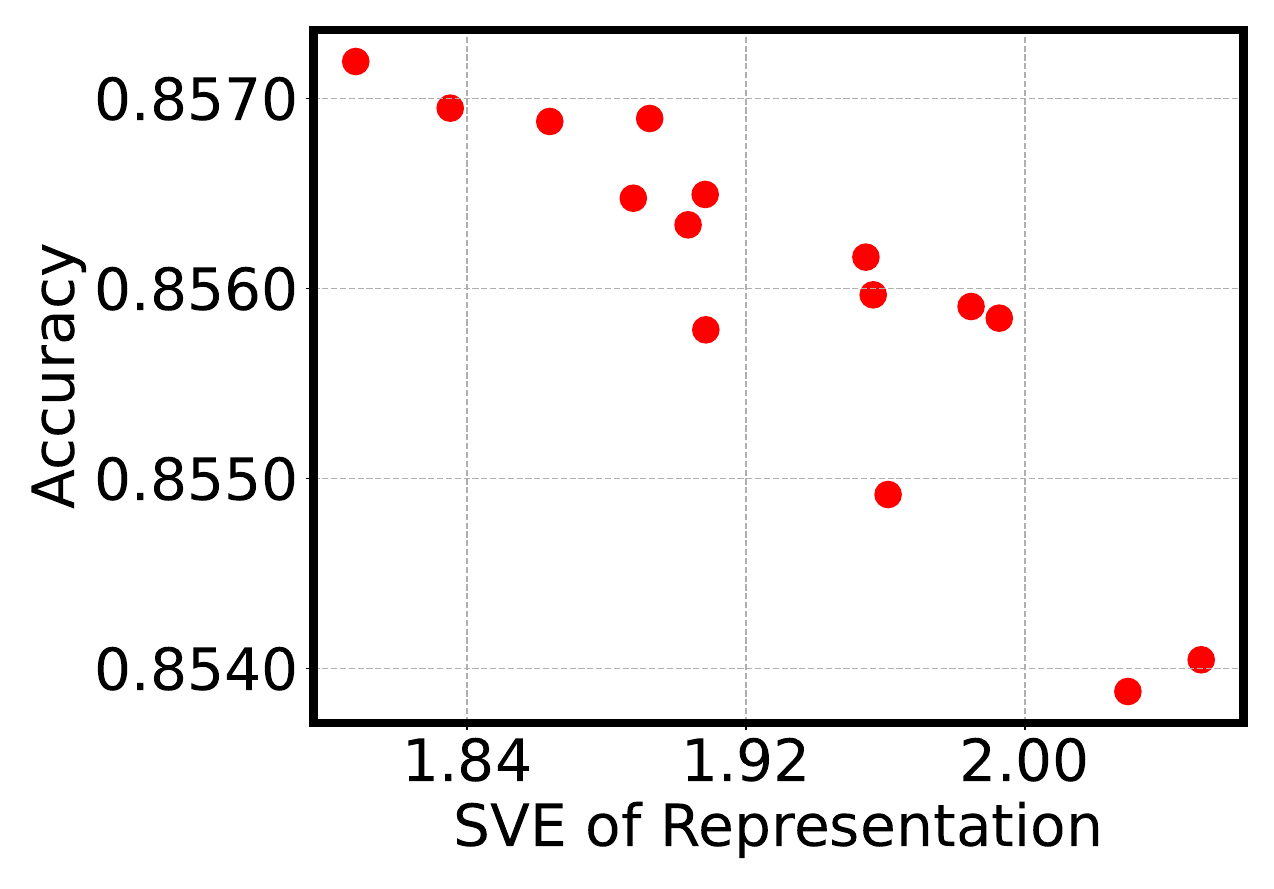}
        % \captionsetup{font=small}
        % \captionsetup{skip=0pt}
        \caption{MLP on AD $\uparrow$}
    \end{subfigure}
    % \captionsetup{font=small}
    % \captionsetup{skip=4pt}
    \caption{The relation between model performance and the corresponding SVE values of representations. For each subfigure, we conduct experiment on multiple random seeds. We find that deep tabular models with higher SVE often yield lower performance.}
    \label{fig:motivation2}
\end{figure*}

(ii) \textit{Representation Redundancy}. To measure the complexity of representation space, we perform singular value decomposition (SVD) on the representations $\mathcal{Z}=\{z_i\}_{i=1}^N\in \mathbb{R}^{N\times D}$ to obtain: $\mathcal{Z}=\textbf{U}\mathbf{\Sigma}\textbf{V}^\intercal$ and introduce singular value entropy (SVE) as follows:

\textbf{Definition 1. Singular Value Entropy.} The singular value entropy (SVE) is defined as the entropy of normalized singular values:
\begin{equation}
\label{SVE}
\begin{aligned}
    \text{SVE} = - \sum_{i=1}^{D}\frac{\sigma_i}{\sum_{j=1}^{D}\sigma_j}\log\frac{\sigma_i}{\sum_{j=1}^{D}\sigma_j},
\end{aligned}
\end{equation}
where $N$ is the number of samples, $D$ is the dimensionality, $\textbf{U}$ and $\textbf{V}$ denote the left and right singular vector matrices, respectively, and $\mathbf{\Sigma}$ denotes the diagonal singular value matrix $\{\sigma_i\}_{i=1}^D$.
The singular value spectrum is widely considered to be related to the generalization performance~\cite{oymak2019generalization, chen2019transferability, xue2022investigating}.
SVE measures the flatness of the singular value distribution~\cite{chen2024understanding}. 
Greater SVE values in the latent space indicate a more comprehensive capture of data structure and a higher level of information content within the latent space.
However, it is observed in Fig.~\ref{fig:motivation2} that existing deep tabular models with higher SVE do not necessarily yield better performance; rather, the models with better performance often exhibit relatively lower SVE.
% However, it is observed in Fig.~\ref{fig:motivation2} that existing deep tabular models with higher SVE do not necessarily yield better performance; rather, the models with the best performance often have relatively smaller SVE.
For each setting, we conduct multiple experiments with random initialization.
This observation suggests that, due to feature heterogeneity, deep tabular models may incorporate redundant information in the representation space, thereby affecting accurate prediction. 


\begin{figure*}[!t]
\centering
\includegraphics[width=\linewidth]{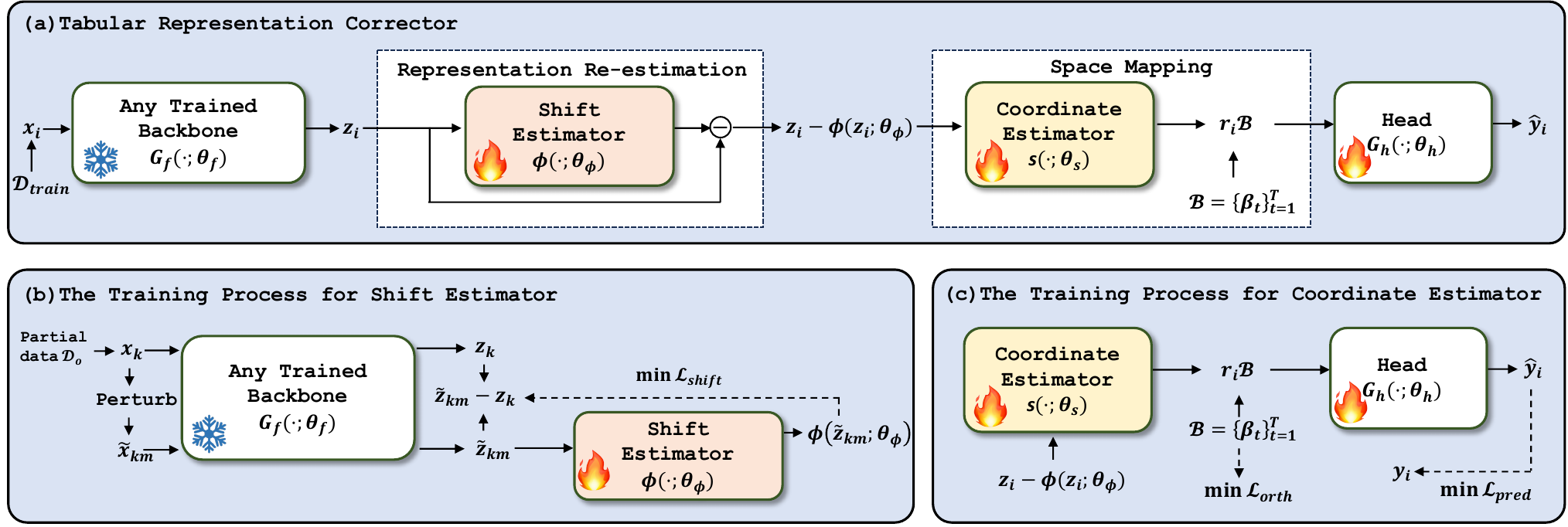}
% \includegraphics[width=\linewidth]{pictures/framework.pdf}
% \captionsetup{font=small}
% \captionsetup{skip=0pt}
\caption{The framework of Tabular Representation Corrector (TRC). Subfigure (a) illustrates the overall TRC framework, subfigure (b) presents the training process for the shift estimator of TRC, and subfigure (c) presents the training process for the coordinate estimator of TRC. Here, $z_i = G_f(x_i;\theta_f)$ is the output of any trained backbone. $z_i$ would be enhanced via two tasks. In subfigure (b), the explicitly perturbed samples are only used for training the shift estimator.  During the test stage, we feed the representations extracted by the existing trained deep tabular model, which are from the test dataset (notably, we do not add any noise to the test dataset), into the shift estimator followed by the coordinate estimator to achieve the calibrated representations of test samples.} 
\label{fig:framework}
\end{figure*}

\textcolor{blue}{Targeting the above two inherent representation issues, we propose two tasks, i.e., (i) \textit{Tabular Representation Re-estimation}, which involves training a shift estimator to calculate and mitigate the inherent shift in the representation $z = G_f(x;\theta_f)$ extracted by a trained deep tabular backbone, thereby re-estimating $z$, and (ii) \textit{Tabular Space Mapping}, 
which transforms the re-estimated representation into a light-embedding vector space via a coordinate estimator, aiming to preserve critical predictive information while reducing redundancy.
These two tasks jointly enhance the frozen representation $z$ without altering $G_f(\cdot;\theta_f)$ thus enjoying high efficiency.
During inference, for each test sample (notably, we do not add any noise to the test dataset), the representation produced by the fixed $G_f(\cdot;\theta_f)$ is sequentially processed by the shift estimator and the coordinate estimator to obtain a calibrated representation.}
Note that \method, as a general representation learning framework, is model-agnostic such that it can be coupled with any trained deep tabular backbone $G_f(\cdot;\theta_f)$ to learn better representations. 
In the following, we will elaborate on the details of \method. 
Fig.~\ref{fig:framework} gives an overview framework.

\subsection{Tabular Representation Re-estimation} 
\label{sec:task1}

\textcolor{blue}{Due to feature heterogeneity, deep tabular models are susceptible to the influence of various types of noise as depicted in Fig.~\ref{fig:motivation1}, which often leads to undesirable shifts in the models' latent space.
This causes the learned representations (sub-optimal representations) to deviate from the ideal optimal ones.}
Directly intervening in the learning process of deep tabular models or designing appropriate pretext tasks is challenging due to the unknown optimal representations.
To alleviate this issue, we propose a task via representation re-estimation to approximate and mitigate inherent shift in the sub-optimal representations of the trained deep tabular models.
This process involves the following stages: (i) \textcolor{blue}{Approximated} Optimal Representations Searching, which identifies a subset of representations containing the least shift information towards the ideal optimal representations; (ii) Inherent Shift Learning, where varying types of random signals are manually injected into the \textcolor{blue}{above selected approximated optimal representations} \textcolor{blue}{to train a robust shift estimator capable of modeling the inherent shift present in sub-optimal representations;} and (iii) \textcolor{blue}{Re-estimation of Sub-optimal Representations}, which \textcolor{blue}{applies the learned shift estimator to remove the estimated shift from each sub-optimal representation, thereby re-estimating the representations.}
% utilizes the shift estimator to compute and mitigate the inherent shift within each sub-optimal representation, thereby re-estimating the representations.

\textbf{\textcolor{blue}{Approximated} Optimal Representations Searching.}
% Latent noises in observations may introduce shift in the representations due to the heterogeneous nature of tabular features, resulting in sub-optimal representation learning.  
By estimating the shift information present in each sub-optimal representation, we can infer the overall shift information within the whole representations of deep tabular models.
To achieve this, we propose searching a subset of samples with relatively optimal representations (\textcolor{blue}{approximated optimal representations}) and then manually introducing shift information into these representations to simulate representations with shift, i.e. simulated sub-optimal representations.
With the simulated sub-optimal representations and their corresponding shift, we can fit a shift estimator to compute the overall shift information, which would be introduced in the subsequent stage.
To obtain the set of samples with \textcolor{blue}{approximated optimal representations}, we first introduce an assumption as follows:

\textbf{Assumption 1.} 
\textit{If we possess a deep tabular model $F(\cdot;\theta) = G_h(G_f(\cdot;\theta_f);\theta_h)$ trained on $\mathcal{D}_{train}$, then there exists a set of samples $\mathcal{D}_o \subseteq \mathcal{D}_{val}$ with lowest gradient norms among $\mathcal{D}_{val}$. $\forall (x,y)\in \mathcal{D}_o$, the following condition holds: 
\begin{equation}
\begin{aligned}
    \mathrm{Rank}(\|\nabla_{\theta_f}\mathcal{L}(F(x;\theta), y)\|_p^q)/|\mathcal{D}_{val}| \leq \tau, 
\end{aligned}
\end{equation}
where $\mathrm{Rank}(\cdot)$ denotes the ranking (integer) in ascending order within $\mathcal{D}_{val}$, $\|\cdot\|_p^q$ denotes the $q$ square of the $L_p$ norm, $\mathcal{L}$ denotes the supervised loss function,  and $\tau$ denotes the threshold that defines $\mathcal{D}_o$.}
% \textbf{Assumption 1.} 
% \textit{If we possess a deep tabular model $F(\cdot;\theta) = G_h(G_f(\cdot;\theta_f);\theta_h)$ trained on $\mathcal{D}_{train}$, then there exists a set of samples $\mathcal{D}_o$ with lowest gradient norms derived from $\mathcal{D}_{val}$, where each sample's representation $z = G_f(x;\theta_f) \in \mathcal{Z}_o$ satisfies the condition $\mathrm{Rank}(\|\nabla_{\theta_f}\mathcal{L}(F(x;\theta), y)\|_p^q)/|\mathcal{D}_{val}| \leq \tau$, with $\mathrm{Rank}(\cdot)$ denoting the ranking (integer) in ascending order within $\mathcal{D}_{val}$, $\|\cdot\|_p^q$ denoting the $q$ square of the $L_p$ norm, $\mathcal{L}$ denoting the supervised loss function,  and $\tau$ denoting the threshold that defines the representations $\mathcal{Z}_o$.}
% \textbf{Assumption 1.} 
% \textit{If we possess a deep tabular model $F(\cdot;\theta) = G_h(G_f(\cdot;\theta_f);\theta_h)$ trained on $\mathcal{D}_{train}$, then there exists a set of representations $\mathcal{A}$ with lowest gradient norms derived from $\mathcal{D}_{val}$, where each representation $G_f(x;\theta_f) \in \mathcal{A}$ satisfies the condition $\mathrm{Rank}(\|\nabla_{\theta_f}\mathcal{L}(F(x;\theta), y)\|_p^q)/|\mathcal{D}_{val}| \leq \tau$, with $\mathrm{Rank}(\cdot)$ denoting the ranking (integer) in ascending order within $\mathcal{D}_{val}$, $\|\cdot\|_p^q$ denoting the $q$ square of the $L_p$ norm, $\mathcal{L}$ denoting the supervised loss function,  and $\tau$ denoting the threshold that defines the representations $\mathcal{A}$.}

After the trained backbone $G_f(\cdot;\theta_f)$ is obtained, we proceed to compute the gradient norms for each sample across $\mathcal{D}_{val}$. 
The searching process is conducted on $\mathcal{D}_{val}$ to prevent overfitting and reduce the time cost.
Based on Assumption 1, we assume that $\mathcal{D}_o$'s corresponding representation set $\mathcal{Z}_o = \{z_k = G_f(x_k;\theta_f)|(x_k,y_k)\in\mathcal{D}_o, k \in \{1,2,...,K\}\}$, where $K$ denotes the size (the threshold $\tau$ is set to 0.01 by default), contains those with the least shift information towards the ideal optimal representations, thus $\mathcal{Z}_o$ could be 
\textcolor{blue}{treated as the approximated optimal representations.}

\textbf{Inherent Shift Learning.}
After acquiring the \textcolor{blue}{approximated} optimal representations $\mathcal{Z}_o = \{z_k = G_f(x_k;\theta_f)|(x_k,y_k)\in\mathcal{D}_o, k \in \{1,2,...,K\}\}$, we guide the training process of the shift estimator $\phi$ by incorporating different types of shift signals into the \textcolor{blue}{approximated} optimal representations $\mathcal{Z}_o$ to generate simulated sub-optimal representations $\tilde{\mathcal{Z}}_o = \{\tilde{z}_{km}\}_{k, m=1}^{K, M}$ and the corresponding shift information $\Delta = \{\Delta_{km}\}_{k, m=1}^{K, M}$, which is as follows: 
\begin{equation}
\label{Shift Learning}
\begin{aligned}
    \min\mathcal{L}_{shift} = \min_{\theta_\phi} \frac{1}{KM} \sum_{k, m=1}^{K, M}\|\phi(\tilde{z}_{km};\theta_\phi) - \Delta_{km}\|_2^2,
\end{aligned}
\end{equation}

\begin{equation}
\label{z_tilde}
\begin{aligned}
    \tilde{z}_{km} = G_f&(\tilde{x}_{km};\theta_f) \\= G_f(\textbf{m}_{km} \odot x_k + (&1-\textbf{m}_{km}) \odot \epsilon_k;\theta_f),
\end{aligned}
\end{equation}  

\begin{equation}
\label{delta_tilde}
\begin{aligned}
    \Delta_{km} = \tilde{z}_{km} - z_k, 
\end{aligned}
\end{equation}
% \begin{equation}
% \footnotesize
% \label{Shift Learning}
% \begin{aligned}
%     % & \min \mathcal{L}_{shift} = \\ & 
%     \min_{\theta_\phi} \sum_{k, m=1}^{K, M}\|\phi(G_f(\textbf{m}^{km} \odot x_k + (1-\textbf{m}^{km}) \odot \epsilon_k;\theta_f);\theta_\phi) - \Delta_{km}\|_2^2, 
% \end{aligned}
% \end{equation}
% where $K$ denotes the number of optimal representations in $\mathcal{A}$, $\textbf{m}^{km} \in \{0,1\}^d$ is a feature mask vector with feature number as $d$ and mask ratio as $\eta$, $\epsilon_k$ is the noise that would be added to the corresponding feature observations $x_k$ of $\mathcal{A}_k$, $\Delta_{km} = \phi(G_f(\textbf{m}^{km} \odot x_k + (1-\textbf{m}^{km}) \odot \epsilon_k;\theta_f);\theta_\phi) - \phi(G_f(x_k;\theta_f);\theta_\phi)$. 
where $\textbf{m}_{km} \in \{0,1\}^d$ is a feature mask vector with feature number as $d$, $\epsilon_k \in \mathbb{R}^d$ is the noise that would be added to the corresponding observation $x_k$ of $z_k$, $M$ is the perturbing times.
Given the feature heterogeneity of tabular data, we propose generating $\epsilon_k$ by sampling each feature element from the empirical distribution of the corresponding column in $\mathcal{D}_{train}$, following previous works~\cite{yoon2020vime, ucar2021subtab, nam2023stunt}.
% As mentioned in the previous stage, through perturbing observations, we manually introduce shift information into the optimal representations $\mathcal{Z}_o$ to obtain simulated sub-optimal representations ($z_k \to \tilde{z}_{km}$) and the corresponding shift.
The shift estimator $\phi$ could be trained under the supervision of the simulated sub-optimal representations $\tilde{\mathcal{Z}}_o$ and their corresponding shift information $\Delta$.
In addition, during training, we also incorporate \textcolor{blue}{approximated} optimal representations $\mathcal{Z}_o$ and their corresponding shift (i.e., 0) as the supervision to prevent excessive shifts in the output of well-optimized representations.
Since the potential shift within representations of real data samples may be diverse, the shift estimator $\phi$ has the challenge of inferring shift from a single type of simulated sub-optimal representations.
To solve it, we suggest to diversify the types of simulated sub-optimal representations by applying $M$ distinct perturbations to the observation $x_k$ to simulate $M$ different sub-optimal representations.
% Specifically, given the $k$-th sample $x_k$, we sample a feature mask vector $\textbf{m}^{km}$ and perturb $x_k$ by $\textbf{m}^{km} \odot x_k + (1-\textbf{m}^{km}) \odot \epsilon_k$, where $\textbf{m}^{ki} \neq \textbf{m}^{kj}$ if $i\neq j$.
Specifically, for $x_k$, we generate a feature mask vector $\textbf{m}_{km}$, where each entry of the vector is set to 1 with probability $\eta_k$ and to 0 otherwise. We then perturb $x_k$ by $\textbf{m}_{km} \odot x_k + (1-\textbf{m}_{km}) \odot \epsilon_k$, where $\textbf{m}_{ki} \neq \textbf{m}_{kj}$ if $i\neq j$.
In addition, for $x_k$, $\eta_k$ is sampled from a uniform distribution to diversify the mask vectors between different samples.
By introducing such varying types of shift signals to $\mathcal{Z}_o$, the shift estimator $\phi$ has the potential to acquire information about the form of the real representation shift. 
% Guided by this information during training, the estimator becomes capable of predicting the latent shift within the sub-optimal representations.

\textbf{\textcolor{blue}{Re-estimation of Sub-optimal Representations.}}
Given the learned shift estimator $\phi$, \textcolor{blue}{we could remove the latent shift $\phi(z_i;\theta_\phi)$ present in sub-optimal representation $z_i = G_f(x_i;\theta_f)$ for any real data sample $x_i$ to achieve representation re-estimation:}
\textcolor{blue}{
\begin{equation}
\label{equ:re-estimation}
\begin{aligned}
    \Phi(z_i;\theta_\phi) = z_i - \phi(z_i;\theta_\phi),
\end{aligned}
\end{equation}
}
% \begin{equation}
% \label{equ:re-estimation}
% \begin{aligned}
%     &\Phi(z_i;\theta_\phi) = z_i - \phi(z_i;\theta_\phi)\\ = G&_f(x_i;\theta_f) - \phi(G_f(x_i;\theta_f);\theta_\phi),
% \end{aligned}
% \end{equation}
where $\Phi(\cdot;\theta_\phi)$ is the re-estimation function.
Through the re-estimation function, we refine the representation space of the trained backbone $G_f(\cdot;\theta_f)$ by removing the latent shift without altering any of the backbone's parameters. Despite the shift removal achieved through this process, the problem of representation redundancy still exists. In the subsequent section, we will introduce another ingeniously designed task to address the issue of representation redundancy.

\subsection{Tabular Space Mapping} 
\label{sec:task2}
The experimental results depicted in Fig.~\ref{fig:motivation2} reveal that in the realm of deep tabular machine learning models, elevated SVE does not consistently correlate with superior performance.
This observation may be attributed to the presence of redundant information.
To address this issue, we introduce a technique designed to map representations into a newly defined light embedding space. 
This space is characterized by clearer and more concise information, aiming to alleviate the influence of redundant information on prediction accuracy while preserving essential information crucial for prediction tasks.
% Additionally, two strategies are applied to the light embedding space: (i) Space Compression, which makes the space information more concise, and (ii) Critical Knowledge Preserving, which contains the key information for prediction.
Additionally, two strategies are applied to regulate the light embedding space: (i) Space Compression, which compresses and transforms the complex knowledge from the original space into a compact space, and (ii) Critical Knowledge Preserving, which contains the key information for prediction.
Since the inherent representation shift information also has the probability to spread into light embedding space, we suggest to apply this task after the previous task.
First, we give the formal definition of light embedding space as follows:

% \textbf{Definition 2. Light Embedding Space (LE-Space).} Given a set of embedding vectors $\mathcal{B} = \{\beta_t\}_{t=1}^T \in \mathbb{R}^{T\times D}$ where $T$ is the size of this set and $\beta_t$ is the $t$-th embedding vector with $D$ as the dimensionality, we define the LE-Space as a space consisting of the embedding vectors $\mathcal{B}$ for representation.
% For example, given a representation in LE-Space with coordinates denoted as $r = \{r^t\}_{t=1}^{T} \in \mathbb{R}^{T\times 1}$, each representation in LE-Space can be formulated as $\sum_{t=1}^{T} r^t \beta_t$.
\textbf{Definition 2. Light Embedding Space (LE-Space).} Given a collection of embedding vectors $\mathcal{B} = \{\beta_t\}_{t=1}^T \in \mathbb{R}^{T\times D}$, where $T$ denotes the number of embeddings and $\beta_t$ represents the $t$-th embedding vector in $D$-dimensional space, we define the LE-Space as the vector space spanned by the set $\mathcal{B}$. 
Formally, any representation in the LE-Space can be expressed as a weighted linear combination of the embedding vectors, given by:
\begin{equation}
\label{equ:le-definition}
\begin{aligned}
    r\mathcal{B} = \sum_{t=1}^{T} r^t \beta_t,
\end{aligned}
\end{equation}
where $r = \{r^t\}_{t=1}^{T} \in \mathbb{R}^{T}$ denotes the coordinates. This formulation ensures that each representation in the LE-Space captures the essential characteristics encoded by the embedding vectors $\mathcal{B}$.

\textbf{Space Compression.}
Since the trained backbone $G_f(\cdot;\theta_f)$ can inherently capture the interactions among features, and after applying the previous task, the original sub-optimal representation $z_i$ have been re-estimated as $\Phi(z_i;\theta_\phi)$ (Eq.~\ref{equ:re-estimation}) to mitigate shift information, it is reasonable to predict $r_i$ for each sample based on the re-estimated representation $\Phi(z_i;\theta_\phi)$.
To obtain the target representation $r_i\mathcal{B}$ within LE-Space, we use a shared coordinate estimator $s(\cdot;\theta_s)$ with learnable $\theta_s$ to calculate the corresponding $r_i$ by:
\begin{equation}
\label{equ:coordinate-estimate}
\begin{aligned}
    r_i = s(\Phi(z_i;\theta_\phi);\theta_s).
\end{aligned}
\end{equation}
The embeddings are random initialized and learnable.
By controlling the size $T$ of the LE-Space to be significantly smaller than the dimensionality $D$ of representation $\Phi(z_i;\theta_\phi)$, we can compress and transfer the information from the original space into the LE-Space.

\textbf{Critical Knowledge Preserving.}
Directly mapping the representations could potentially result in a collapse of LE-Space learning, where part of the embedding vectors in the LE-Space become similar, thus diminishing the space's representational capacity.
To diversify these embedding vectors, we define the orthogonality loss to encourage the cosine similarity of any pairs of embedding vectors to be close to 0:
% To maintain orthogonality for these embedding vectors, we define the orthogonality loss to encourage the cosine similarity of any pairs of embedding vectors to be close to 0:

\begin{equation}
\label{regulariz_loss}
\begin{aligned}
    \min \mathcal{L}_{orth} = \min_{\mathcal{B}} \left(\frac{\|A\|_{1}}{\|A\|_{2}^{2}} + (\|A\|_{1}-T)^2\right),
\end{aligned}
\end{equation}
where $A \in [0, 1]^{T\times T}$, $A_{ij} = \|\cos(\beta_i, \beta_j)\|_1$, and $T$ is the number of embeddings. 
The first term $\frac{\|A\|_{1}}{\|A\|_{2}^{2}}$ encourages sparsity in $A$, meaning that each element $A_{ij} = \|\cos(\beta_i, \beta_j)\|_1$ should be close to 0 (indicating orthogonality between $\beta_i$ and $\beta_j$) or 1. 
The second term promotes $\|A\|_{1} \to T$, given that $A_{ii} = \|\cos(\beta_i, \beta_i)\|_1 = 1$ for all $i \in \{1,2,...,T\}$, implying that the off-diagonal elements in the $A$ to be close to 0.
Furthermore, to maintain the key information for accurate prediction, we transfer the label information into the explicitly defined embedding vectors of LE-Space by:
\begin{equation}
\label{task specific loss}
\begin{aligned}
    \min\mathcal{L}_{pred} = \min_{\theta_s, \theta_\phi, \mathcal{B}, \theta_h} \frac{1}{N}\sum_{i=1}^N \mathcal{L}(G_h(r_i\mathcal{B};\theta_h), y_i),
\end{aligned}
\end{equation}  
where $r_i = s(\Phi(z_i;\theta_\phi);\theta_s)$ (Eq.~\ref{equ:coordinate-estimate}), $G_h(\cdot;\theta_h)$ is the re-initialized projection head. 
In the new space, the information is compressed compared to the original space as the size $T$ of embedding vectors is significantly smaller than the dimensionality $D$ of representation $\Phi(z_i;\theta_\phi)$, while essential information crucial for prediction has been preserved. As a result, the new space contains less redundant information when contrasted with the original space. 
\textcolor{blue}{The detailed discussion is provided in Appendix J.}
% \textcolor{blue}{The detailed discussion is provided in Appendix~\ref{appendix:theoretical analysis and discussion}.}

\begin{algorithm}
% \small
\caption{\method algorithm workflow.}
\label{algorithm}
\begin{algorithmic}[1]
\Require dataset $\mathcal{D}_{train}$, $\mathcal{D}_{val}$, trained $F(\cdot;\theta) = G_h\left(G_f(\cdot;\theta_f); \theta_h\right)$, shift estimator $\phi(\cdot;\theta_\phi)$, re-estimation function $\Phi(\cdot;\theta_\phi)$, coordinate estimator $s(\cdot;\theta_s)$, embedding vectors $\mathcal{B}$, number of batches $B$;
\State Compute $\|\nabla_{\theta_f}\mathcal{L}(F(x;\theta), y)\|_p^q$ over $\mathcal{D}_{val}$ and identify $\mathcal{Z}_o$ based on \textbf{Assumption 1};
\State Generate simulated sub-optimal representations $\tilde{\mathcal{Z}}_o$ and the corresponding shift information $\Delta$;
\State Re-initialize the parameters of $G_h(\cdot;\theta_h)$;
\While{$\theta_\phi, \mathcal{B}, \theta_s$ and $\theta_h$ has not converged}
    \State Compute $\mathcal{L}_{shift}$ (Eq.~\ref{Shift Learning});
    \State Update $\theta_\phi$ by minimizing $\mathcal{L}_{shift}$ through gradient descent;
    \For{$i \gets 1$ to $B$}
        \State Sample minibatch;
        \State Obtain representation $z = G_f(x;\theta_f)$;
        \State Re-estimate representation by $\Phi(z;\theta_\phi)$;
        \State Calculate $r$ by $s(\Phi(z;\theta_\phi);\theta_s)$;
        \State Transform the representation of $\Phi(z;\theta_\phi)$ to $r\mathcal{B}$;
        \State Compute $\mathcal{L}_{orth}$ (Eq.~\ref{regulariz_loss}) and $\mathcal{L}_{pred}$ (Eq.~\ref{task specific loss});
        \State Update $\theta_\phi, \mathcal{B}, \theta_s$ and $\theta_h$ by minimizing $\mathcal{L}_{orth} + \mathcal{L}_{pred}$ through gradient descent.
    \EndFor
\EndWhile
\Ensure $\phi(\cdot;\theta_\phi), s(\cdot;\theta_s), \mathcal{B}, G_h(\cdot;\theta_h)$.
\end{algorithmic}
\end{algorithm}

\subsection{Framework Overview}
Altogether, \method requires learning the parameters of $\theta_\phi, \mathcal{B}, \theta_s$ and $\theta_h$.
During each training iteration, \method first aims to minimize $\mathcal{L}_{shift}$ to update $\theta_\phi$. 
Subsequently, \method aims to minimize the combined objective of $\mathcal{L}_{orth} + \mathcal{L}_{pred}$ to update $\theta_\phi, \mathcal{B}, \theta_s$, and $\theta_h$. 
These two optimization processes, which correspond to Tabular Representation Re-estimation and Tabular Space Mapping respectively, are iteratively alternated to refine the representations of deep tabular model without modifying the backbone's parameters $\theta_f$ in a model-agnostic manner.
We provide the training process of \method in Algorithm~\ref{algorithm}.
\textcolor{blue}{For clarity, we also summarize the key notations used throughout the paper in Table VIII of Appendix A. This includes the symbols, their definitions, and dimensionalities where applicable.}
% \textcolor{blue}{For clarity, we also summarize the key notations used throughout the paper in Table~\ref{appendix:notation} of Appendix~\ref{appendix:NOMENCLATURE}. This includes the symbols, their definitions, and dimensionalities where applicable.}

Specifically, \method first identifies a set of samples $\mathcal{D}_o$ with lowest gradient norms and obtains the corresponding representations $\mathcal{Z}_o$ as the approximated optimal representations.
We then manually generate the simulated sub-optimal representations $\tilde{\mathcal{Z}}_o$ and the corresponding shift information $\Delta$ based on $\mathcal{Z}_o$.
$\tilde{\mathcal{Z}}_o$ and $\Delta$ are used to supervise the training process of the shift estimator $\phi(\cdot;\theta_\phi)$ via $\mathcal{L}_{shift}$ (Eq.~\ref{Shift Learning}).
For a real data sample $x_i$, \method removes the latent shift $\phi(z_i;\theta_\phi)$ on representation $z_i$ to re-estimate it by $\Phi(z_i;\theta_\phi) = z_i - \phi(z_i;\theta_\phi)$ (Eq.~\ref{equ:re-estimation}), thereby alleviating the issue of representation shift.
After that, the coordinate estimator $s(\cdot;\theta_s)$ is used to calculate the coordinate $r_i$ (Eq.~\ref{equ:coordinate-estimate}) to transform $\Phi(z_i;\theta_\phi)$ to $r_i\mathcal{B}$ within LE-Space to compress the information.
In addition, \method diversifies the embedding vectors $\mathcal{B}$ via $\mathcal{L}_{orth}$ (Eq.~\ref{regulariz_loss}) and optimizes the $\mathcal{L}_{pred}$ (Eq.~\ref{task specific loss}) to preserve the critical knowledge for prediction.
Therefore, \method alleviates the issue of representation redundancy.

During inference, we feed the representations extracted by the existing trained deep tabular backbone, which are from the test dataset (notably, we do not add any noise to the test dataset), into the shift estimator followed by the coordinate estimator to achieve the calibrated representations of test samples. 
Note that \method, as a general representation learning framework, is model-agnostic such that it can be coupled with any trained deep tabular backbone $G_f(\cdot;\theta_f)$ to learn better representations.

% Subsequently, the representations are re-estimated using $\Phi(G_f(\cdot;\theta_f);\theta_\phi) = G_f(\cdot;\theta_f) - \phi(G_f(\cdot;\theta_f);\theta_\phi)$. Then, the re-estimated representations by $\Phi(G_f(\cdot;\theta_f);\theta_\phi)$ are transformed into the LE-Space. \method then aims to minimize $\mathcal{L}_{orthogonalization} + \mathcal{L}_{pred}$ \textcolor{red}{directly use equation} with respect to $\theta_\phi, \mathcal{B}, \theta_s$, and $\theta_h$. 
% The above two optimization processes are iteratively alternated.
% The proposed two tasks jointly refine the deep tabular model without modifying the backbone's parameters $\theta_f$.
% Altogether, \method aims to minimize the following objective function w.r.t. $\theta_\phi, \mathcal{B}, \theta_s$ and $\theta_h$ :
% \begin{equation}
% \begin{aligned}
%     \mathcal{O} = \mathcal{L}_{shift} + \mathcal{L}_{orthogonalization} + \mathcal{L}_{pred}.
% \end{aligned}
% \end{equation}
% We provide the algorithm of \method and the full implementation details in Appendix~\ref{appendix:TPL details}.

\section{Experiment \& Analysis}
\label{sec: experiment}
\subsection{Experimental Setup}

\textbf{Datasets.} 
We consider a variety of tabular datasets, including regression, binary and multi-class classification.
% In this paper, we consider a variety of tabular datasets with heterogeneous features, including regression, binary classification and multi-class classification.
Specifically, the datasets include: Combined Cycle Power Plant (CO)~\cite{Chary2021PredictionOF}, Diamonds (DI)~\cite{chen2023trompt}, Qsar Fish Toxicity (QS)~\cite{Cassotti2015ASQ}, California Housing  (CA)~\cite{pace1997sparse}, Pol (PO)~\cite{Grinsztajn2022WhyDT}, Superconductivty Data (SU)~\cite{Hamidieh2018ADS}, Adult (AD) ~\cite{kohavi1996scaling}, Australian (AU)~\cite{schafl2023hopular}, Gesture Phase (GE)~\cite{madeo2013gesture}, \textcolor{blue}{Year (YE)~\cite{Bertin-Mahieux2011} and Covertype (COV)~\cite{blackard1999comparative}}.
% We split each dataset into training, validation, and test sets by the ratio of 6:2:2. 
% The dataset properties and pre-processing information are provided in Appendix~\ref{appendix:datasets details}.
The dataset properties are summarized in Table~\ref{table:data properties}.
To handle categorical features, we adopt an integer encoding scheme, where each category within a column is uniquely mapped to an integer to index the embedding in the lookup table. We maintain consistent embedding dimensions for all categorical features. For numerical features, we apply the column-wise normalization method.
In regression tasks, we also apply the normalization to the labels. 
To ensure fair comparisons, we adhere to identical preprocessing procedures and use the quantile transformation following ~\cite{gorishniy2021revisiting} for each dataset.
We split each dataset into training, validation, and test sets by the ratio of 6:2:2.
The used datasets are collected in the public OpenML-AutoML Benchmark (\url{https://openml.github.io/automlbenchmark/}) and UCI datasets (\url{https://archive.ics.uci.edu/datasets}).
Following previous studies~\cite{gorishniy2021revisiting}, we use Root Mean Squared Error (RMSE) (lower is better) to evaluate the regression tasks, Accuracy (higher is better) to evaluate binary and multiclass classification tasks. 

\begin{table}[!t]
    \tiny
    \centering
	\caption{Tabular data properties. ``Objects'' indicates the dataset size. RMSE denotes Root Mean Square Error for regression, Accuracy is used for binary and multiclass classification. 
 % The used datasets are collected in the public OpenML-AutoML Benchmark \url{} and UCI datasets \url{https://archive.ics.uci.edu/datasets}.
 }
    \label{table:data properties}
    \setlength{\tabcolsep}{0.8mm}{
    \begin{tabular}{cccccccccccc}
    \toprule
      & CO & DI & QS & CA & PO & SU & AD & AU & GE & \textcolor{blue}{YE} & \textcolor{blue}{COV}\\
    \midrule
    Objects & 9568 & 53940 & 908 & 20640 & 15000 & 21263 & 48842 & 690 & 9873 & \textcolor{blue}{515345} & \textcolor{blue}{581012}\\
    Numerical & 4 & 6 & 6 & 8 & 48 & 81 & 6 & 14 & 32 & \textcolor{blue}{90} & \textcolor{blue}{54} \\
    Categorical & 0 & 3 & 0 & 0 & 0 & 0 & 8 & 0 & 0 & \textcolor{blue}{0} & \textcolor{blue}{0} \\
    Classes & - & - & - & - & - & - & 2 & 2 & 5 & \textcolor{blue}{-} & \textcolor{blue}{7} \\
    Metric & RMSE & RMSE & RMSE & RMSE & RMSE & RMSE & Accuracy & Accuracy & Accuracy & \textcolor{blue}{RMSE} & \textcolor{blue}{Accuracy} \\
    \bottomrule
    \end{tabular}}
\end{table}

\begin{table*}[!t]
    \caption{Results of overall performance. ``i'' denotes in-learning and ``p'' denotes pre-learning for baseline models. 
    % Lower evaluation metric is better for regression, and higher evaluation metric is better for classification. 
    The best results are highlighted in bold.
    The improvement of \method over baselines is statistically significant at the 95\% confidence level.}
    \label{tab:Overall Performance}
    \scriptsize
    % \footnotesize
    \centering
    % \resizebox{\linewidth}{!}{
    \setlength{\tabcolsep}{3.9mm}{
    \begin{tabular}{cccccccccc}
    \toprule
      & CO $\downarrow$ & DI $\downarrow$ & QS $\downarrow$ & CA $\downarrow$ & PO $\downarrow$ & SU $\downarrow$ & AD $\uparrow$ & AU $\uparrow$ & GE $\uparrow$ \\
    \midrule
    MLP (i) & 3.961 & 564.373 & 0.875 & 0.505 & 10.892 & 10.379 & 0.856 & 0.870 & 0.578 \\
    +\textbf{TRC} & \bfseries 3.899 & \bfseries 558.330 & \bfseries 0.825 & \bfseries 0.502 & \bfseries 10.593 & \bfseries 10.326 & \bfseries 0.858 & \bfseries 0.891 & \bfseries 0.580 \\
    \midrule
    DCN2 (i) & 4.016 & 591.958 & 1.027 & 0.495 & 8.805 & 10.674 & 0.856 & 0.841 & 0.564 \\
    +\textbf{TRC} & \bfseries 3.858 & \bfseries 574.453 & \bfseries 0.879 & \bfseries 0.493 & \bfseries 8.790 & \bfseries 10.584 & \bfseries 0.857 & \bfseries 0.862 & \bfseries 0.584 \\
    \midrule
    SNN (i) & 11.789 & 1530.293 & 1.018 & 0.896 & 18.517 & 25.498 & 0.847 & 0.848 & 0.550 \\
    +\textbf{TRC} & \bfseries 6.880 & \bfseries 693.369 & \bfseries 0.870 & \bfseries 0.699 & \bfseries 11.202 & \bfseries 15.537 & \bfseries 0.847 & \bfseries 0.877 & \bfseries 0.573 \\
    \midrule
    ResNet (i) & 3.982 & \bfseries 606.282 & 0.872 & 0.517 & 10.812 & 11.163 & 0.847 & 0.870 & 0.587 \\
    +\textbf{TRC} & \bfseries 3.919 & 615.736 & \bfseries 0.845 & \bfseries 0.508 & \bfseries 10.205 & \bfseries 10.720 & \bfseries 0.849 & \bfseries 0.877 & \bfseries 0.597 \\
    \midrule
    AutoInt (i) & 4.020 & 562.169 & 0.860 & 0.487 & 6.450 & 11.193 & 0.854 & 0.848 & 0.598 \\
    +\textbf{TRC} & \bfseries 3.960 & \bfseries 557.423 & \bfseries 0.829 & \bfseries 0.477 & \bfseries 5.885 & \bfseries 10.826 & \bfseries 0.858 & \bfseries 0.891 & \bfseries 0.606 \\
    \midrule
    TANGOS (i) & 4.687 & 965.169 & 0.945 & 0.573 & 9.401 & 10.849 & \bfseries0.84 & 0.855 & 0.569 \\
    +\textbf{TRC} & \bfseries4.478 & \bfseries886.904 & \bfseries0.914 & \bfseries0.538 & \bfseries9.271 & \bfseries10.520 & 0.839 & \bfseries0.882 & \bfseries0.590 \\
    \midrule
     % \textcolor{red}{TANGOS(i)}&\textcolor{red}{4.529}&\textcolor{red}{965.169}&\textcolor{red}{0.945}&\textcolor{red}{0.550}&\textcolor{red}{9.401}&\textcolor{red}{\bfseries10.497}&\textcolor{red}{0.839}&\textcolor{red}{0.841}&\textcolor{red}{0.582}\\
    \textcolor{blue}{FT-Transformer} (i) & 3.709 & 551.190 & 0.823 & 0.469 & 2.919 & 10.410 & 0.858 & 0.848 & 0.611 \\
    +\textbf{TRC} & \bfseries 3.648 & \bfseries 543.653 & \bfseries 0.780 & \bfseries 0.462 & \bfseries 2.728 & \bfseries 10.223 & \bfseries 0.862 & \bfseries 0.884 & \bfseries 0.624 \\
    \midrule
    PTaRL (i) & 3.668 & 550.582 & 0.813 & 0.447 & 2.533 & 10.199 & 0.872 & 0.879 & 0.583 \\
    +\textbf{TRC} & \bfseries 3.423 & \bfseries 543.671 & \bfseries 0.775 & \bfseries 0.427 & \bfseries 2.412 & \bfseries 9.724 & \bfseries 0.889 & \bfseries 0.890 & \bfseries 0.619 \\
    \midrule
    SCARF (p) & 3.856 & 579.610 & 0.863 & \bfseries 0.520 & 8.310 & \bfseries 10.324 & 0.858 & 0.862 & \bfseries 0.589 \\
    +\textbf{TRC} & \bfseries 3.847 & \bfseries 577.803 & \bfseries 0.808 & \bfseries 0.520 & \bfseries 8.151 & 10.355 & \bfseries 0.859 & \bfseries 0.884 & \bfseries 0.589 \\
    \midrule
    SAINT (p) & 4.022 & 597.207 & 0.827 & 0.508 & 4.415 & 13.095 & 0.857 & 0.870 & 0.549 \\
    +\textbf{TRC} & \bfseries 3.903 & \bfseries 557.951 & \bfseries 0.818 & \bfseries 0.491 & \bfseries 4.335 & \bfseries 12.443 & \bfseries 0.858 & \bfseries 0.884 & \bfseries 0.595 \\
    \midrule
    VIME (p) & 5.218 & 945.238 & 1.018 & 0.679 & 10.914 & 15.645 & 0.768 & 0.812 & 0.473 \\
    +\textbf{TRC} & \bfseries 4.371 & \bfseries 612.454 & \bfseries 0.950 & \bfseries 0.645 & \bfseries 7.489 & \bfseries 15.028 & \bfseries 0.846 & \bfseries 0.884 & \bfseries 0.479 \\
    \midrule
    MLP w/ SSL-Rec (p) & 3.92 & 595.378 & 0.872 & \textbf{0.498} & 10.052 & 10.408 & 0.855 & 0.862 & 0.578 \\
    +\textbf{TRC} & \bfseries3.859 & \bfseries581.511 & \bfseries0.825 & 0.499 & \bfseries9.796 & \bfseries10.335 & \bfseries0.856 & \bfseries0.884 & \bfseries0.595 \\
    \midrule
    MLP w/ SSL-Contrastive (p) & 3.956 & 570.54 & 0.872 & 0.506 & 13.305 & 10.478 & 0.855 & 0.862 & 0.576 \\
    +\textbf{TRC} & \bfseries3.923 & \bfseries562.671 & \bfseries0.825 & \bfseries0.502 & \bfseries12.803 & \bfseries10.436 & \bfseries0.856 & \bfseries0.884 & \bfseries0.587 \\
    \bottomrule
    \end{tabular}}
    % \vspace{1em}
\end{table*}

\begin{figure*}[!t]
    \centering
    \captionsetup[sub]{skip=0pt}
    \begin{subfigure}[b]{0.48\linewidth}
        \centering
        \includegraphics[width=\linewidth]{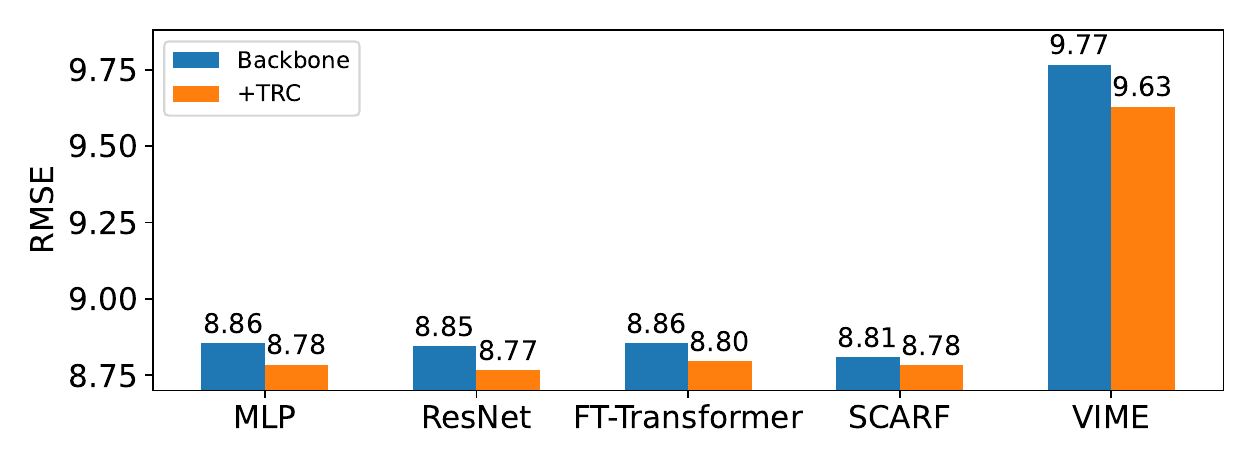}
        \caption{YE dataset $\downarrow$}
    \end{subfigure}
    \begin{subfigure}[b]{0.48\linewidth}
        \centering
        \includegraphics[width=\linewidth]{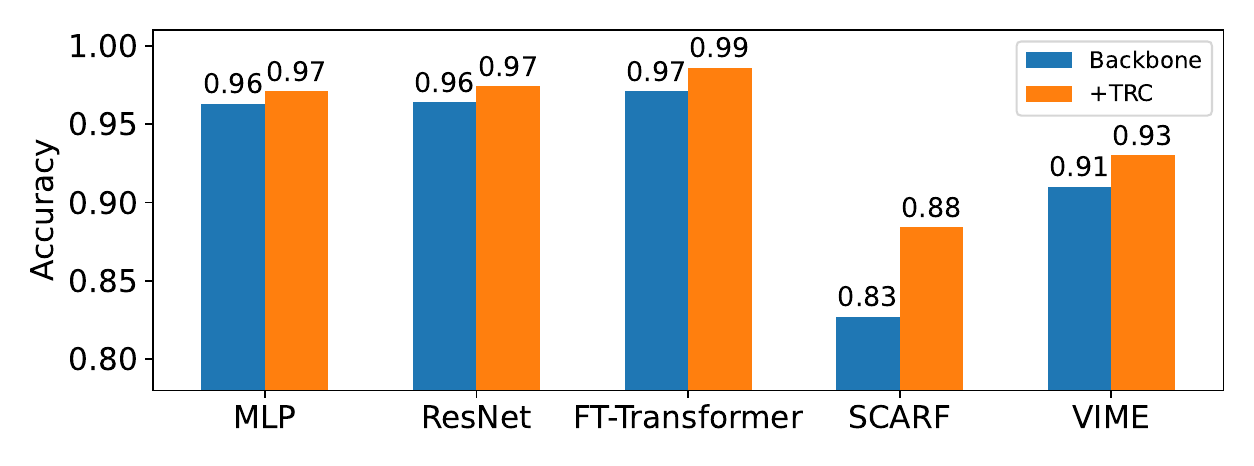}
        \caption{COV dataset $\uparrow$}
    \end{subfigure}
    \caption{\textcolor{blue}{Performance improvement of TRC over various backbone models on large scale datasets.}}
    \label{fig:scalibility on large datasets}
\end{figure*}

\textbf{Baseline Deep Tabular Models.}
As the \method is a model-agnostic paradigm, we include 13 mainstream deep tabular models, which include in-learning and pre-learning paradigms, to test \method's applicability and effectiveness to different predictors with diverse architectures. The baseline models include: MLP~\cite{taud2018multilayer}, DCN2~\cite{wang2021dcn}, SNN~\cite{klambauer2017self}, ResNet~\cite{he2016deep}, AutoInt~\cite{song2019autoint}, TANGOS~\cite{jeffares2022tangos}, FT-Transformer~\cite{gorishniy2021revisiting}, PTaRL~\cite{ye2024ptarl}, SCARF~\cite{bahriscarf}, SAINT~\cite{somepalli2022saint} and VIME~\cite{yoon2020vime}.
In addition, the authors of~\cite{rubachev2022revisiting} indicate that pre-training MLP with several self-supervised learning objectives could achieve promising results. Therefore, we also include two of them as additional baseline methods.
We implement all deep tabular baseline models' hyperparameters following their original papers.
The detailed descriptions about these models are as follows:

\textbf{In-learning Style.}
\begin{itemize}[leftmargin=*]
\item 
MLP~\cite{taud2018multilayer}. The Multilayer Perceptron (MLP) is a type of feedforward artificial neural network that consists of multiple layers of interconnected nodes, commonly used as a baseline model for supervised learning tasks in tabular domain.
\item 
DCN2~\cite{wang2021dcn}. This method consists of an MLP-like module and the feature crossing
module (a combination of linear layers and multiplications).
\item 
SNN~\cite{klambauer2017self}. An MLP-like architecture with the SELU activation that maintains zero mean and unit variance,
enabling training deeper models.
\item 
ResNet~\cite{he2016deep}. The key innovation is the use of residual connections, also known as skip connections or shortcut connections. These connections enable the network to effectively train deep neural networks, which was challenging before due to the vanishing gradient problem. 
In this paper, we use the ResNet for tabular data~\cite{gorishniy2021revisiting}.
\item 
AutoInt~\cite{song2019autoint}. This method transforms features to embeddings and applies a series of
attention-based transformations to the embeddings based on Transformer~\cite{vaswani2017attention}. 
\item 
TANGOS~\cite{jeffares2022tangos}. This method proposes a new regularization technique for tabular data, encouraging neurons to focus on sparse, non-overlapping input features. We adopt MLP as the base model for TANGOS.
\item 
FT-Transformer~\cite{gorishniy2021revisiting}. This method is introduced by \textit{Gorishniy et al.}~\cite{gorishniy2021revisiting} to further improve AutoInt through better token embeddings.
\item PTaRL~\cite{ye2024ptarl}. This method constructs a new projection space and uses Optimal Transport~\cite{peyre2017computational} to project data samples into this space, enabling the learning of disentangled representations with constraints designed for tabular data. We adopt FT-Transformer as the base model for PTaRL.
\end{itemize}

\textbf{Pre-learning Style.}
\begin{itemize}[leftmargin=*] 
\item
SCARF~\cite{bahriscarf}. This method extends the SimCLR framework to tabular data by using contrastive learning.
\item 
SAINT~\cite{somepalli2022saint}. This method adopts attention over both rows and columns, and it includes an enhanced embedding method. Contrastive learning is used during pretraining.
\item 
VIME~\cite{yoon2020vime}. This method utilizes self- and semi-supervised learning by recovering the corrupted inputs for tabular data.
\item
MLP w/ SSL-Rec~\cite{rubachev2022revisiting}. This method aims to reconstruct the original input, given the corrupted input (the reconstruction loss is computed for all columns).
\item 
MLP w/ SSL-Contrastive~\cite{rubachev2022revisiting}. This method uses InfoNCE loss, considering corrupted inputs as positives for original inputs and the rest of the batch as negatives.
\end{itemize}

\textbf{Implementation Details.}
Our \method is a model-agnostic method that aims to enhance the representations of any deep tabular model $F(\cdot;\theta)$ without altering its internal architecture. 
To achieve it, \method includes two tasks, i.e., Tabular Representation Re-estimation and Tabular Space Mapping.
In the first task, for the $q$ square of the $L_p$ norm, we set $q=1$ and $p=1$.
The threshold $\tau$ that defines the representations $\mathcal{Z}_o$ is set to 0.01.
The shift estimator $\phi$, which is used to calculate the potential shifts present in the representations, is a simple 2-layer fully-connected MLP.
We propose generating $\epsilon_k$ by sampling each feature element from the empirical distribution of the corresponding column in $\mathcal{D}_{train}$, following previous works~\cite{yoon2020vime, ucar2021subtab, nam2023stunt}.
For mask vector $\textbf{m}_{km}$, each entry of the vector is set to 1 with probability $\eta_k$ and to 0 otherwise, where $\eta_k$ is sampled from a uniform distribution $[0.1, 0.3]$. 
The perturbing times $M$ for constructing the simulated sub-optimal representations is set to 3, unless otherwise specified.
For the second task, the coordinate estimator $s$, which is used to calculate the coefficients for embedding vectors, is a simple linear layer with softmax activation function.
The default number of embeddings $T$ is set to 10, unless otherwise specified.
The batch size is set to 128 and AdamW is used as the optimizer, with a learning rate of 1e-4, weight decay of 1e-5.
A patience of 10 is kept for early stopping.
We implement all deep tabular backbone models' hyperparameters following their original papers.
% Following the common practice of previous studies, we use Root Mean Squared Error (RMSE) (lower is better) to evaluate the regression tasks, Accuracy (higher is better) to evaluate binary and multiclass classification tasks. 
All experiments were conducted on the Ubuntu 20.04.4 LTS operating system, Intel(R) Xeon(R) Gold 5220 CPU @ 2.20GHz with single NVIDIA A40 48GB GPU and 512GB of
RAM. The framework is implemented with Python 3.11.5 and PyTorch 2.0.2.
To reduce the effect of randomness, the reported performance is averaged over 10 independent runs.

\begin{table*}[t]
    \caption{\textcolor{blue}{Analysis on the effects of different components of \method.} The best results are highlighted in bold, and the second best results are underscored.}
    \label{tab:ablation}
    \footnotesize
    \centering
    % \resizebox{\linewidth}{!}{
    \setlength{\tabcolsep}{3.1mm}{
    \begin{tabular}{ccccccccccccc}
    \toprule
    Backbone & TR & SC & DE & CO $\downarrow$ & DI $\downarrow$ & QS $\downarrow$ & CA $\downarrow$ & PO $\downarrow$ & SU $\downarrow$ & AD $\uparrow$ & AU $\uparrow$ & GE $\uparrow$ \\
    \midrule
    \multirow{6}{*}{MLP} &   &   &   & 3.961 & 564.373 & 0.875 & 0.505 & 10.892 & 10.379 & 0.856 & 0.87 & \underline{0.578} \\
     & \checkmark &   &   & 3.916 & 563.649 & 0.868 & \textbf{0.502} & 10.605 & \underline{10.328} & \textbf{0.858} & 0.87 & 0.571 \\
     &   & \checkmark &   & 3.92 & 564.141 & 0.845 & 0.505 & 10.631 & 10.349 & \textbf{0.858} & 0.862 & 0.568 \\
     & \checkmark & \checkmark &   & \underline{3.908} & \underline{558.52} & 0.852 & \textbf{0.502} & \underline{10.602} & 10.341 & \textbf{0.858} & 0.87 & 0.575 \\
     &   & \checkmark & \checkmark & 3.922 & 565.129 & \underline{0.841} & 0.505 & 10.622 & 10.363 & \textbf{0.858} & \underline{0.877} & 0.577 \\
     & \checkmark & \checkmark & \checkmark & \textbf{3.899} & \textbf{558.33} & \textbf{0.825} & \textbf{0.502} & \textbf{10.593} & \textbf{10.326} & \textbf{0.858} & \textbf{0.891} & \textbf{0.58} \\
    \bottomrule
    \end{tabular}}
    % \vspace{0.5em}
\end{table*}

\begin{table*}[t]
    \footnotesize
    \centering
    \caption{Performance of backbone w/ TRC and deeper backbone. The depths of backbone layers are increased only for ``deeper backbone''.}
    \label{tab:deeper comparison}
    \setlength{\tabcolsep}{4.1mm}{
    \begin{tabular}{cccccccccc}
    \toprule
 & CO $\downarrow$ & DI $\downarrow$ & QS $\downarrow$ & CA $\downarrow$ & PO $\downarrow$ & SU $\downarrow$ & AD $\uparrow$ & AU $\uparrow$ & GE $\uparrow$ \\
\midrule
MLP & 3.961 & 564.373 & 0.875 & 0.505 & 10.892 & 10.379 & 0.856 & 0.870 & 0.578 \\
deeper MLP & 3.942 & 562.888 & 0.871 & 0.510 & 11.583 & 10.471 & 0.856 & 0.862 & 0.572 \\
MLP+\textbf{TRC} & \textbf{3.899} & \textbf{558.330} & \textbf{0.825} & \textbf{0.502} & \textbf{10.593} & \textbf{10.326} & \textbf{0.858} & \textbf{0.891} & \textbf{0.580} \\
\midrule
% DCN2 & 4.016 & 591.958 & 1.027 & 0.495 & 8.805 & 10.674 & 0.856 & 0.841 & 0.564 \\
% deeper DCN2 & 3.972 & 595.765 & 0.984 & 0.496 & 9.165 & 10.675 & 0.853 & 0.842 & 0.575 \\
% DCN2+TRC & \textbf{3.858} & \textbf{574.453} & \textbf{0.879} & \textbf{0.493} & \textbf{8.790} & \textbf{10.584} & \textbf{0.857} & \textbf{0.862} & \textbf{0.584} \\
% \midrule
% SNN & 11.789 & 1530.293 & 1.018 & 0.896 & 18.517 & 25.498 & 0.847 & 0.848 & 0.550 \\
% deeper SNN & 10.614 & 1042.509 & 1.121 & 0.851 & 17.460 & 25.039 & 0.847 & 0.833 & 0.553 \\
% SNN+TRC & \textbf{6.880} & \textbf{693.369} & \textbf{0.870} & \textbf{0.699} & \textbf{11.202} & \textbf{15.537} & \textbf{0.847} & \textbf{0.877} & \textbf{0.573} \\
% \midrule
% ResNet & 3.982 & \textbf{606.282} & 0.872 & 0.517 & 10.812 & 11.163 & 0.847 & 0.870 & 0.587 \\
% deeper ResNet & 3.973 & 657.823 & 0.877 & 0.532 & 10.728 & 12.008 & 0.842 & 0.862 & 0.582 \\
% ResNet+TRC & \textbf{3.919} & 615.736 & \textbf{0.845} & \textbf{0.508} & \textbf{10.205} & \textbf{10.720} & \textbf{0.849} & \textbf{0.877} & \textbf{0.597} \\
% \midrule
% AutoInt & 4.020 & 562.169 & 0.860 & 0.487 & 6.450 & 11.193 & 0.854 & 0.848 & 0.598 \\
% deeper AutoInt & 4.099 & 562.383 & 0.875 & 0.488 & 6.201 & 11.265 & 0.853 & 0.862 & 0.577 \\
% AutoInt+TRC & \textbf{3.960} & \textbf{557.423} & \textbf{0.829} & \textbf{0.477} & \textbf{5.885} & \textbf{10.826} & \textbf{0.858} & \textbf{0.891} & \textbf{0.606} \\
% \midrule
\textcolor{blue}{FT-Transformer} & 3.709 & 551.190 & 0.823 & 0.469 & 2.919 & 10.410 & 0.858 & 0.848 & 0.611 \\
deeper \textcolor{blue}{FT-Transformer} & 3.704 & 559.363 & 0.816 & 0.474 & 2.890 & 10.442 & 0.851 & 0.856 & 0.613 \\
\textcolor{blue}{FT-Transformer}+\textbf{TRC} & \textbf{3.648} & \textbf{543.653} & \textbf{0.780} & \textbf{0.462} & \textbf{2.728} & \textbf{10.223} & \textbf{0.862} & \textbf{0.884} & \textbf{0.624} \\
% \midrule
% SAINT & 4.022 & 597.207 & 0.827 & 0.508 & 4.415 & 13.095 & 0.857 & 0.870 & 0.549 \\
% deeper SAINT& 4.012 & 585.623 & 0.836 & 0.509 & 4.549 & 14.681 & 0.853 & 0.857 & 0.572 \\
% SAINT+TRC & \textbf{3.903} & \textbf{557.951} & \textbf{0.818} & \textbf{0.491} & \textbf{4.335} & \textbf{12.443} & \textbf{0.858} & \textbf{0.884} & \textbf{0.595} \\
% \midrule
% VIME & 5.218 & 945.238 & 1.018 & 0.679 & 10.914 & 15.645 & 0.768 & 0.812 & 0.473 \\
% deeper VIME & 5.348 & 910.491 & 0.999 & 0.671 & 9.984 & 15.332 & 0.743 & 0.827 & 0.436 \\
% VIME+TRC & \textbf{4.371} & \textbf{612.454} & \textbf{0.950} & \textbf{0.645} & \textbf{7.489} & \textbf{15.028} & \textbf{0.846} & \textbf{0.884} & \textbf{0.479} \\
\bottomrule
    \end{tabular}}
    % \label{tab:deeper comparison full}
\end{table*}

\subsection{\textcolor{blue}{Main Results}}

\textbf{Overall Performance.}
Table~\ref{tab:Overall Performance} has shown the performance comparison of 13 different backbones and their \method-enhanced versions over 9 real-world tabular datasets.
From the table, we can observe that \method achieves consistent improvements over the baseline deep models in all settings, including both regression and classification scenarios. 
On average, \method achieves a relative performance improvement of approximately 5.1\% across all scenarios.
Specifically, for regression, \method realizes a relative performance improvement of about 6.5\% across all datasets, while for classification, \method achieves a relative performance improvement of approximately 2.4\% across all datasets.
% For regression, \method achieves a relative performance improvement of approximately 6.5\% across all datasets.
% For classification, \method achieves a relative performance improvement of approximately 2.4\% across all datasets.
Even in cases where \method does not outperform the baseline, its performance remains comparable. Additionally, \method is able to enhance performance even over the best-performing baseline model for each dataset.
Moreover, we conduct Wilcoxon signed-rank test (with $\alpha=0.05$) ~\cite{woolson2007wilcoxon} to measure the improvement significance.
The results show that in all settings, the improvement of \method over deep tabular models is statistically significant at the 95\% confidence level (with p-value $=6.07\mathrm{e}-9$).
This demonstrates the superior adaptability and generalization ability of \method to different models and tasks.
\textcolor{blue}{We further include experiments to verify the scalability of TRC on large scale tabular datasets, as illustrated in Fig.~\ref{fig:scalibility on large datasets}.
The results show that TRC could still enhance the performance of backbone models on the large scale datasets.}

\begin{table}[t!]
\centering
\footnotesize
\centering
% \captionsetup{font=small}
\caption{\textcolor{blue}{Performance comparison between various fine-tuning paradigms and TRC. ``i'' denotes in-learning and ``p'' denotes pre-learning for
baseline models.}}
\label{table:comparison with fine-tuning}
\setlength{\tabcolsep}{3.0mm}{
    \begin{tabular}{ccccc}
    \toprule
                 & CO $\downarrow$ & DI $\downarrow$ & QS $\downarrow$ & AU $\uparrow$ \\ \midrule
    MLP (i)                & 3.961                         & 564.373          & 0.875                & 0.870           \\ 
+Linear Head            & 3.952                         & 563.945          & 0.870                & 0.870           \\
+Linear Head \& Backbone & 3.937                         & 563.304          & 0.880                & 0.862           \\
+MLP Head               & 3.950                         & 562.076          & 0.862                & 0.870           \\
+MLP Head \& Backbone    & 3.934                         & 562.427          & 0.857                & 0.870           \\
+LoRA                   & 3.941                         & 563.314          & 0.875                & 0.870           \\ 
+\textbf{TRC}           & \textbf{3.899}                & \textbf{558.330} & \textbf{0.825}       & \textbf{0.891}  \\ \midrule
DCN2 (i)               & 4.016                         & 591.958          & 1.027                & 0.841           \\ 
+Linear Head            & 4.173                         & 589.030          & 0.958                & 0.843           \\
+Linear Head \& Backbone & 4.009                         & 592.881          & 1.029                & 0.845           \\
+MLP Head               & 4.077                         & 586.749          & 0.906                & 0.847           \\
+MLP Head \& Backbone    & 4.088                         & 591.743          & 0.910                & 0.842           \\
+LoRA                   & 4.001                         & 591.903          & 1.021                & 0.849           \\ 
+\textbf{TRC}           & \textbf{3.858}                & \textbf{574.453} & \textbf{0.879}       & \textbf{0.862}  \\ \midrule
% AutoInt (i)            & 4.020                         & 562.169          & 0.860                & 0.848           \\
% Linear Head            & 4.158                         & 560.129          & 0.850                & 0.851           \\
% Linear Head + Backbone & 4.106                         & 562.994          & 0.867                & 0.871           \\
% MLP Head               & 4.090                         & 563.511          & 0.858                & 0.862           \\
% MLP Head + Backbone    & 4.112                         & 560.663          & 0.853                & 0.870           \\
% LoRA                   & 4.107                         & 567.822          & 0.860                & 0.848           \\
% \textbf{TRC}           & \textbf{3.960}                & \textbf{557.423} & \textbf{0.829}       & \textbf{0.891}  \\ \midrule
FT-Transformer (i)     & 3.709                         & 551.190          & 0.823                & 0.848           \\ 
+Linear Head            & 3.695                         & 552.416          & 0.828                & 0.828           \\
+Linear Head \& Backbone & 3.710                         & 554.259          & 0.816                & 0.870           \\
+MLP Head               & 3.698                         & 549.648          & 0.816                & 0.832           \\
+MLP Head \& Backbone    & 3.706                         & 556.086          & 0.802                & 0.870           \\
+LoRA                   & 3.683                         & 550.241          & 0.823                & 0.862           \\ 
+\textbf{TRC}           & \textbf{3.648}                & \textbf{543.653} & \textbf{0.780}       & \textbf{0.884}  \\ \midrule
SCARF (p)              & 3.856                         & 579.610          & 0.863                & 0.862           \\ 
+Linear Head            & 3.870                         & 578.389          & 0.842                & 0.867           \\
+Linear Head \& Backbone & 3.871                         & 579.757          & 0.856                & 0.870           \\
+MLP Head               & 3.864                         & 578.565          & 0.850                & 0.862           \\
+MLP Head \& Backbone    & 3.872                         & 604.720          & 0.851                & 0.862           \\
+LoRA                   & 3.874                         & 589.828          & 0.863                & 0.870           \\ 
+\textbf{TRC}           & \textbf{3.847}                & \textbf{577.803} & \textbf{0.808}       & \textbf{0.884}  \\ 
% SAINT (p)              & 4.022                         & 597.207          & 0.827                & 0.870           \\
% Linear Head            & 4.121                         & 594.298          & 0.829                & 0.877           \\
% Linear Head + Backbone & 3.992                         & 576.602          & 0.835                & 0.874           \\
% MLP Head               & 3.962                         & 590.147          & 0.829                & 0.870           \\
% MLP Head + Backbone    & 4.094                         & 613.463          & 0.844                & 0.874           \\
% LoRA                   & 4.012                         & 583.750          & 0.826                & 0.870           \\
% \textbf{TRC}           & \textbf{3.903}                & \textbf{557.951} & \textbf{0.818}       & \textbf{0.884}  \\
\bottomrule
\end{tabular}}
\end{table}

\textbf{Ablation Study and Comparison with Deeper Backbone Models.}
We further conduct ablation study to demonstrate the effectiveness of key components of \method.
Specifically, we denote the task of \underline{t}abular representation \underline{r}e-estimation as ``TR'', and for the task of tabular space mapping, we denote the process of \underline{s}pace \underline{c}ompression as ``SC'' and the strategy for \underline{d}iversifying the \underline{e}mbedding vectors as ``DE''.
Given that the strategy for diversifying the embedding vectors should be applied after the process of space compression, we conduct a comparative analysis between the backbone integrated with \method and its five variants.
The results presented in Table~\ref{tab:ablation} underscore that the joint effects of all of these components are crucial to achieve consistently good performance, thereby underscoring the effectiveness of \method.
More results are detailed in Appendix B.
% More results are detailed in Appendix~\ref{appendix:ablation study}.
Besides, \method is appended to the output of the backbone $G_f(\cdot;\theta_f)$ while keeping the parameters fixed. 
We compare the performance of the backbone coupled with \method against backbone with deeper layers (additional 3 layers).
As shown in Table~\ref{tab:deeper comparison}, increasing the number of layers in backbone model does not necessarily lead to improved performance; in some cases, it even results in performance degradation. Furthermore, the comparison between deeper backbone and backbone augmented with \method highlights that the well-designed tasks can significantly enhance model representations, which is not simply achieved by increasing model depth.
% More results are provided in Appendix B.
More results are provided in Appendix C.
% More results are provided in Appendix~\ref{appendix:deeper backbone}.

\textbf{\textcolor{blue}{Comparison with Fine-tuning Paradigm.}}
\textcolor{blue}{We also compare TRC with some fine-tuning methods. Specifically, given any trained backbone models, we compare TRC against five different fine-tuning strategies in Table~\ref{table:comparison with fine-tuning}: 
i) Linear Head: we simply attach a new linear head to a trained backbone and train only this new head; ii) Linear Head \& Backbone: we use the same architecture as strategy i, but we train both new head and the whole backbone; iii) MLP Head: we replace the linear head of strategy i with a two-layer MLP as the new head, and also train this new head only; iv) MLP Head \& Full Backbone:  we use the same architecture as strategy iii, but we train both new head and the whole backbone; v) LoRA (Low-Rank Adaptation): we use the parameter-efficient fine-tuning strategy LoRA (Low-Rank Adaptation), that injects trainable low-rank matrices into the whole frozen backbone parameters to adapt models with minimal additional parameters. In practice, we inject low-rank matrices into all linear layers of the backbone and freeze the original weights.
For all strategies, we set the learning rate to 1e-4, which matches the learning rate used for TRC.  
The results show that none of these fine-tuning strategies including LoRA achieve performance comparable to TRC.
The possible reason is that only training the new head or fine-tuning the backbone with conventional supervised training loss (already used by the trained backbone) can not provide additional useful information for representation. 
Different from fine-tuning methods, TRC is a representation correction method that tailor-designed for solving the inherent deep tabular backbone representations issues in a post-hoc and model-agnostic way, without altering backbone parameters.
Specifically, we build a shift estimator to calculate the inherent shift of tabular representations and design a space mapping method to alleviate the influence of redundant information. To end this, we further design two unsupervised losses (please refer to Eq.~\ref{Shift Learning} and Eq.~\ref{regulariz_loss}).
}
% We need to stress that attaching a new head to a trained backbone and training this new head (or backbone + new head) only with conventional supervised training loss (called fine-tuning) can not achieve satisfactory performance, where we report the experimental results in Table~\ref{table:comparison with fine-tuning} for a fair comparison. We can observe that 
% the fine-tuning performance is similar to that of the well-trained model and worse than ours. 
% The possible reason is that only training the new head or fine-tuning the backbone with conventional supervised training loss (already used by the trained backbone) can not provide additional useful information for representation. 
% Different from these fine-tuning methods, TRC enhances the representations extracted by well-trained deep models via two tailor-designed tasks. Specifically, we build a shift estimator to calculate the inherent shift of tabular representations and design a space mapping method to alleviate the influence of redundant information. To end this, we further design two unsupervised losses (please refer to Eq.~\ref{Shift Learning} and Eq.~\ref{regulariz_loss}). 

\subsection{\textcolor{blue}{Further Analysis}}
\begin{figure}[t!]
    \centering
    \hfill
    \begin{subfigure}[b]{0.47\linewidth}
        \centering
        \includegraphics[width=\linewidth]{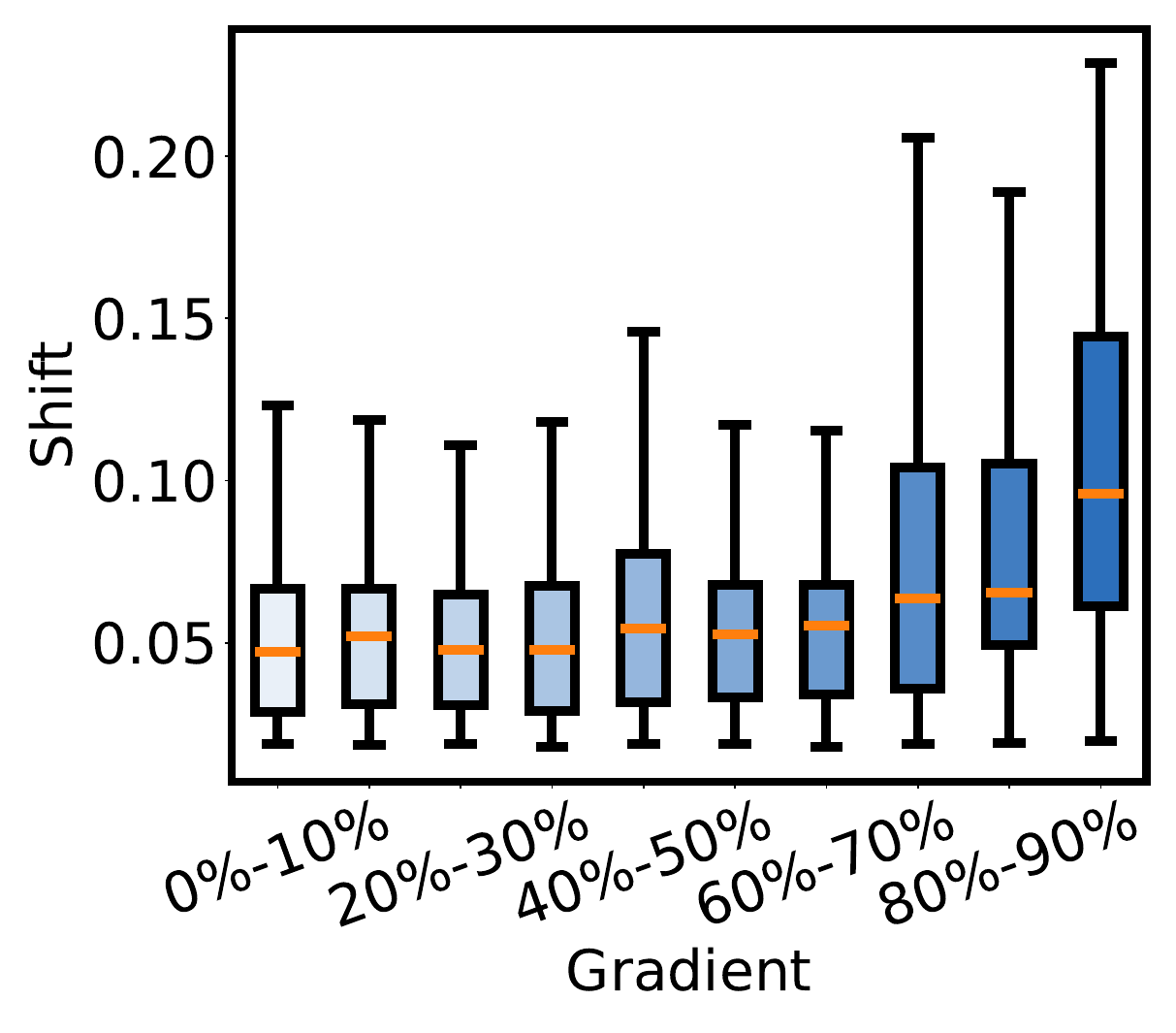}
        % \captionsetup{skip=0pt}
        \caption{MLP on CO dataset}
    \end{subfigure}
    \hfill
    \begin{subfigure}[b]{0.47\linewidth}
        \centering
        \includegraphics[width=\linewidth]{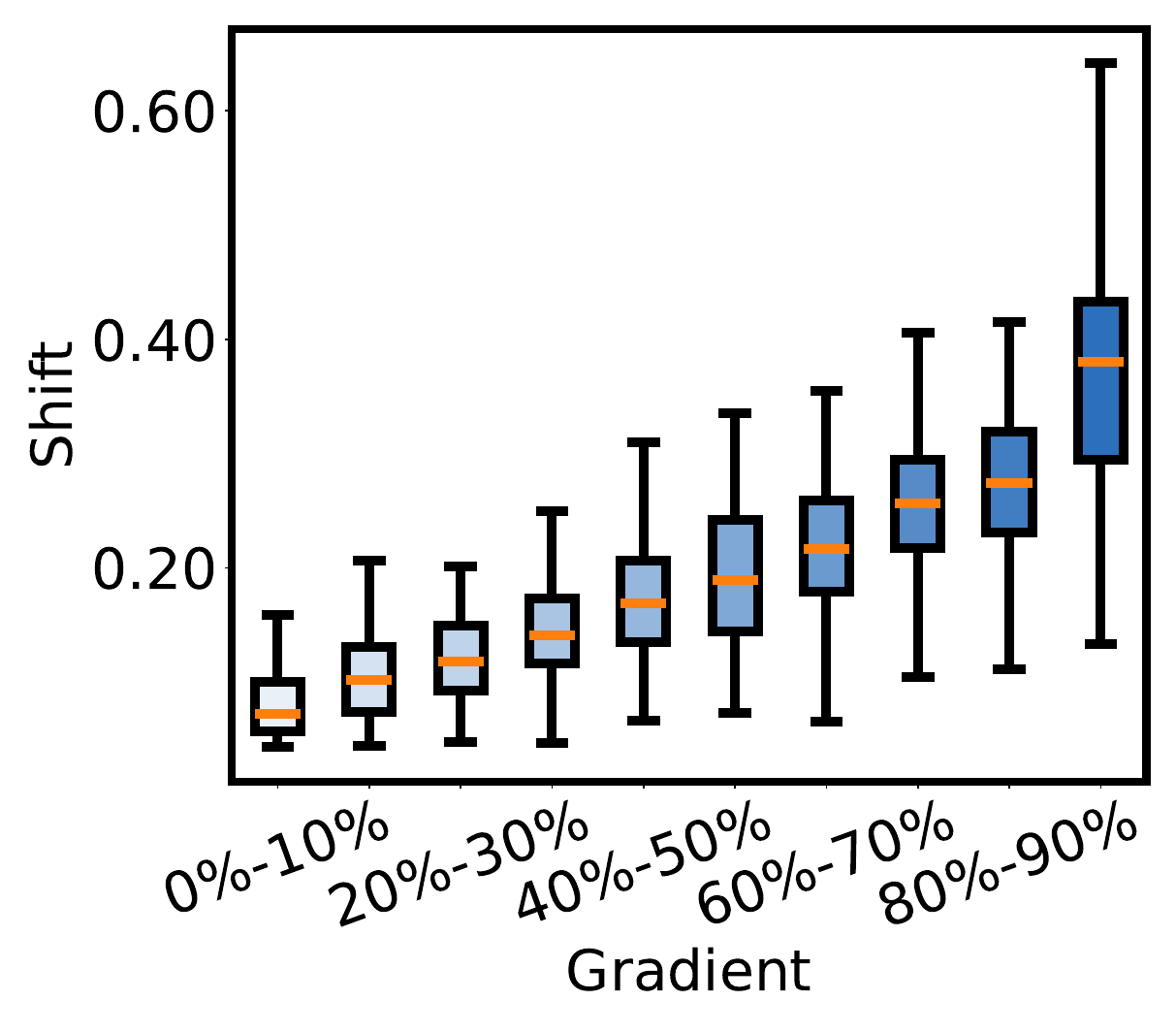}
        % \captionsetup{skip=0pt}
        \caption{ResNet on CO dataset}
    \end{subfigure}
    \hfill
    
    \hfill
    \begin{subfigure}[b]{0.47\linewidth}
        \centering
        \includegraphics[width=\linewidth]{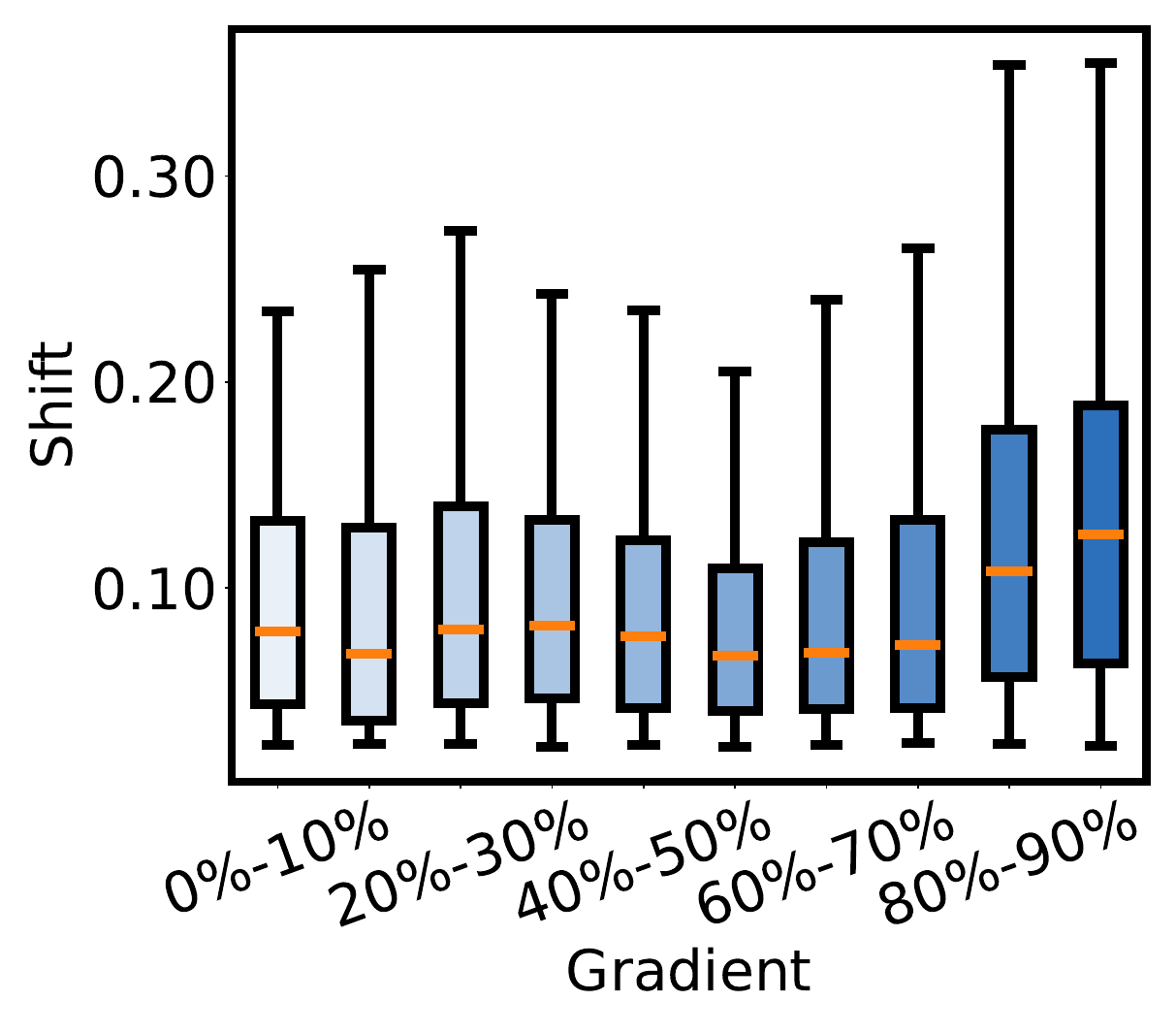}
        % \captionsetup{skip=0pt}
        \caption{MLP on CA dataset}
    \end{subfigure}
    \hfill
    \begin{subfigure}[b]{0.47\linewidth}
        \centering
        \includegraphics[width=\linewidth]{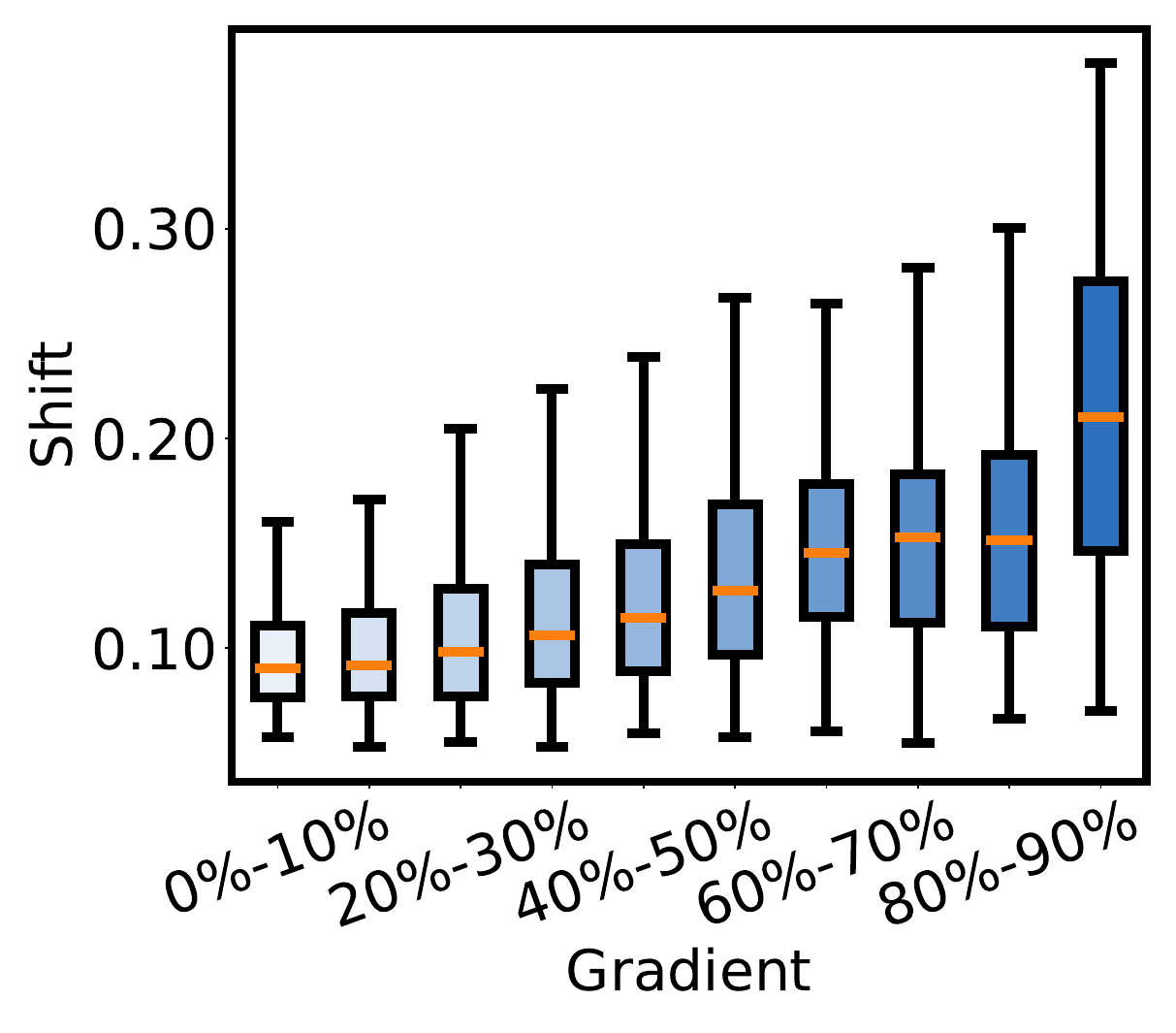}
        % \captionsetup{skip=0pt}
        \caption{ResNet on CA dataset}
    \end{subfigure}
    \hfill
    % \captionsetup{skip=4pt}
    \caption{Comparison of the $L_1$ norm of the shift information estimated by \method on data samples with different gradients. For each data sample, the gradient is computed by $\|\nabla_{\theta_f}\mathcal{L}(x, y;\theta)\|_1$, where $\theta$ is the parameters of the trained deep tabular model. We partition the $L_1$ norm of gradients into 10 equally spaced intervals from low to high and count the number of samples in each interval.}
    \label{fig:gradient for representation}
    % \vspace{-1em}
\end{figure}

\textbf{\method enables the learning of the inherent shift by representation re-estimation.}
Fig.~\ref{fig:gradient for representation} illustrates the estimated shift information within the representations of trained deep tabular backbones.
For the representations of the backbone $G_f(\cdot;\theta_f)$, those with larger gradients are associated with greater shift information learned by TRC.
Through the task of \textit{tabular representation re-estimation}, the shift information contained in sub-optimal representations can be gradually eliminated.
% Owing to feature heterogeneity, observations in the dataset are susceptible to varying degrees of noise. 
% Such noise within the observations can lead the learned representations by deep tabular machine learning models to deviate from the optimal ideal distribution, thereby decreasing the model performance. 

\begin{figure*}[!t]
% \vspace{-1em}
    \centering
    \hfill
    \begin{subfigure}[b]{0.23\linewidth}
        \centering
        \includegraphics[width=\linewidth]{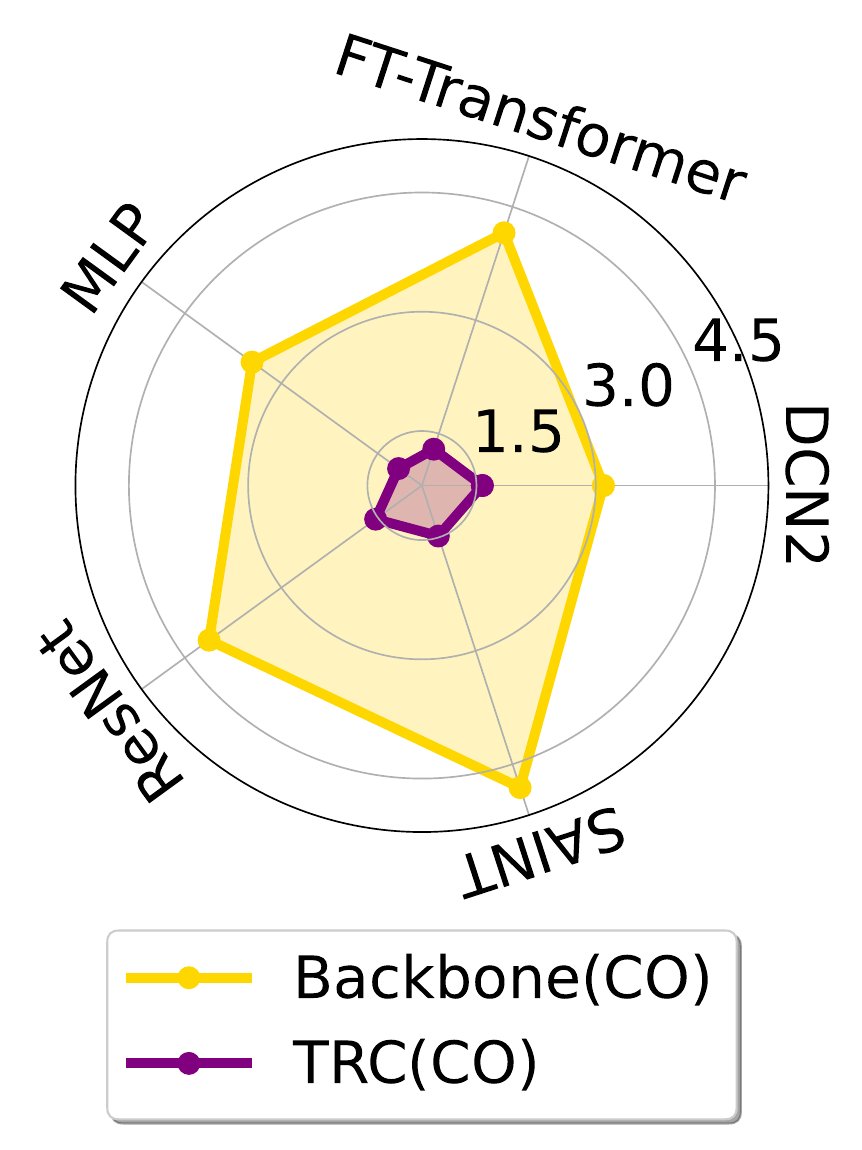}
        % \captionsetup{skip=0pt}
        \caption{SVE on CO dataset}
    \end{subfigure}
    \hfill
    \begin{subfigure}[b]{0.24\linewidth}
        \centering
        \includegraphics[width=\linewidth]{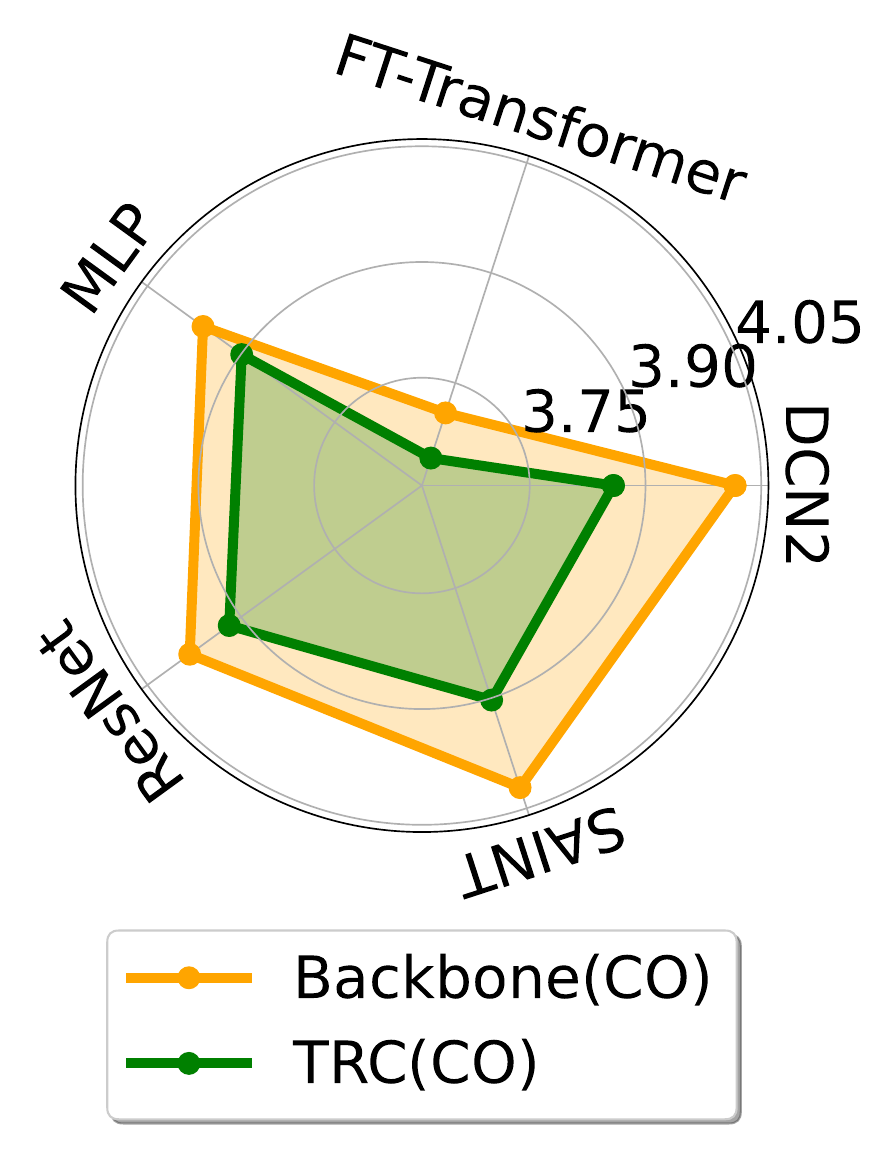}
        % \captionsetup{skip=0pt}
        \caption{RMSE ($\downarrow$) on CO dataset}
    \end{subfigure}
    \hfill
    \begin{subfigure}[b]{0.23\linewidth}
        \centering
        \includegraphics[width=\linewidth]{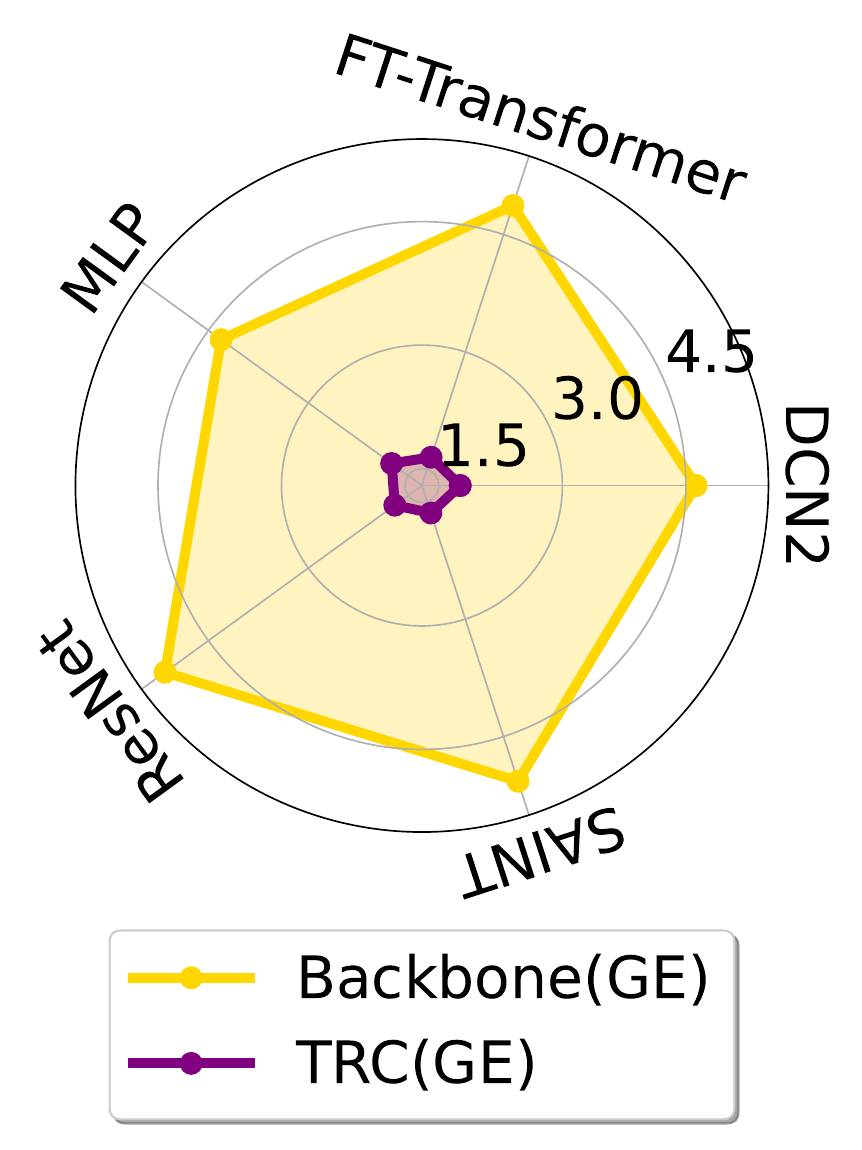}
        % \captionsetup{skip=0pt}
        \caption{SVE on GE dataset}
    \end{subfigure}
    \hfill
    \begin{subfigure}[b]{0.23\linewidth}
        \centering
        \includegraphics[width=\linewidth]{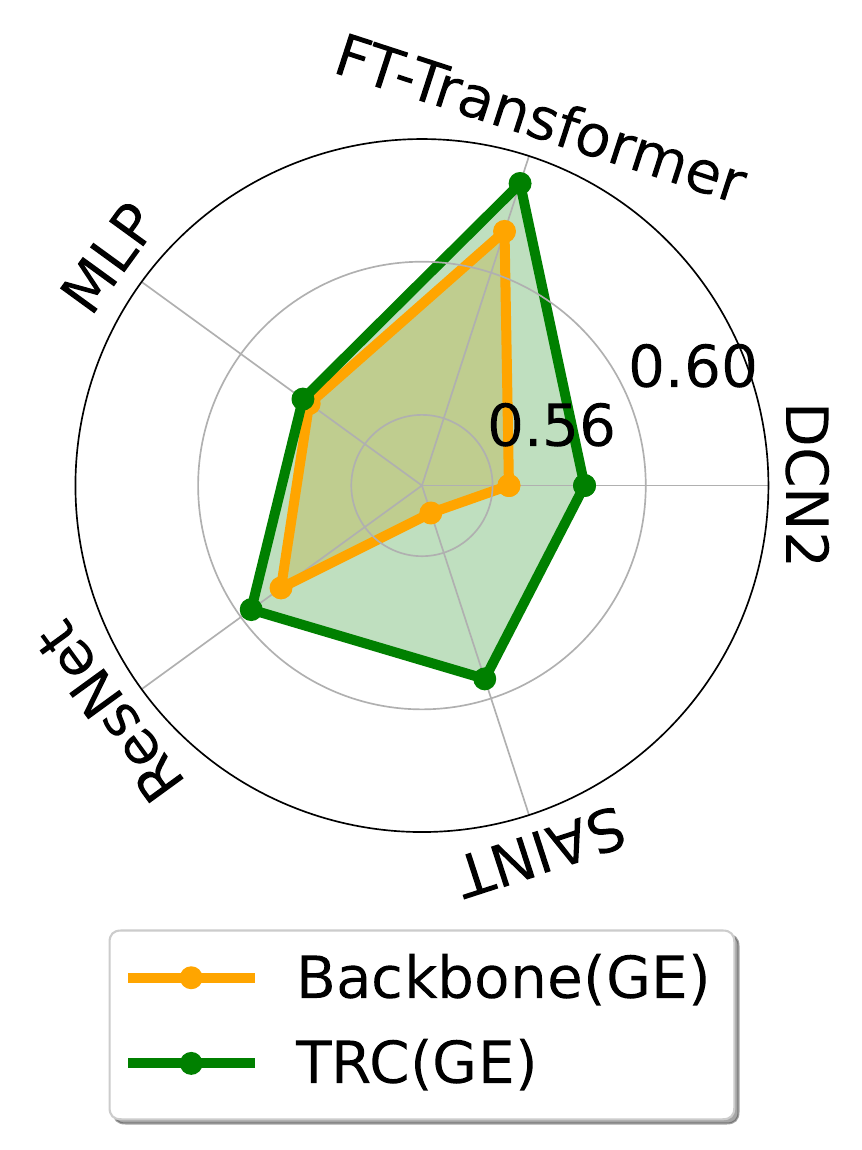}
        % \captionsetup{skip=4pt}
        \caption{Accuracy ($\uparrow$) on GE dataset}
    \end{subfigure}
    \caption{SVE and performance changes of backbone models w/ and w/o \method on different datasets. Subfigures (a) and (c) show the Singular Value Entropy (SVE) (Eq.~\ref{SVE}) of representation space w/ and w/o \method on different datasets. Subfigures (b) and (d) show the corresponding performance.}
    \label{fig:sve and performance}
\end{figure*}

\textbf{\method reduces the redundant information by space mapping.}
Subfigure (a) of Fig.~\ref{fig:sve and performance} shows the representation SVE of deep tabular backbone $G_f(\cdot;\theta_f)$ w/ and w/o \method, and subfigure (b) shows the corresponding performance. 
Similar to subfigures (c) and (d).
We could find that with \method, the SVE value of the representations decreased, and the performance improved.
This demonstrates that \method reduces the redundant information via \textit{tabular space mapping} to enhance the model performance.

\begin{table}[t!]
\centering
\footnotesize
% \captionsetup{font=small}
\caption{The influence of using a small portion (1\%) of validation set for training. ``Train'' indicates only training data are used for training model, ``Val'' indicates additional small portion of validation data are used for training.}
\label{table:influence of a small portion of validation data}
\setlength{\tabcolsep}{1.9mm}{
    \begin{tabular}{ccccccc}
    \toprule
                 & \multicolumn{2}{c}{CO $\downarrow$} & \multicolumn{2}{c}{DI $\downarrow$} & \multicolumn{2}{c}{AU $\uparrow$} \\ \cmidrule(lr){2-3} \cmidrule(lr){4-5} \cmidrule(lr){6-7}
                 & Train           & Val               & Train             & Val                 & Train             & Val                \\ \midrule                 
    MLP      & 3.961           & 3.968             & 564.373           & 565.616             & 0.870              & 0.870               \\
    +\textbf{TRC}& \textbf{3.902}  & \textbf{3.899}    & \textbf{559.020}  & \textbf{558.330}   & \textbf{0.891}     & \textbf{0.891}      \\ \midrule
    % DCN2 (i)     & 4.016           & 3.992             & 591.958           & 589.953             & 1.027             & 1.028                & 0.841              & 0.849               \\
    % +\textbf{TRC}& \textbf{3.846}  & \textbf{3.858}    & \textbf{574.190}  & \textbf{574.453}    & \textbf{0.870}    & \textbf{0.879}       & \textbf{0.859}     & \textbf{0.862}      \\ \midrule
    % AutoInt (i)  & 4.020           & 4.019             & 562.169           & 561.472             & 0.860             & 0.858                & 0.848              & 0.855               \\
    % +\textbf{TRC}& \textbf{3.956}  & \textbf{3.960}    & \textbf{555.852}  & \textbf{557.423}    & \textbf{0.825}    & \textbf{0.829}       & \textbf{0.887}     & \textbf{0.891}      \\ \midrule
    \textcolor{blue}{FT-Transformer} & 3.709           & 3.726             & 551.190           & 550.228                  & 0.848              & 0.848               \\
    +\textbf{TRC}& \textbf{3.647}  & \textbf{3.648}    & \textbf{544.570}  & \textbf{543.653}     & \textbf{0.877}     & \textbf{0.884}      \\ 
    % SCARF (p)    & 3.856           & 3.862             & 579.610           & 580.673             & 0.863             & 0.856                & 0.862              & 0.865               \\
    % +\textbf{TRC}& \textbf{3.843}  & \textbf{3.847}    & \textbf{576.799}  & \textbf{577.803}    & \textbf{0.818}    & \textbf{0.808}       & \textbf{0.889}     & \textbf{0.884}      \\ \midrule
    % SAINT (p)    & 4.022           & 4.134             & 597.207           & 596.034             & 0.827             & 0.831                & 0.870              & 0.879               \\
    % +\textbf{TRC}& \textbf{3.899}  & \textbf{3.903}    & \textbf{557.392}  & \textbf{557.951}    & \textbf{0.812}    & \textbf{0.818}       & \textbf{0.890}     & \textbf{0.884}      \\
    \bottomrule
    \end{tabular}}
% \vspace{-1em}
\end{table}

\textbf{The performance improvement comes from the effectiveness of the \method, rather than from additional training data.}
Only a small portion (1\%) of the validation data is used to train our shift estimator, and these additional data are not used in the remaining learning or test process. We provide additional experiment results in Table~\ref{table:influence of a small portion of validation data} to show whether the improvements come from additional training data or the proposed modifications.  Here, we consider two settings. One setting is that we use a portion of the training data rather than the additional validation data to train the shift estimator, making it fair when being compared with existing baselines. Another setting is that we also add a small portion (1\%) of the validation data into the training set when training baseline models. As listed in Table~\ref{table:influence of a small portion of validation data}, ours still outperforms the corresponding baseline even without using additional validation data. In addition, the baseline shows similar performance whether or not additional data is used, and \method similarly demonstrates consistent performance regardless of the use of additional data.
Therefore, we demonstrate that the performance improvement comes from the effectiveness of the \method, rather than from additional training data. 

\begin{figure}[t!]
    \centering
    \hfill
    \begin{subfigure}[b]{0.47\linewidth}
        \centering
        \includegraphics[width=\linewidth]{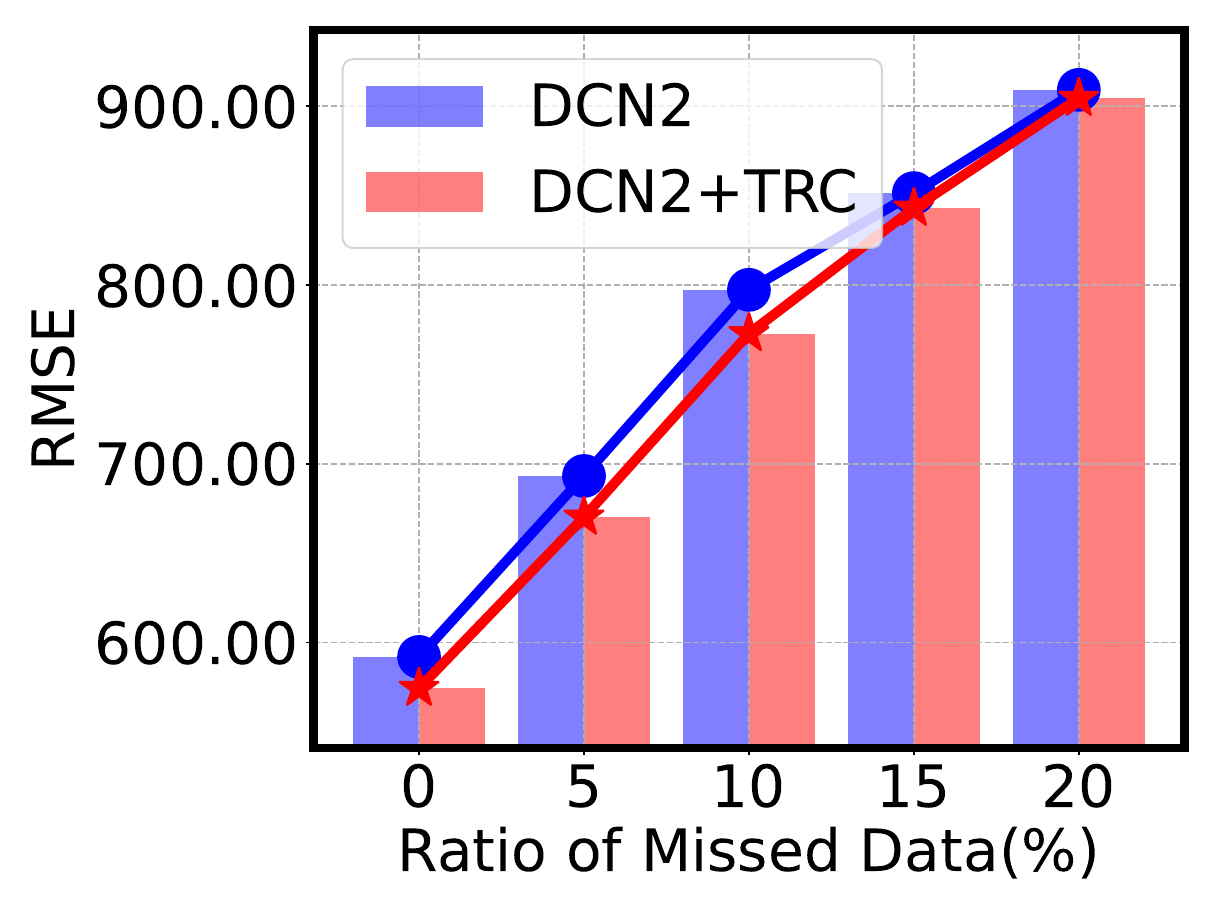}
        % \captionsetup{skip=0pt}
        \caption{RMSE on DI dataset $\downarrow$}
    \end{subfigure}    
    \hfill
    \begin{subfigure}[b]{0.47\linewidth}
        \centering
        \includegraphics[width=\linewidth]{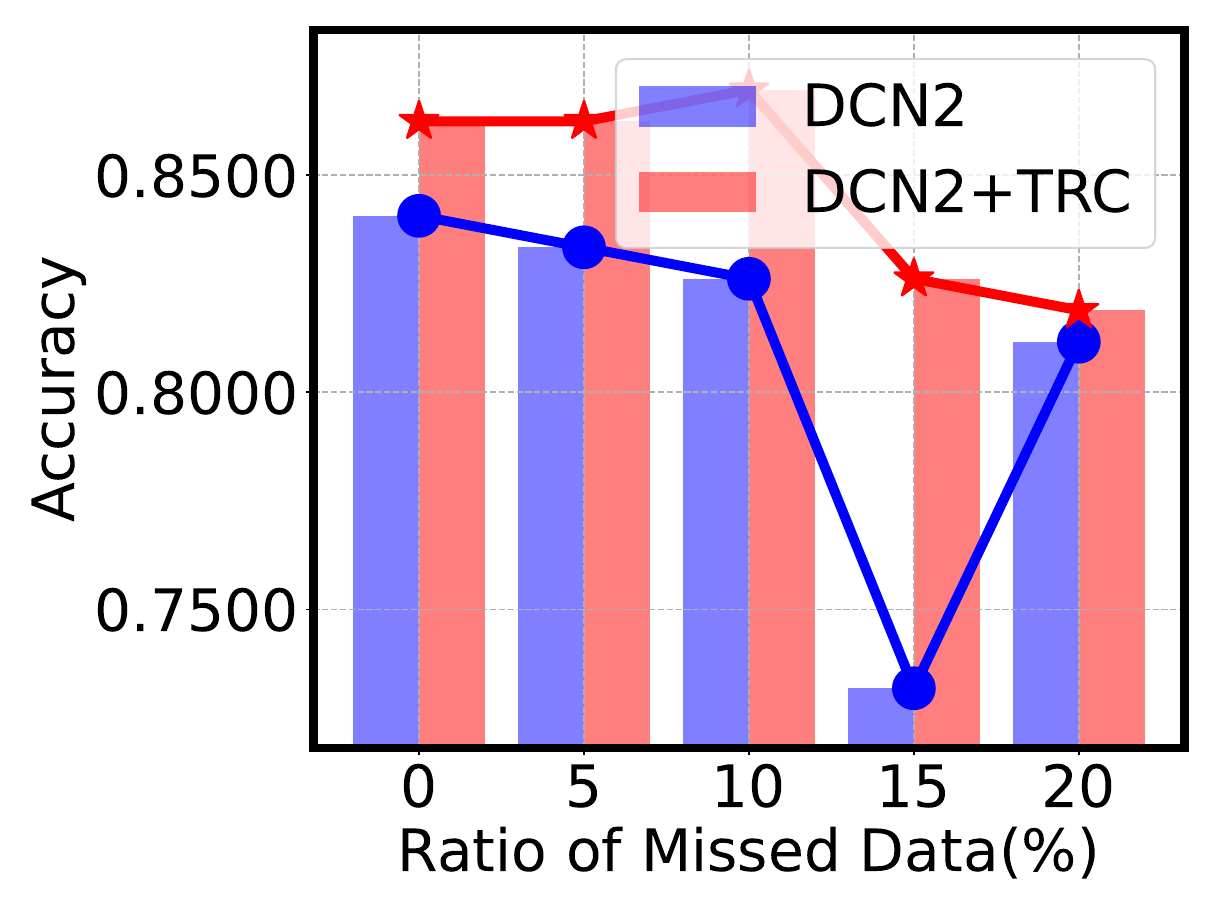}
        \caption{ACC on AU dataset $\uparrow$}
    \end{subfigure}   
    \hfill
    
    \hfill
    \begin{subfigure}[b]{0.47\linewidth}
        \centering
        \includegraphics[width=\linewidth]{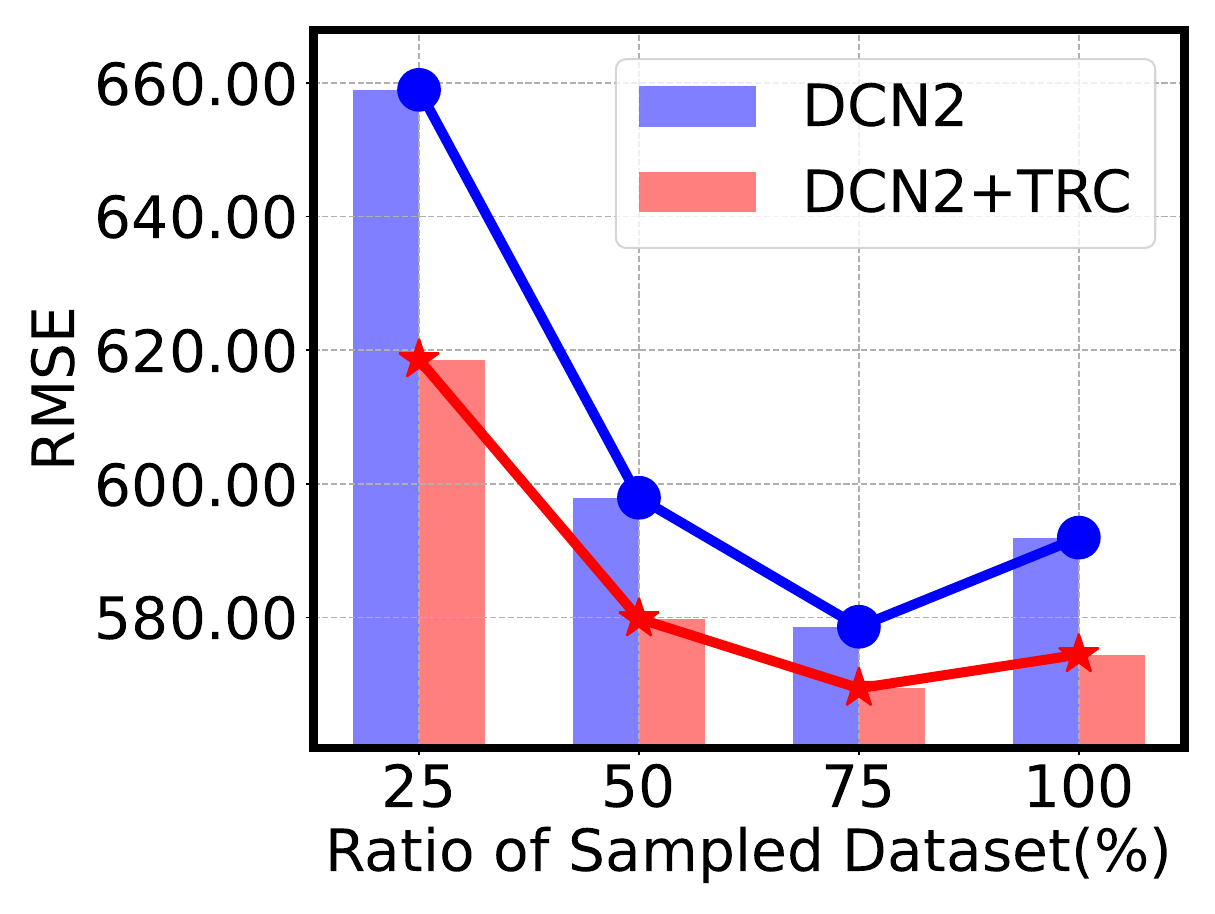}
        % \captionsetup{skip=0pt}
        \caption{RMSE on DI dataset $\downarrow$}
    \end{subfigure}   
    \hfill
    \begin{subfigure}[b]{0.47\linewidth}
        \centering
        \includegraphics[width=\linewidth]{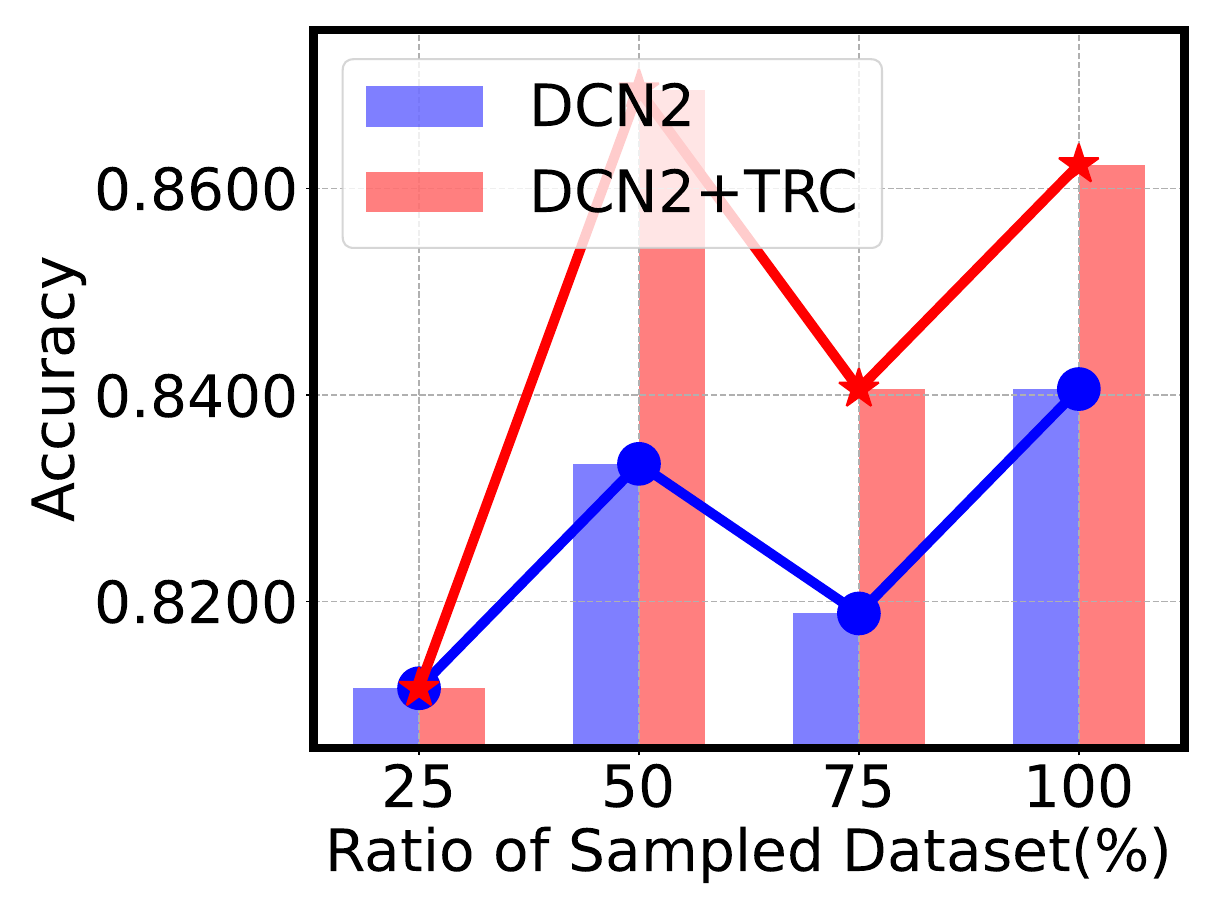}
        \caption{ACC on AU dataset $\uparrow$}
    \end{subfigure}
    \hfill
    \caption{The results of \method training with missing values and fewer samples. The left two subfigures indicate the scenarios with missing values, while the right two indicate the scenarios where the number of training samples is reduced.}
    \label{fig:missing value and few sample}
    % \vspace{-1em}
\end{figure}

\begin{figure*}[t!]
    \centering
    \begin{subfigure}[b]{0.23\linewidth}
        \centering
        \includegraphics[width=\linewidth]{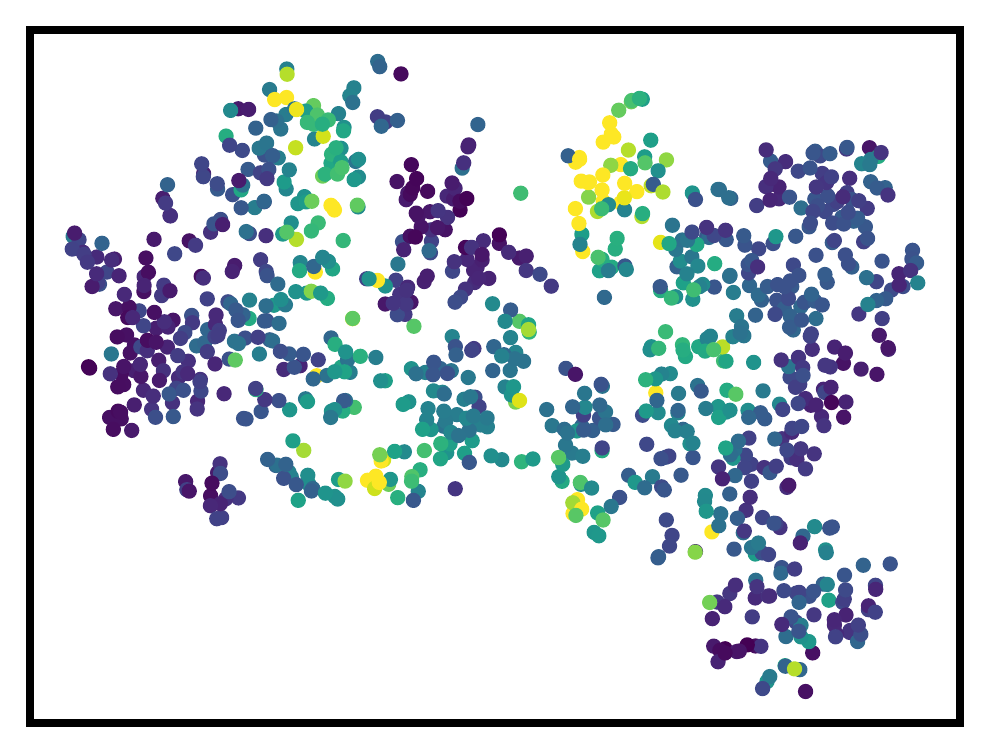}
        % \captionsetup{font=small}
        % \captionsetup{skip=0pt}
        \caption{Baseline}
    \end{subfigure}
    % \hfill
    \begin{subfigure}[b]{0.23\linewidth}
        \centering
        \includegraphics[width=\linewidth]{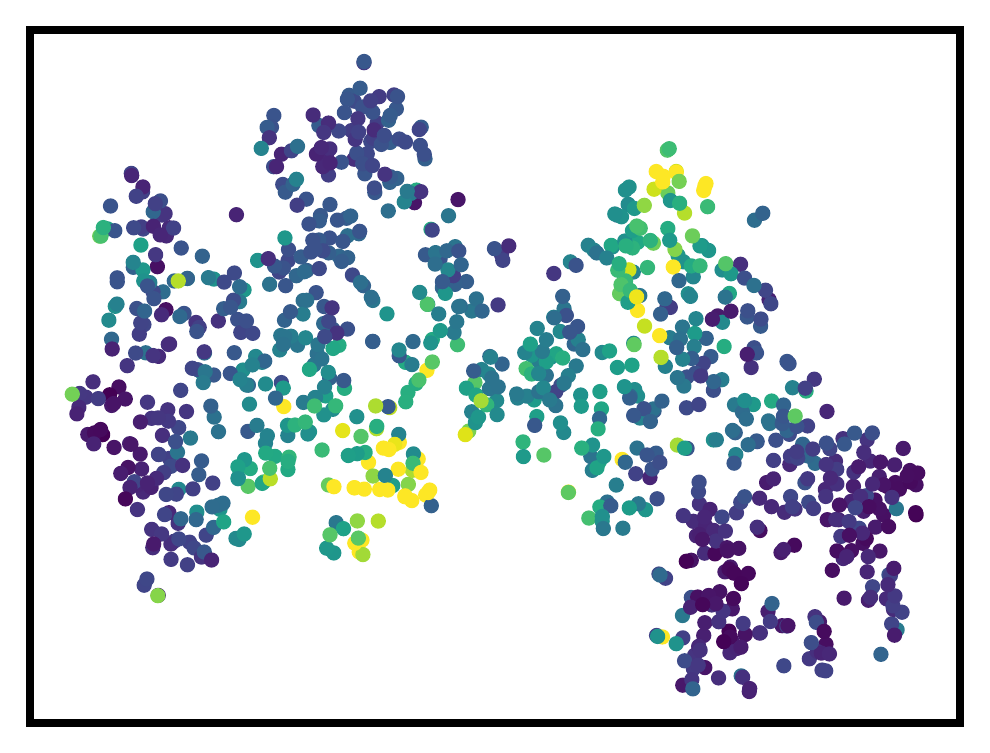}
        % \captionsetup{font=small}
        % \captionsetup{skip=0pt}
        \caption{w/ FT}
    \end{subfigure}
    % \hfill
    \begin{subfigure}[b]{0.23\linewidth}
        \centering
        \includegraphics[width=\linewidth]{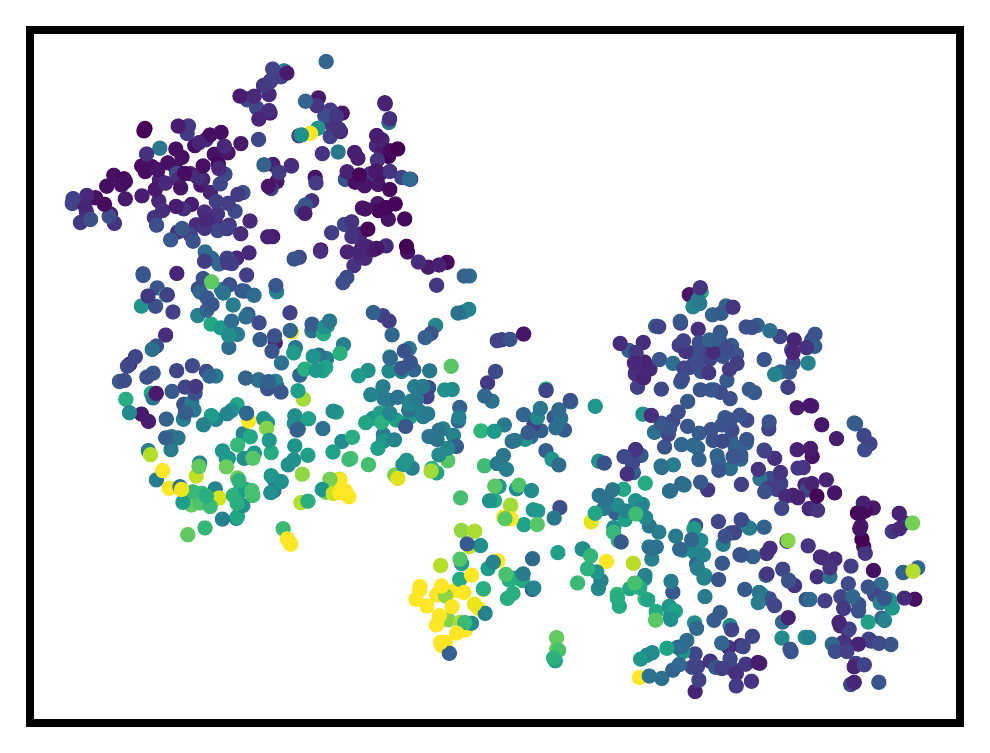}
        % \captionsetup{font=small}
        % \captionsetup{skip=0pt}
        \caption{w/ Task 1}
    \end{subfigure}
    % \hfill
    \begin{subfigure}[b]{0.23\linewidth}
        \centering
        \includegraphics[width=\linewidth]{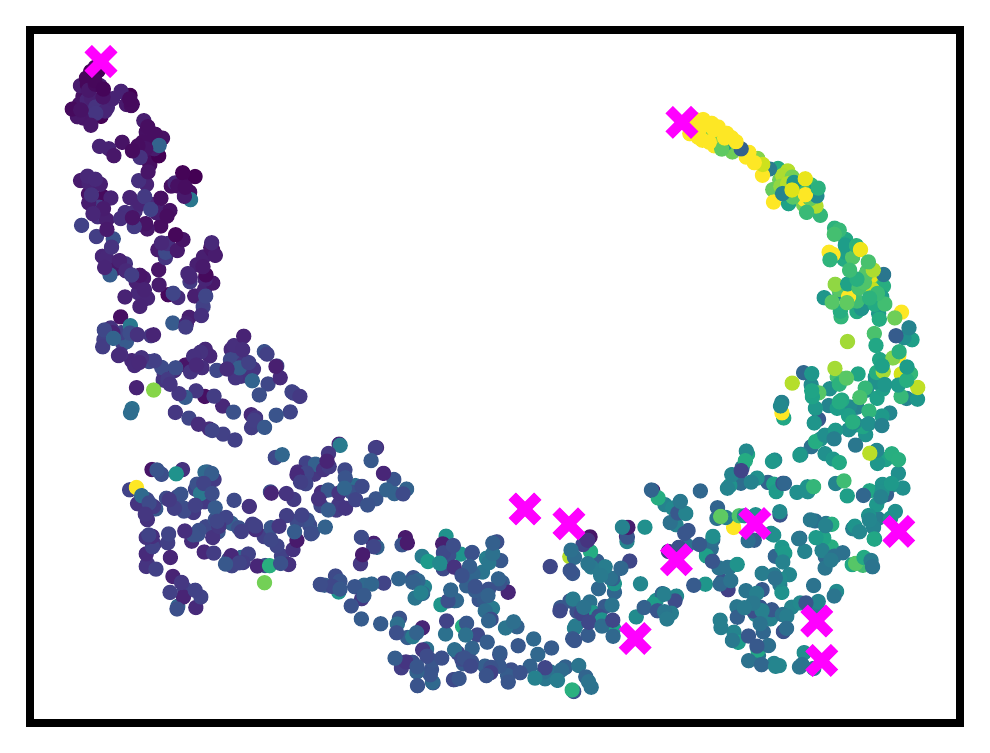}
        % \captionsetup{font=small}
        % \captionsetup{skip=0pt}
        \caption{w/ Task 1, 2 (TRC)}
    \end{subfigure}
    % \hfill
    \begin{subfigure}[b]{0.23\linewidth}
        \centering
        \includegraphics[width=\linewidth]{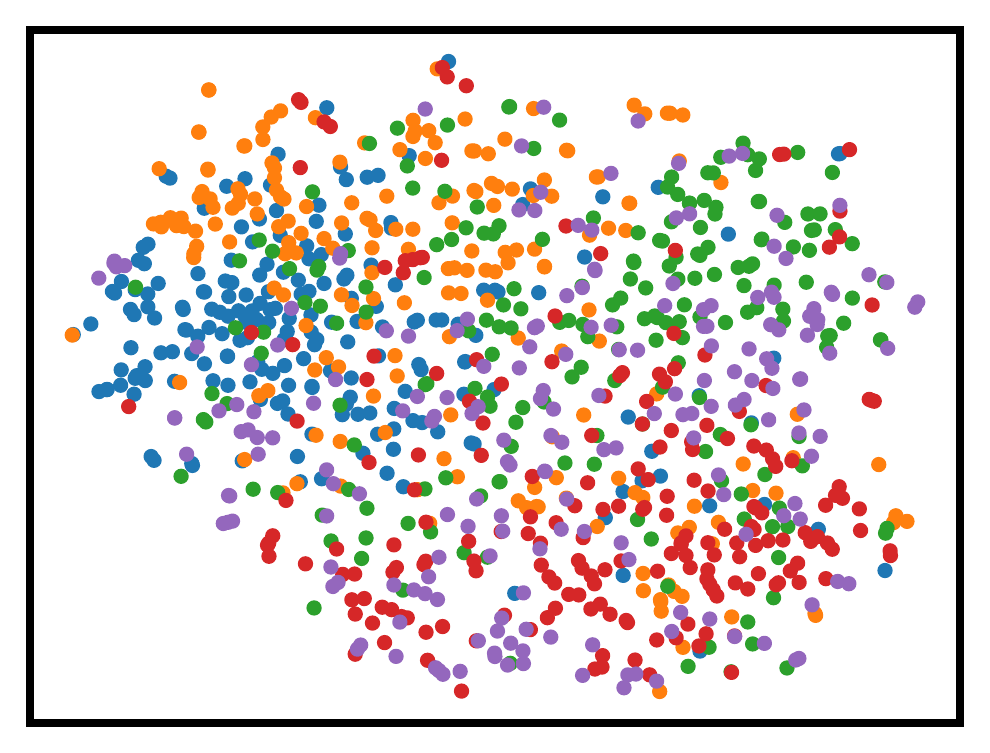}
        % \captionsetup{font=small}
        % \captionsetup{skip=0pt}
        \caption{Baseline}
    \end{subfigure}
    % \hfill
    \begin{subfigure}[b]{0.23\linewidth}
        \centering
        \includegraphics[width=\linewidth]{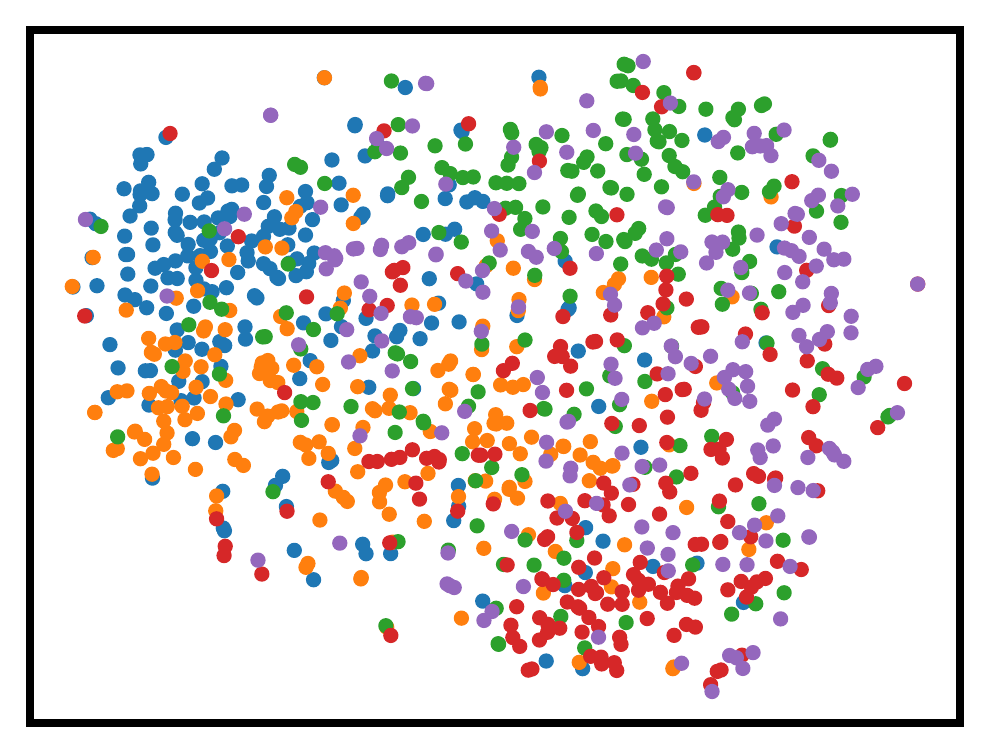}
        % \captionsetup{font=small}
        % \captionsetup{skip=0pt}
        \caption{w/ FT}
    \end{subfigure}
    % \hfill
    \begin{subfigure}[b]{0.23\linewidth}
        \centering
        \includegraphics[width=\linewidth]{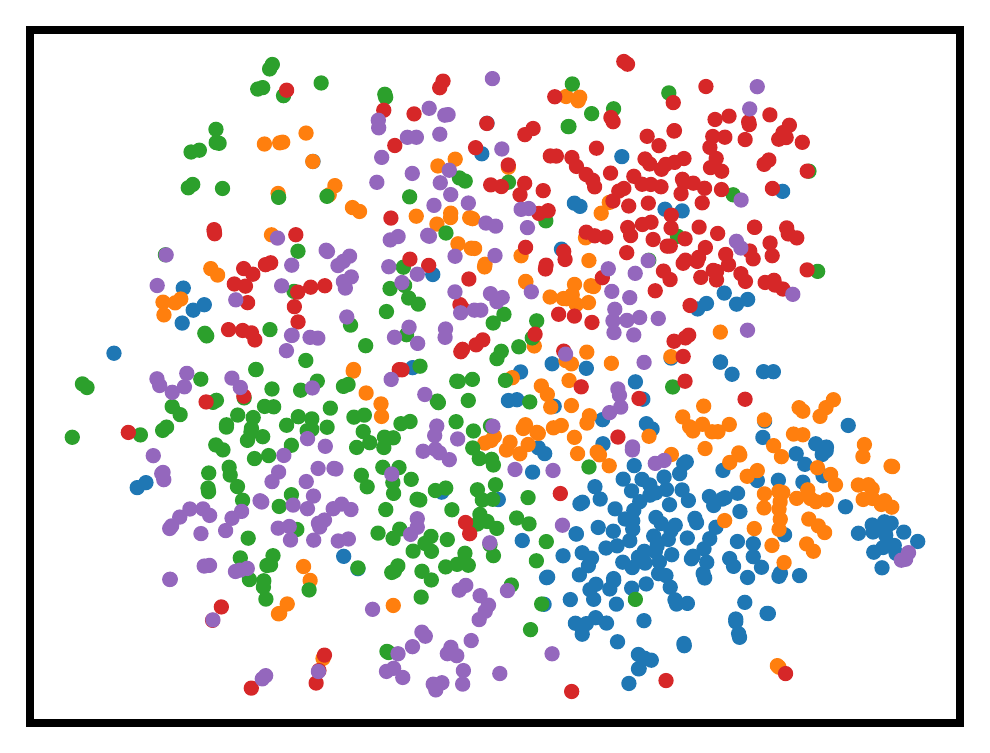}
        % \captionsetup{font=small}
        % \captionsetup{skip=0pt}
        \caption{w/ Task 1}
    \end{subfigure}
    % \hfill
    \begin{subfigure}[b]{0.23\linewidth}
        \centering
        \includegraphics[width=\linewidth]{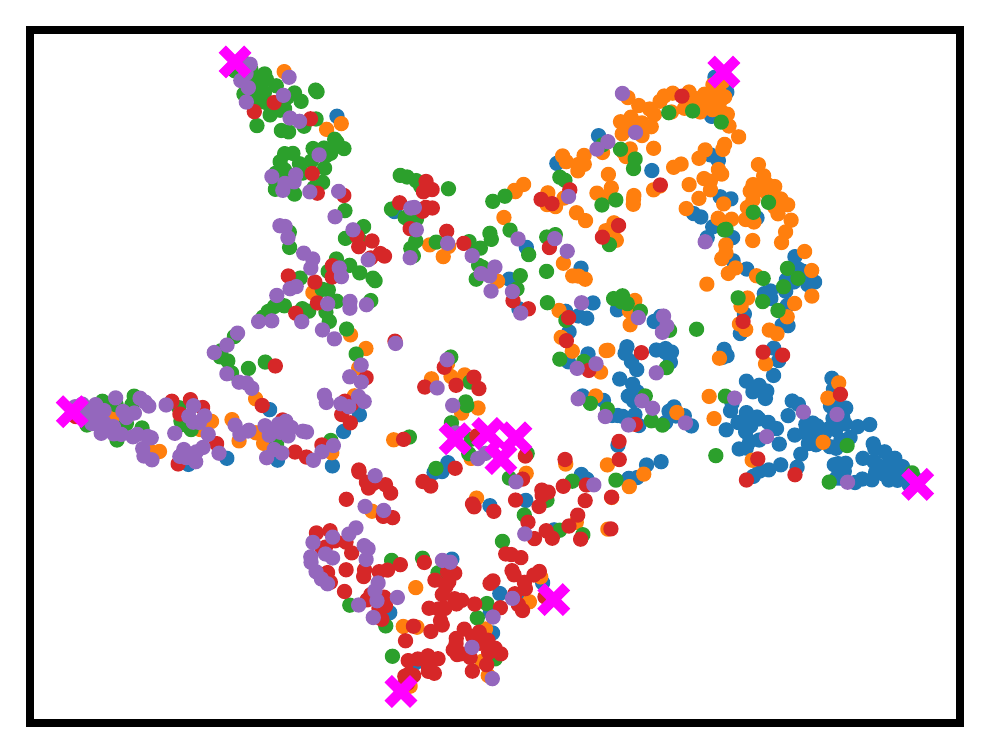}
        % \captionsetup{font=small}
        % \captionsetup{skip=0pt}
        \caption{w/ Task 1, 2 (TRC)}
    \end{subfigure}
    % \hfill
    % \captionsetup{skip=0pt}
    \caption{The TSNE visualization of learned representations of deep tabular model, deep tabular model w/ FT (fine-tuning), deep tabular model w/ Task 1 (Tabular Representation Re-estimation) and deep tabular model w/ Task 1, 2 (Tabular Representation Re-estimation and Tabular Space Mapping) (i.e., TRC). Here, the backbone model is ResNet. The first row represents the CA dataset (regression), while the second row represents the GE dataset (classification). Different colors indicate different labels. Marker=``x'' indicates embedding vectors.}
    \label{fig:visualization with embeddings}
    % \vspace{-1em}
\end{figure*}

\textcolor{blue}{\textbf{The performance of \method under missing values and reduced training data.}}
% \textbf{\method continues to enhance the performance of deep tabular models even under challenging conditions where training samples contain missing values and the overall number of training samples is reduced.}
We systematically increase the proportion of missing values in features and reduce the proportion of training samples separately.  
The results in Fig.~\ref{fig:missing value and few sample} indicate that as the dataset's missing values escalate or the training sample size diminishes, the difficulty of model training intensifies, which is reflected in a decline in model performance. Nevertheless, by leveraging \method, we are able to consistently enhance the capabilities of the deep tabular model, thereby alleviating the negative impact of data incompleteness and scarcity.
% More results can be found in Appendix C.
More results can be found in the Appendix D.
% More results can be found in the Appendix~\ref{appendix:missing value and few sample}.

\begin{figure}[t!]
    \centering
    % \hfill
    % \begin{subfigure}[b]{0.24\linewidth}
    %     \centering
    %     \includegraphics[width=\linewidth]{pictures/parameters/california_housing.pdf}
    %     \caption{CA dataset}
    % \end{subfigure}
    % \hfill
    % \begin{subfigure}[b]{0.23\linewidth}
    %     \centering
    %     \includegraphics[width=\linewidth]{pictures/parameters/combined_cycle_power_plant.pdf}
    %     % \captionsetup{skip=0pt}
    %     \caption{CO dataset}
    % \end{subfigure}
    % \hfill
    \begin{subfigure}[b]{0.46\linewidth}
        \centering
        \includegraphics[width=\linewidth]{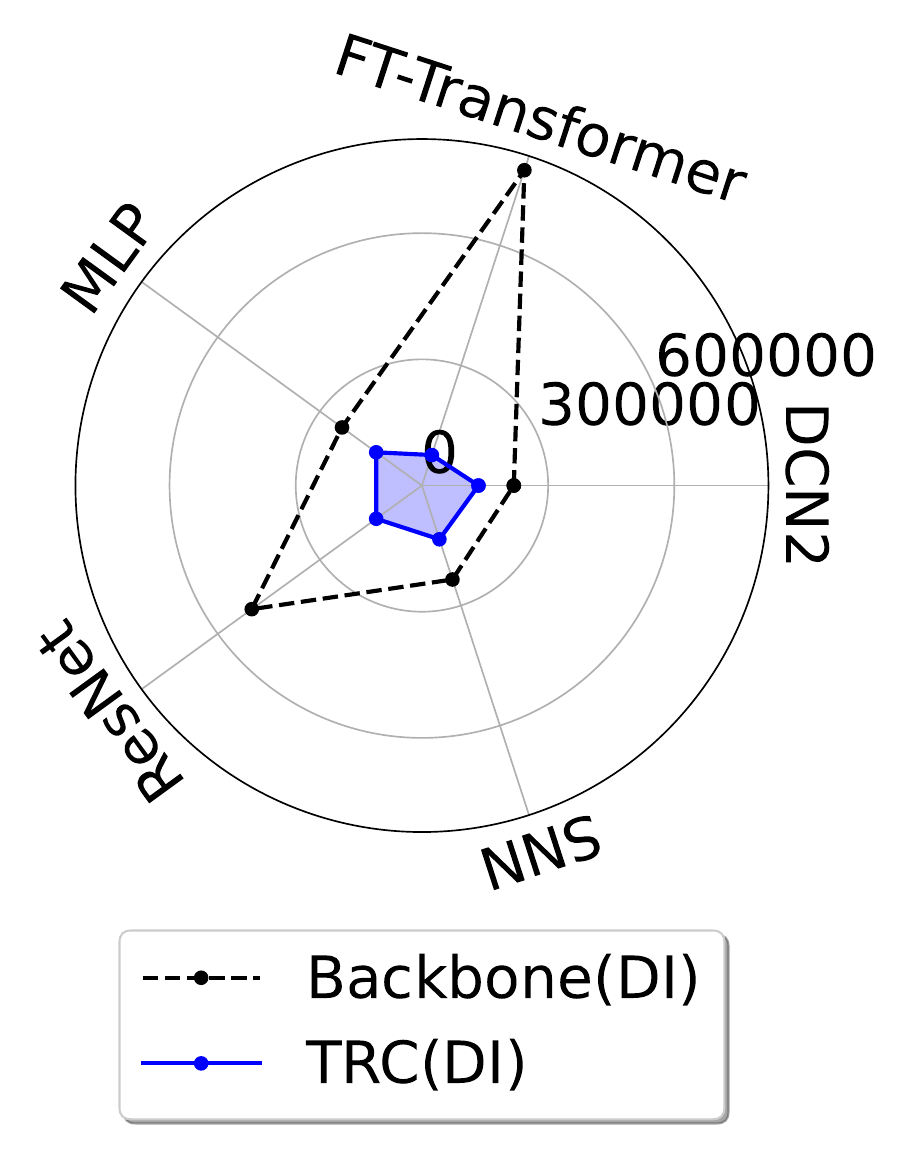}
        % \captionsetup{skip=0pt}
        \caption{DI dataset}
    \end{subfigure}
    % \hfill
    % \begin{subfigure}[b]{0.24\linewidth}
    %     \centering
    %     \includegraphics[width=\linewidth]{pictures/parameters/qsar_fish_toxicity.pdf}
    %     \caption{QS dataset}
    % \end{subfigure}
    % \centering
    % \hfill
    % \begin{subfigure}[b]{0.23\linewidth}
    %     \centering
    %     \includegraphics[width=\linewidth]{pictures/parameters/superconductivty_data.pdf}
    %     % \captionsetup{skip=0pt}
    %     \caption{SU dataset}
    % \end{subfigure}
    % \hfill
    % \begin{subfigure}[b]{0.24\linewidth}
    %     \centering
    %     \includegraphics[width=\linewidth]{pictures/parameters/pol.pdf}
    %     \caption{PO dataset}
    % \end{subfigure}
    % \hfill
    \begin{subfigure}[b]{0.46\linewidth}
        \centering
        \includegraphics[width=\linewidth]{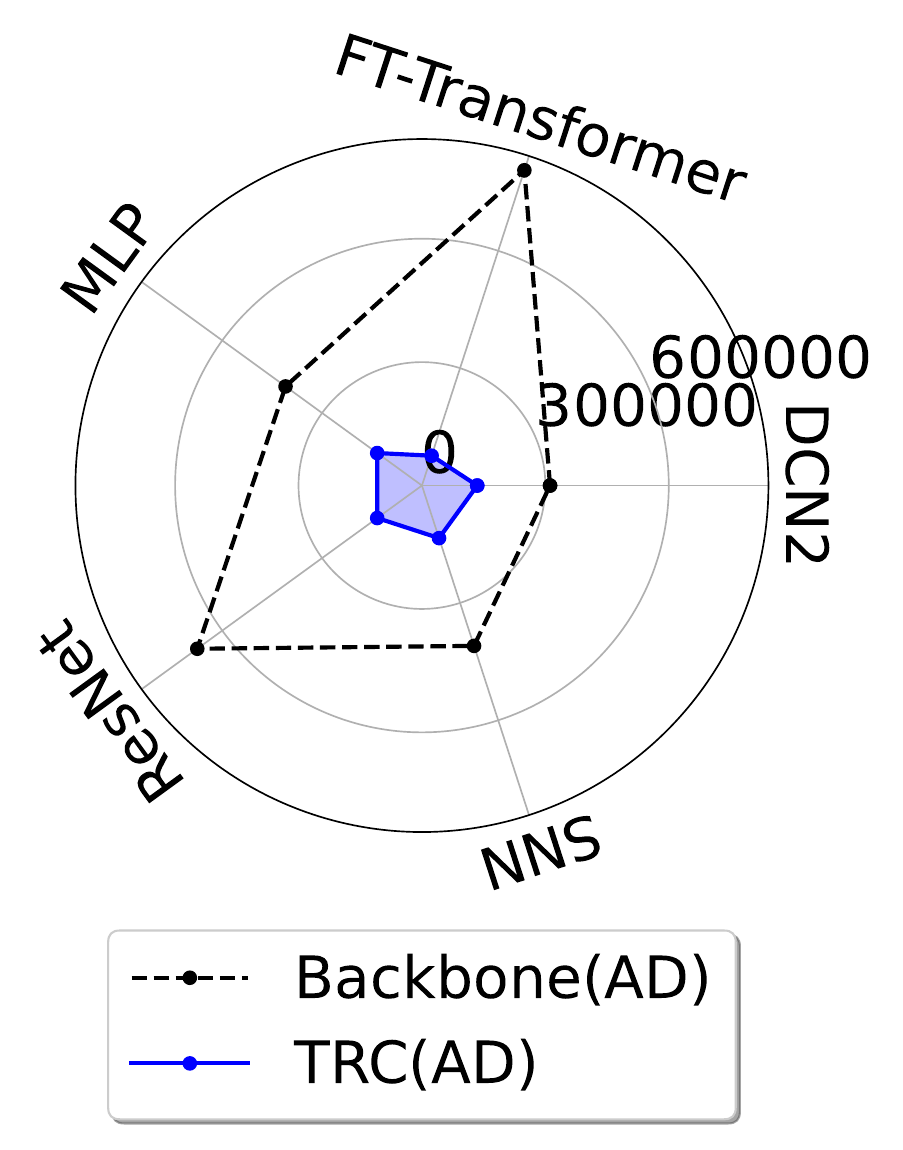}
        % \captionsetup{skip=0pt}
        \caption{AD dataset}
    \end{subfigure}
    % \hfill
    % \begin{subfigure}[b]{0.24\linewidth}
    %     \centering
    %     \includegraphics[width=\linewidth]{pictures/parameters/GesturePhaseSegmentationProcessed.pdf}
    %     \caption{GE dataset}
    % \end{subfigure}
    % \captionsetup{skip=4pt}
    \caption{The comparison of parameters between \method and different deep tabular models. The parameters for \method do not include the parameters for backbone models.}
    % \vspace{-4mm}
    \label{fig:parameters analysis}
\end{figure}

\textbf{Visualization.}
The TSNE visualization of learned representations in Fig.~\ref{fig:visualization with embeddings} demonstrates that TRC could well calibrate representations.
Even only with Tabular Representation Re-estimation, the representations are more organized compared to the baseline and fine-tuning approaches, as shown in subfigure (c).
We also visualize the learned embedding vectors in subfigures (d) and (h).
The embedding vectors are learnable, and we encourage the embedding vectors to be orthogonal through an orthogonality loss (Eq.~\ref{regulariz_loss}), as orthogonality is known to be helpful in improving discriminative ability~\cite{liu2023learning, jiang2024uncovering}.
Our findings reveal that these embedding vectors are extensively distributed across the latent space, contributing to the effective modeling of representations. In the regression tasks, which involve continuous labels and necessitate fine-grained information for discriminative representations, the application of TRC leads to a marked improvement in the ability to differentiate between various continuous labels. Consequently, the arrangement of these representations becomes more structured and organized.
In the classification tasks, after applying TRC, representations are separated into several non-overlapping regions, where representations with similar labels are clustered together, demonstrating the method's efficacy in enhancing the separability of the data points.

\textbf{Parameters analysis and Computational efficiency.}
The proposed \method provides a parameter-efficient, cost-effective technique to enhance the representations of deep tabular backbones without altering any of their parameters.
As illustrated in Fig.~\ref{fig:parameters analysis}, the parameter quantity of \method is significantly lower compared to deep tabular models, particularly for the FT-Transformer model, which inherently has a large number of parameters. 
More results are provided in Appendix E.
% More results are provided in Appendix~\ref{appendix:parameter analysis}.
In addition, since \method does not need to retrain the deep tabular backbone, the training time of \method is noticeably reduced compared to the deep tabular model in most cases. 
\textcolor{blue}{The inference cost of TRC is also much smaller compared to deep tabular backbones, thus will not introduce high inference overhead to the existing backbones.}
Details of \textcolor{blue}{training and inference} time cost are provided in \textcolor{blue}{Table XI and Table XII in Appendix F}.
% Details of \textcolor{blue}{training and inference} time cost are provided in \textcolor{blue}{Table~\ref{appendix:time cost} and Table~\ref{appendix:infer time cost} in Appendix~\ref{appendix:computational efficiency}}.

% \textbf{Sensitivity Analysis.}
% We incorporate the sensitivity analysis for the number of embedding vectors, the weight of loss function $\mathcal{L}_{orth}$, the ratio of selected optimal representations, and the perturbing times for constructing simulated sub-optimal representations in Inherent Shift Learning in Fig.~\ref{fig:sensitivity}.
% More results are provided in Appendix F.

\begin{figure}[!t]
\centering
\includegraphics[width=0.95\linewidth]{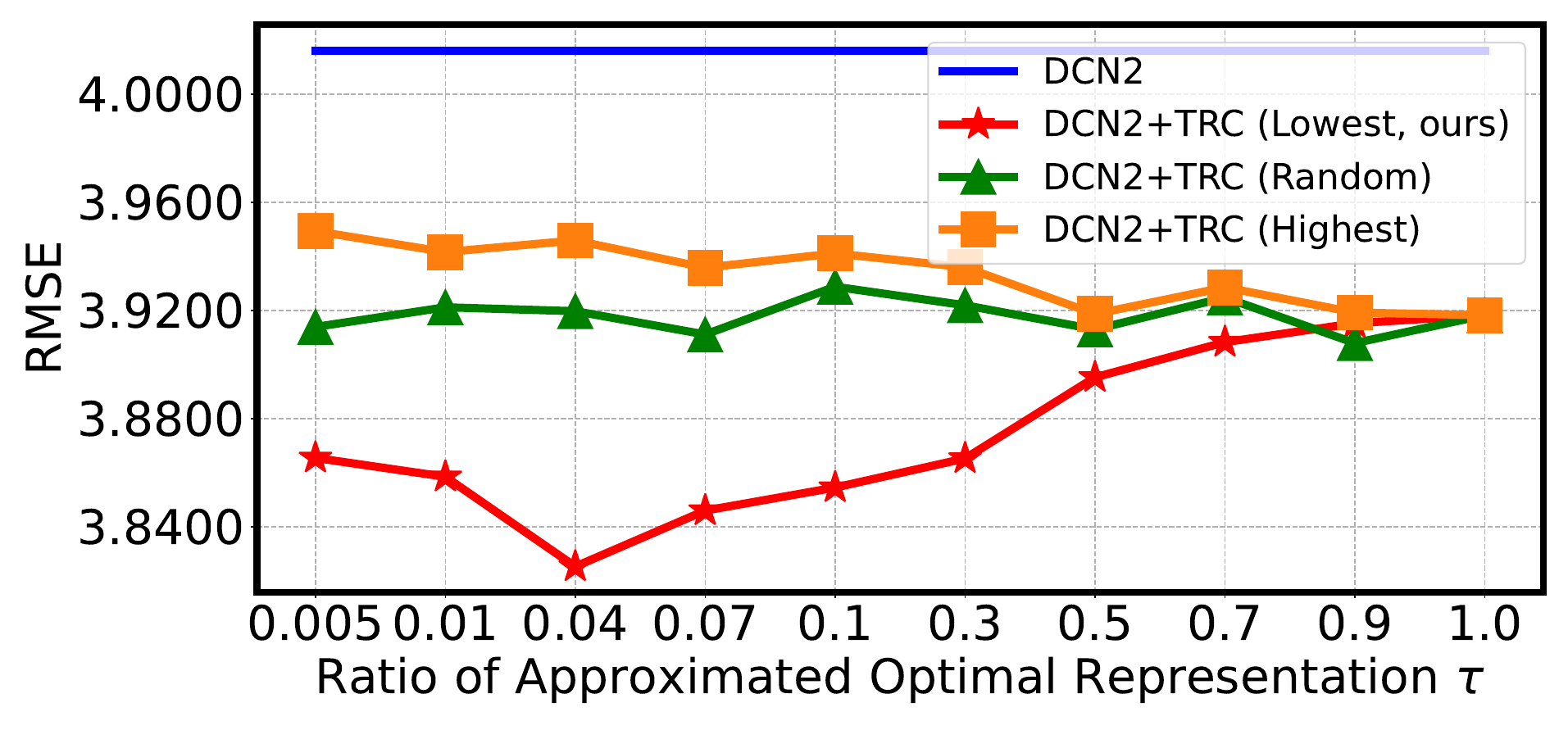}
\caption{\textcolor{blue}{Comparison with different approximated optimal representation selection strategies across a wide range of threshold $\tau$ on CO dataset $\downarrow$.}} 
\label{fig:tau}
\end{figure}

\textcolor{blue}{\textbf{The Threshold of Selected Approximated Optimal Representations $\tau$.}
To verify the robustness of TRC to violations of proposed assumption and the choice of the threshold $\tau$, we compare TRC against two variants: (i) TRC (Random) that randomly selects samples and (ii) TRC (Highest) that selects samples with the highest gradient norms.
We provide these comparisons across a wide range of threshold $\tau$ in Fig.~\ref{fig:tau}.
Here, the backbone model is DCN2. 
Additional results are provided in Fig. 21 in Appendix I.
% Additional results are provided in Fig.~\ref{appendix:s3} in Appendix~\ref{appendix:sensitivity analysis}.
The results show that TRC benefits from selecting samples with the lowest gradient norms, as hypothesized.
When the assumption is violated, e.g., by selecting random samples or those with the highest gradient norms, TRC still improves the performance of the backbone model, albeit to a lesser extent.
When the threshold $\tau$ is relatively low (e.g., less than 0.3), the performance of TRC is stable. 
As $\tau$ increases further, samples with high gradient norms are also selected and the corresponding performance degrades slightly, but TRC could still enhance the performance of backbone model. 
Overall, TRC is robust to the proposed assumption and choice of the threshold $\tau$.
We suggest to select the samples with the lowest gradient norms using a conservative threshold (less than 0.3), which balances performance gain and computational cost.}

% \begin{figure}[h!]
% % \vspace{-1em}
%     \centering
    
%     % \hfill
%     \begin{subfigure}[b]{0.95\linewidth}
%         \centering
%         \includegraphics[width=\linewidth]{pictures/robustness/tau_new/combined_cycle_power_plant_DCN2_tau.pdf}
%         % \caption{CO dataset}
%     \end{subfigure}
%     % \hfill
%     \begin{subfigure}[b]{0.95\linewidth}
%         \centering
%         \includegraphics[width=\linewidth]{pictures/robustness/T_new/combined_cycle_power_plant_DCN2_T.pdf}
%         % \caption{CO dataset}
%     \end{subfigure}
%     \caption{\textcolor{blue}{Sensitivity analysis w.r.t. the threshold $\tau$ that defines the representations $\mathcal{Z}_o$ and the number of embedding vectors $T$, the weight for $\mathcal{L}_{orth}$.}}
%     \label{fig:sensitivity}
% \end{figure}

\textbf{Additional Experiments.}
\textcolor{blue}{
% We incorporate the sensitivity analysis for the number of embedding vectors in Fig.~\ref{fig:T} with detailed discussion in Appendix~\ref{appendix:theoretical analysis and discussion}.
A representative case of the sensitivity analysis on the number of embedding vectors is provided in Fig.~\ref{fig:T}. 
Additional results are presented in Fig.22 of Appendix I, and a more in-depth discussion is provided in Appendix J.
% Additional results are presented in Fig.~\ref{appendix:s4} of Appendix~\ref{appendix:sensitivity analysis}, and a more in-depth discussion is provided in Appendix~\ref{appendix:theoretical analysis and discussion}.
} 
Additional sensitivity analysis w.r.t. the weight of loss function $\mathcal{L}_{orth}$, and the perturbing times for constructing simulated sub-optimal representations in Inherent Shift Learning in Appendix I.
% Additional sensitivity analysis w.r.t. the weight of loss function $\mathcal{L}_{orth}$, and the perturbing times for constructing simulated sub-optimal representations in Inherent Shift Learning in Appendix~\ref{appendix:sensitivity analysis}.
The comparison between \method and tree-based methods is detailed in Table~\ref{table:tree}.
We also provide the visualizations of learned coefficients $r$ (Eq.~\ref{equ:coordinate-estimate}) for embedding vectors in Appendix G.
% We also provide the visualizations of learned coefficients $r$ (Eq.~\ref{equ:coordinate-estimate}) for embedding vectors in Appendix~\ref{appendix:visualization}.
\textcolor{blue}{The discussion about how well the computed shift by shift estimator can model the real inherent shift is provided in Appendix K, and the limitation discussion about \textbf{Assumption 1} is presented in Appendix L.}
% \textcolor{blue}{The discussion about how well the computed shift by shift estimator can model the real inherent shift is provided in Appendix~\ref{appendix:simulated shift and inherent shift}, and the limitation discussion about \textbf{Assumption 1} is presented in Appendix~\ref{appendix:limitations of assumption 1}.}

\begin{figure}[!t]
\centering
\includegraphics[width=0.95\linewidth]{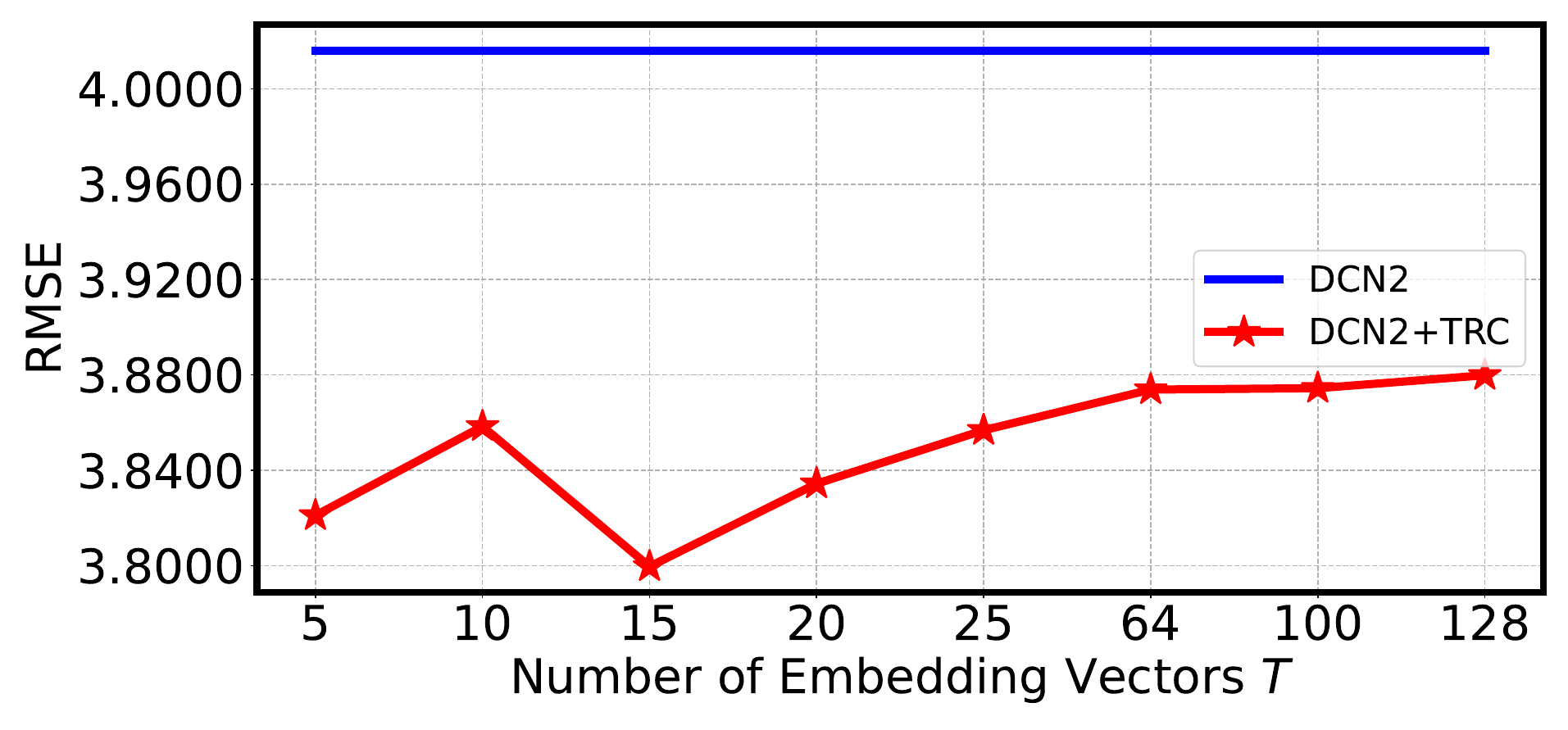}
\caption{\textcolor{blue}{The performance of \method with varying number of embedding vectors on CO dataset $\downarrow$.}}
\label{fig:T}
\end{figure}

\begin{table}[t]
\centering
\footnotesize
\caption{Comparison with tree-based methods.}
\label{table:tree}
\setlength{\tabcolsep}{2.9mm}{
\begin{tabular}{ccccc}
\toprule
& DI $\downarrow$ & QS $\downarrow$ & PO $\downarrow$ & AU $\uparrow$ \\
\midrule
\textcolor{blue}{FT-Transformer}+\textbf{TRC} & \textbf{543.653} & \textbf{0.780} & \textbf{2.728} & \textbf{0.884} \\
\midrule
CatBoost~\cite{prokhorenkova2018catboost} & 546.827 & 0.817 & 4.641 & 0.855 \\
\midrule
XGBoost~\cite{chen2016xgboost} & 559.680 & 0.820 & 4.936 & 0.862 \\
\bottomrule
\end{tabular}}
\end{table}

\section{Conclusion}
\label{sec:conclusion}
In this paper, we introduce a novel deep Tabular Representation Corrector, TRC, to enhance trained tabular models through two specific tasks without altering their parameters in a model-agnostic way. 
% Unlike existing in-learning or pre-learning approaches, post-learning offers a cost-effective, parameter-efficient technique to enhance representations for deep tabular models.
% Unlike existing approaches that directly intervene in the learning process of predictive models (in-learning) or rely on prior knowledge to design pretext tasks for pre-training (pre-learning), post-learning provides a cost-effective, parameter-efficient technique to enhance representations for deep tabular models.
Specifically, \method could solve two inherent representation issues, i.e., \textit{representation shift} and \textit{representation redundancy}.
We propose two tasks, i.e., (i) \textit{Tabular Representation Re-estimation}, that involves training a shift estimator to calculate the inherent shift of tabular representations to subsequently mitigate it, thereby re-estimating the representations, and (ii) \textit{Tabular Space Mapping}, that transforms the re-estimated representations into a light-embedding vector space while preserves crucial predictive information to minimize redundancy.
The empirical results on various real world tasks demonstrated the effectiveness of \method for tabular deep learning.
Our work can shed some light on developing better algorithms for similar tasks.
% Although the primary focus of this work is tabular data, the flexibility of \method opens up exciting possibilities for extensions. 
% % For example, \method can be extended to other domains like time series and graphs, leaving further exploration for future work.
% For example, \method can be extended to other domains such as time series, graph and so on.
% We leave the extensions of \method as a future work.

% \section*{Acknowledgment}
% The authors would like to thank the anonymous referees for their valuable comments. In this work, Hangting Ye, Peng Wang, Dandan Guo and Yi Chang are supported by the National Key R\&D Program of China under Grant (No. 2023YFF0905400) and the National Natural Science Foundation of China (No. U2341229, No. 62306125, No. 623B2043).
% % The preferred spelling of the word ``acknowledgment'' in America is without 
% % an ``e'' after the ``g''. Avoid the stilted expression ``one of us (R. B. 
% % G.) thanks $\ldots$''. Instead, try ``R. B. G. thanks$\ldots$''. Put sponsor 
% % acknowledgments in the unnumbered footnote on the first page.

% \balance

\bibliographystyle{IEEEtran}
\bibliography{reference}

\clearpage
\appendix
\section{Appendix / supplemental material}

\textcolor{blue}{
\subsection{Nomenclature}
\label{appendix:NOMENCLATURE}
For clarity, we summarize the key notations used throughout the paper in Table~\ref{appendix:notation}. This includes the symbols, their definitions, and dimensionalities where applicable.}

\begin{table}[h!]
\caption{\textcolor{blue}{Notation.}}
\label{appendix:notation}
\scriptsize
\centering
\resizebox{\linewidth}{!}{
\begin{tabular}{ll}
\textbf{Symbols} & \textbf{Description}  \\
$x, y$ & Observed sample and the corresponding label.  \\
$N$ & Number of observed samples in a dataset.  \\
$\mathcal{D}_{train}$, $\mathcal{D}_{val}$, $\mathcal{D}_{test}$ & Training set, validation set, test set. \\
$F(\cdot;\theta)$ & Deep tabular model.  \\
$G_f(\cdot;\theta_f)$ & Existing deep tabular backbone. \\
$G_h(\cdot;\theta_h)$ & Prediction head. \\ 
$z$ & Representation extracted by $G_f(\cdot;\theta_f)$. \\
$D$ & Dimensionality of $z$. \\
$\mathcal{L}$ & Loss function. \\
$\mathcal{Z}=\{z_i\}_{i=1}^N\in \mathbb{R}^{N\times D}$ & Representations of $N$ samples. \\
$\text{SVE}$ & Singular Value Entropy. \\
$\mathcal{Z}_o$ & Approximated optimal representations. \\
$\phi$ & Shift estimator. \\
$\tilde{\mathcal{Z}}_o$ & Simulated sub-optimal representations derived from $\mathcal{Z}_o$. \\
$\Delta$ & $\tilde{\mathcal{Z}}_o$'s corresponding shift. \\
$\Phi(\cdot;\theta_\phi)$ & Re-estimation function. \\
$\mathcal{B} = \{\beta_t\}_{t=1}^T \in \mathbb{R}^{T\times D}$ & Embedding vectors. \\
$T$ & Number of embedding vectors. \\
$\beta_t$ & $t$-th embedding vector. \\
LE-Space & Light Embedding Space consists of embedding vectors. \\
$s(\cdot;\theta_s)$ & Coordinate estimator. \\
$r$ & Coordinate of representation in the LE-Space. \\ 
$A \in [0, 1]^{T\times T}$ & Cosine similarity between embedding vectors. \\
\end{tabular}}
% \vspace{30em}
\end{table}

% \begin{table*}[b!]
% \caption{\textcolor{blue}{Notation.}}
% \label{appendix:notation}
% % \scriptsize
% \centering
% \begin{tabular}{ll}
% \textbf{Symbols} & \textbf{Description}  \\
% $x, y$ & Observed sample and the corresponding label.  \\
% $N$ & Number of observed samples in a dataset.  \\
% $\mathcal{D}_{train}$, $\mathcal{D}_{val}$, $\mathcal{D}_{test}$ & Training set, validation set, test set. \\
% $F(\cdot;\theta)$ & Deep tabular model.  \\
% $G_f(\cdot;\theta_f)$ & Existing deep tabular backbone. \\
% $G_h(\cdot;\theta_h)$ & Prediction head. \\ 
% $z$ & Representation extracted by $G_f(\cdot;\theta_f)$. \\
% $D$ & Dimensionality of $z$. \\
% $\mathcal{L}$ & Loss function. \\
% $\mathcal{Z}=\{z_i\}_{i=1}^N\in \mathbb{R}^{N\times D}$ & Representations of $N$ samples. \\
% $\text{SVE}$ & Singular Value Entropy. \\
% $\mathcal{Z}_o$ & Approximated optimal representations. \\
% $\phi$ & Shift estimator. \\
% $\tilde{\mathcal{Z}}_o$ & Simulated sub-optimal representations derived from $\mathcal{Z}_o$. \\
% $\Delta$ & $\tilde{\mathcal{Z}}_o$'s corresponding shift. \\
% $\Phi(\cdot;\theta_\phi)$ & Re-estimation function. \\
% $\mathcal{B} = \{\beta_t\}_{t=1}^T \in \mathbb{R}^{T\times D}$ & Embedding vectors. \\
% $T$ & Number of embedding vectors. \\
% $\beta_t$ & $t$-th embedding vector. \\
% LE-Space & Light Embedding Space consists of embedding vectors. \\
% $s(\cdot;\theta_s)$ & Coordinate estimator. \\
% $r$ & Coordinate of representation in the LE-Space. \\ 
% $A \in [0, 1]^{T\times T}$ & Cosine similarity between embedding vectors. \\

% \end{tabular}
% \vspace{30em}
% \end{table*}

% \clearpage
\subsection{Ablation Study}
\label{appendix:ablation study}
We further conduct ablation study to demonstrate the effectiveness of key components of \method.
Specifically, we denote the task of \underline{t}abular representation \underline{r}e-estimation as ``TR'', and for the task of tabular space mapping, we denote the process of \underline{s}pace \underline{c}ompression as ``SC'' and the strategy for \underline{d}iversifying the \underline{e}mbedding vectors as ``DE''.
Given that the strategy for diversifying the embedding vectors should be applied after the process of space compression, we conduct a comparative analysis between the backbone integrated with \method and its five variants in Table~\ref{appendix:ablation}.

\begin{table}[h]
    \caption{\textcolor{blue}{Analysis on the effects of different components of \method.} The best results are highlighted in bold, and the second best results are underscored. \textcolor{blue}{This
table serves as an extension of Table III in manuscript.}}
%     \caption{\textcolor{blue}{Analysis on the effects of different components of \method.} The best results are highlighted in bold, and the second best results are underscored. \textcolor{blue}{This
% table serves as an extension of Table~\ref{tab:ablation} in manuscript.}}
    \label{appendix:ablation}
    \scriptsize
    \centering
    \resizebox{\linewidth}{!}{
    \begin{tabular}{ccccccccccccc}
\toprule
     & TR & SC & DE & CA $\downarrow$ & CO $\downarrow$ & DI $\downarrow$ & QS $\downarrow$ & SU $\downarrow$ & PO $\downarrow$ & AD $\uparrow$ & AU $\uparrow$ & GE $\uparrow$ \\
    \midrule
    \multirow{6}{*}{MLP} &   &   &   & 0.505 & 3.961 & 564.373 & 0.875 & 10.379 & 10.892 & 0.856 & 0.87 & \underline{0.578} \\
     & \checkmark &   &   & \textbf{0.502} & 3.916 & 563.649 & 0.868 & \underline{10.328} & 10.605 & \textbf{0.858} & 0.87 & 0.571 \\
     &   & \checkmark &   & 0.505 & 3.92 & 564.141 & 0.845 & 10.349 & 10.631 & \textbf{0.858} & 0.862 & 0.568 \\
     & \checkmark & \checkmark &   & \textbf{0.502} & \underline{3.908} & \underline{558.52} & 0.852 & 10.341 & \underline{10.602} & \textbf{0.858} & 0.87 & 0.575 \\
     &   & \checkmark & \checkmark & 0.505 & 3.922 & 565.129 & \underline{0.841} & 10.363 & 10.622 & \textbf{0.858} & \underline{0.877} & 0.577 \\
     & \checkmark & \checkmark & \checkmark & \textbf{0.502} & \textbf{3.899} & \textbf{558.33} & \textbf{0.825} & \textbf{10.326} & \textbf{10.593} & \textbf{0.858} & \textbf{0.891} & \textbf{0.58} \\ 
     \midrule
    \multirow{6}{*}{DCN2} &   &   &   & 0.495 & 4.016 & 591.958 & 1.027 & 10.674 & 8.805 & 0.856 & 0.841 & 0.564 \\
     & \checkmark &   &   & \textbf{0.493} & 3.912 & 579.461 & 0.941 & \textbf{10.521} & 8.795 & \textbf{0.858} & 0.848 & \underline{0.575} \\
     &   & \checkmark &   & 0.498 & 4.016 & 580.454 & \underline{0.86} & 10.634 & \textbf{8.757} & \textbf{0.858} & 0.848 & 0.565 \\
     & \checkmark & \checkmark &   & 0.495 & \textbf{3.844} & \underline{574.712} & 0.884 & 10.727 & 8.782 & \textbf{0.858} & 0.841 & 0.563 \\
     &   & \checkmark & \checkmark & 0.496 & 4.023 & 580.247 & \textbf{0.853} & 10.634 & \underline{8.764} & \textbf{0.858} & \underline{0.855} & 0.567 \\
     & \checkmark & \checkmark & \checkmark & \textbf{0.493} & \underline{3.858} & \textbf{574.453} & 0.879 & \underline{10.584} & 8.79 & 0.857 & \textbf{0.862} & \textbf{0.584} \\
     \midrule
    \multirow{6}{*}{SNN} &   &   &   & 0.896 & 11.789 & 1530.293 & 1.018 & 25.498 & 18.517 & \textbf{0.847} & 0.848 & 0.55 \\
     & \checkmark &   &   & 0.88 & 11.369 & 1226.754 & 0.921 & 25.265 & 15.86 & \textbf{0.847} & 0.855 & \underline{0.572} \\
     &   & \checkmark &   & 0.748 & \textbf{5.701} & 783.015 & 0.889 & 17.738 & 12.001 & \textbf{0.847} & 0.833 & 0.543 \\
     & \checkmark & \checkmark &   & 0.743 & 6.889 & \underline{696.473} & \underline{0.886} & \textbf{15.512} & \underline{11.422} & \textbf{0.847} & \underline{0.862} & 0.561 \\
     &   & \checkmark & \checkmark & \underline{0.742} & 7.306 & 760.66 & 0.891 & 17.711 & 12.002 & \textbf{0.847} & 0.855 & 0.546 \\
     & \checkmark & \checkmark & \checkmark & \textbf{0.699} & \underline{6.88} & \textbf{693.369} & \textbf{0.87} & \underline{15.537} & \textbf{11.202} & \textbf{0.847} & \textbf{0.877} & \textbf{0.573} \\
     \midrule
    \multirow{6}{*}{ResNet} &   &   &   & 0.517 & 3.982 & \underline{606.282} & 0.872 & 11.163 & 10.812 & 0.847 & 0.87 & \underline{0.587} \\
     & \checkmark &   &   & \underline{0.512} & 3.925 & \textbf{597.975} & 0.866 & 10.94 & 10.405 & \underline{0.849} & 0.862 & 0.58 \\
     &   & \checkmark &   & 0.513 & 3.96 & 612.993 & \textbf{0.83} & 10.944 & \textbf{10.134} & \underline{0.849} & 0.87 & 0.576 \\
     & \checkmark & \checkmark &   & \underline{0.512} & \underline{3.922} & 606.482 & 0.85 & \underline{10.755} & 10.15 & \textbf{0.85} & 0.862 & 0.578 \\
     &   & \checkmark & \checkmark & 0.513 & 3.96 & 617.098 & \underline{0.841} & 10.951 & \underline{10.137} & \underline{0.849} & \textbf{0.877} & 0.586 \\
     & \checkmark & \checkmark & \checkmark & \textbf{0.508} & \textbf{3.919} & 615.736 & 0.845 & \textbf{10.72} & 10.205 & \underline{0.849} & \textbf{0.877} & \textbf{0.597} \\
    \midrule
    \multirow{6}{*}{AutoInt} &   &   &   & 0.487 & 4.02 & 562.169 & 0.86 & 11.193 & 6.45 & 0.854 & 0.848 & 0.598 \\
     & \checkmark &   &   & 0.481 & 4.018 & 562.677 & 0.849 & 11.089 & 6.19 & \textbf{0.858} & 0.87 & \underline{0.605} \\
     &   & \checkmark &   & 0.48 & 4.001 & 559.535 & 0.835 & 10.86 & \underline{5.694} & 0.857 & 0.855 & 0.591 \\
     & \checkmark & \checkmark &   & \textbf{0.477} & \underline{3.971} & \textbf{555.787} & \textbf{0.828} & 10.972 & 5.88 & 0.856 & \textbf{0.891} & 0.59 \\
     &   & \checkmark & \checkmark & 0.48 & 4.003 & 559.341 & 0.842 & \underline{10.856} & \textbf{5.685} & 0.856 & 0.87 & 0.595 \\
     & \checkmark & \checkmark & \checkmark & \textbf{0.477} & \textbf{3.96} & \underline{557.423} & \underline{0.829} & \textbf{10.826} & 5.885 & \textbf{0.858} & \textbf{0.891} & \textbf{0.606} \\
    \midrule
    \multirow{6}{*}{FT-Transformer} &   &   &   & 0.469 & 3.709 & 551.19 & 0.823 & 10.41 & 2.919 & 0.858 & 0.848 & 0.611 \\
     & \checkmark &   &   & 0.463 & 3.685 & 549.075 & 0.819 & 10.295 & \textbf{2.71} & 0.861 & \underline{0.862} & 0.61 \\
     &   & \checkmark &   & 0.464 & 3.668 & 549.684 & 0.812 & 10.338 & 2.759 & 0.861 & 0.855 & 0.62 \\
     & \checkmark & \checkmark &   & \textbf{0.462} & \underline{3.659} & \textbf{543.504} & 0.806 & \underline{10.272} & 2.76 & \textbf{0.862} & \underline{0.862} & \underline{0.622} \\
     &   & \checkmark & \checkmark & 0.464 & 3.668 & 549.923 & \underline{0.797} & 10.342 & 2.763 & 0.861 & 0.855 & 0.618 \\
     & \checkmark & \checkmark & \checkmark & \textbf{0.462} & \textbf{3.648} & \underline{543.653} & \textbf{0.78} & \textbf{10.223} & \underline{2.728} & \textbf{0.862} & \textbf{0.884} & \textbf{0.624} \\
      \midrule
    \multirow{6}{*}{SCARF} &   &   &   & 3.856 & 579.61 & 0.863 & \textbf{0.52} & \underline{7.292} & 10.24 & 0.858 & 0.87 & \textbf{0.593} \\
     & \checkmark &   &   & \underline{3.839} & 585.659 & 0.864 & \textbf{0.52} & 7.584 & 10.296 & 0.858 & \textbf{0.884} & 0.584 \\
     &   & \checkmark &   & 3.841 & 588.05 & \underline{0.829} & 0.521 & 7.485 & \underline{10.235} & 0.858 & 0.877 & 0.586 \\
     & \checkmark & \checkmark &   & \textbf{3.827} & \textbf{571.684} & 0.837 & \textbf{0.52} & \textbf{7.212} & 10.292 & 0.858 & \textbf{0.884} & 0.587 \\
     &   & \checkmark & \checkmark & 3.841 & \underline{572.493} & 0.835 & 0.523 & 7.325 & \textbf{10.201} & \textbf{0.859} & 0.87 & 0.584 \\
     & \checkmark & \checkmark & \checkmark & 3.847 & 577.803 & \textbf{0.808} & \textbf{0.52} & 7.531 & 10.355 & \textbf{0.859} & \textbf{0.884} & \underline{0.589} \\
    \midrule
    \multirow{6}{*}{SAINT} &   &   &   & 0.508 & 4.022 & 597.207 & 0.827 & 13.095 & 4.415 & \underline{0.857} & 0.87 & 0.549 \\
     & \checkmark &   &   & \textbf{0.491} & 3.911 & 559.944 & 0.822 & 12.868 & 4.403 & \underline{0.857} & 0.862 & 0.579 \\
     &   & \checkmark &   & 0.495 & \textbf{3.869} & 565.402 & \textbf{0.81} & 12.813 & 4.358 & \underline{0.857} & 0.87 & 0.581 \\
     & \checkmark & \checkmark &   & 0.492 & 3.911 & \textbf{556.744} & 0.823 & 12.85 & \textbf{4.329} & 0.856 & \textbf{0.884} & \underline{0.586} \\
     &   & \checkmark & \checkmark & 0.495 & \textbf{3.869} & 565.393 & 0.821 & \underline{12.81} & 4.365 & \underline{0.857} & 0.87 & 0.584 \\
     & \checkmark & \checkmark & \checkmark & \textbf{0.491} & 3.903 & \underline{557.951} & \underline{0.818} & \textbf{12.443} & \underline{4.335} & \textbf{0.858} & \textbf{0.884} & \textbf{0.595} \\
    
    \midrule
    \multirow{6}{*}{VIME} &   &   &   & 0.679 & 5.218 & 945.238 & 1.018 & 15.645 & 10.914 & 0.768 & 0.812 & 0.473 \\
     & \checkmark &   &   & \underline{0.643} & 4.378 & 707.749 & 0.952 & 15.486 & \underline{7.365} & \textbf{0.847} & 0.87 & 0.473 \\
     &   & \checkmark &   & \textbf{0.636} & \textbf{4.174} & \underline{629.678} & 0.96 & 15.537 & \textbf{6.861} & 0.846 & 0.87 & 0.472 \\
     & \checkmark & \checkmark &   & 0.645 & \underline{4.342} & 638.677 & 0.957 & \underline{15.073} & 7.443 & \textbf{0.847} & \underline{0.877} & 0.472 \\
     &   & \checkmark & \checkmark & 0.648 & 4.378 & 644.659 & \textbf{0.949} & 15.544 & 7.799 & 0.846 & \underline{0.877} & \underline{0.476} \\
     & \checkmark & \checkmark & \checkmark & 0.645 & 4.371 & \textbf{612.454} & \underline{0.95} & \textbf{15.028} & 7.489 & 0.846 & \textbf{0.884} & \textbf{0.479} \\
\bottomrule
    \end{tabular}}
    % \vspace{-1em}
\end{table}

% \clearpage
\subsection{Comparison with Deeper Backbone}

Besides, \method is appended to the output of the backbone $G_f(\cdot;\theta_f)$ while keeping the parameters fixed. 
We compare the performance of the backbone coupled with \method against backbone with deeper layers (additional 3 layers) in Table~\ref{appendix:deeper}.
Increasing the number of layers in backbone model does not necessarily lead to improved performance; in some cases, it even results in performance degradation. Furthermore, the comparison between deeper backbone and backbone augmented with \method highlights that the well-designed tasks can significantly enhance model representations, which is not simply achieved by increasing model depth.

\label{appendix:deeper backbone}
\begin{table}[h!]
    \centering
    \caption{Performance of backbone w/ TRC and deeper backbone. The depths of backbone layers are increased only for ``deeper backbone''. \textcolor{blue}{This
table serves as an extension of Table IV in manuscript.}}
%     \caption{Performance of backbone w/ TRC and deeper backbone. The depths of backbone layers are increased only for ``deeper backbone''. \textcolor{blue}{This
% table serves as an extension of Table~\ref{tab:deeper comparison} in manuscript.}}
    \label{appendix:deeper}
    \scriptsize
    \resizebox{\linewidth}{!}{
    \begin{tabular}{cccccccccc}
    \toprule
 & CO $\downarrow$ & DI $\downarrow$ & QS $\downarrow$ & CA $\downarrow$ & PO $\downarrow$ & SU $\downarrow$ & AD $\uparrow$ & AU $\uparrow$ & GE $\uparrow$ \\
\midrule
MLP & 3.961 & 564.373 & 0.875 & 0.505 & 10.892 & 10.379 & 0.856 & 0.870 & 0.578 \\
deeper MLP & 3.922 & 562.888 & 0.851 & 0.510 & 11.583 & \textbf{10.271} & 0.856 & 0.862 & 0.572 \\
MLP+\textbf{TRC} & \textbf{3.899} & \textbf{558.330} & \textbf{0.825} & \textbf{0.502} & \textbf{10.593} & 10.326 & \textbf{0.858} & \textbf{0.891} & \textbf{0.580} \\
\midrule
DCN2 & 4.016 & 591.958 & 1.027 & 0.495 & 8.805 & 10.674 & 0.856 & 0.841 & 0.564 \\
deeper DCN2 & 3.872 & 575.765 & 0.904 & \textbf{0.486} & 9.165 & 10.675 & \textbf{0.858} & \textbf{0.862} & 0.575 \\
DCN2+\textbf{TRC} & \textbf{3.858} & \textbf{574.453} & \textbf{0.879} & 0.493 & \textbf{8.790} & \textbf{10.584} & 0.857 & \textbf{0.862} & \textbf{0.584} \\
\midrule
SNN & 11.789 & 1530.293 & 1.018 & 0.896 & 18.517 & 25.498 & \textbf{0.847} & 0.848 & 0.550 \\
deeper SNN & 10.614 & 1042.509 & 1.121 & 0.851 & 14.460 & \textbf{15.039} & \textbf{0.847} & 0.833 & 0.553 \\
SNN+\textbf{TRC} & \textbf{6.880} & \textbf{693.369} & \textbf{0.870} & \textbf{0.699} & \textbf{11.202} & 15.537 & \textbf{0.847} & \textbf{0.877} & \textbf{0.573} \\
\midrule
ResNet & 3.982 & \textbf{606.282} & 0.872 & 0.517 & 10.812 & 11.163 & 0.847 & 0.870 & 0.587 \\
deeper ResNet & 3.943 & 657.823 & 0.847 & 0.512 & 10.628 & 11.008 & \textbf{0.850} & 0.862 & \textbf{0.600} \\
ResNet+\textbf{TRC} & \textbf{3.919} & 615.736 & \textbf{0.845} & \textbf{0.508} & \textbf{10.205} & \textbf{10.720} & 0.849 & \textbf{0.877} & 0.597 \\
\midrule
AutoInt & 4.020 & 562.169 & 0.860 & 0.487 & 6.450 & 11.193 & 0.854 & 0.848 & 0.598 \\
deeper AutoInt & \textbf{3.899} & 562.383 & \textbf{0.815} & 0.488 & \textbf{5.201} & 11.265 & 0.857 & 0.862 & 0.577 \\
AutoInt+\textbf{TRC} & 3.960 & \textbf{557.423} & 0.829 & \textbf{0.477} & 5.885 & \textbf{10.826} & \textbf{0.858} & \textbf{0.891} & \textbf{0.606} \\
\midrule
FT-Transformer & 3.709 & 551.190 & 0.823 & 0.469 & 2.919 & 10.410 & 0.858 & 0.848 & 0.611 \\
deeper FT-Transformer & \textbf{3.604} & 559.363 & 0.786 & 0.464 & 2.850 & \textbf{10.142} & 0.860 & 0.870 & 0.623 \\
FT-Transformer+\textbf{TRC} & 3.648 & \textbf{543.653} & \textbf{0.780} & \textbf{0.462} & \textbf{2.728} & 10.223 & \textbf{0.862} & \textbf{0.884} & \textbf{0.624} \\
\midrule
SAINT & 4.022 & 597.207 & 0.827 & 0.508 & 4.415 & 13.095 & 0.857 & 0.870 & 0.549 \\
deeper SAINT& 3.912 & 585.623 & \textbf{0.806} & 0.501 & 41.349 & 34.681 & \textbf{0.860} & 0.877 & 0.572 \\
SAINT+\textbf{TRC} & \textbf{3.903} & \textbf{557.951} & 0.818 & \textbf{0.491} & \textbf{4.335} & \textbf{12.443} & 0.858 & \textbf{0.884} & \textbf{0.595} \\
\midrule
VIME & 5.218 & 945.238 & 1.018 & 0.679 & 10.914 & 15.645 & 0.768 & 0.812 & 0.473 \\
deeper VIME & \textbf{4.348} & 690.491 & 0.969 & 0.651 & \textbf{5.984} & 15.332 & 0.843 & 0.877 & 0.436 \\
VIME+\textbf{TRC} & 4.371 & \textbf{612.454} & \textbf{0.950} & \textbf{0.645} & 7.489 & \textbf{15.028} & \textbf{0.846} & \textbf{0.884} & \textbf{0.479} \\
\bottomrule
    \end{tabular}}
    % \label{tab:deeper comparison full}
\end{table}

\subsection{Training with Missing Value and Fewer Samples}
\label{appendix:missing value and few sample}
\begin{figure}[h!]
    \centering
    \begin{subfigure}[b]{0.47\linewidth}
        \centering
        \includegraphics[width=\linewidth]{pictures/scene/missing_ratio/diamonds_DCN2.pdf}
        \caption{DCN2 on DI}
    \end{subfigure}   
    \hfill
    \begin{subfigure}[b]{0.47\linewidth}
        \centering
        \includegraphics[width=\linewidth]{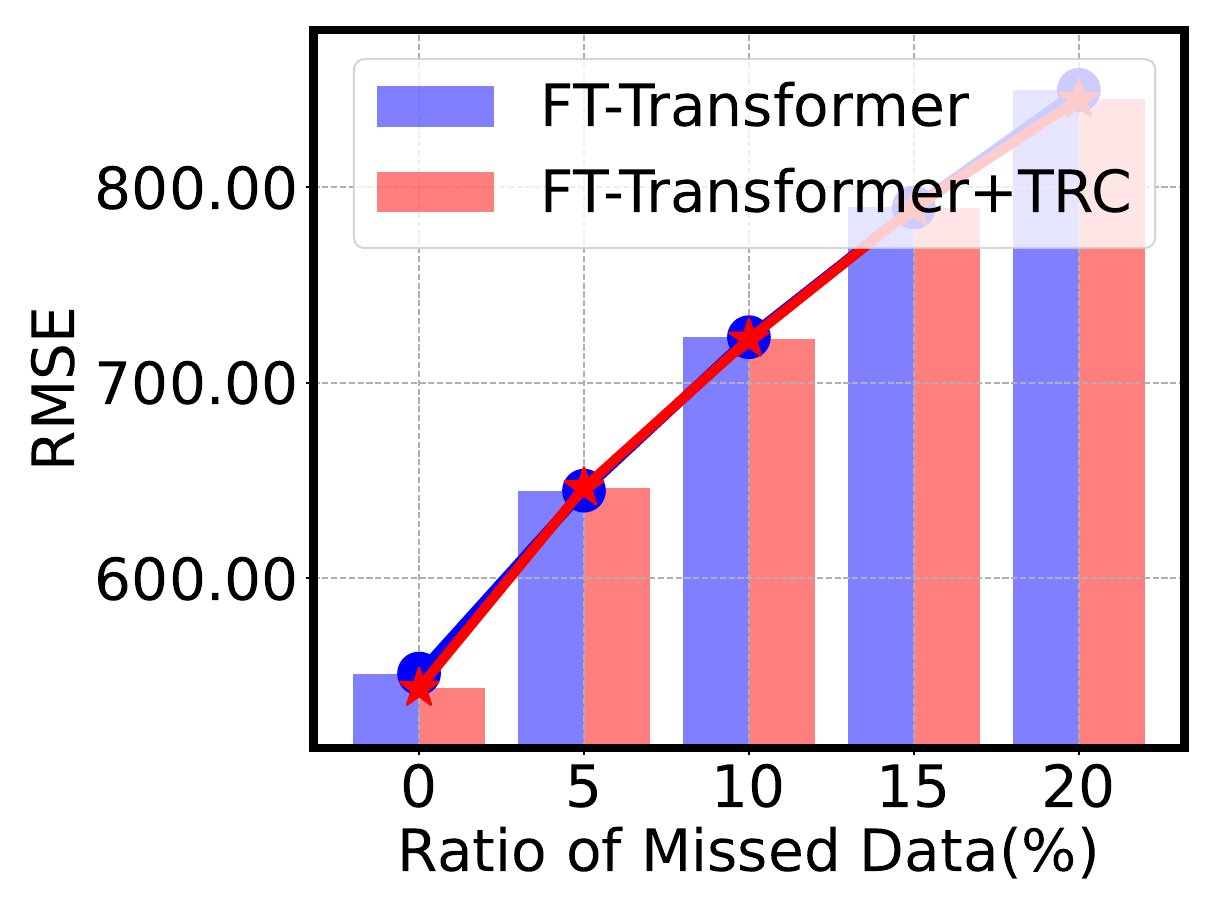}
        \caption{FT-Transformer on DI}
    \end{subfigure}   
    \hfill
    \begin{subfigure}[b]{0.47\linewidth}
        \centering
        \includegraphics[width=\linewidth]{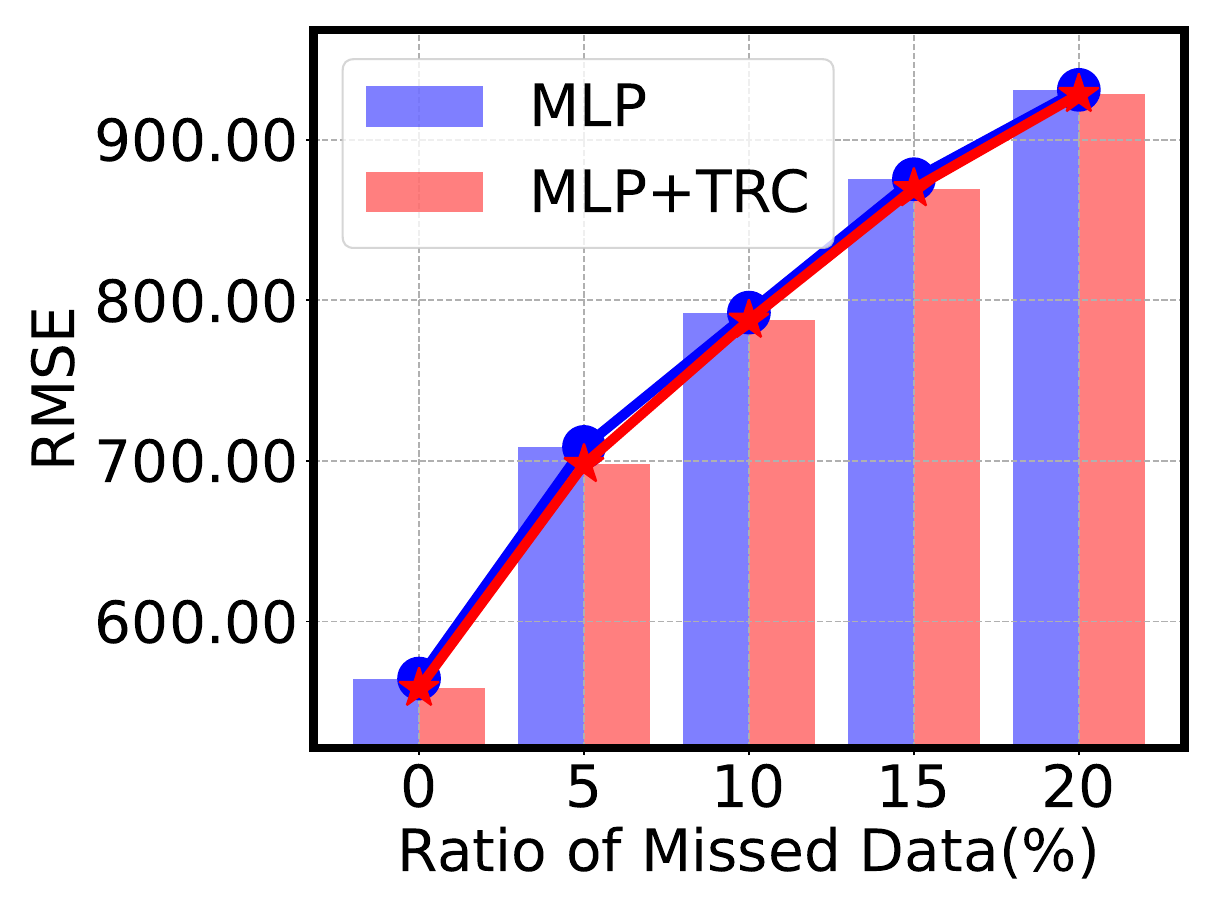}
        \caption{MLPs on DI}
    \end{subfigure}   
    \hfill
    \begin{subfigure}[b]{0.47\linewidth}
        \centering
        \includegraphics[width=\linewidth]{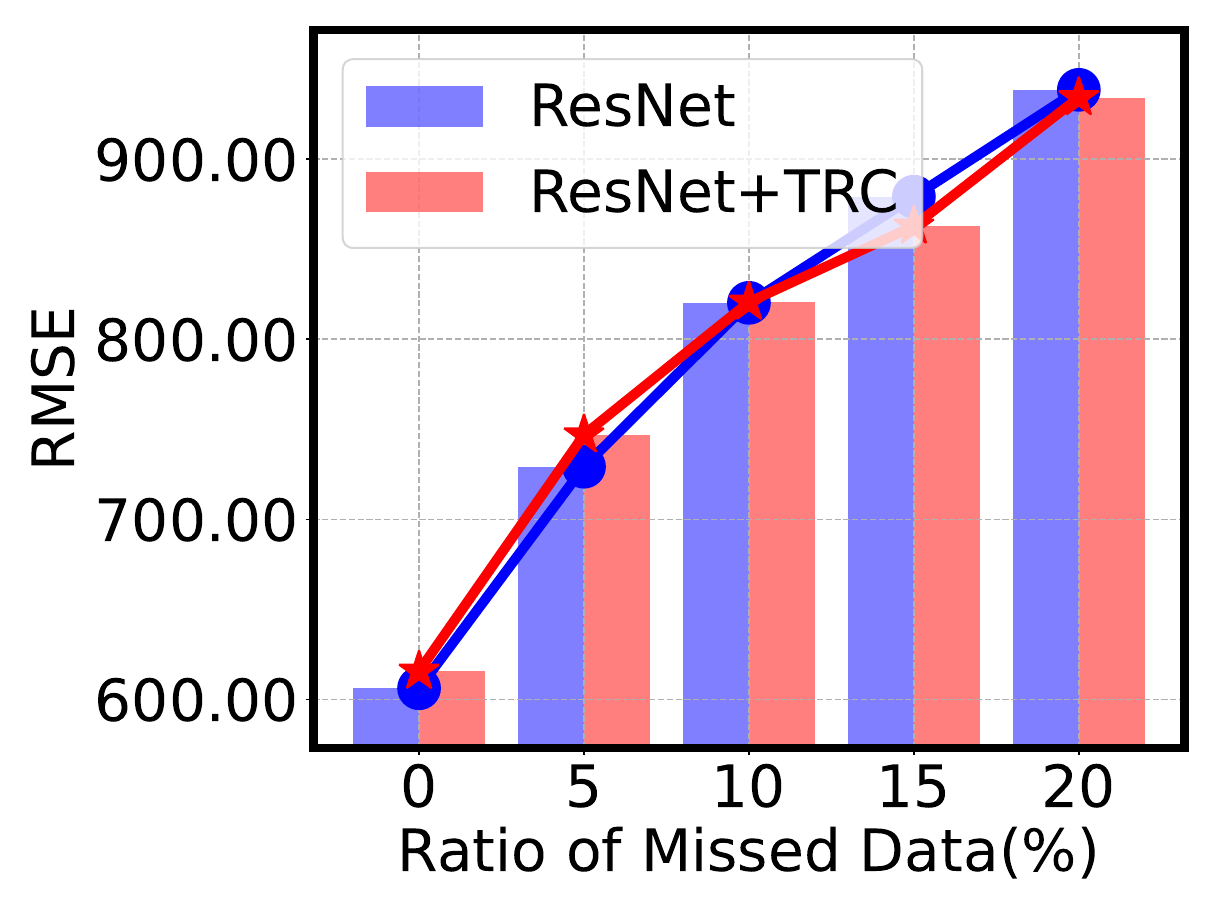}
        \caption{ResNet on DI}
    \end{subfigure}

    \begin{subfigure}[b]{0.47\linewidth}
        \centering
        \includegraphics[width=\linewidth]{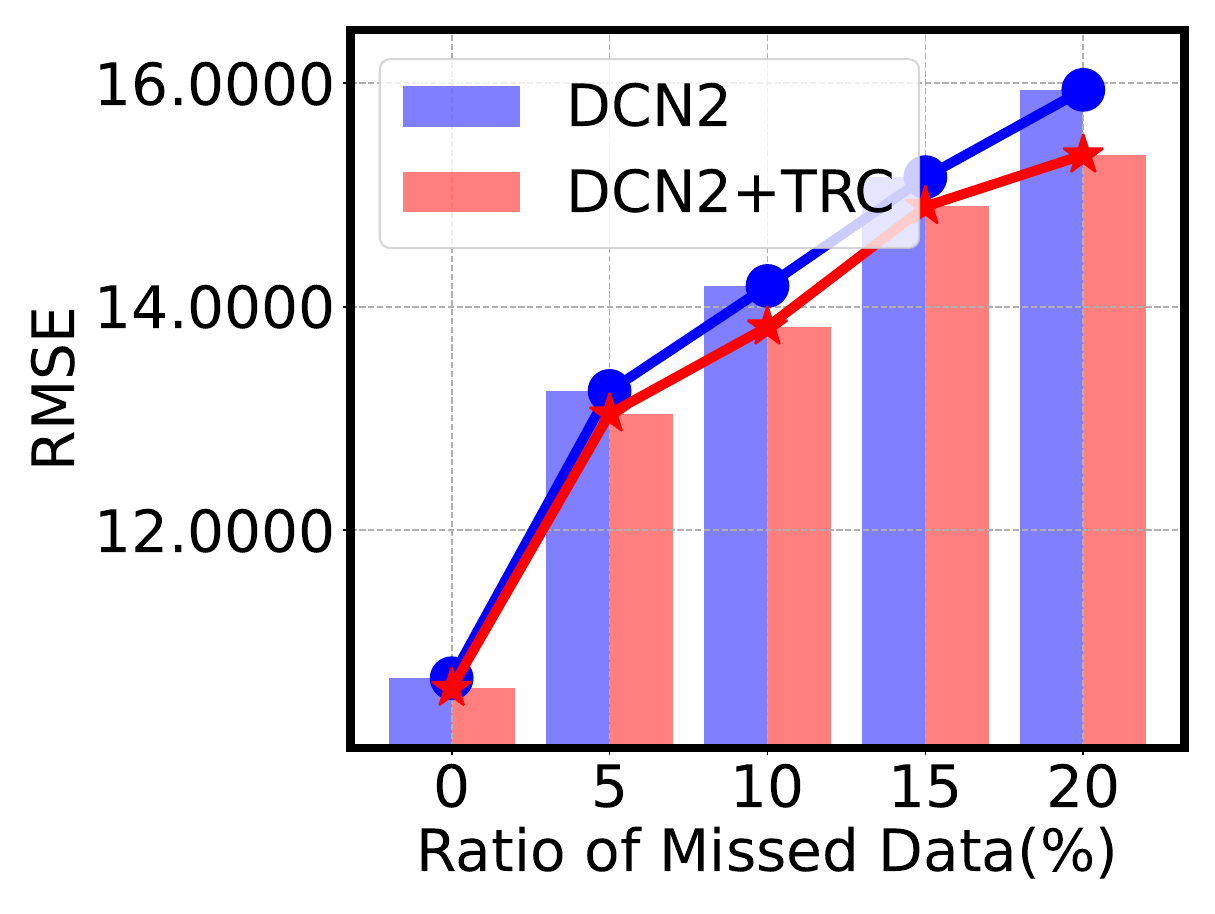}
        \caption{DCN2 on SU}
    \end{subfigure}   
    \hfill
    \begin{subfigure}[b]{0.47\linewidth}
        \centering
        \includegraphics[width=\linewidth]{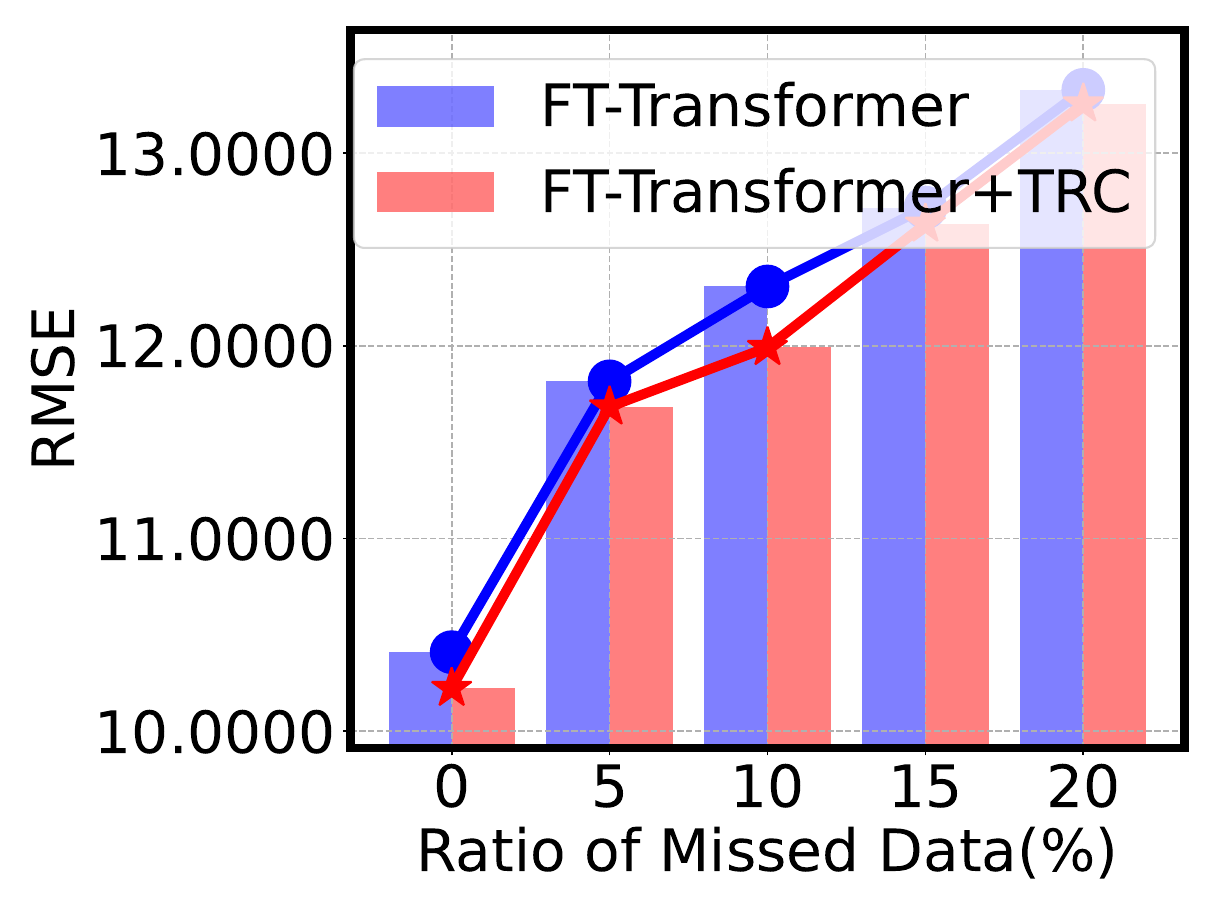}
        \caption{FT-Transformer on SU}
    \end{subfigure}   
    \hfill
    \begin{subfigure}[b]{0.47\linewidth}
        \centering
        \includegraphics[width=\linewidth]{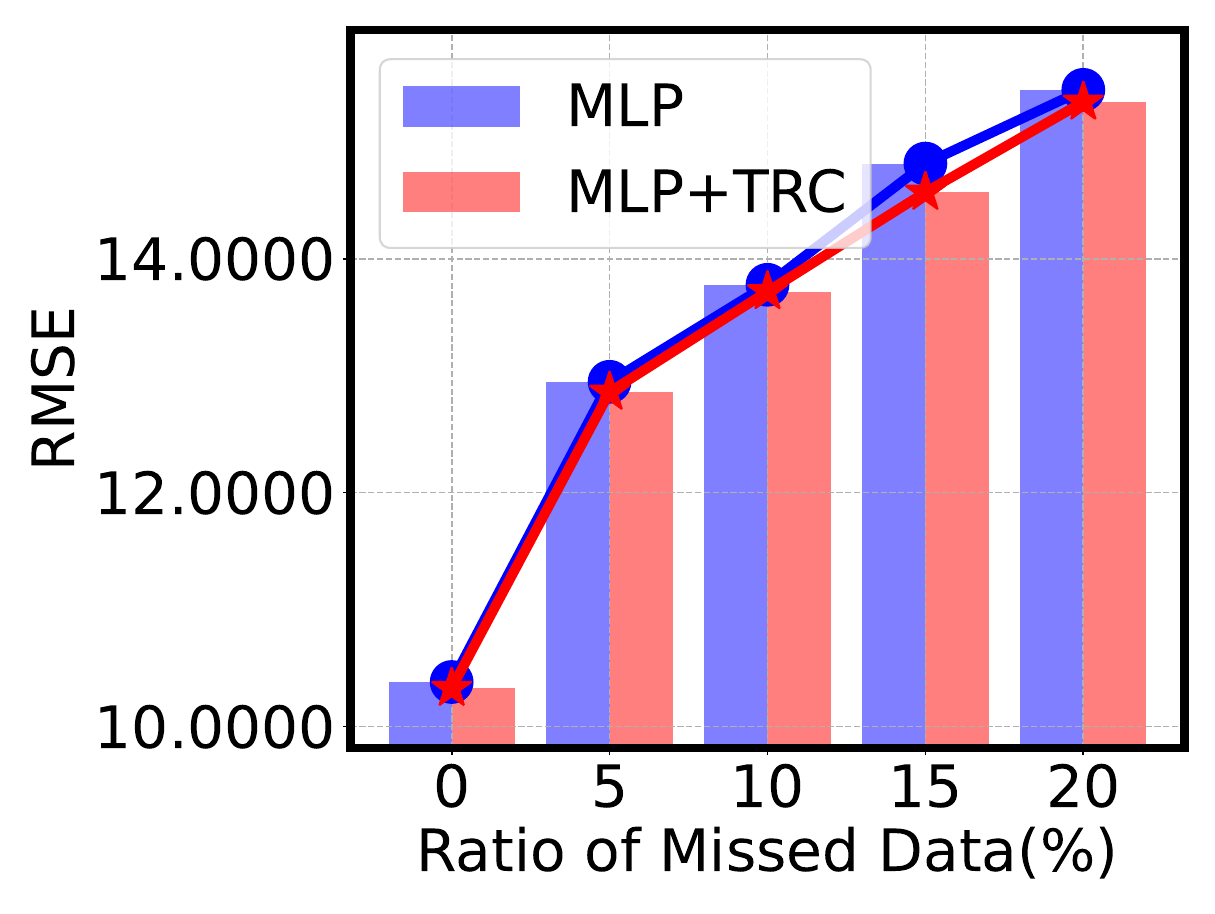}
        \caption{MLPs on SU}
    \end{subfigure}   
    \hfill
    \begin{subfigure}[b]{0.47\linewidth}
        \centering
        \includegraphics[width=\linewidth]{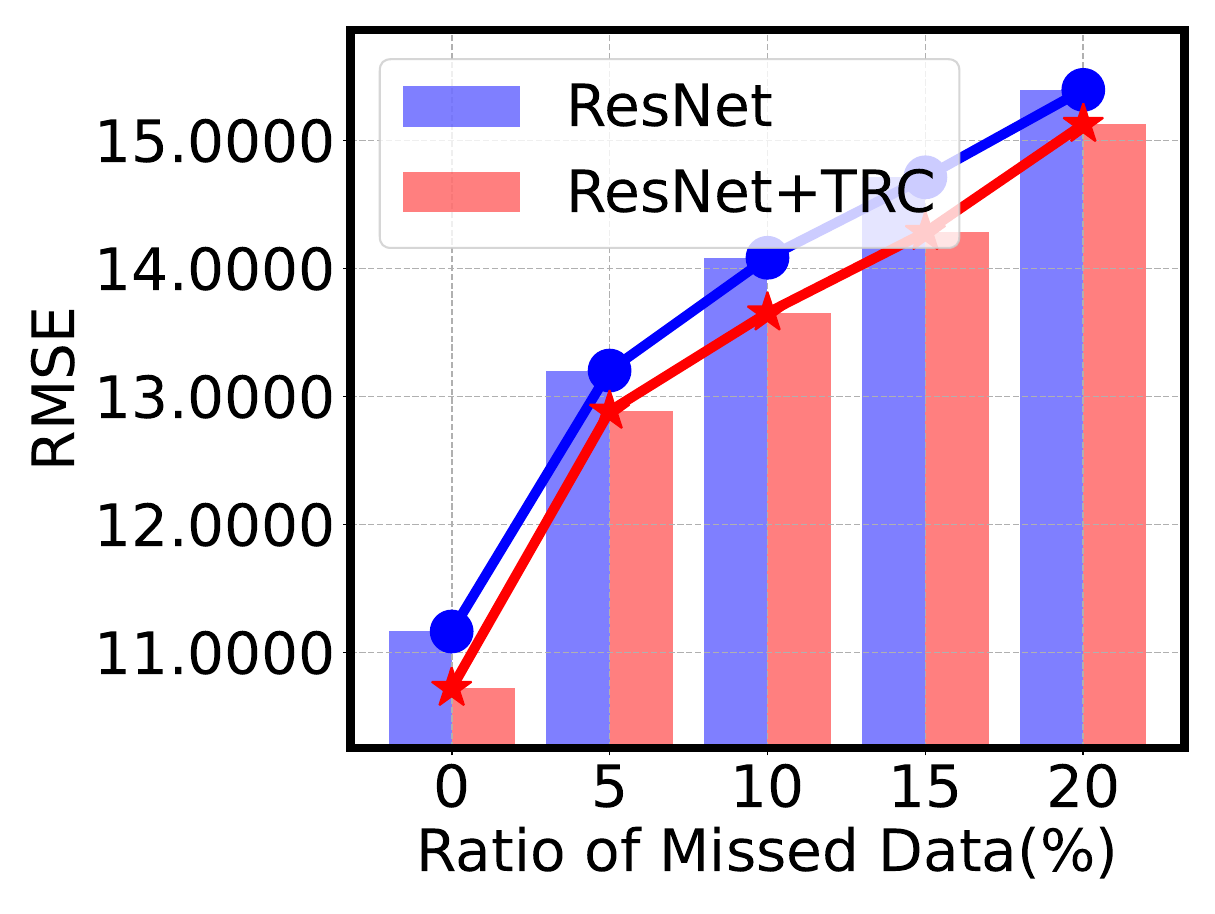}
        \caption{ResNet on SU}
    \end{subfigure}
    \caption{The results of \method training with missing values. \textcolor{blue}{This
figure serves as an extension of Fig. 7 in manuscript.}}
%     \caption{The results of \method training with missing values. \textcolor{blue}{This
% figure serves as an extension of Fig.~\ref{fig:missing value and few sample} in manuscript.}}
    \label{appendix:missing}
\vspace{4em}
\end{figure}

\begin{figure}[h!]
\vspace{4.7em}
    \centering
    \begin{subfigure}[b]{0.47\linewidth}
        \centering
        \includegraphics[width=\linewidth]{pictures/scene/few_sample/diamonds_DCN2.pdf}
        \caption{DCN2 on DI}
    \end{subfigure}   
    \hfill
    \begin{subfigure}[b]{0.47\linewidth}
        \centering
        \includegraphics[width=\linewidth]{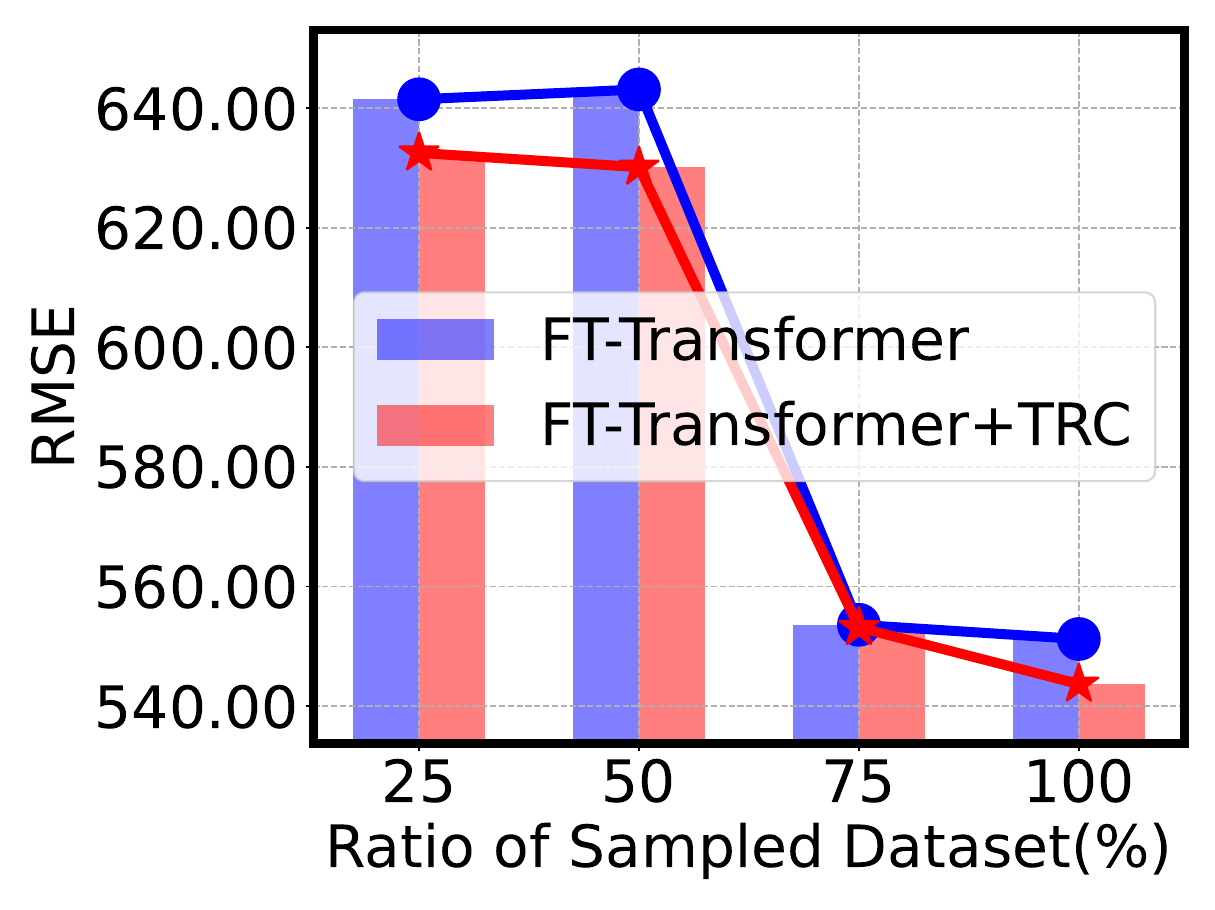}
        \caption{FT-Transformer on DI}
    \end{subfigure}   
    \hfill
    \begin{subfigure}[b]{0.47\linewidth}
        \centering
        \includegraphics[width=\linewidth]{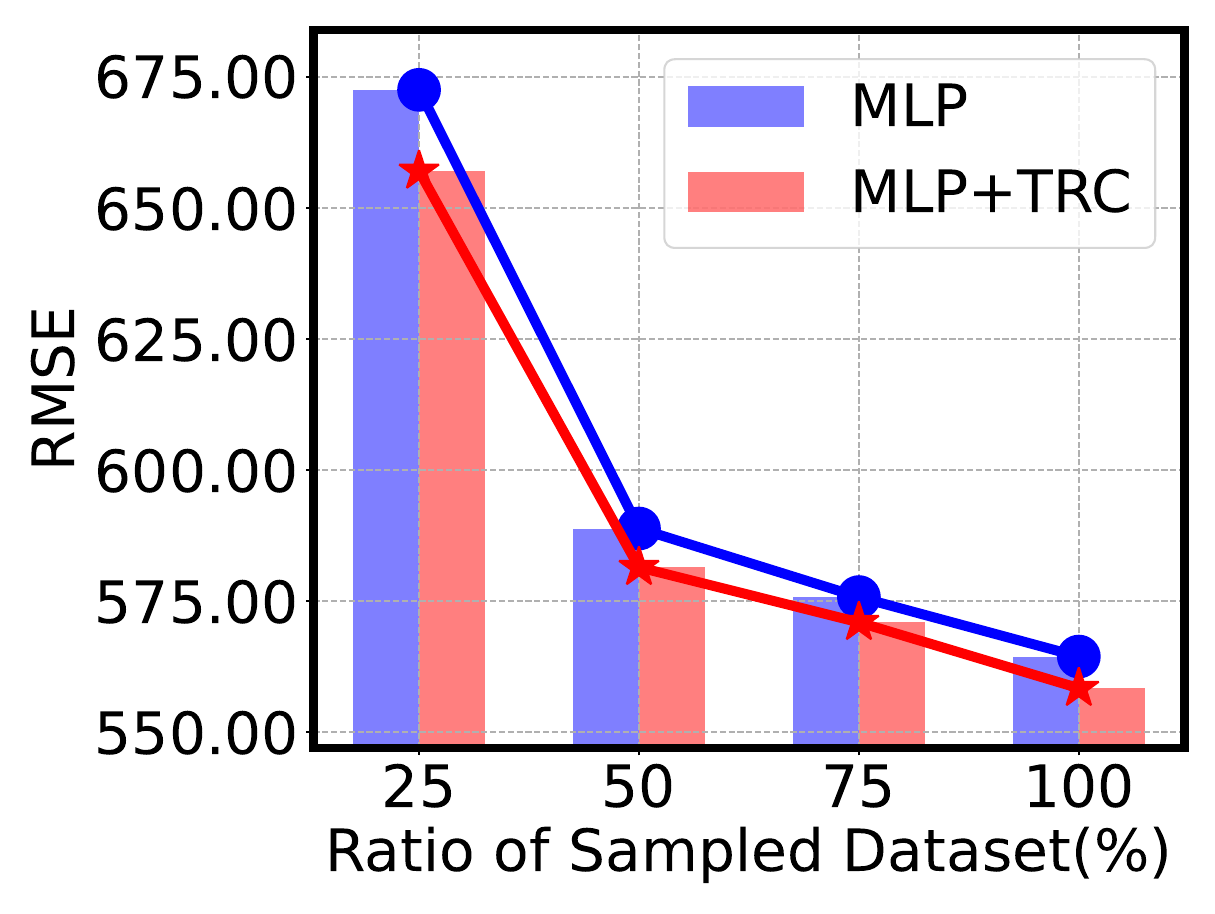}
        \caption{MLPs on DI}
    \end{subfigure}   
    \hfill
    \begin{subfigure}[b]{0.47\linewidth}
        \centering
        \includegraphics[width=\linewidth]{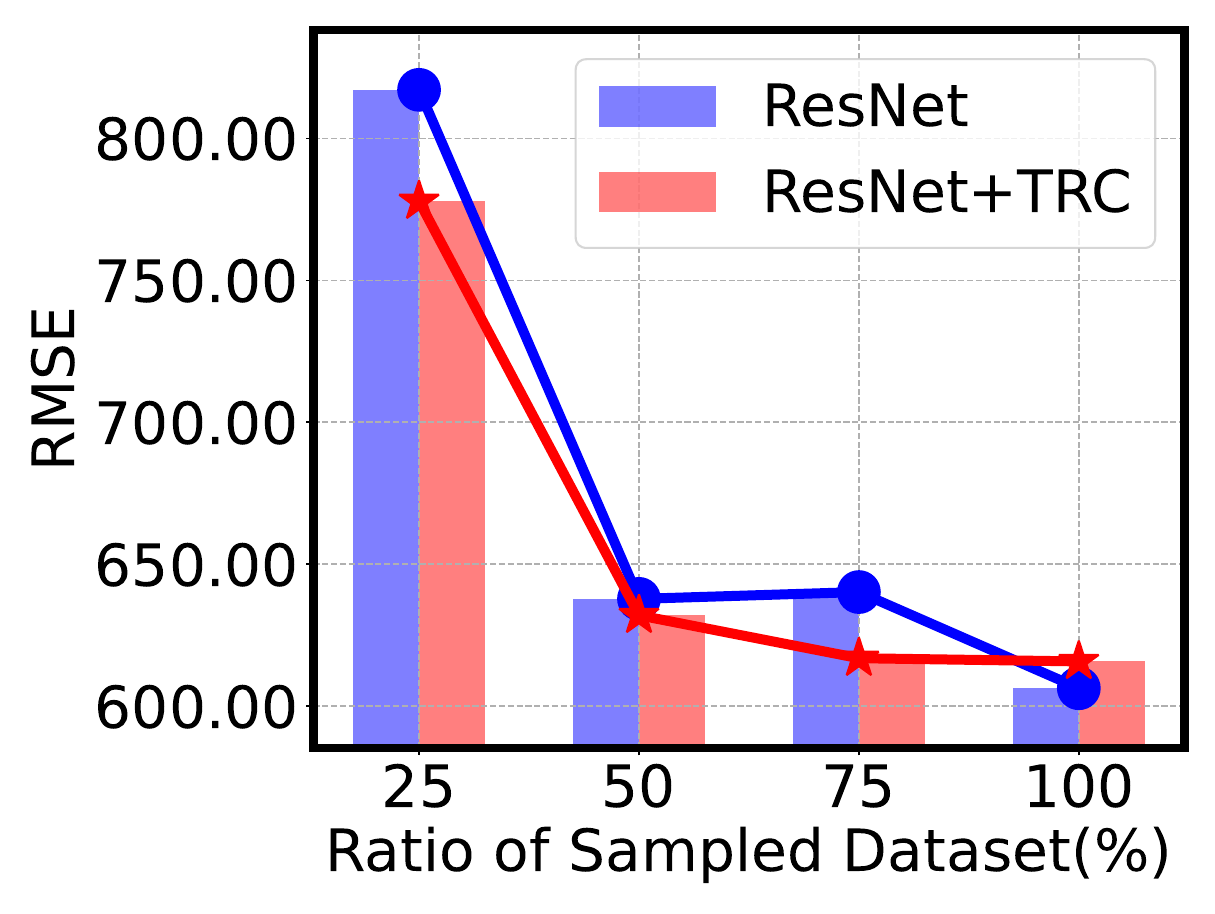}
        \caption{ResNet on DI}
    \end{subfigure}

    \begin{subfigure}[b]{0.47\linewidth}
        \centering
        \includegraphics[width=\linewidth]{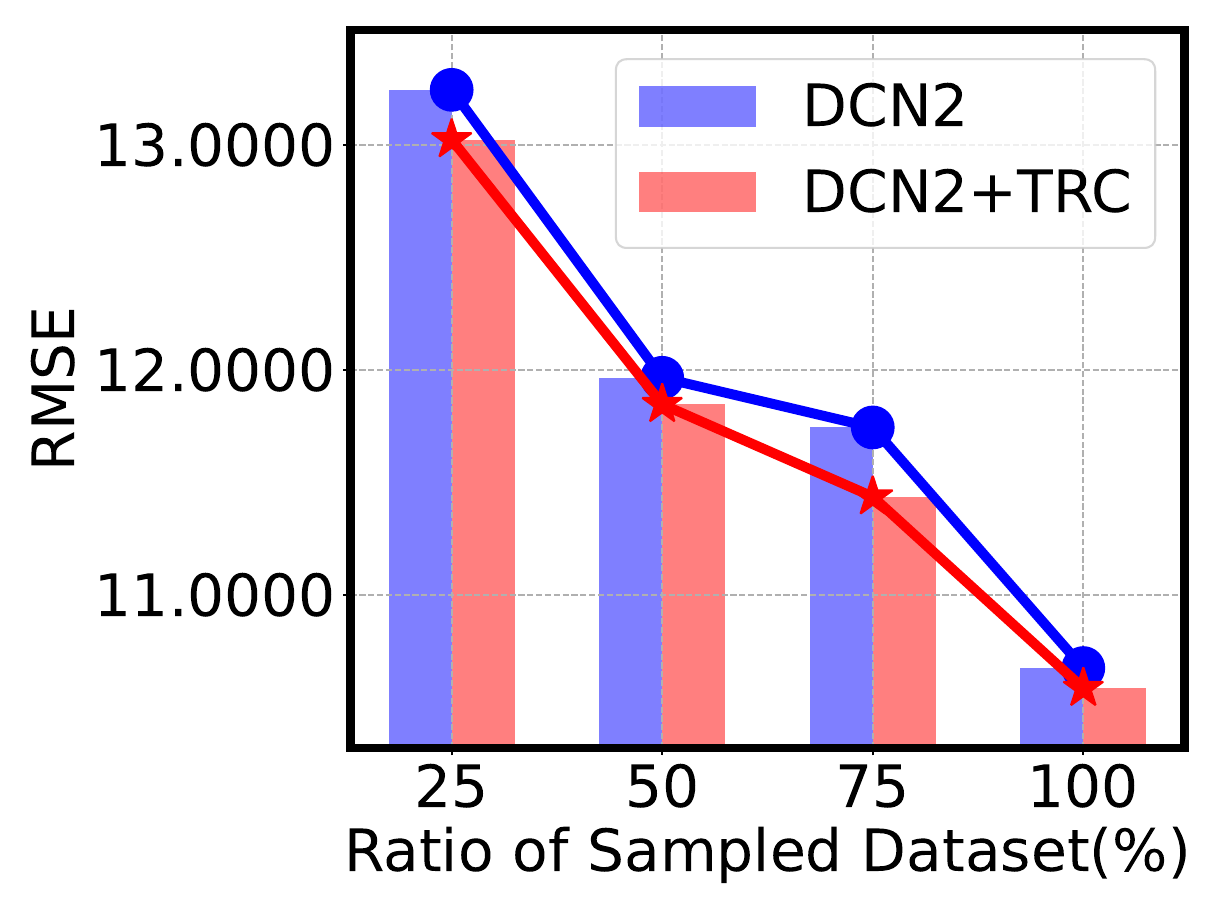}
        \caption{DCN2 on SU}
    \end{subfigure}   
    \hfill
    \begin{subfigure}[b]{0.47\linewidth}
        \centering
        \includegraphics[width=\linewidth]{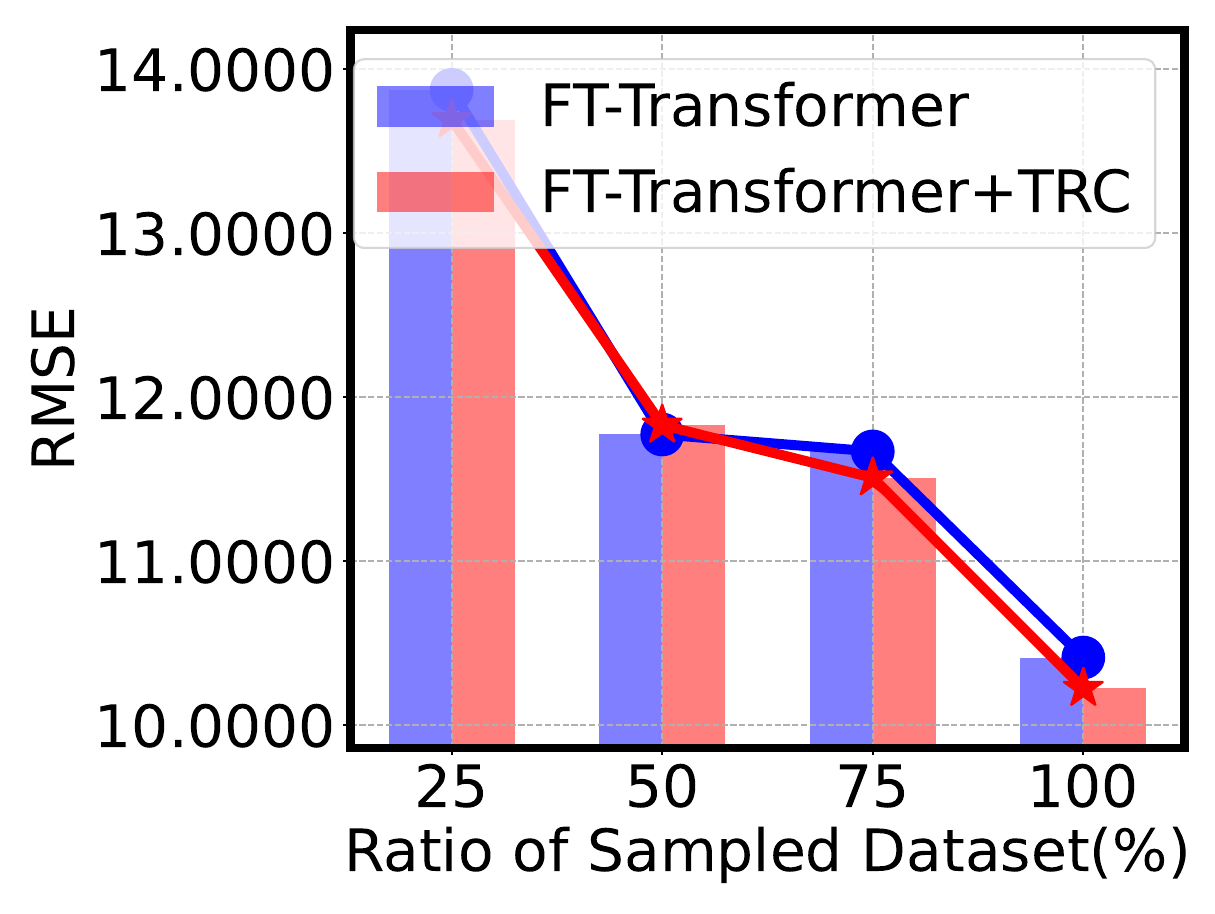}
        \caption{FT-Transformer on SU}
    \end{subfigure}   
    \hfill
    \begin{subfigure}[b]{0.47\linewidth}
        \centering
        \includegraphics[width=\linewidth]{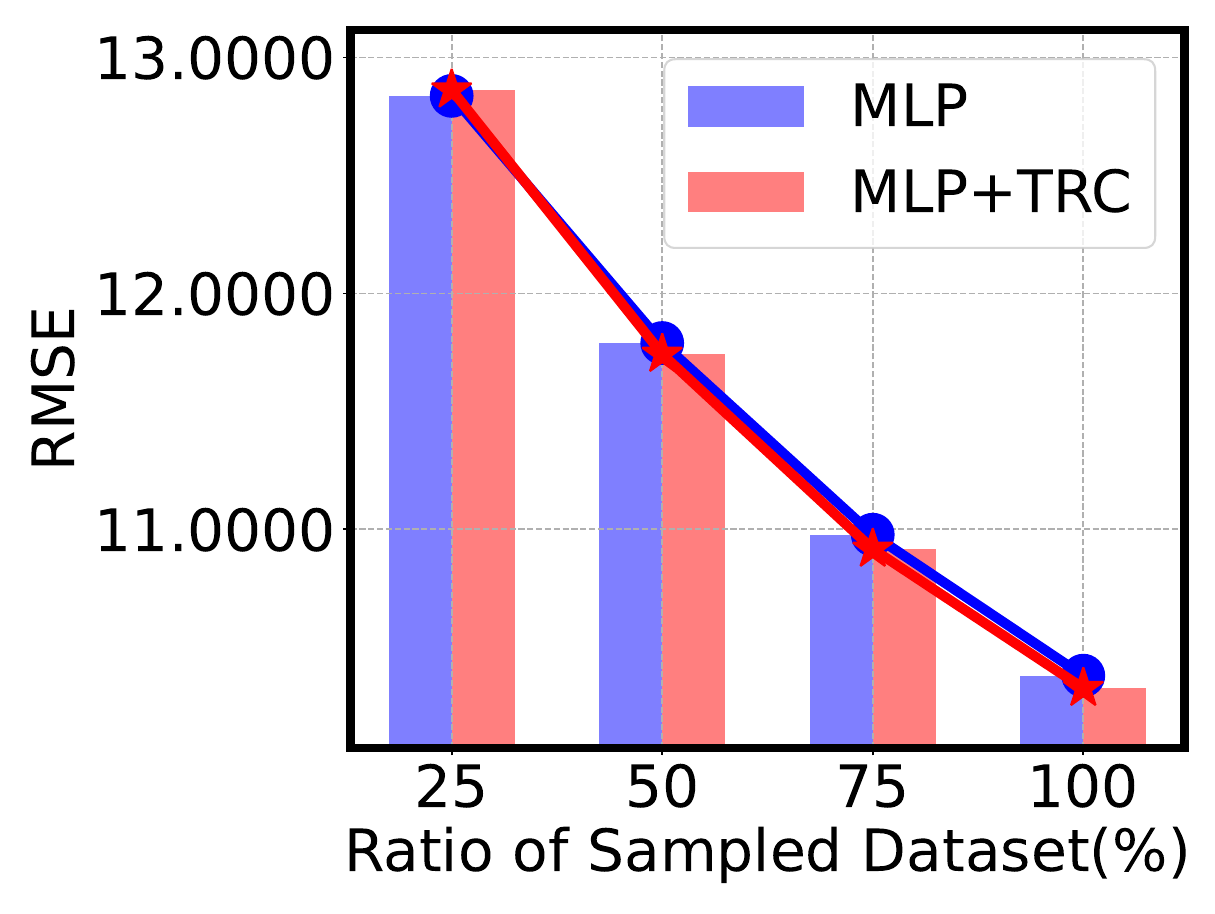}
        \caption{MLPs on SU}
    \end{subfigure}   
    \hfill
    \begin{subfigure}[b]{0.47\linewidth}
        \centering
        \includegraphics[width=\linewidth]{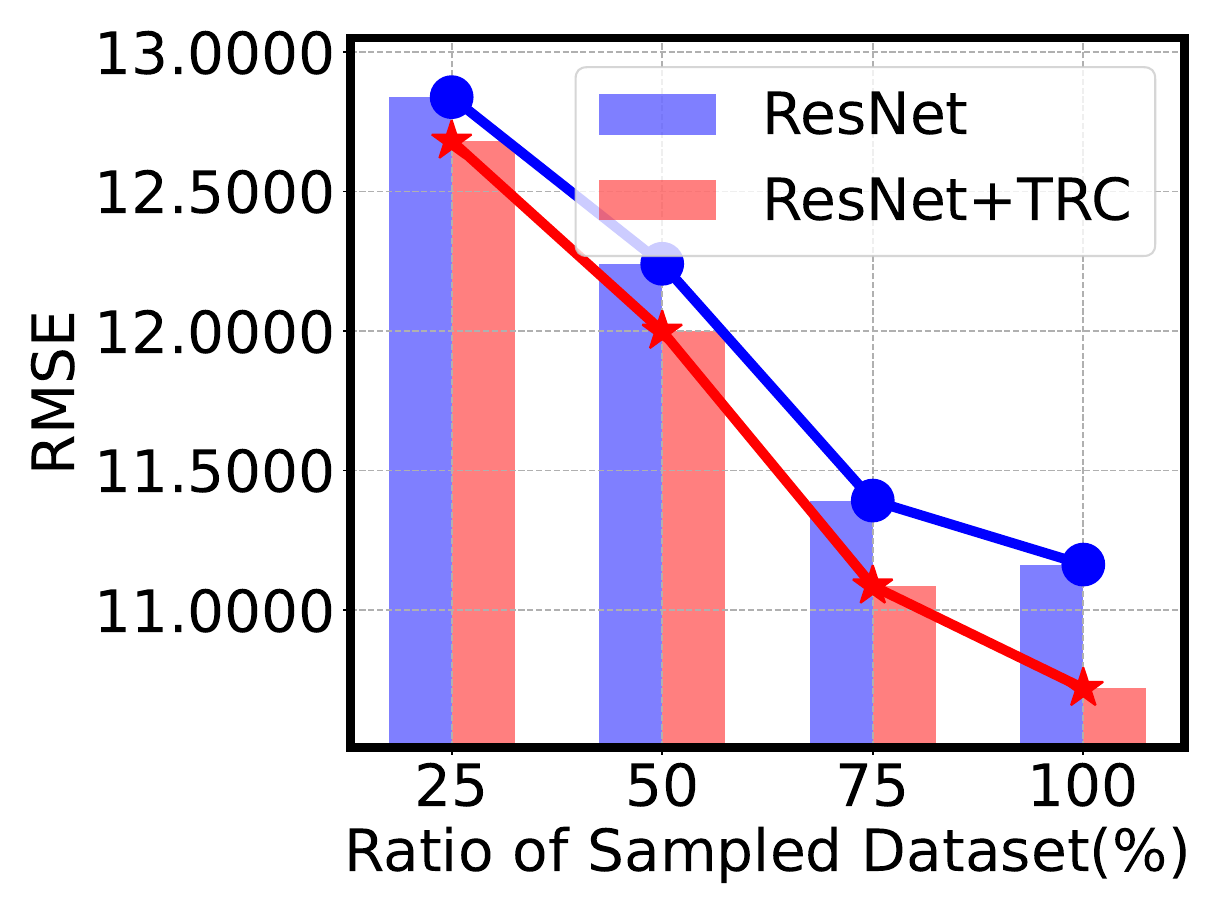}
        \caption{ResNet on SU}
    \end{subfigure}
    \caption{The results of \method training with fewer samples. \textcolor{blue}{This
figure serves as an extension of Fig. 7 in manuscript.}}
%     \caption{The results of \method training with fewer samples. \textcolor{blue}{This
% figure serves as an extension of Fig.~\ref{fig:missing value and few sample} in manuscript.}}
    \label{appendix:fewer}
\end{figure}

\method continues to enhance the performance of deep tabular models even under challenging conditions where training samples contain missing values and the overall number of training samples is reduced.
We systematically increase the proportion of missing values in features and reduce the proportion of training samples separately.  
The results in Fig~\ref{appendix:missing} and Fig~\ref{appendix:fewer} indicate that as the dataset's missing values escalate or the training sample size diminishes, the difficulty of model training intensifies, which is reflected in a decline in model performance. Nevertheless, by leveraging \method, we are able to consistently enhance the capabilities of the deep tabular model, thereby alleviating the negative impact of data incompleteness and scarcity.

% \textbf{Missing Value Scene.}

% \textbf{Few Sample Scene.}

\subsection{Parameter Analysis}
\label{appendix:parameter analysis}
\begin{figure*}[h!]
    \centering
    % \hfill
    \begin{subfigure}[b]{0.23\linewidth}
        \centering
        \includegraphics[width=\linewidth]{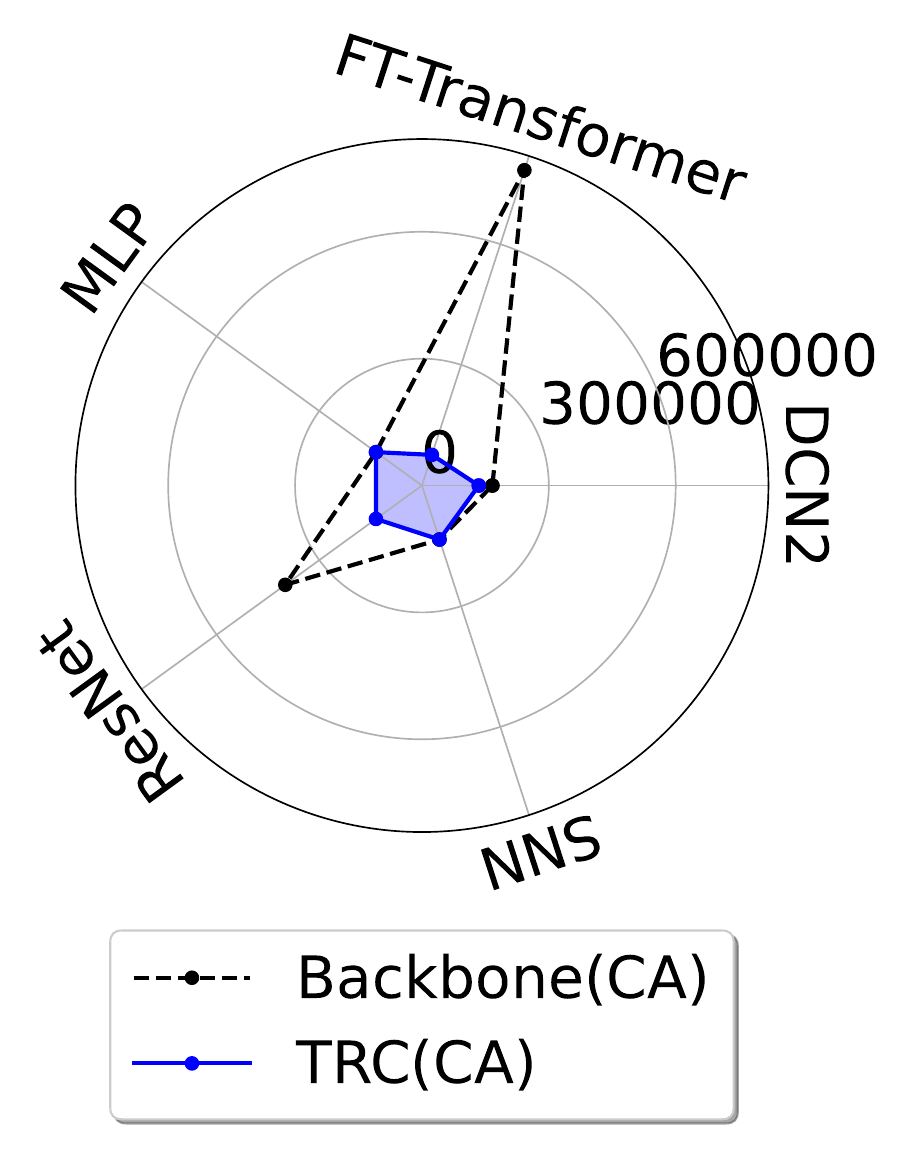}
        \caption{CA dataset}
    \end{subfigure}
    % \hfill
    \begin{subfigure}[b]{0.23\linewidth}
        \centering
        \includegraphics[width=\linewidth]{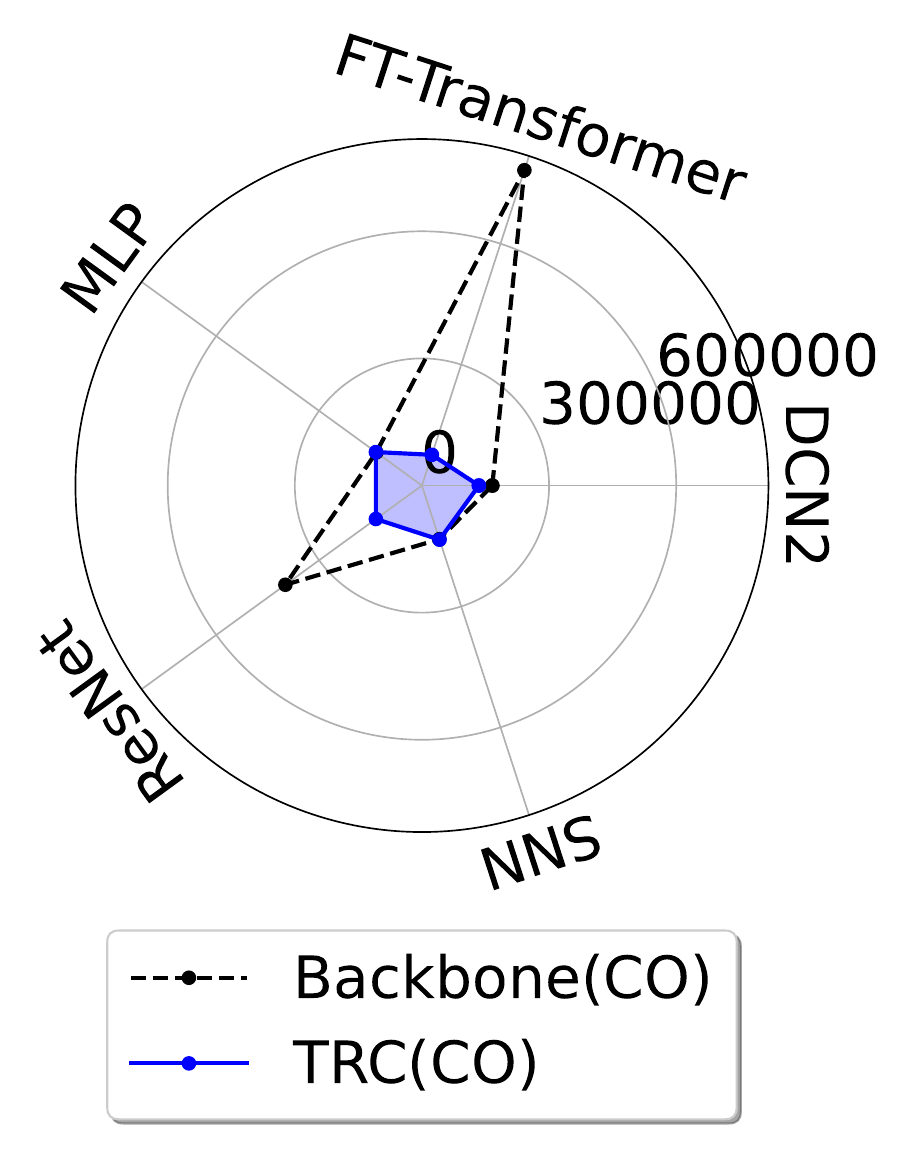}
        % \captionsetup{skip=0pt}
        \caption{CO dataset}
    \end{subfigure}
    % \hfill
    \begin{subfigure}[b]{0.23\linewidth}
        \centering
        \includegraphics[width=\linewidth]{pictures/parameters/diamonds.pdf}
        % \captionsetup{skip=0pt}
        \caption{DI dataset}
    \end{subfigure}
    % \hfill
    \begin{subfigure}[b]{0.23\linewidth}
        \centering
        \includegraphics[width=\linewidth]{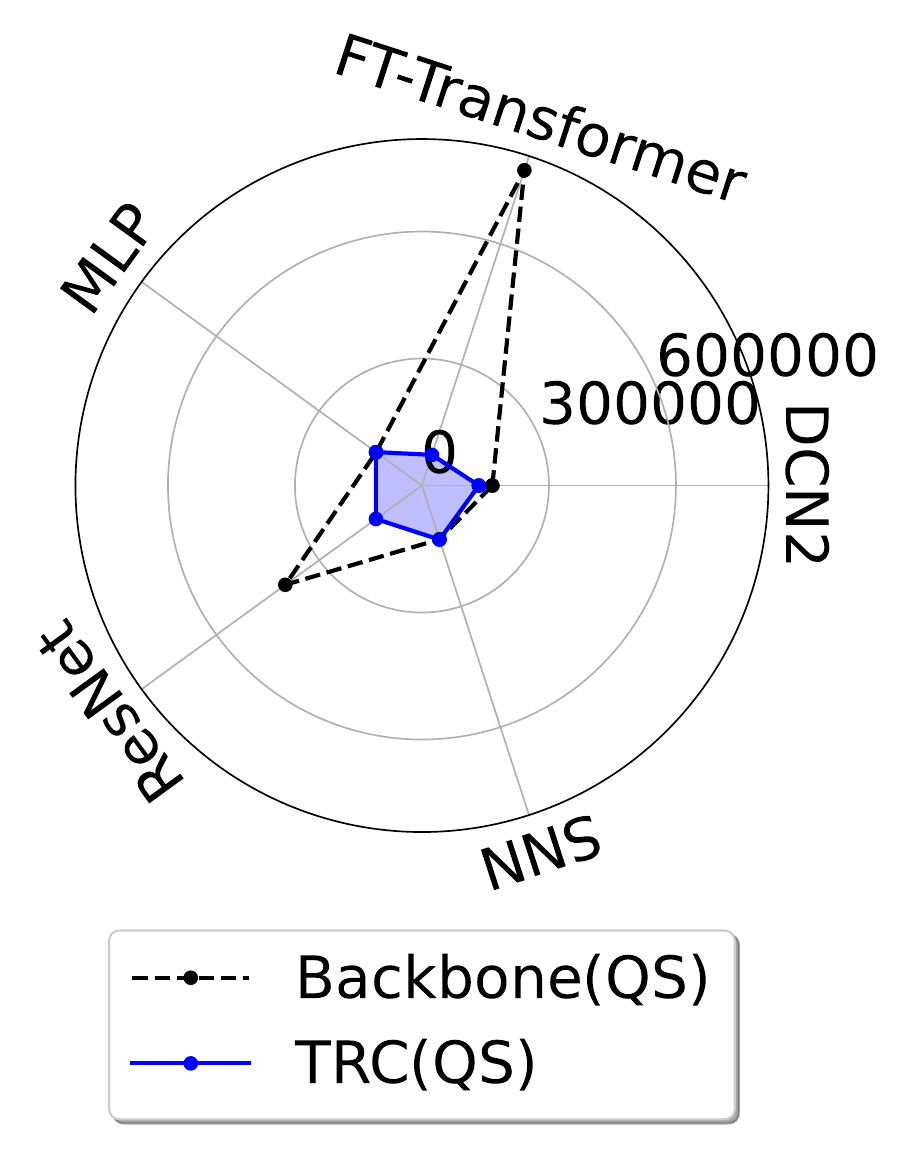}
        \caption{QS dataset}
    \end{subfigure}
    \centering
    % \hfill
    \begin{subfigure}[b]{0.23\linewidth}
        \centering
        \includegraphics[width=\linewidth]{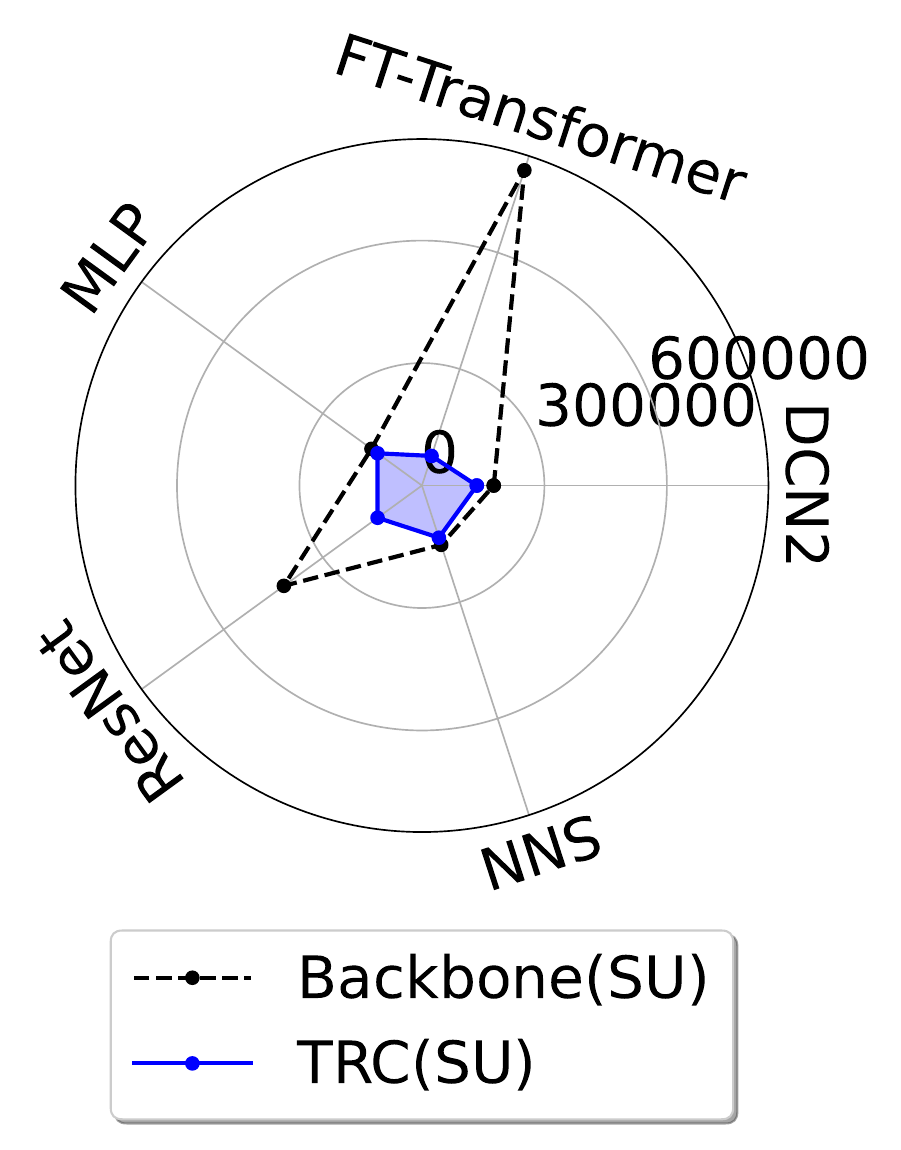}
        % \captionsetup{skip=0pt}
        \caption{SU dataset}
    \end{subfigure}
    % \hfill
    \begin{subfigure}[b]{0.23\linewidth}
        \centering
        \includegraphics[width=\linewidth]{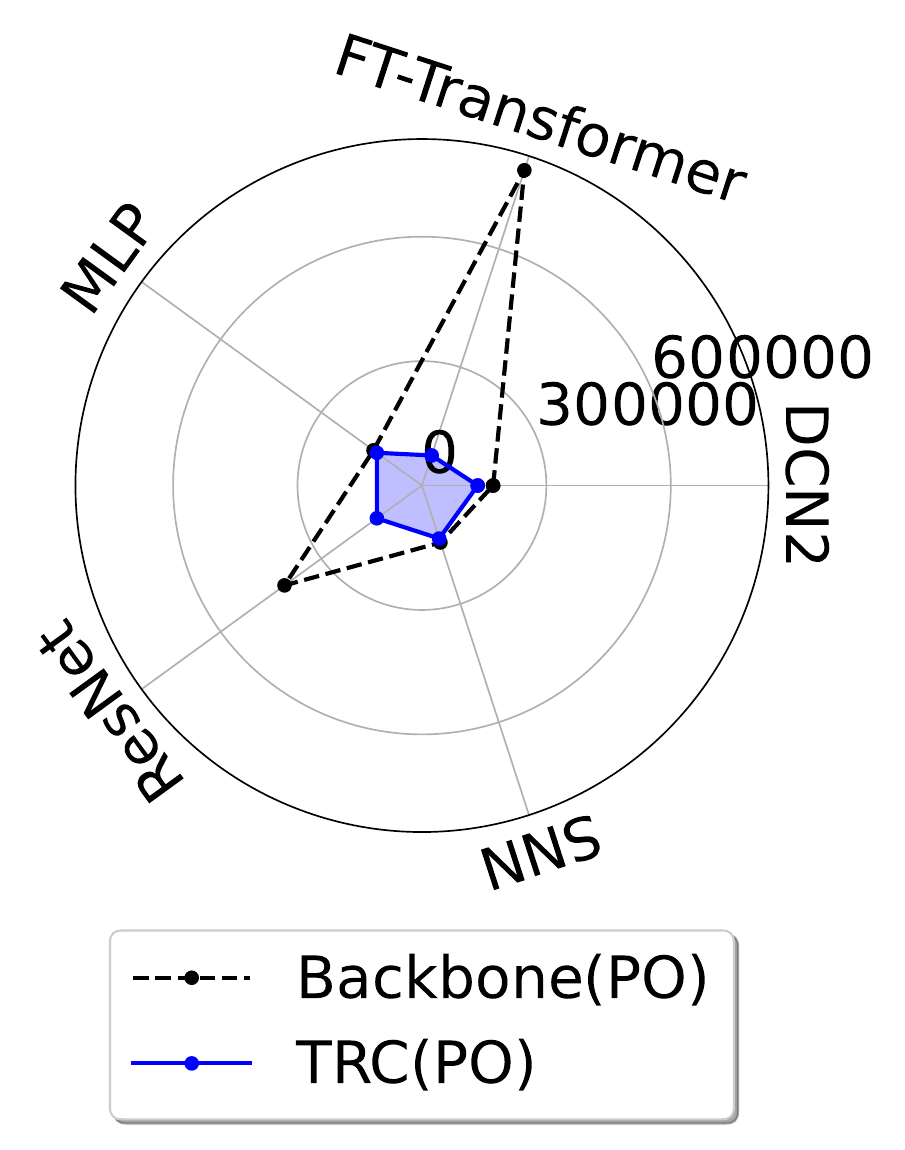}
        \caption{PO dataset}
    \end{subfigure}
    % \hfill
    \begin{subfigure}[b]{0.23\linewidth}
        \centering
        \includegraphics[width=\linewidth]{pictures/parameters/adult.pdf}
        % \captionsetup{skip=0pt}
        \caption{AD dataset}
    \end{subfigure}
    % \hfill
    \begin{subfigure}[b]{0.23\linewidth}
        \centering
        \includegraphics[width=\linewidth]{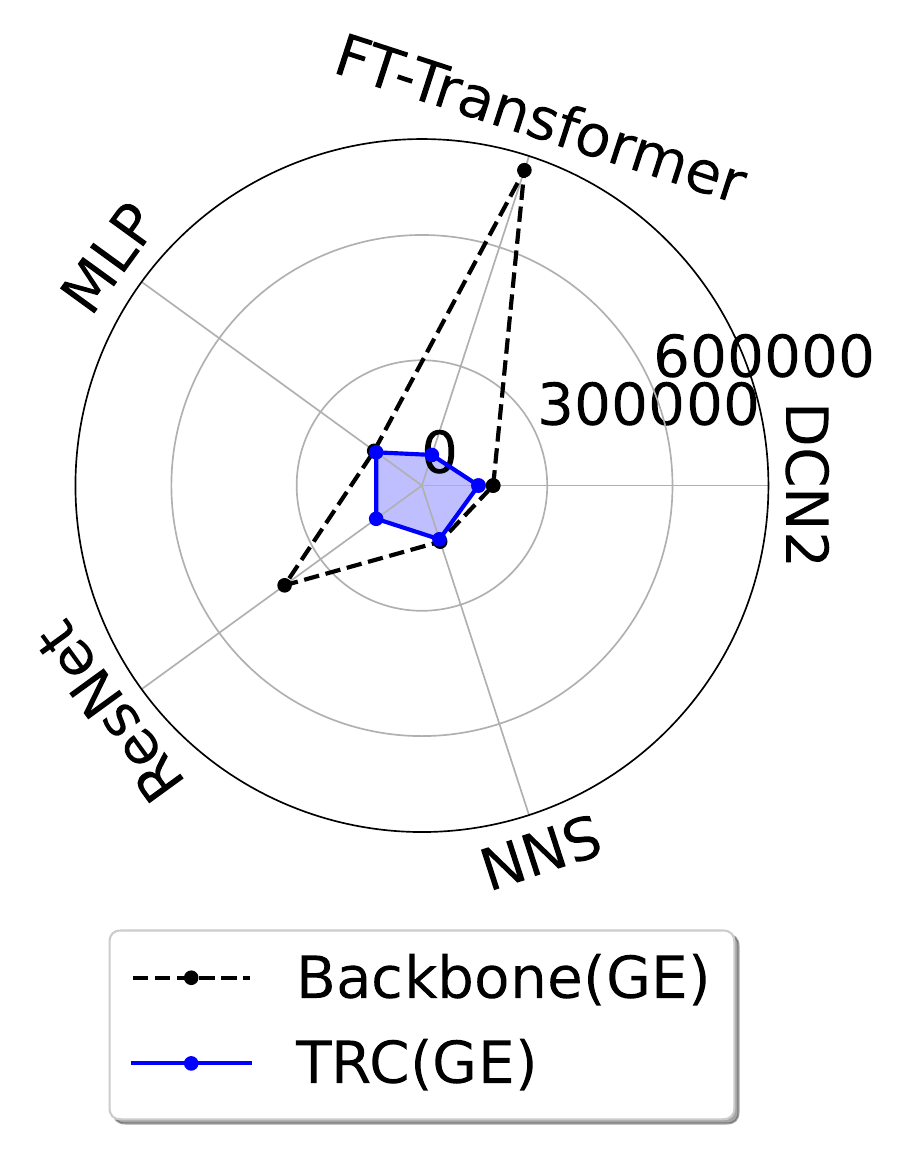}
        \caption{GE dataset}
    \end{subfigure}
    % \captionsetup{skip=0pt}
    \caption{The comparison of parameters between \method and deep tabular models. \textcolor{blue}{This
figure serves as an extension of Fig. 9 in manuscript.}}
%     \caption{The comparison of parameters between \method and deep tabular models. \textcolor{blue}{This
% figure serves as an extension of Fig.~\ref{fig:parameters analysis} in manuscript.}}
    \label{appendix:parameter}
    
\end{figure*}

The proposed \method provides a parameter-efficient, cost-effective technique to enhance the representations of deep tabular backbones without altering any of their parameters.
As shown in Fig.~\ref{appendix:parameter}, the parameter quantity of \method is significantly lower compared to deep tabular models, particularly for the FT-Transformer model, which inherently has a large number of parameters. 

\subsection{Computational Efficiency}
\label{appendix:computational efficiency}
We provide \textcolor{blue}{training and inference} time cost in Table~\ref{appendix:time cost} and \textcolor{blue}{Table~\ref{appendix:infer time cost}}.
\textcolor{blue}{All experiments were conducted on the Ubuntu 20.04.4 LTS operating system, Intel(R) Xeon(R) Gold 5220 CPU @ 2.20GHz with single NVIDIA A40 48GB GPU and 512GB of RAM.}
Since \method does not need to retrain the deep tabular backbone, the training time of \method is noticeably reduced compared to the deep tabular model in most cases. 
\textcolor{blue}{In addition, the inference cost of TRC is also much smaller compared to deep tabular backbones, thus will not introduce high inference overhead to the existing backbones.
The relatively higher inference time for both backbone and TRC observed on the YE and COV datasets are are primarily due to their large sizes, as shown in Table I of manuscript.
}

\clearpage

\begin{table*}[h!]
    \caption{\textcolor{blue}{Time cost (seconds) per training epoch on the full training set. The training time for TRC does not include the training cost for backbone model.
    }}
    \scriptsize
    \label{appendix:time cost}
    \centering
    \setlength{\tabcolsep}{1.4mm}{
    \begin{tabular}{cccccccccccc}
\toprule
 & CO  & DI  & QS  & CA  & PO  & SU  & AD  & AU & GE & YE & COV\\
\midrule
MLP & 0.129 & 0.719 & 0.013 & 0.267 & 0.189 & 0.285 & 0.566 & 0.011 & 0.123 & 9.259 & 9,442 \\
\textbf{TRC} & 0.171 & 0.878 & 0.020 & 0.348 & 0.248 & 0.372 & 0.688 & 0.017 & 0.167 & 13.086 & 13.818\\
\midrule
% DCN2 & 0.216 & 1.179 & 0.022 & 0.454 & 0.324 & 0.487 & 0.946 & 0.018 & 0.212 \\
% \textbf{TRC} & 0.170 & 0.881 & 0.023 & 0.350 & 0.251 & 0.375 & 0.701 & 0.020 & 0.169 \\
% \midrule
% SNN & 0.133 & 0.758 & 0.014 & 0.274 & 0.199 & 0.299 & 0.610 & 0.011 & 0.130 \\
% \textbf{TRC} & 0.169 & 0.876 & 0.021 & 0.343 & 0.249 & 0.370 & 0.696 & 0.017 & 0.167 \\
% \midrule
ResNet & 0.188 & 1.092 & 0.020 & 0.399 & 0.282 & 0.424 & 0.864 & 0.016 & 0.184 & 15.285 & 13.948 \\
\textbf{TRC} & 0.168 & 0.880 & 0.021 & 0.354 & 0.249 & 0.374 & 0.698 & 0.018 & 0.167 & 13.629 & 12.909\\
\midrule
% AutoInt & 0.322 & 1.783 & 0.034 & 0.694 & 0.560 & 1.356 & 1.410 & 0.029 & 0.323 \\
% TRC & 0.183 & 1.047 & 0.027 & 0.401 & 0.442 & 0.867 & 0.854 & 0.025 & 0.260 \\
% \midrule
% TANGOS & 2.635 & 15.640 & 0.302 & 6.148 & 6.921 &6.919 &14.832 &0.262& 3.043 \\
% \textbf{TRC} & 0.278 &1.046 &0.036 & 0.561 &0.382 &0.550 &1.701 &0.029 & 0.283\\
% \midrule
FT-Transformer & 0.424 & 2.348 & 0.045 & 0.917 & 0.685 & 1.397 & 1.857 & 0.038 & 0.431 & 68.678 & 52.186 \\
\textbf{TRC} & 0.174 & 0.891 & 0.027 & 0.356 & 0.259 & 0.382 & 0.705 & 0.024 & 0.174 & 13.200 & 13.194 \\
% \midrule
% PTARL & 0.231 & 1.475 & 0.025 & 0.469 & 0.361 & 1.065 & 2.541 & 0.040 & 0.281 \\
% TRC & 0.250 & 1.384 & 0.032 & 0.546 & 0.382 & 0.547 & 1.391 & 0.028 & 0.254 \\
% \midrule
% SAINT & 0.667 & 4.350 & 0.092 & 1.746 & 3.937 & 12.095 & 3.584 & 0.084 & 1.641 \\
% TRC & 0.168 & 0.731 & 0.047 & 0.303 & 0.261 & 0.401 & 0.592 & 0.048 & 0.185 \\
\midrule
SCARF & 8.972 & 3.368 & 0.043 & 1.435 & 0.873 & 1.331 & 2.485 & 0.034 & 0.593 & 22.253 & 22.083 \\
\textbf{TRC} & 4.882 & 1.145 & 0.032 & 0.452 & 0.317 & 0.474 & 0.908 & 0.027 & 0.216 & 12.974 & 13.113 \\
\midrule
VIME & 0.619 & 2.974 & 0.272 & 1.061 & 1.197 & 2.169 & 2.972 & 0.261 & 0.776 & 43.997 & 36.498 \\
\textbf{TRC} & 0.270 & 1.332 & 0.037 & 0.512 & 0.398 & 0.582 & 1.099 & 0.028 & 0.260 & 11.700 & 13.356 \\
% \midrule
% MLP w/ SSL-Rec & 0.151 & 1.384 & 0.018 & 0.341 & 0.304 & 0.468 & 1.814 & 0.014 & 0.208 \\
% TRC & 0.228 & 1.178 & 0.036 & 0.447 & 0.344 & 0.466 & 1.104 & 0.023 & 0.237 \\
% \midrule
% MLP w/ SSL-Contrastive & 0.169 & 1.122 & 0.064 & 0.394 & 0.381 & 0.432 & 0.993 & 0.056 & 0.201 \\
% \textbf{TRC} & 0.205 & 1.252 & 0.026 & 0.450 & 0.323 & 0.451 & 1.079 & 0.022 & 0.211 \\ 
\bottomrule
\end{tabular}}
\end{table*}

\begin{table*}[!h]
\caption{\textcolor{blue}{
Inference time cost (seconds) on the full test set. The inference time for TRC does not include the inference time for backbone model.
}}
    \scriptsize
    \label{appendix:infer time cost}
    \centering
    \setlength{\tabcolsep}{1.4mm}{
\begin{tabular}{cccccccccccc}
\toprule
              & CO  & DI  & QS  & CA  & PO  & SU  & AD  & AU & GE & YE & COV\\ \midrule
MLP           & 0.007 & 0.053    & 0.003                & 0.015               & 0.012 & 0.013                  & 0.081 & 0.003      & 0.009                             & 0.177 & 0.397    \\
\textbf{TRC}  & 0.006 & 0.032    & 0.002                & 0.013               & 0.011 & 0.011                  & 0.053 & 0.002      & 0.006                             & 0.151 & 0.297    \\ \midrule
ResNet        & 0.016 & 0.111    & 0.006                & 0.042               & 0.030 & 0.034                  & 0.147 & 0.006      & 0.020                             & 0.433 & 0.970    \\
\textbf{TRC}  & 0.005 & 0.033    & 0.002                & 0.015               & 0.010 & 0.012                  & 0.043 & 0.002      & 0.006                             & 0.154 & 0.341    \\ \midrule
FT-Transformer & 0.042 & 0.275    & 0.012                & 0.130               & 0.131 & 0.180                  & 0.512 & 0.012      & 0.083                             & 2.704 & 4.600    \\
\textbf{TRC}  & 0.005 & 0.030    & 0.001                & 0.014               & 0.010 & 0.012                  & 0.051 & 0.001      & 0.007                             & 0.142 & 0.318    \\ \midrule
Scarf         & 0.007 & 0.051    & 0.003                & 0.016               & 0.012 & 0.014                  & 0.080 & 0.003      & 0.009                             & 0.175 & 0.413    \\
\textbf{TRC}  & 0.006 & 0.034    & 0.002                & 0.014               & 0.009 & 0.012                  & 0.047 & 0.002      & 0.007                             & 0.137 & 0.337    \\ \midrule
VIME          & 0.008 & 0.073    & 0.004                & 0.019               & 0.015 & 0.015                  & 0.149 & 0.004      & 0.011                             & 0.222 & 0.480    \\
\textbf{TRC}  & 0.005 & 0.030    & 0.002                & 0.013               & 0.010 & 0.010                  & 0.046 & 0.002      & 0.006                             & 0.138 & 0.268    \\
\bottomrule
\end{tabular}}
\end{table*}

% \clearpage
% \subsection{The Relationship between SVE and the Number of Embedding Vectors}
% We incorporate the sensitivity analysis for the number of embedding vectors, the weight of loss function $\mathcal{L}_{orth}$, the ratio of selected optimal representations, and the perturbing times for constructing simulated sub-optimal representations in Inherent Shift Learning in Fig.~\ref{appendix:s1}, Fig.~\ref{appendix:s2}, Fig.~\ref{appendix:s3}, and Fig.~\ref{appendix:s4}.
% \begin{figure}[h]
%     \centering
%     \begin{subfigure}[b]{0.48\linewidth}
%         \centering
%         \includegraphics[width=\linewidth]{pictures/sve_proto/california_housing.pdf}
%         \caption{CA dataset}
%     \end{subfigure}
%     \hfill
%     \begin{subfigure}[b]{0.48\linewidth}
%         \centering
%         \includegraphics[width=\linewidth]{pictures/sve_proto/combined_cycle_power_plant.pdf}
%         \caption{CO dataset}
%     \end{subfigure}
    
%     \begin{subfigure}[b]{0.48\linewidth}
%         \centering
%         \includegraphics[width=\linewidth]{pictures/sve_proto/adult.pdf}
%         \caption{AD dataset}
%     \end{subfigure}
%     \hfill
%     \begin{subfigure}[b]{0.48\linewidth}
%         \centering
%         \includegraphics[width=\linewidth]{pictures/sve_proto/GesturePhaseSegmentationProcessed.pdf}
%         \caption{GE dataset}
%     \end{subfigure}
%     \centering

%     \caption{Increasing the number of embedding vectors leads to larger SVE values of representations.}
% \end{figure}

\clearpage
\subsection{Visualization}
\label{appendix:visualization}
\begin{figure}[h!]
    \centering
    \captionsetup[sub]{font=scriptsize}
    % \begin{subfigure}[b]{0.24\linewidth}
    %     \centering
    %     \includegraphics[width=\linewidth]{pictures/embedding_tsne/adult_AutoInt.pdf}
    %     \caption{AutoInt on AD dataset}
    % \end{subfigure}
    % \begin{subfigure}[b]{0.24\linewidth}
    %     \centering
    %     \includegraphics[width=\linewidth]{pictures/embedding_tsne/adult_DCN2.pdf}
    %     \caption{DCN2 on AD dataset}
    % \end{subfigure}
    % \begin{subfigure}[b]{0.24\linewidth}
    %     \centering
    %     \includegraphics[width=\linewidth]{pictures/embedding_tsne/adult_FTTransformer.pdf}
    %     \caption{FT-Transformer on AD dataset}
    % \end{subfigure}
    % \begin{subfigure}[b]{0.24\linewidth}
    %     \centering
    %     \includegraphics[width=\linewidth]{pictures/embedding_tsne/adult_MLP.pdf}
    %     \caption{MLP on AD dataset}
    % \end{subfigure}
    % \begin{subfigure}[b]{0.24\linewidth}
    %     \centering
    %     \includegraphics[width=\linewidth]{pictures/embedding_tsne/adult_ResNet.pdf}
    %     \caption{ResNet on AD dataset}
    % \end{subfigure}
    % \begin{subfigure}[b]{0.24\linewidth}
    %     \centering
    %     \includegraphics[width=\linewidth]{pictures/embedding_tsne/adult_Saint.pdf}
    %     \caption{SAINT on AD dataset}
    % \end{subfigure}
    % \begin{subfigure}[b]{0.24\linewidth}
    %     \centering
    %     \includegraphics[width=\linewidth]{pictures/embedding_tsne/adult_SNN.pdf}
    %     \caption{SNN on AD dataset}
    % \end{subfigure}
    % \begin{subfigure}[b]{0.24\linewidth}
    %     \centering
    %     \includegraphics[width=\linewidth]{pictures/embedding_tsne/adult_VIME.pdf}
    %     \caption{VIME on AD dataset}
    % \end{subfigure}
    \begin{subfigure}[b]{0.24\linewidth}
        \centering
        \includegraphics[width=\linewidth]{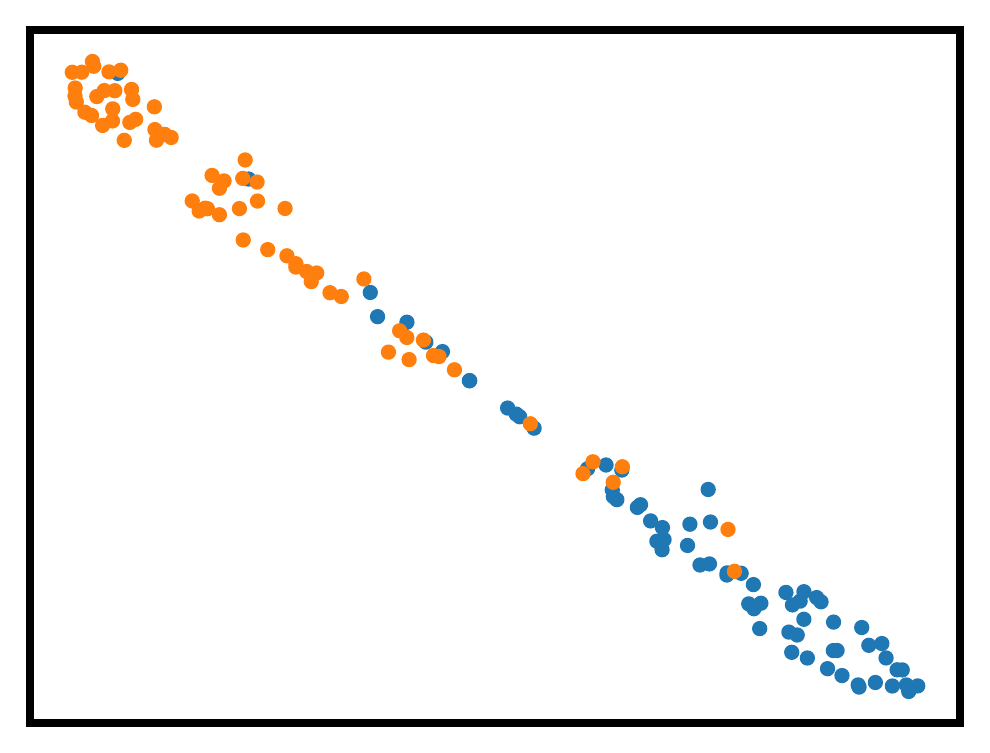}
        \caption{AutoInt on AU dataset}
    \end{subfigure}
    \begin{subfigure}[b]{0.24\linewidth}
        \centering
        \includegraphics[width=\linewidth]{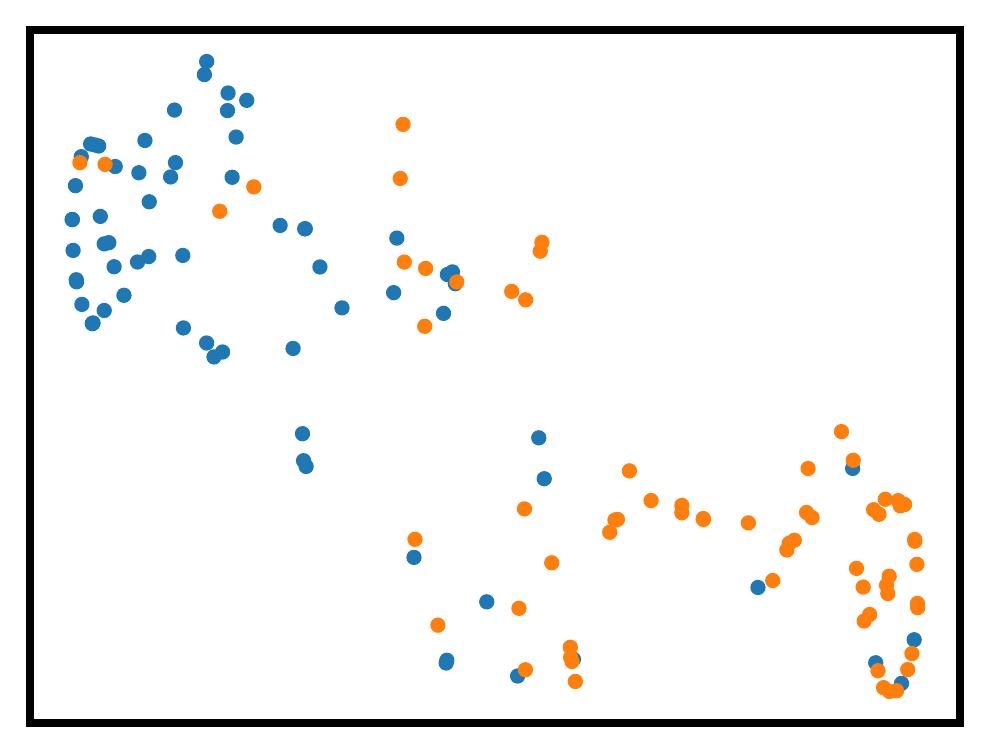}
        \caption{DCN2 on AU dataset}
    \end{subfigure}
    \begin{subfigure}[b]{0.24\linewidth}
        \centering
        \includegraphics[width=\linewidth]{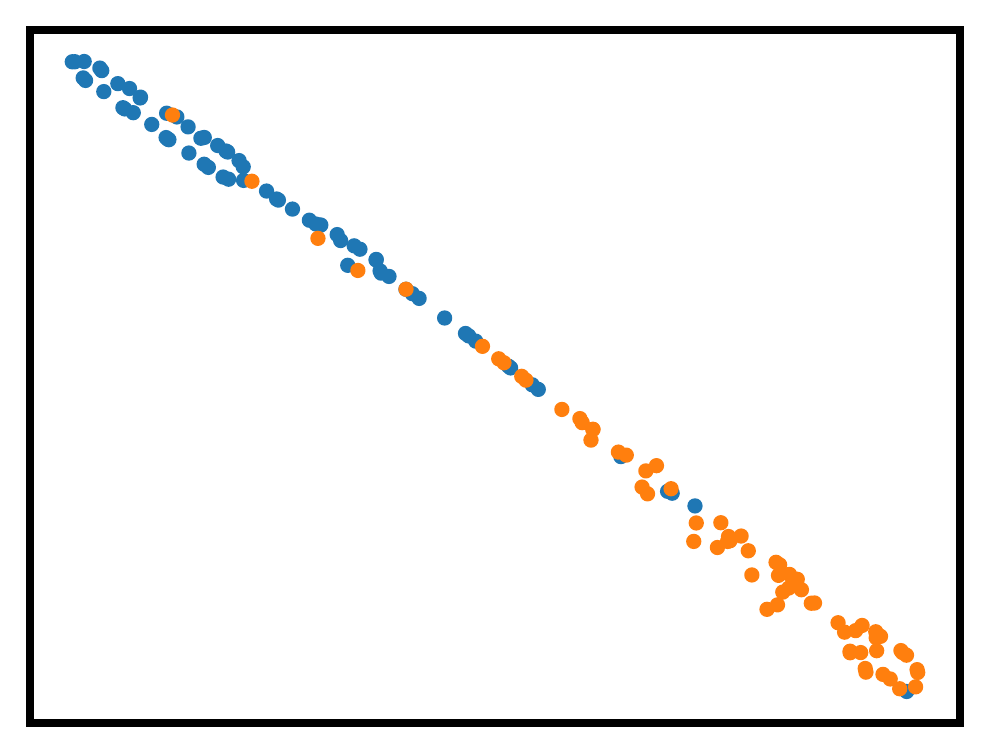}
        \caption{FT-Transformer on AU dataset}
    \end{subfigure}
    \begin{subfigure}[b]{0.24\linewidth}
        \centering
        \includegraphics[width=\linewidth]{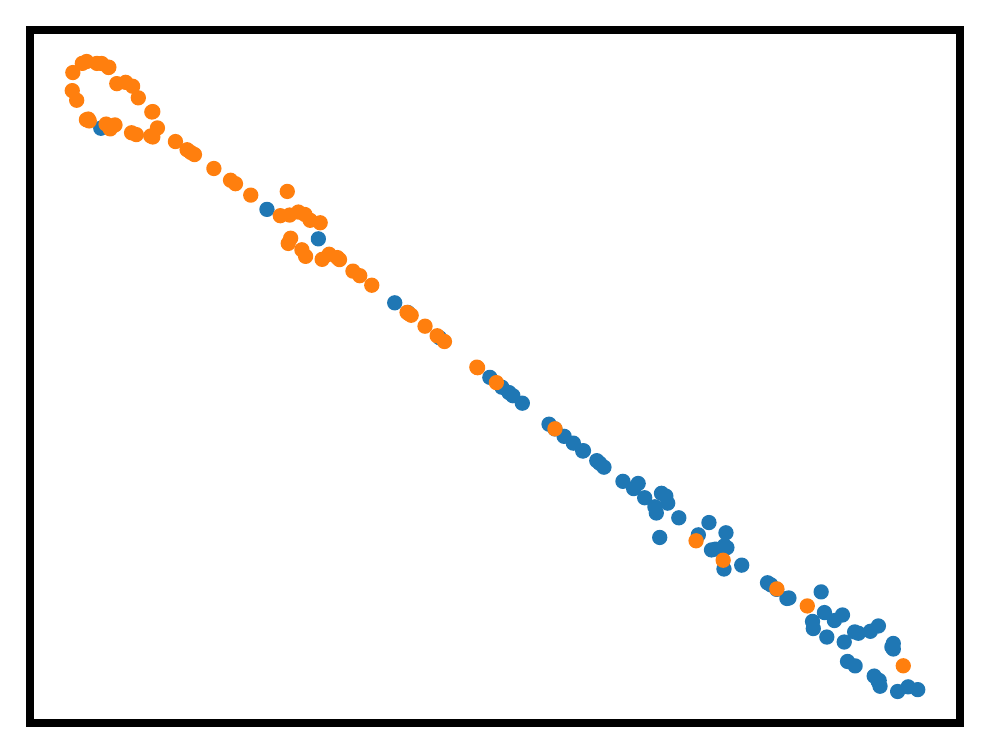}
        \caption{MLP on AU dataset}
    \end{subfigure}

    \begin{subfigure}[b]{0.24\linewidth}
        \centering
        \includegraphics[width=\linewidth]{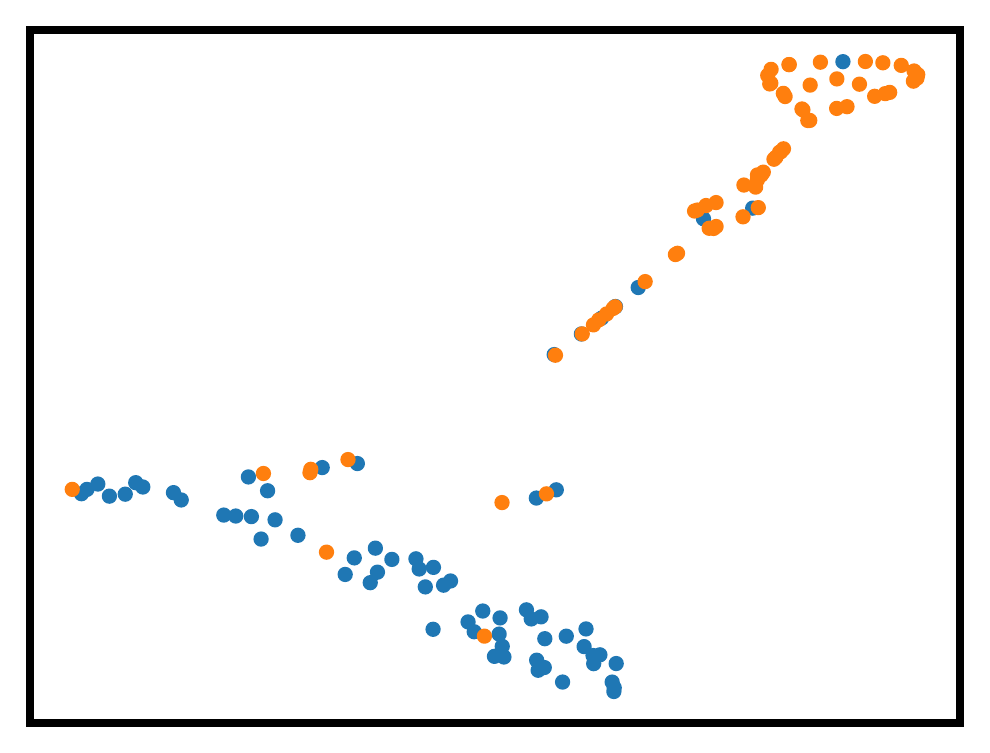}
        \caption{ResNet on AU dataset}
    \end{subfigure}
    \begin{subfigure}[b]{0.24\linewidth}
        \centering
        \includegraphics[width=\linewidth]{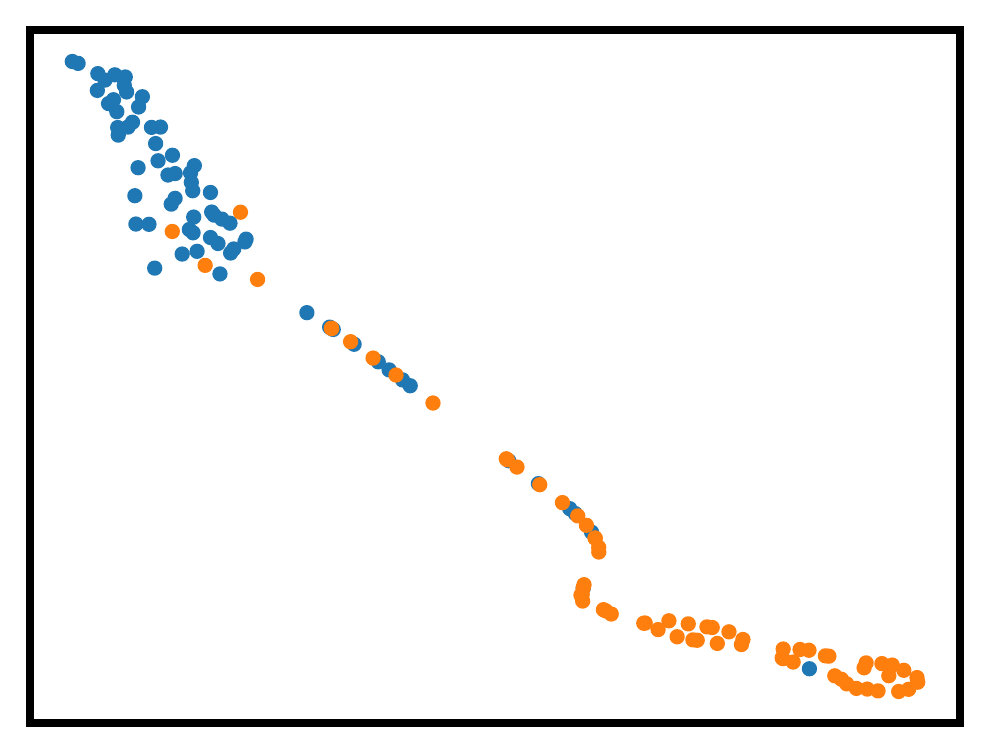}
        \caption{SAINT on AU dataset}
    \end{subfigure}
    \begin{subfigure}[b]{0.24\linewidth}
        \centering
        \includegraphics[width=\linewidth]{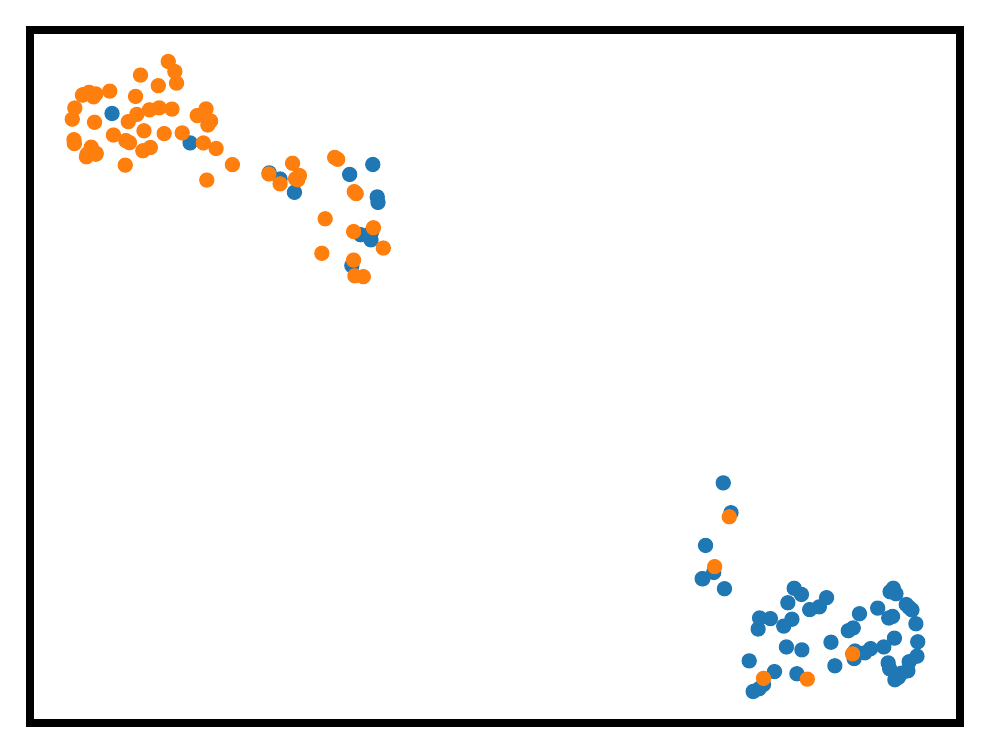}
        \caption{SNN on AU dataset}
    \end{subfigure}
    \begin{subfigure}[b]{0.24\linewidth}
        \centering
        \includegraphics[width=\linewidth]{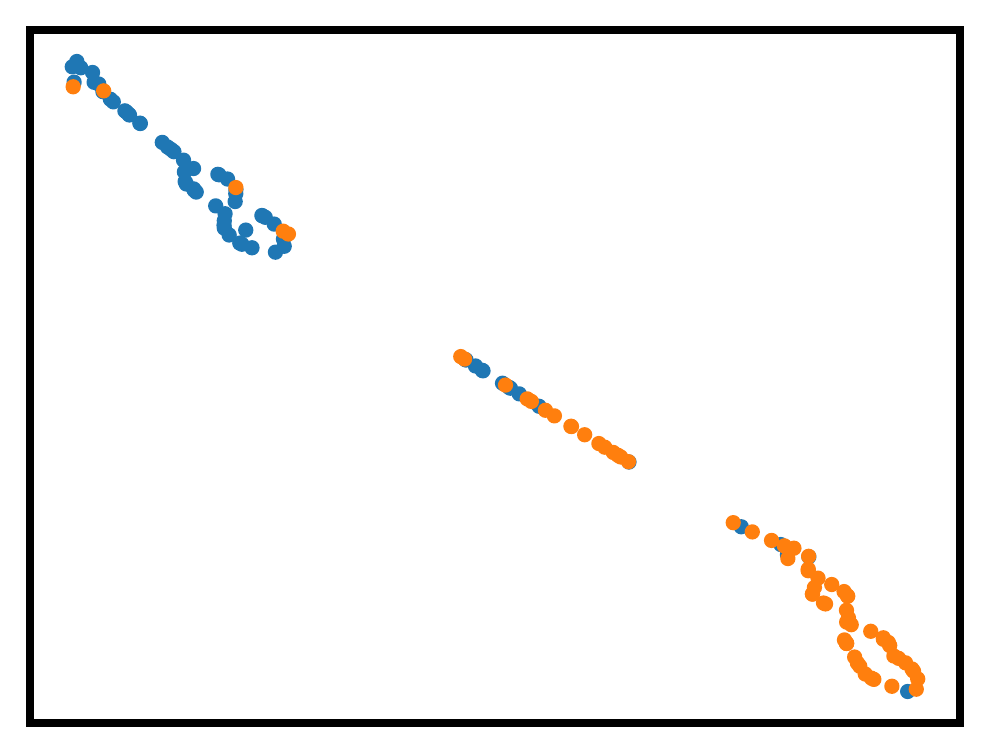}
        \caption{VIME on AU dataset}
    \end{subfigure}
    \begin{subfigure}[b]{0.24\linewidth}
        \centering
        \includegraphics[width=\linewidth]{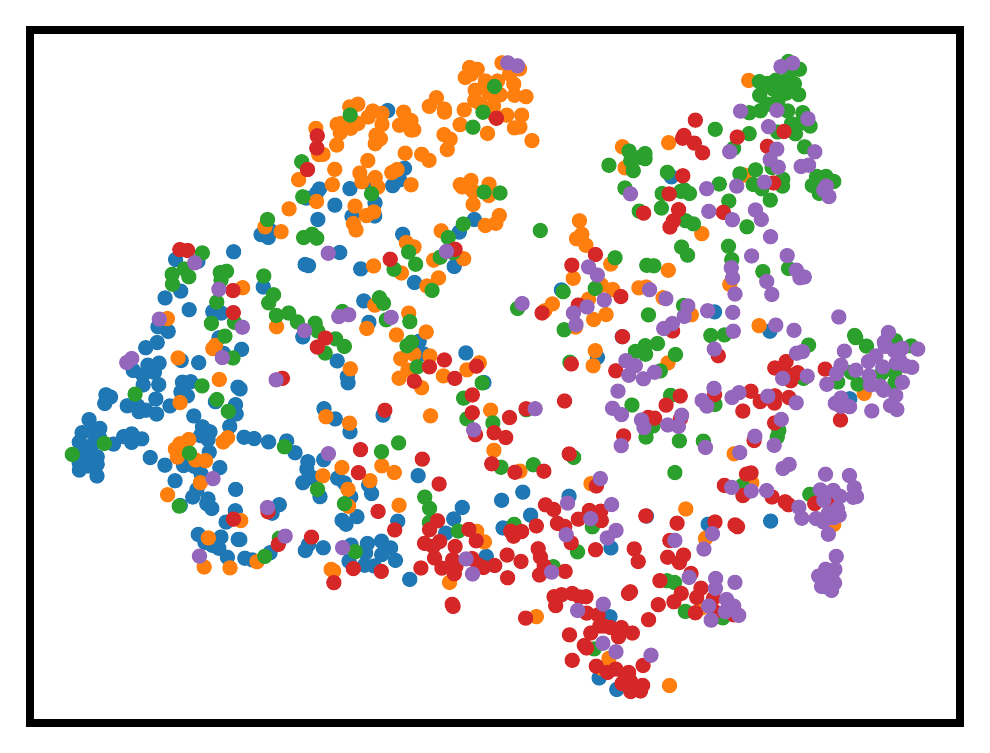}
        \caption{AutoInt on GE dataset}
    \end{subfigure}
    \begin{subfigure}[b]{0.24\linewidth}
        \centering
        \includegraphics[width=\linewidth]{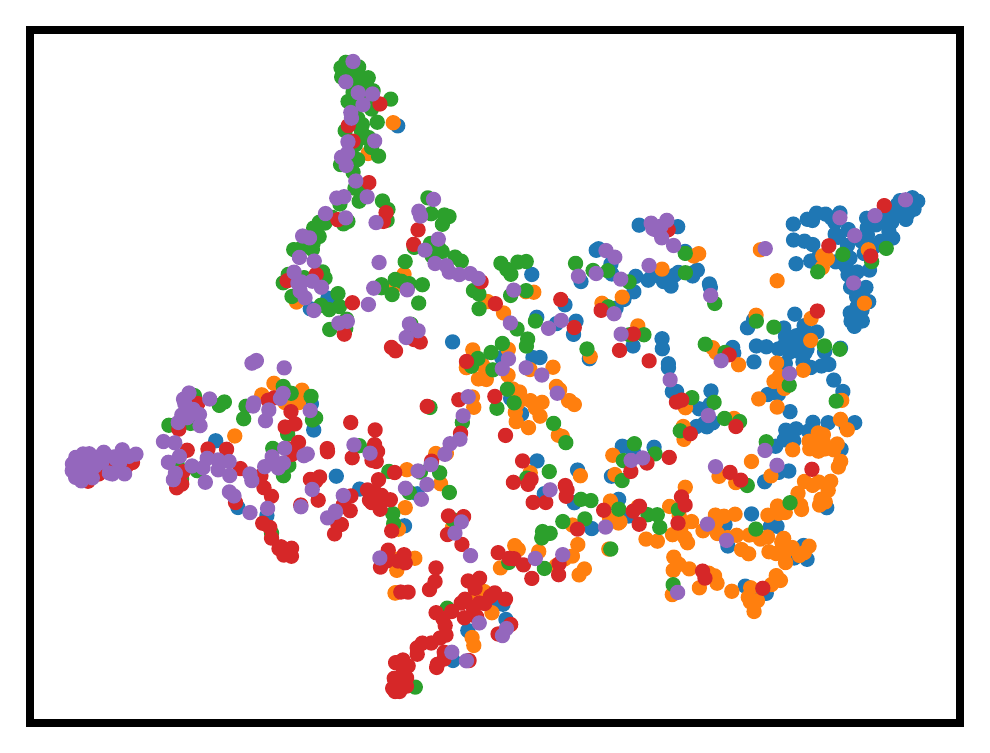}
        \caption{DCN2 on GE dataset}
    \end{subfigure}
    \begin{subfigure}[b]{0.24\linewidth}
        \centering
        \includegraphics[width=\linewidth]{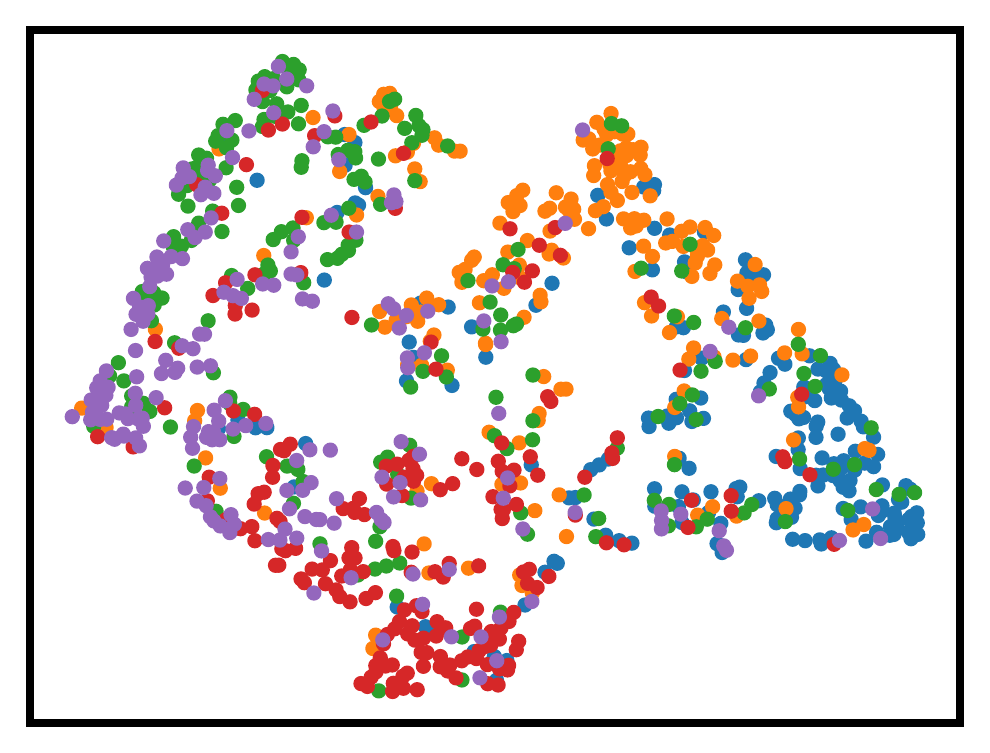}
        \caption{FT-Transformer on GE dataset}
    \end{subfigure}
    \begin{subfigure}[b]{0.24\linewidth}
        \centering
        \includegraphics[width=\linewidth]{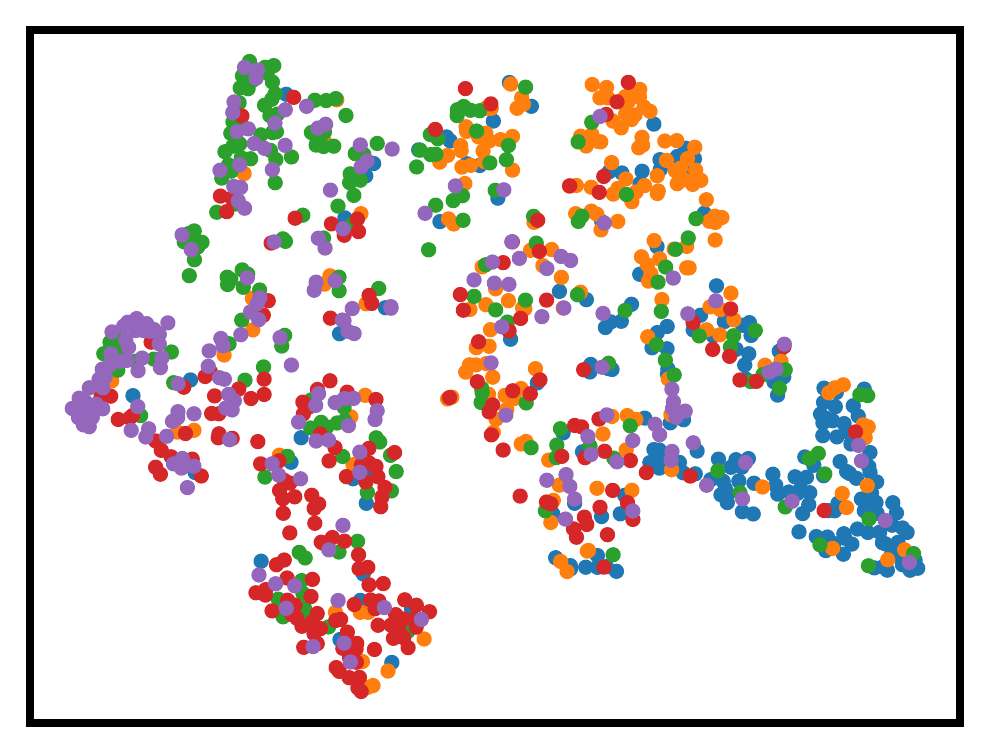}
        \caption{MLP on GE dataset}
    \end{subfigure}
    \begin{subfigure}[b]{0.24\linewidth}
        \centering
        \includegraphics[width=\linewidth]{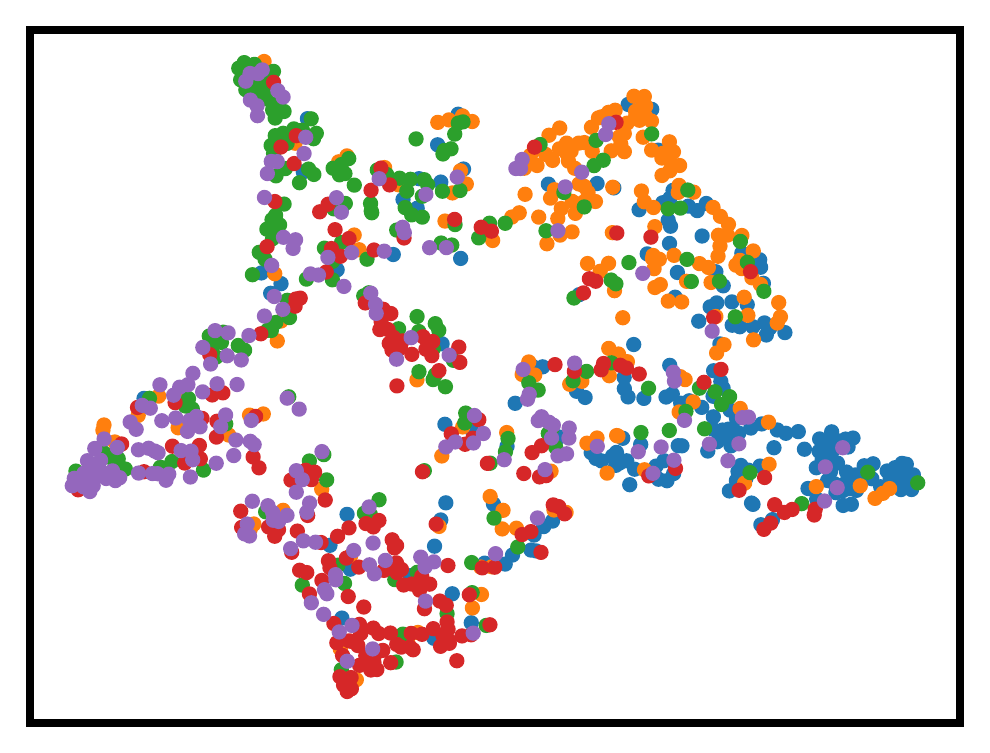}
        \caption{ResNet on GE dataset}
    \end{subfigure}
    \begin{subfigure}[b]{0.24\linewidth}
        \centering
        \includegraphics[width=\linewidth]{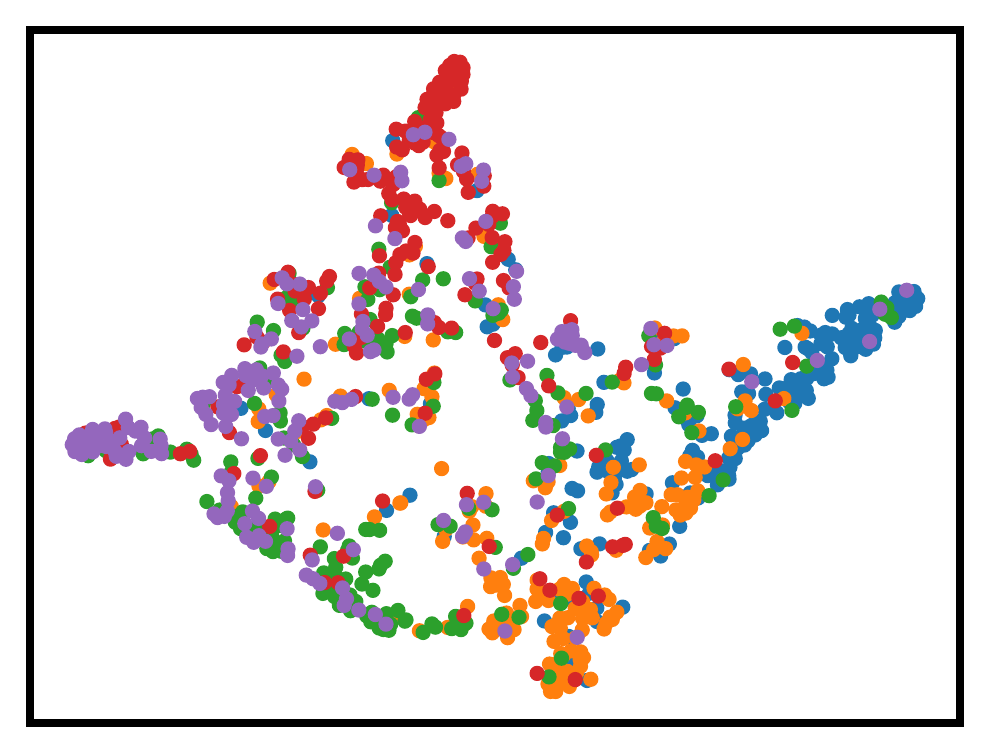}
        \caption{SAINT on GE dataset}
    \end{subfigure}
    \begin{subfigure}[b]{0.24\linewidth}
        \centering
        \includegraphics[width=\linewidth]{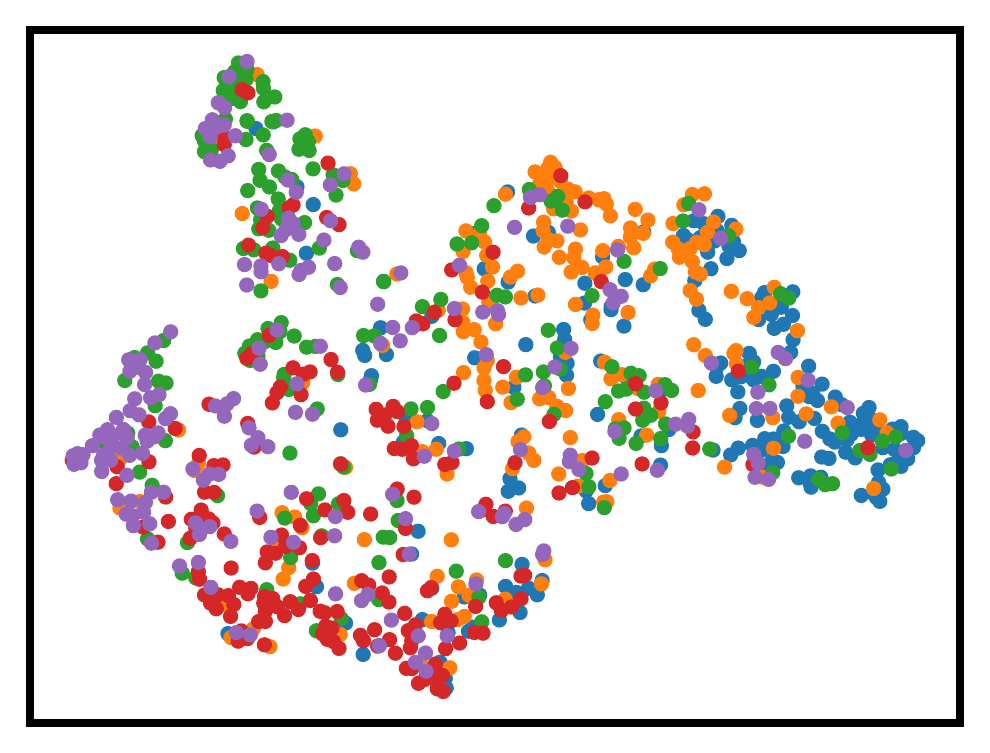}
        \caption{SNN on GE dataset}
    \end{subfigure}
    \begin{subfigure}[b]{0.24\linewidth}
        \centering
        \includegraphics[width=\linewidth]{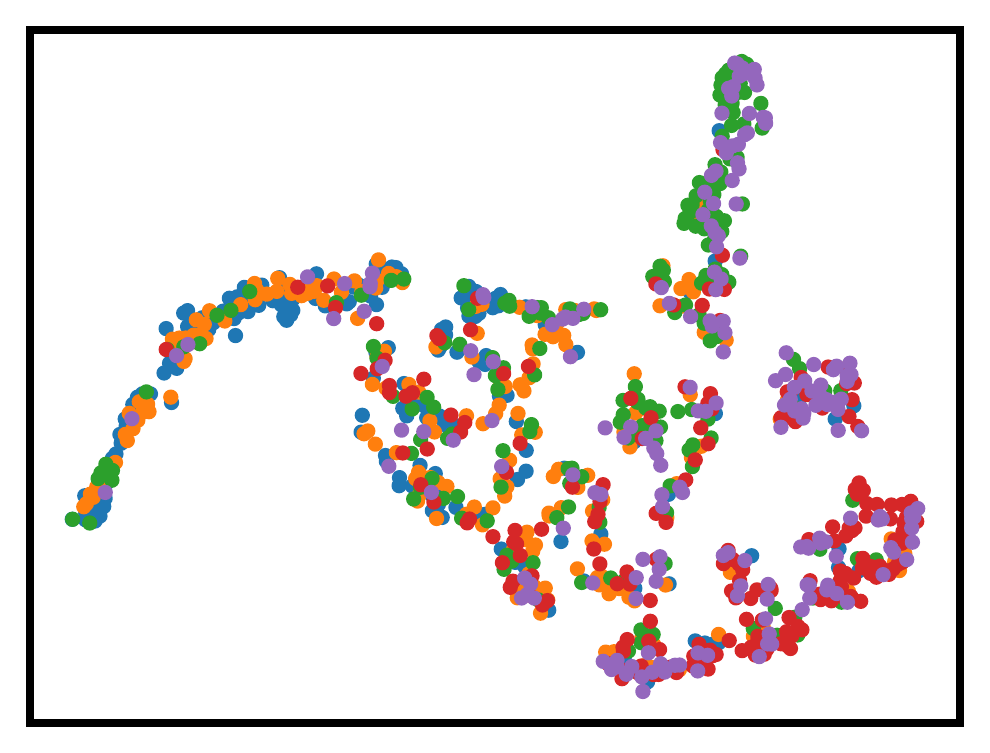}
        \caption{VIME on GE dataset}
    \end{subfigure}
    \caption{The TSNE visualization of coefficients for embedding vectors.}
    \label{appendix:coefficients}
\end{figure}

\begin{figure}[h!]
    \centering
    \captionsetup[sub]{font=scriptsize}
    \begin{subfigure}[b]{0.24\linewidth}
        \centering
        \includegraphics[width=\linewidth]{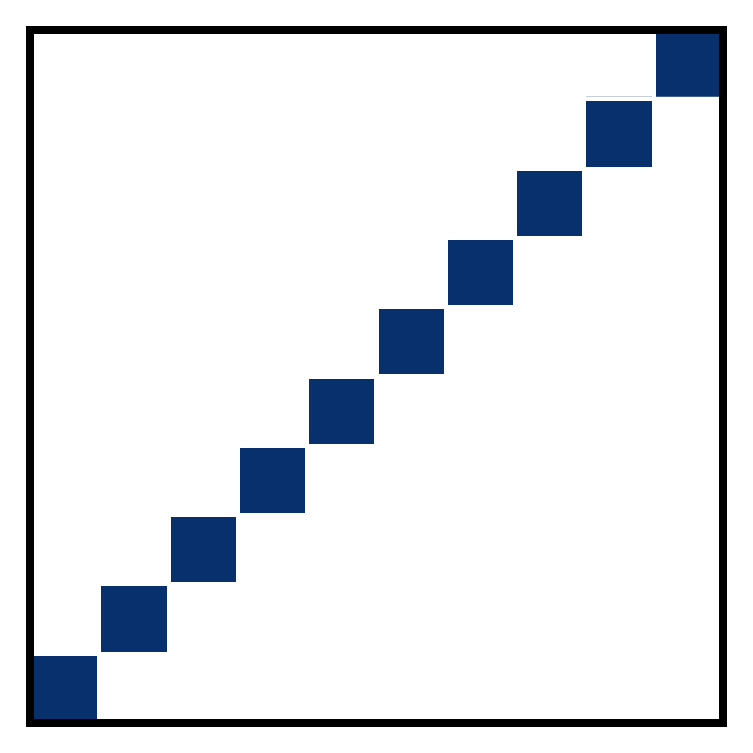}
        \caption{CA dataset}
    \end{subfigure}   
    \hfill
    \begin{subfigure}[b]{0.24\linewidth}
        \centering
        \includegraphics[width=\linewidth]{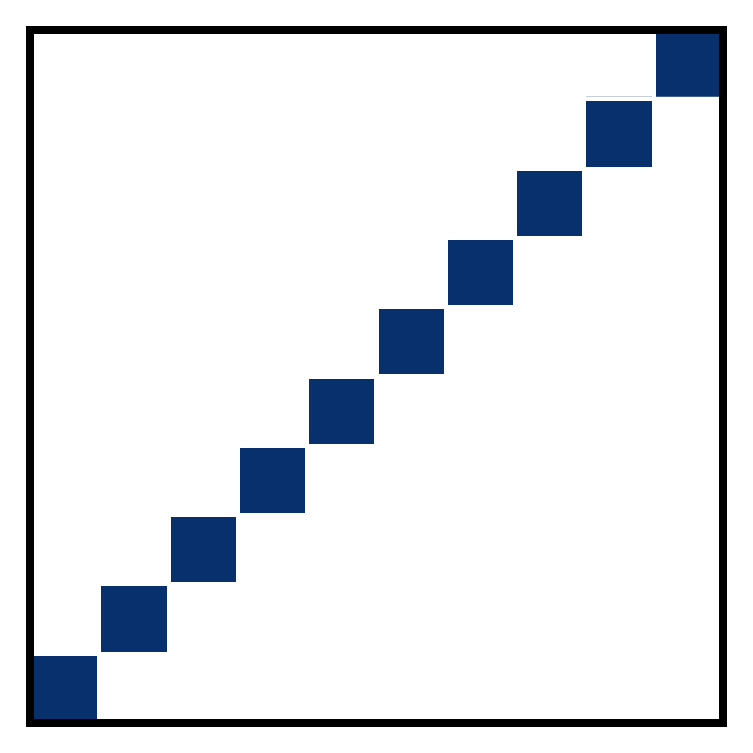}
        \caption{CO dataset}
    \end{subfigure}   
    \hfill
    \begin{subfigure}[b]{0.24\linewidth}
        \centering
        \includegraphics[width=\linewidth]{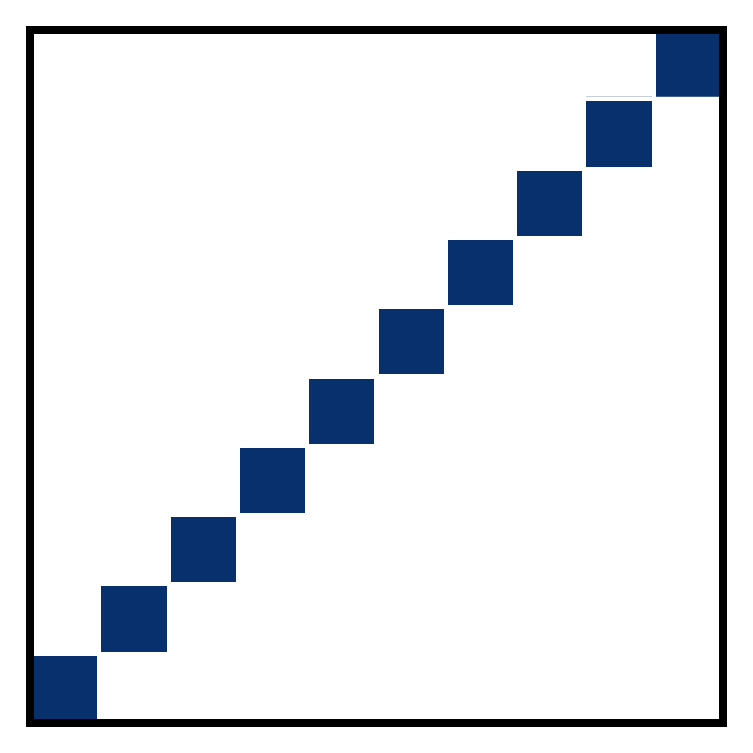}
        \caption{SU dataset}
    \end{subfigure}   
    \hfill
    \begin{subfigure}[b]{0.24\linewidth}
        \centering
        \includegraphics[width=\linewidth]{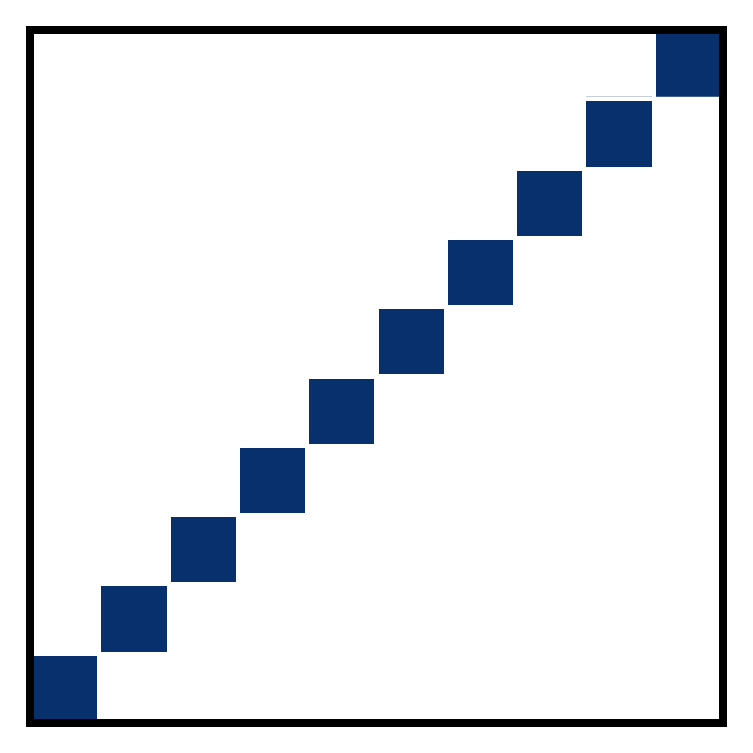}
        \caption{AD dataset}
    \end{subfigure}   
    \caption{The visualization of the similarity matrix for embedding vectors on FT-Transformer.}
    \label{appendix:orth}
\end{figure}

\begin{figure}[!h]
    \centering
    \captionsetup[sub]{font=scriptsize}
    % \begin{subfigure}[b]{0.24\linewidth}
    %     \centering
    %     \includegraphics[width=\linewidth]{pictures/embedding_coeff/adult_AutoInt.pdf}
    %     \caption{AutoInt on AD dataset}
    % \end{subfigure}
    % \begin{subfigure}[b]{0.24\linewidth}
    %     \centering
    %     \includegraphics[width=\linewidth]{pictures/embedding_coeff/adult_DCN2.pdf}
    %     \caption{DCN2 on AD dataset}
    % \end{subfigure}
    % \begin{subfigure}[b]{0.24\linewidth}
    %     \centering
    %     \includegraphics[width=\linewidth]{pictures/embedding_coeff/adult_FTTransformer.pdf}
    %     \caption{FT-Transformer on AD dataset}
    % \end{subfigure}
    % \begin{subfigure}[b]{0.24\linewidth}
    %     \centering
    %     \includegraphics[width=\linewidth]{pictures/embedding_coeff/adult_MLP.pdf}
    %     \caption{MLP on AD dataset}
    % \end{subfigure}
    % \begin{subfigure}[b]{0.24\linewidth}
    %     \centering
    %     \includegraphics[width=\linewidth]{pictures/embedding_coeff/adult_ResNet.pdf}
    %     \caption{ResNet on AD dataset}
    % \end{subfigure}
    % \begin{subfigure}[b]{0.24\linewidth}
    %     \centering
    %     \includegraphics[width=\linewidth]{pictures/embedding_coeff/adult_Saint.pdf}
    %     \caption{SAINT on AD dataset}
    % \end{subfigure}
    % \begin{subfigure}[b]{0.24\linewidth}
    %     \centering
    %     \includegraphics[width=\linewidth]{pictures/embedding_coeff/adult_SNN.pdf}
    %     \caption{SNN on AD dataset}
    % \end{subfigure}
    % \begin{subfigure}[b]{0.24\linewidth}
    %     \centering
    %     \includegraphics[width=\linewidth]{pictures/embedding_coeff/adult_VIME.pdf}
    %     \caption{VIME on AD dataset}
    % \end{subfigure}
    \begin{subfigure}[b]{0.24\linewidth}
        \centering
        \includegraphics[width=\linewidth]{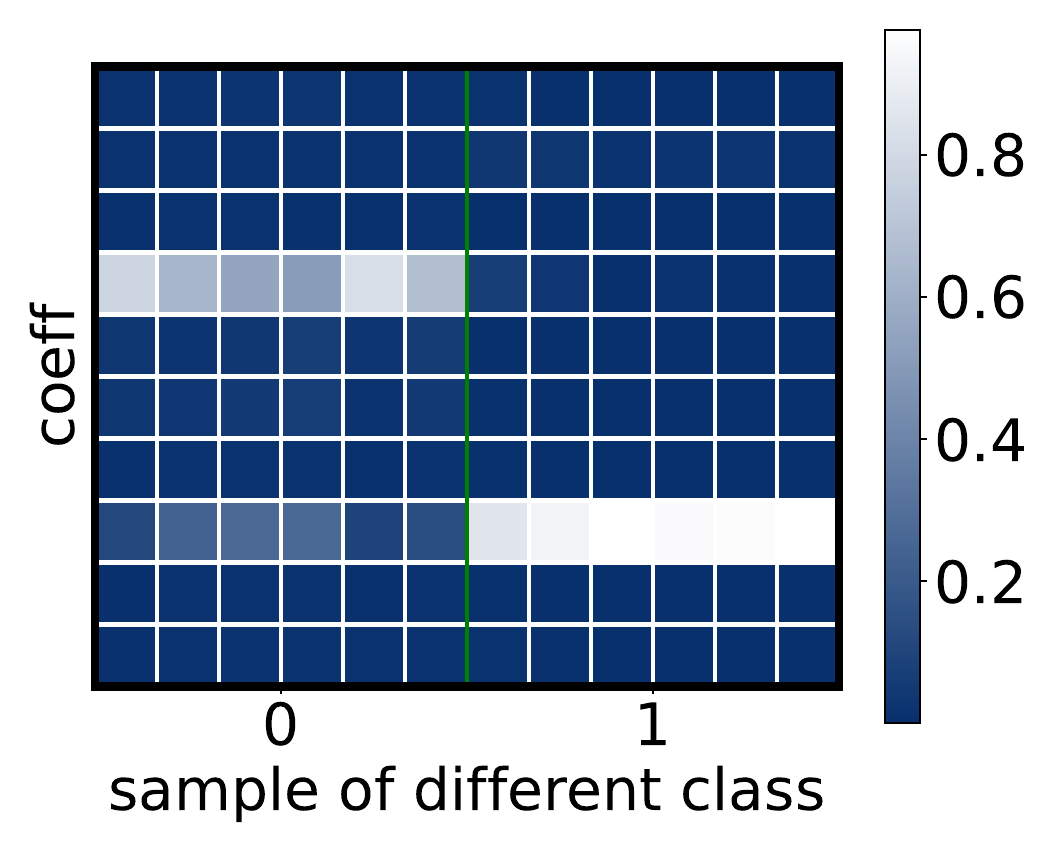}
        \caption{AutoInt on AU dataste}
    \end{subfigure}
    \begin{subfigure}[b]{0.24\linewidth}
        \centering
        \includegraphics[width=\linewidth]{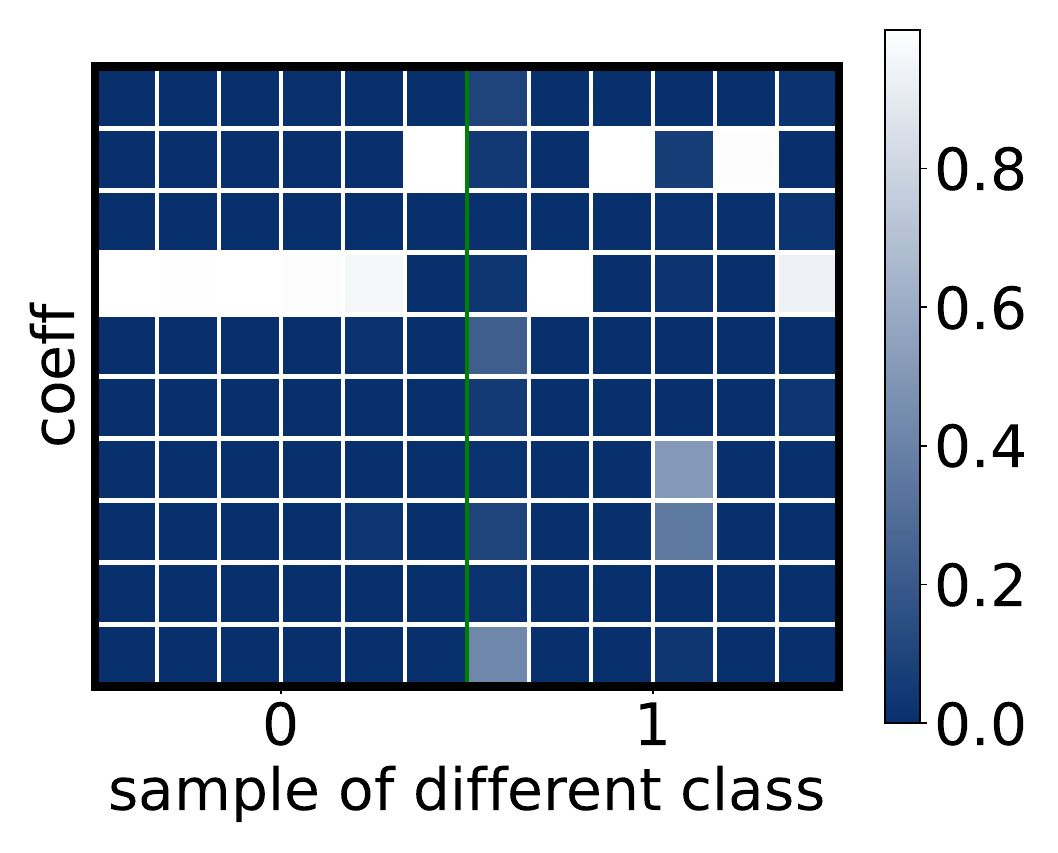}
        \caption{DCN2 on AU dataset}
    \end{subfigure}
    \begin{subfigure}[b]{0.24\linewidth}
        \centering
        \includegraphics[width=\linewidth]{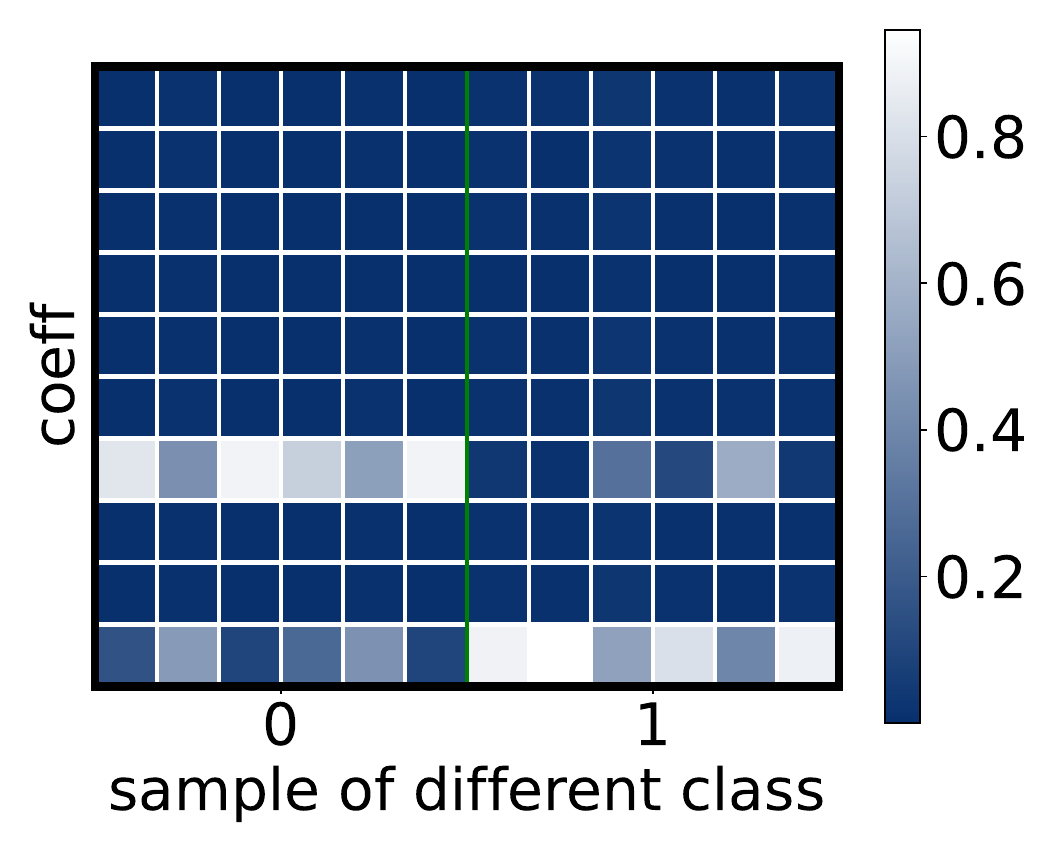}
        \caption{FT-Transformer on AU dataset}
    \end{subfigure}
    \begin{subfigure}[b]{0.24\linewidth}
        \centering
        \includegraphics[width=\linewidth]{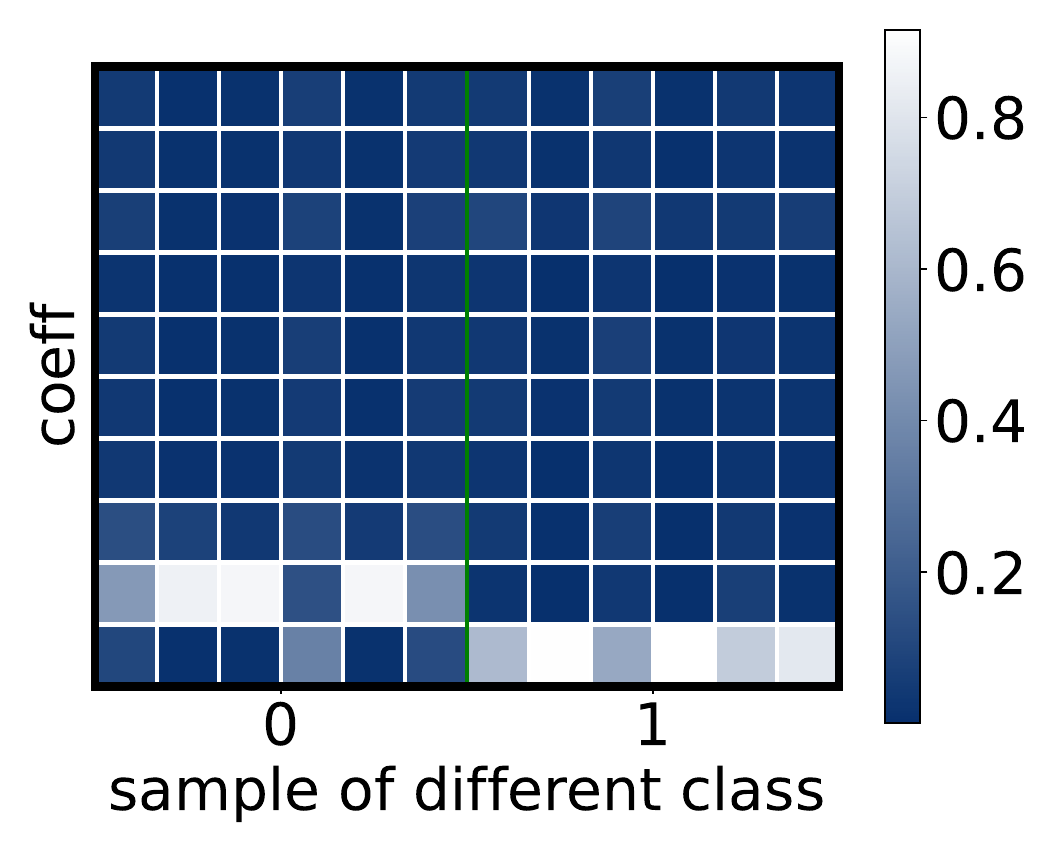}
        \caption{MLP on AU dataset}
    \end{subfigure}

    \begin{subfigure}[b]{0.24\linewidth}
        \centering
        \includegraphics[width=\linewidth]{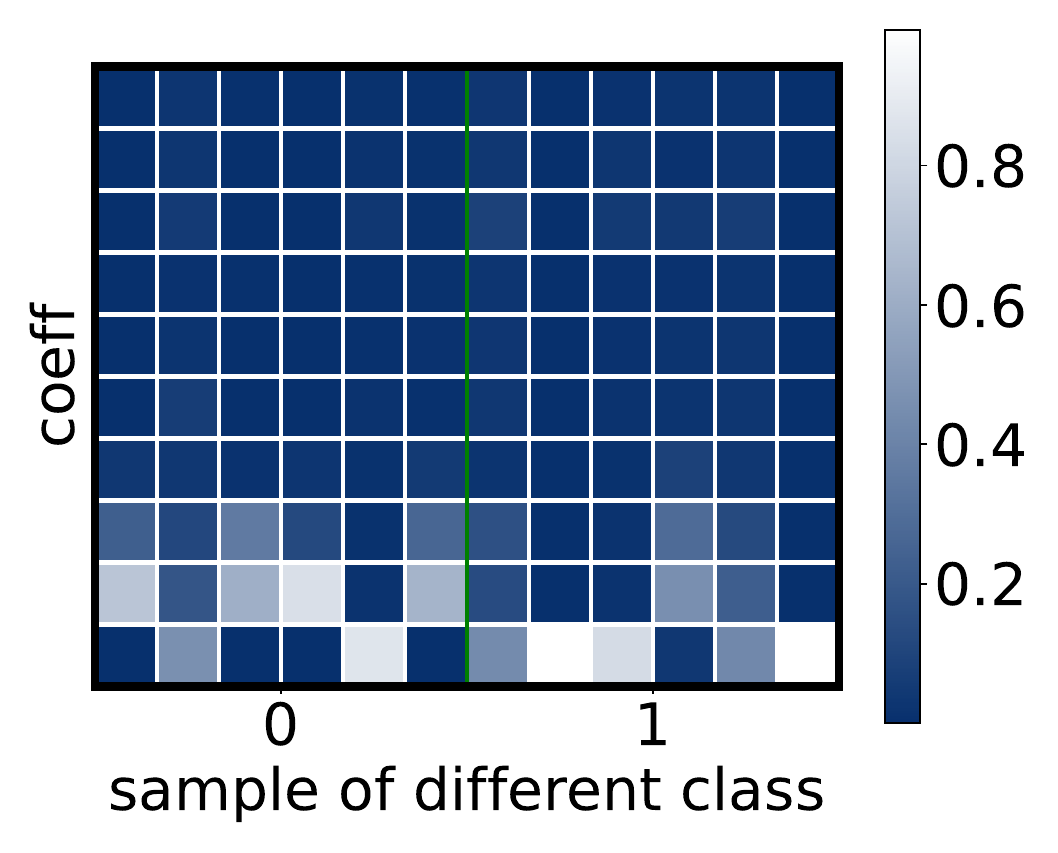}
        \caption{ResNet on AU dataset}
    \end{subfigure}
    \begin{subfigure}[b]{0.24\linewidth}
        \centering
        \includegraphics[width=\linewidth]{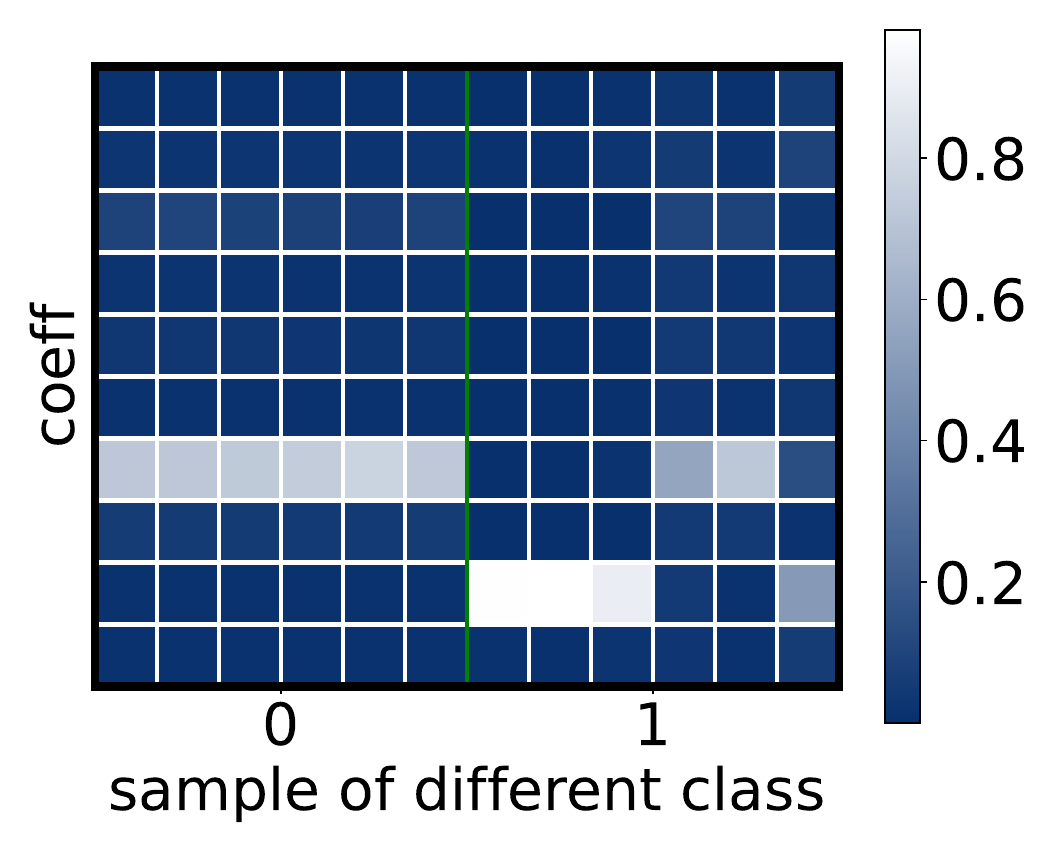}
        \caption{SAINT on AU dataset}
    \end{subfigure}
    \begin{subfigure}[b]{0.24\linewidth}
        \centering
        \includegraphics[width=\linewidth]{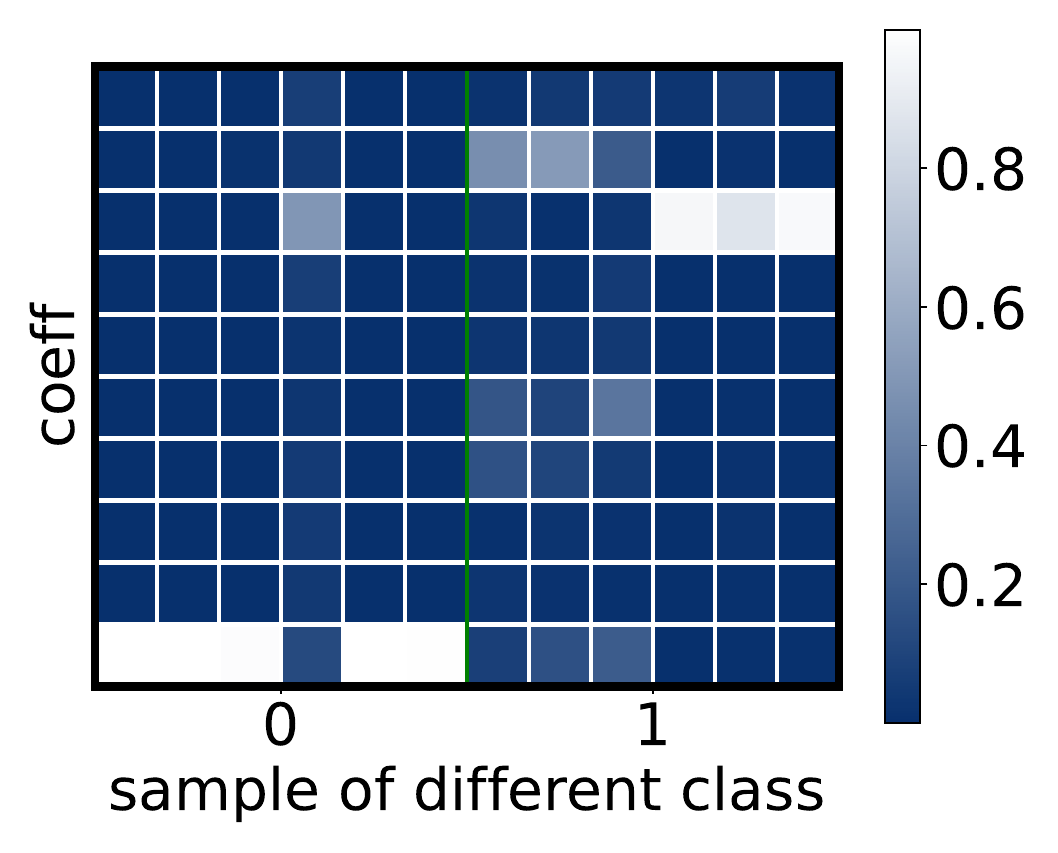}
        \caption{SNN on AU dataset}
    \end{subfigure}
    \begin{subfigure}[b]{0.24\linewidth}
        \centering
        \includegraphics[width=\linewidth]{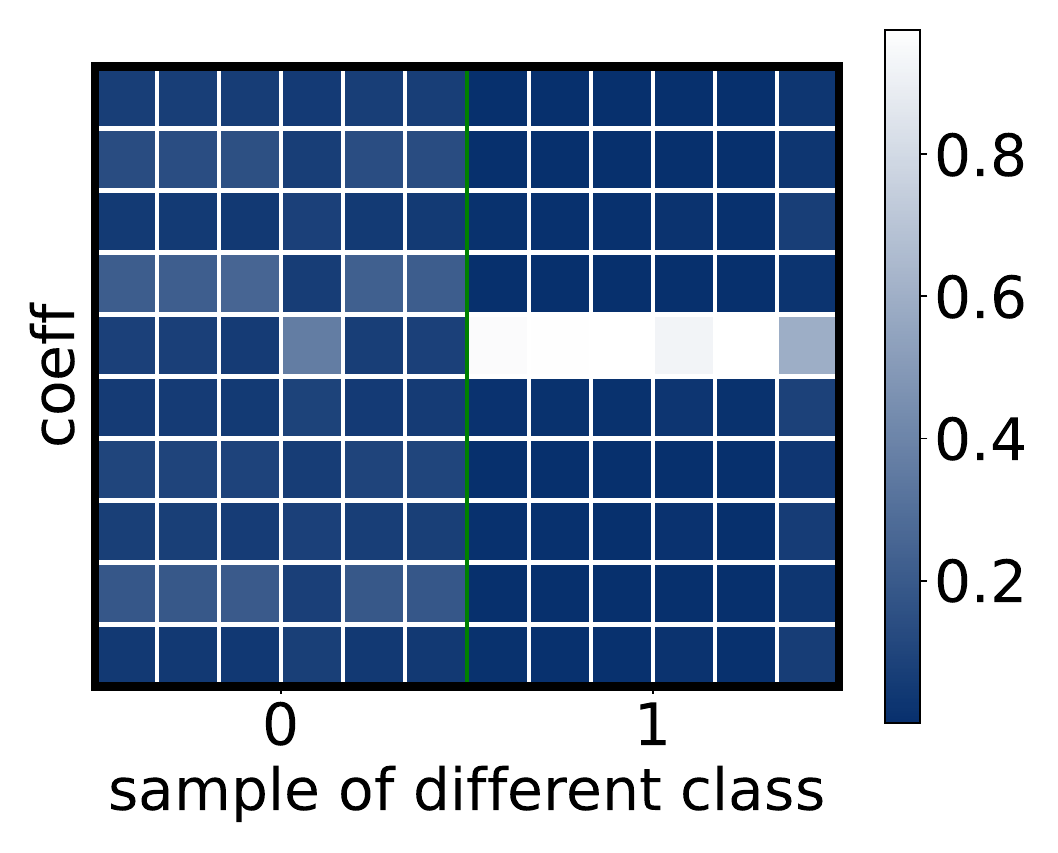}
        \caption{VIME on AU dataset}
    \end{subfigure}
    \begin{subfigure}[b]{0.24\linewidth}
        \centering
        \includegraphics[width=\linewidth]{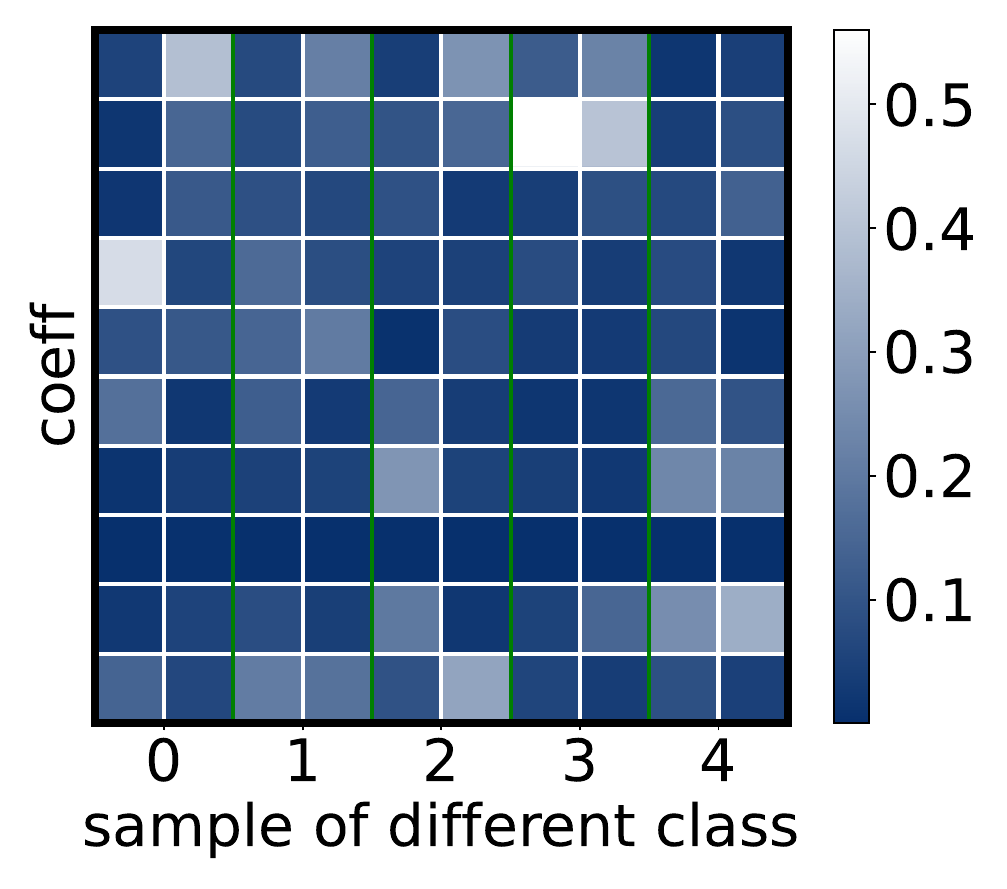}
        \caption{AutoInt on GE dataset}
    \end{subfigure}
    \begin{subfigure}[b]{0.24\linewidth}
        \centering
        \includegraphics[width=\linewidth]{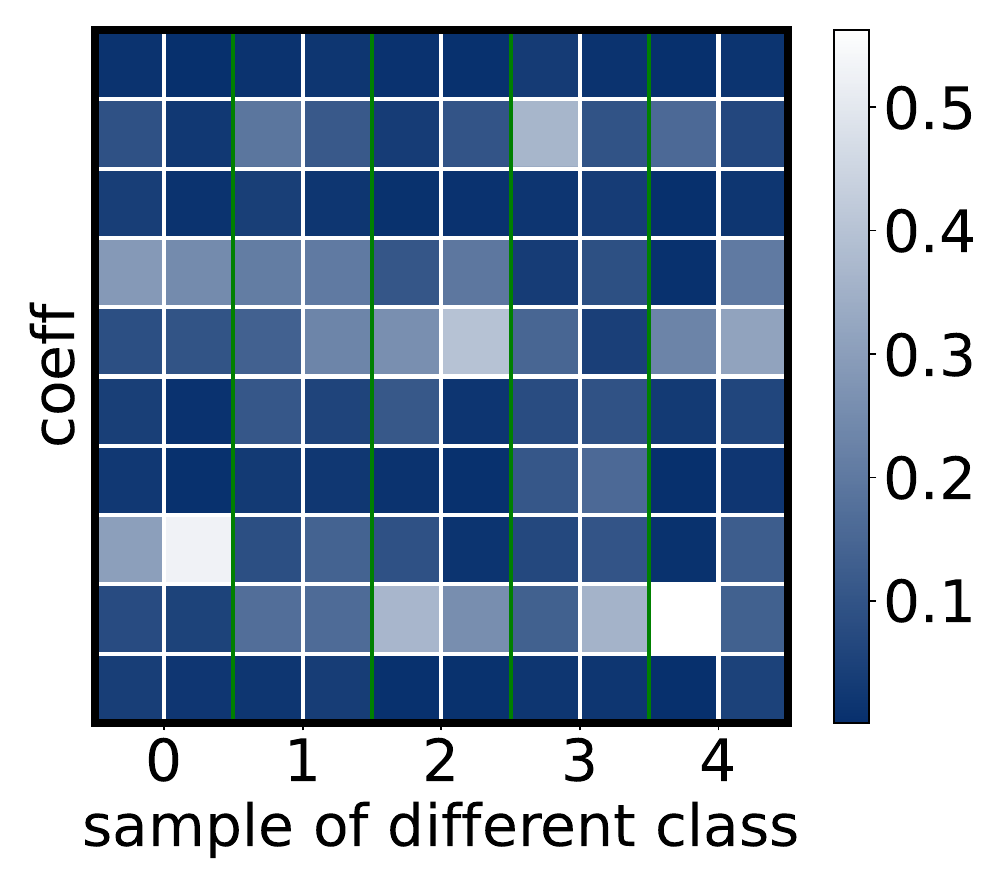}
        \caption{DCN2 on GE dataset}
    \end{subfigure}
    \begin{subfigure}[b]{0.24\linewidth}
        \centering
        \includegraphics[width=\linewidth]{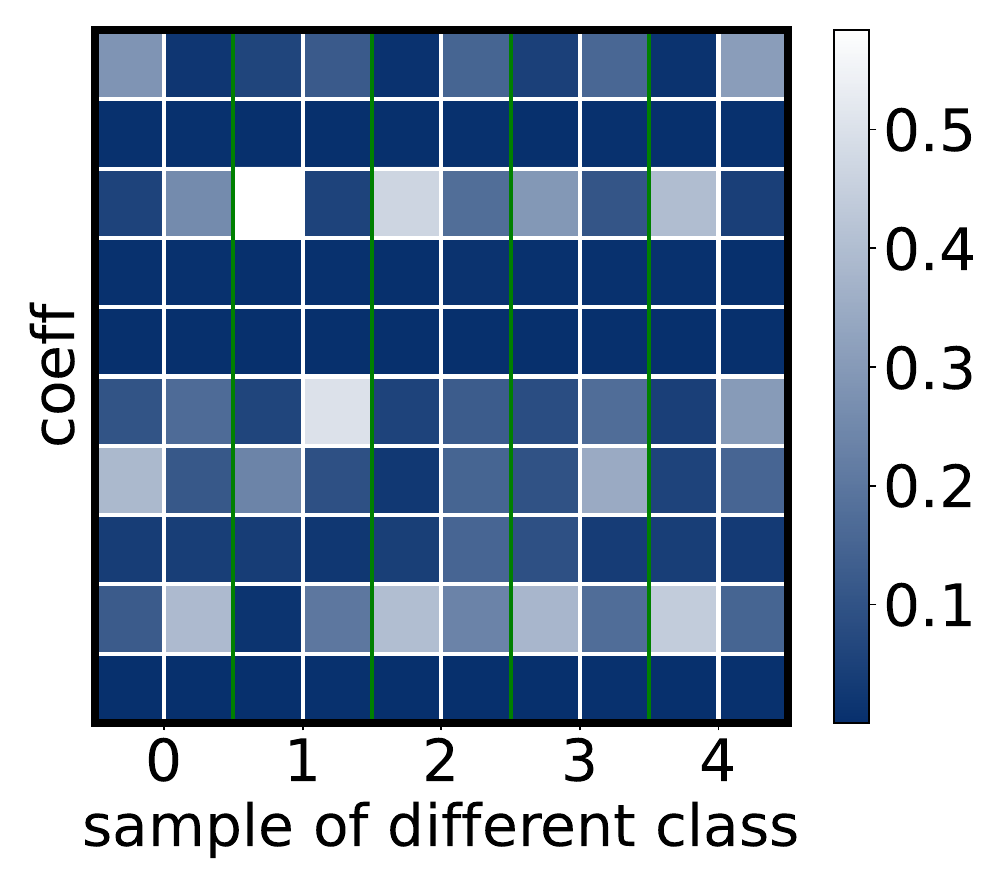}
        \caption{FT-Transformer on GE dataset}
    \end{subfigure}
    \begin{subfigure}[b]{0.24\linewidth}
        \centering
        \includegraphics[width=\linewidth]{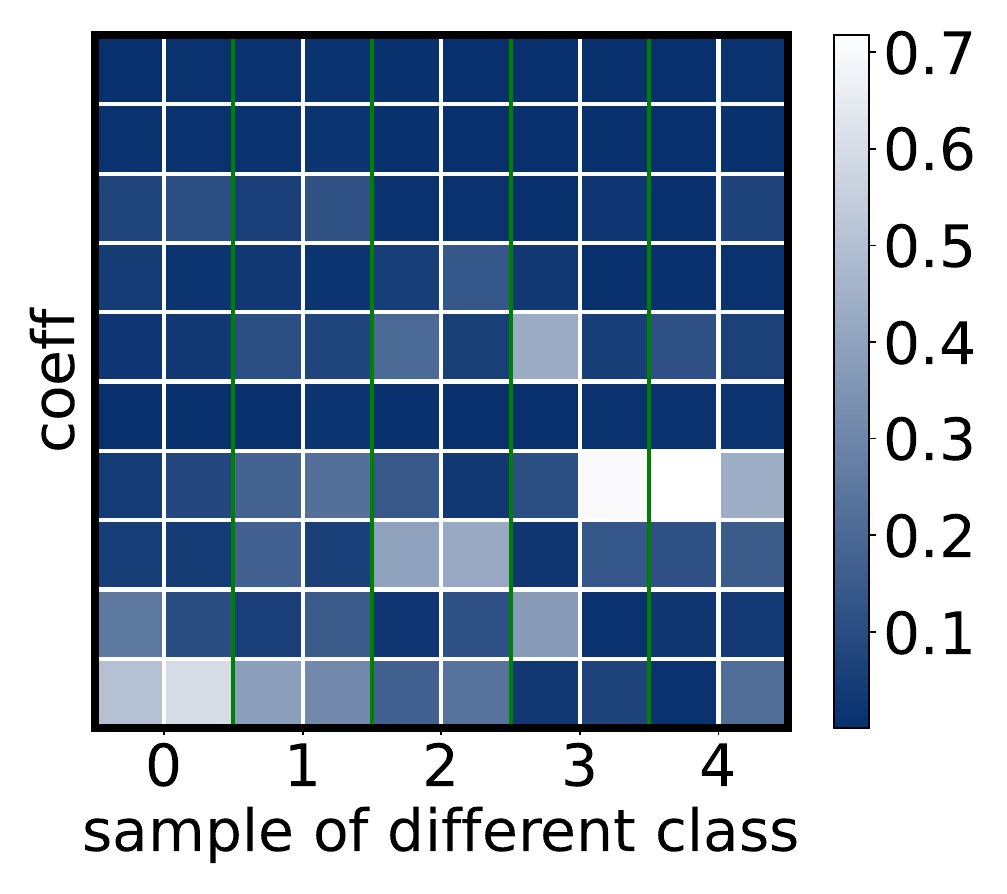}
        \caption{MLP on GE dataset}
    \end{subfigure}
    \begin{subfigure}[b]{0.24\linewidth}
        \centering
        \includegraphics[width=\linewidth]{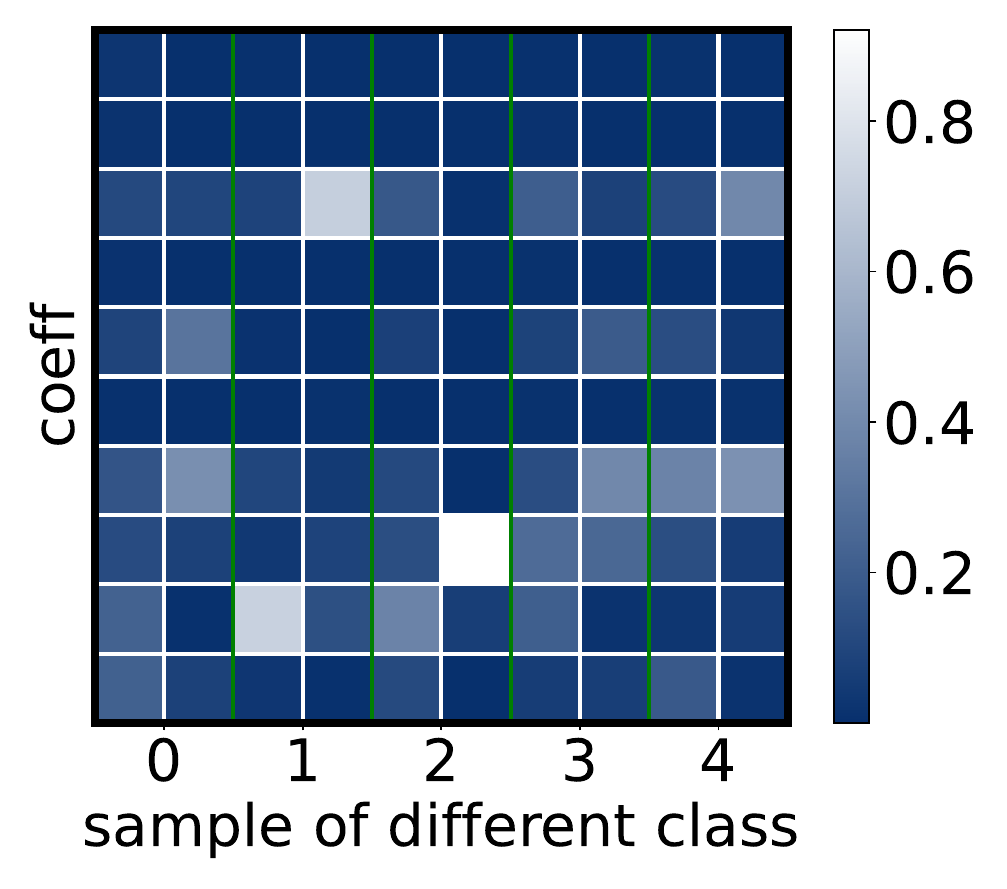}
        \caption{ResNet on GE dataset}
    \end{subfigure}
    \begin{subfigure}[b]{0.24\linewidth}
        \centering
        \includegraphics[width=\linewidth]{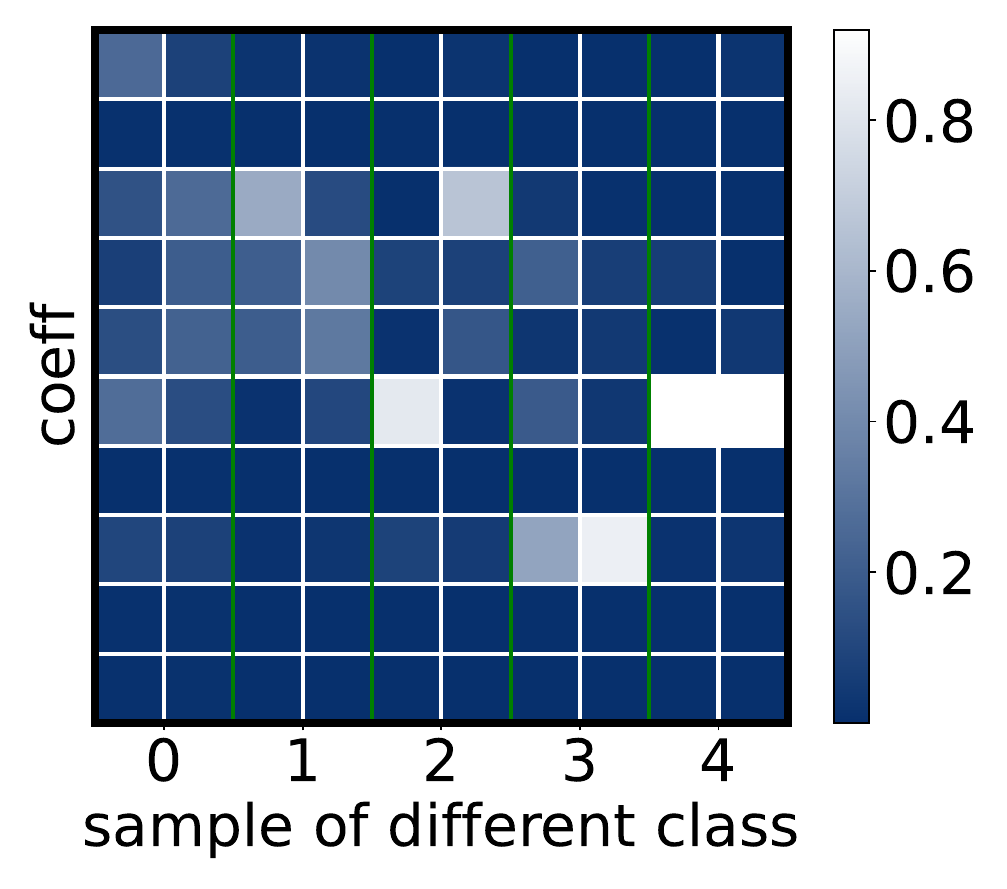}
        \caption{SAINT on GE dataset}
    \end{subfigure}
    \begin{subfigure}[b]{0.24\linewidth}
        \centering
        \includegraphics[width=\linewidth]{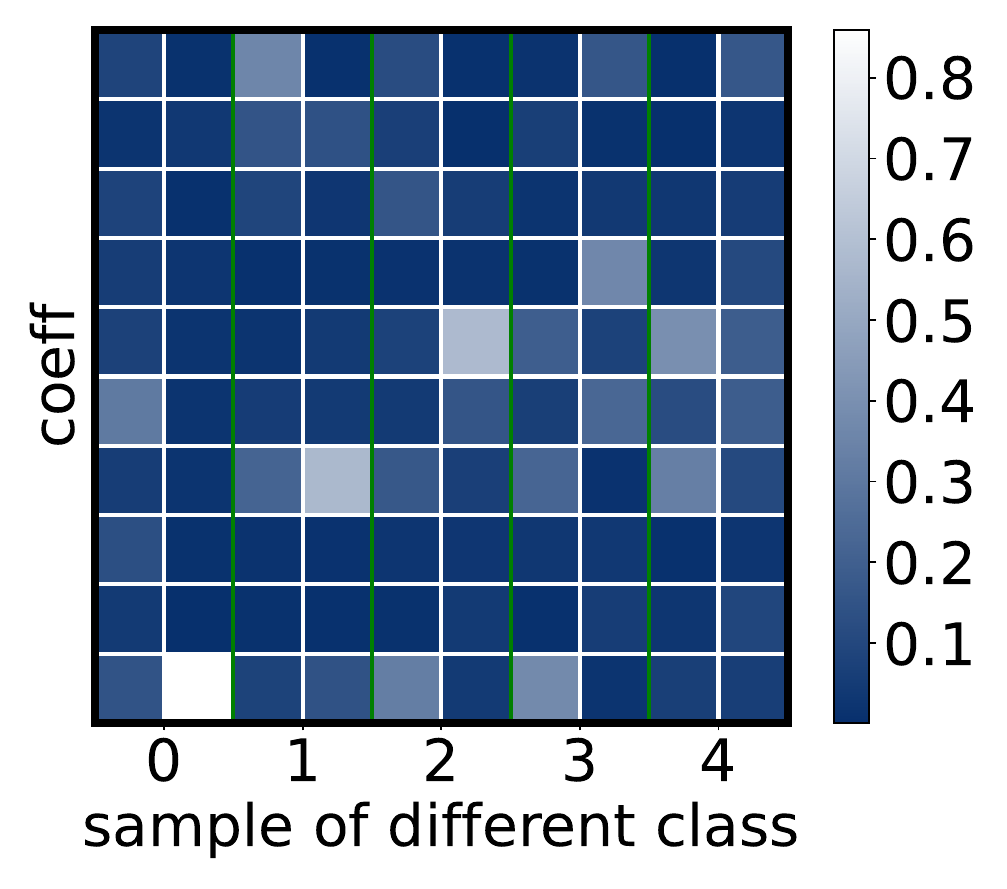}
        \caption{SNN on GE dataset}
    \end{subfigure}
    \begin{subfigure}[b]{0.24\linewidth}
        \centering
        \includegraphics[width=\linewidth]{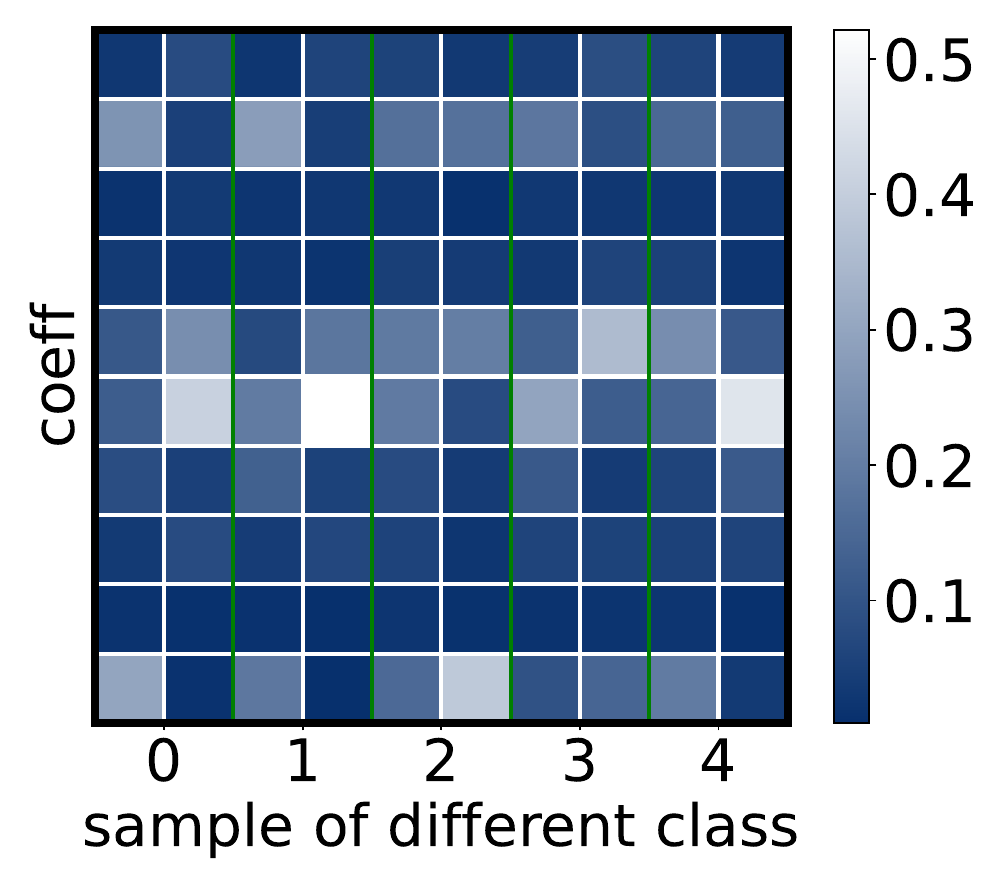}
        \caption{VIME on GE dataset}
    \end{subfigure}
    \caption{The visualization of coefficients for embedding vectors.}
    \label{appendix:heatmap coefficients}
\end{figure}

We provide the TSNE visualization of coefficients for embedding vectors in Fig.~\ref{appendix:coefficients}, the visualization of the similarity matrix for embedding vectors in Fig.~\ref{appendix:orth}, and the heatmap of coefficients for embedding vectors in Fig.~\ref{appendix:heatmap coefficients}.
% \begin{figure}
%     \centering
%     \hfill
%     \begin{subfigure}[b]{0.24\linewidth}
%         \centering
%         \includegraphics[width=\linewidth]{pictures/Tsne/california_housing_ResNet.pdf}
%         % \captionsetup{skip=0pt}
%         \caption{ResNet on CA}
%     \end{subfigure}
%     \hfill
%     \begin{subfigure}[b]{0.24\linewidth}
%         \centering
%         \includegraphics[width=\linewidth]{pictures/Tsne/california_housing_ResNet_TPL.pdf}
%         % \captionsetup{skip=0pt}
%         \caption{ResNet w/ TRC on CA}
%     \end{subfigure}
%     \hfill
%     \begin{subfigure}[b]{0.24\linewidth}
%         \centering
%         \includegraphics[width=\linewidth]{pictures/Tsne/GesturePhaseSegmentationProcessed_ResNet.pdf}
%         % \captionsetup{skip=0pt}
%         \caption{ResNet on GE}
%     \end{subfigure}
%     \hfill
%     \begin{subfigure}[b]{0.24\linewidth}
%         \centering
%         \includegraphics[width=\linewidth]{pictures/Tsne/GesturePhaseSegmentationProcessed_ResNet_TPL.pdf}
%         % \captionsetup{skip=0pt}
%         \caption{ResNet w/ TRC on GE}
%     \end{subfigure}
%     \hfill
%     % \captionsetup{skip=0pt}
%     \caption{The TSNE visualization of learned representations of deep tabular machine learning models w/o and w/ \method. Different colors indicate different labels. Marker=``x'' indicates embedding vectors.}
%     \label{appendix:representation}
% \end{figure}

% \clearpage
\newpage
\subsection{The Relationship between SVE and the Number of Embedding Vectors}
\begin{figure*}[h!]
    \centering
    \begin{subfigure}[b]{0.24\linewidth}
        \centering
        \includegraphics[width=\linewidth]{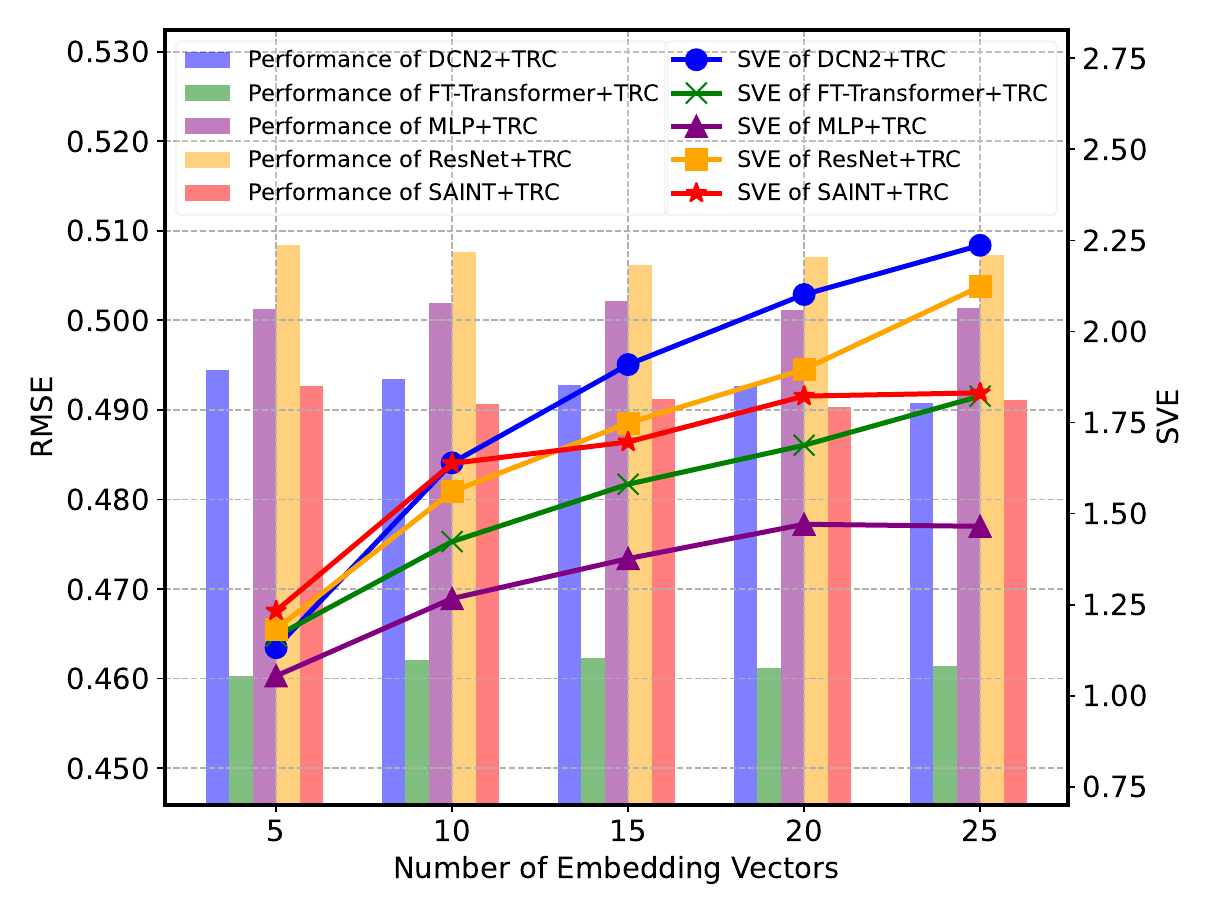}
        \caption{CA dataset}
    \end{subfigure}
    \hfill
    \begin{subfigure}[b]{0.24\linewidth}
        \centering
        \includegraphics[width=\linewidth]{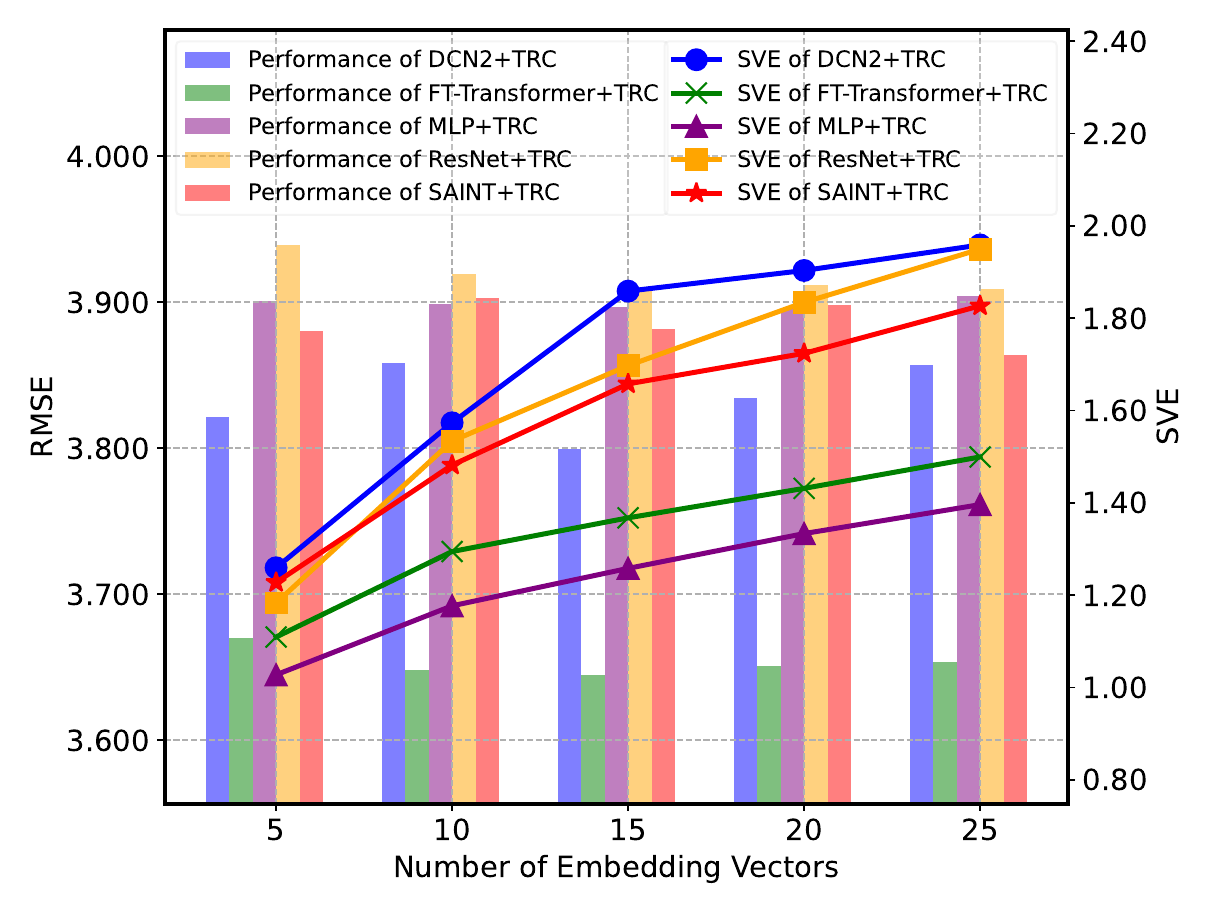}
        \caption{CO dataset}
    \end{subfigure}
    \begin{subfigure}[b]{0.24\linewidth}
        \centering
        \includegraphics[width=\linewidth]{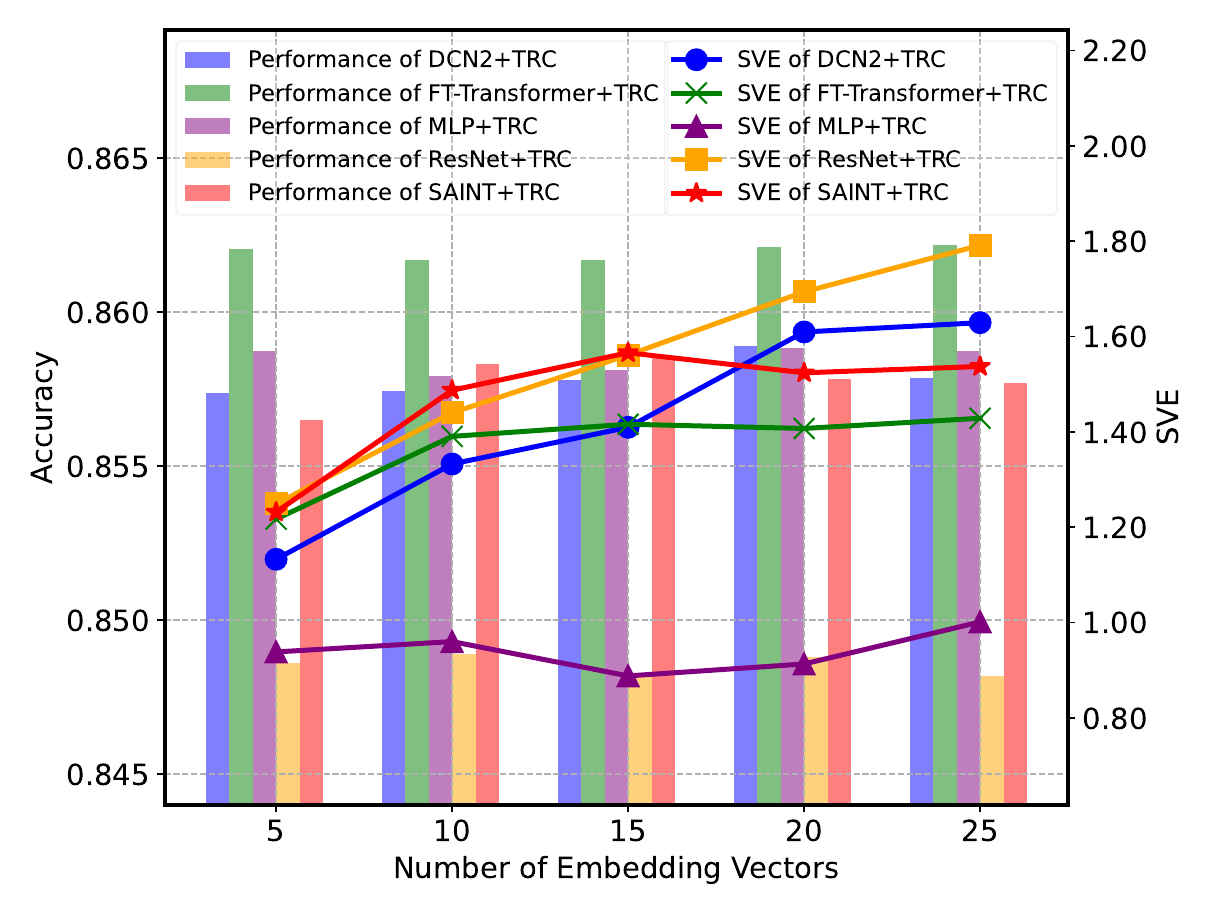}
        \caption{AD dataset}
    \end{subfigure}
    \hfill
    \begin{subfigure}[b]{0.24\linewidth}
        \centering
        \includegraphics[width=\linewidth]{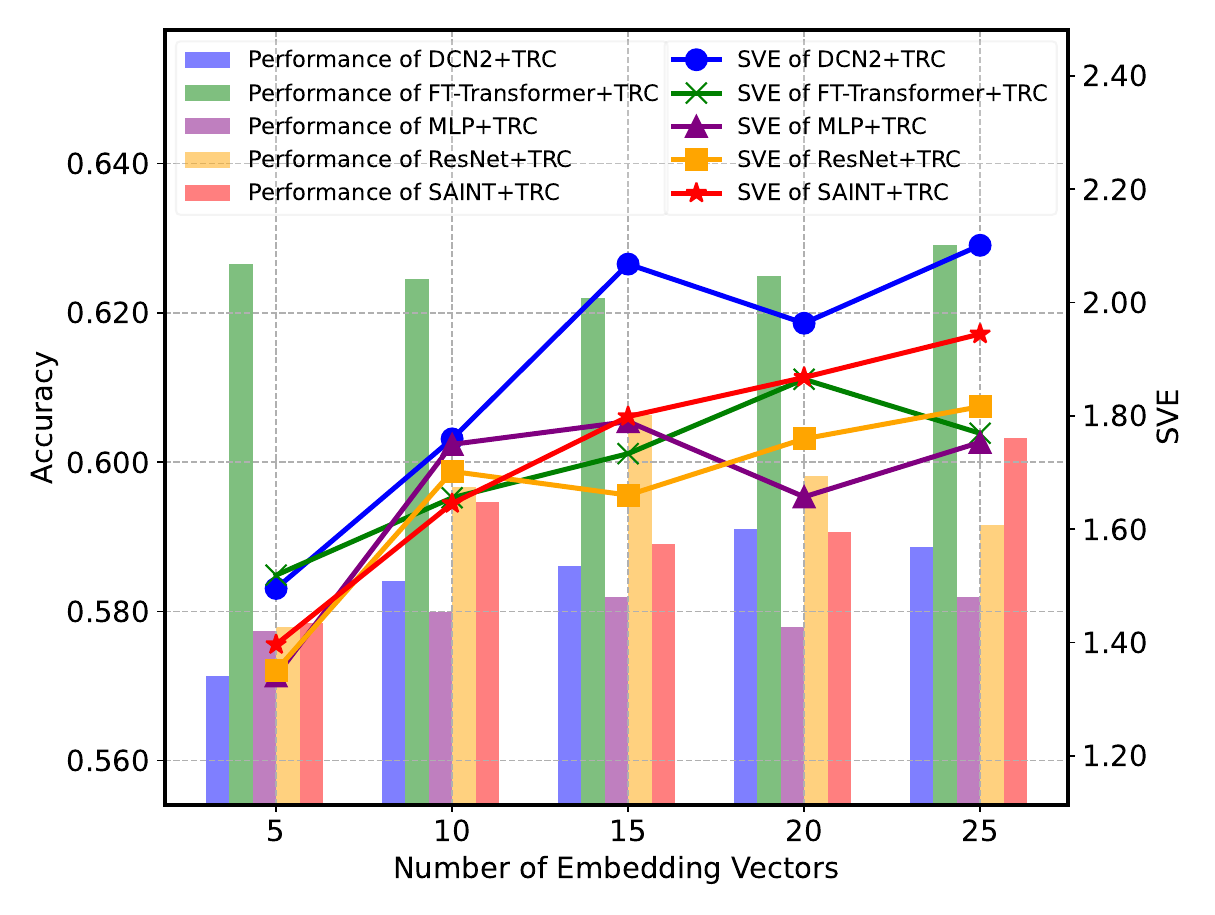}
        \caption{GE dataset}
    \end{subfigure}
    \centering

    \caption{Increasing the number of embedding vectors leads to larger SVE values of representations.}
    \label{appendix:T_SVE}
\end{figure*}

We provide the relationship between SVE and the number of embedding vectors in Fig.~\ref{appendix:T_SVE}.

% \clearpage
\subsection{Sensitivity Analysis}
\label{appendix:sensitivity analysis}

\begin{figure*}[h!]
    \centering
    
    \hfill
    \begin{subfigure}[b]{0.24\linewidth}
        \centering
        \includegraphics[width=\linewidth]{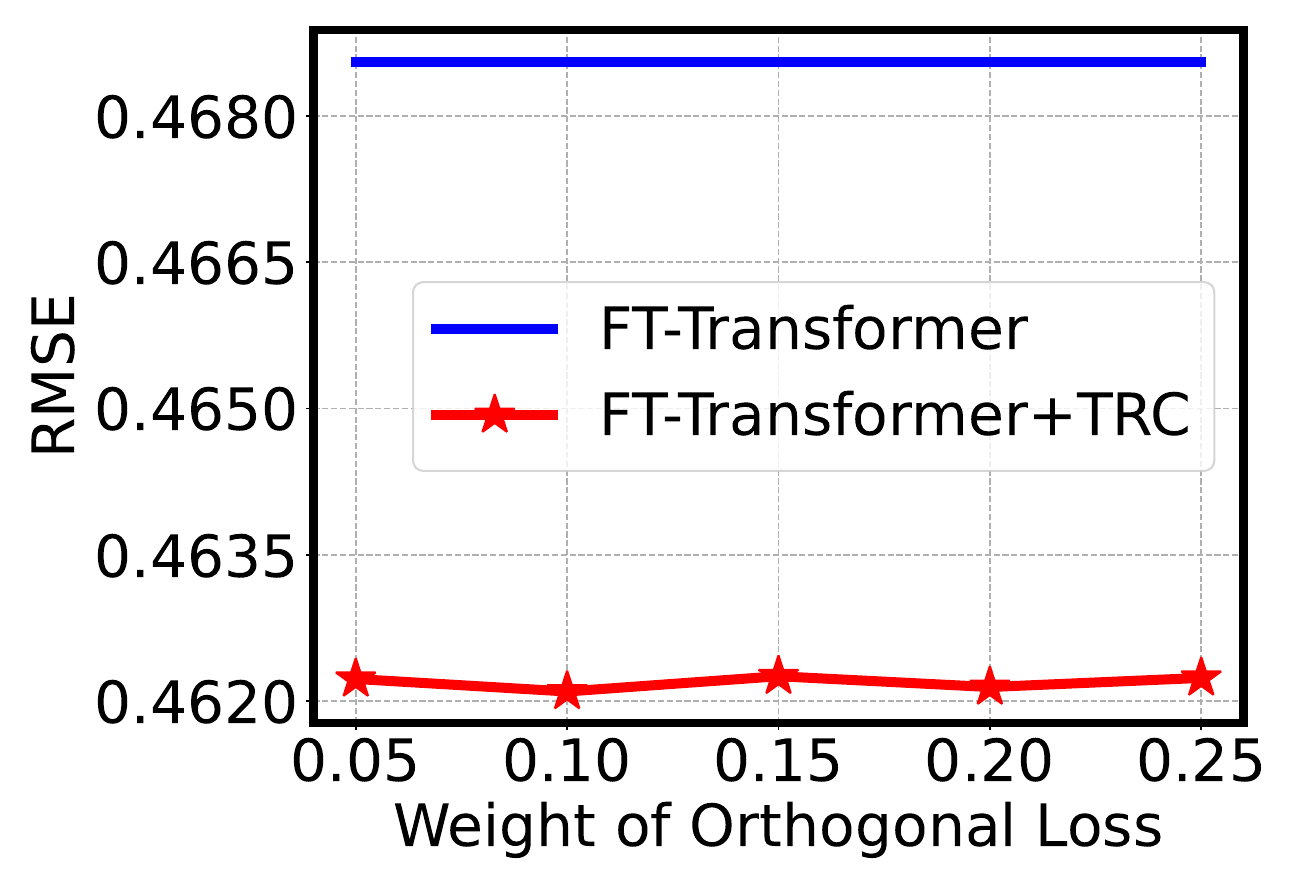}
        \caption{CA dataset $\downarrow$}
    \end{subfigure}
    \hfill
    \begin{subfigure}[b]{0.24\linewidth}
        \centering
        \includegraphics[width=\linewidth]{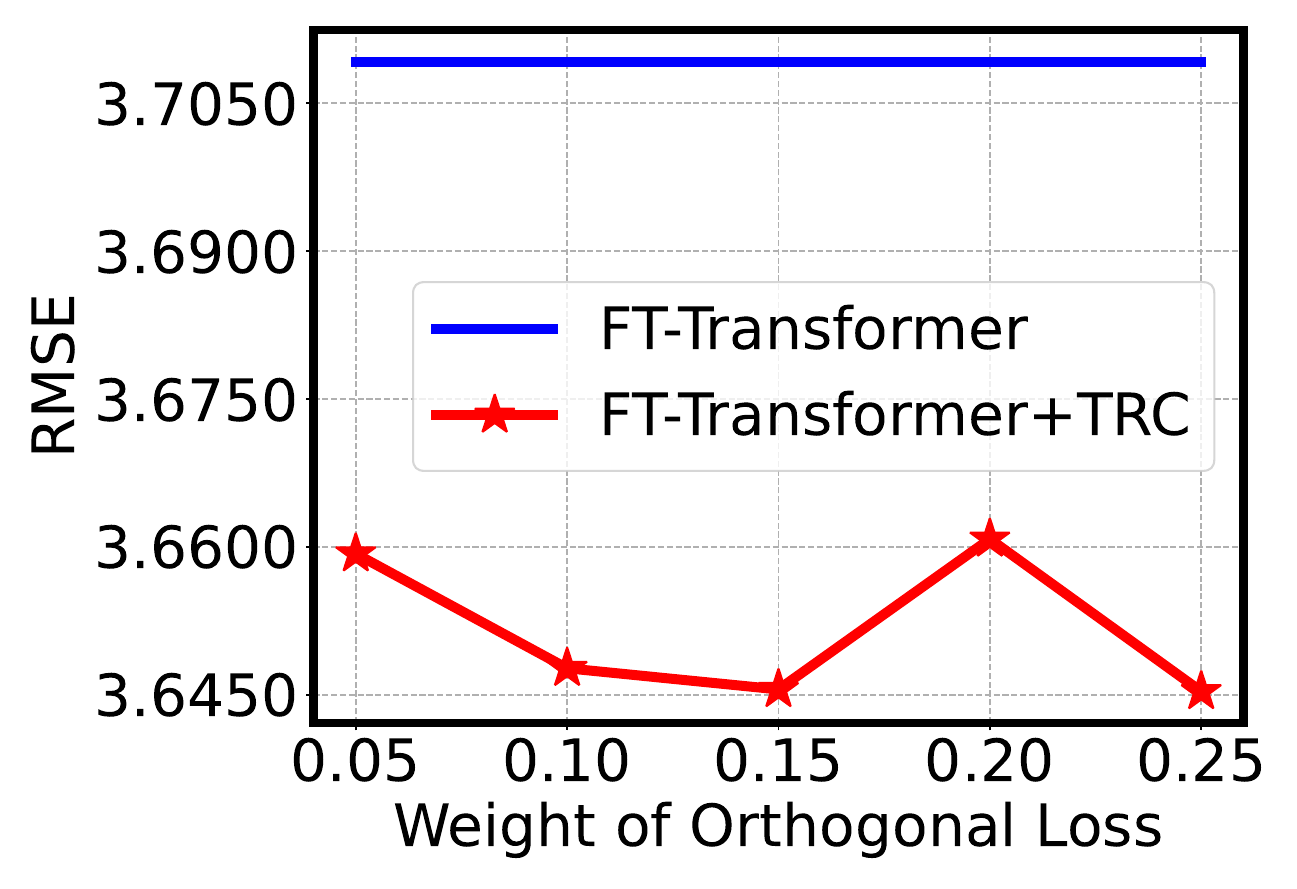}
        \caption{CO dataset $\downarrow$}
    \end{subfigure}
    \hfill
    \begin{subfigure}[b]{0.24\linewidth}
        \centering
        \includegraphics[width=\linewidth]{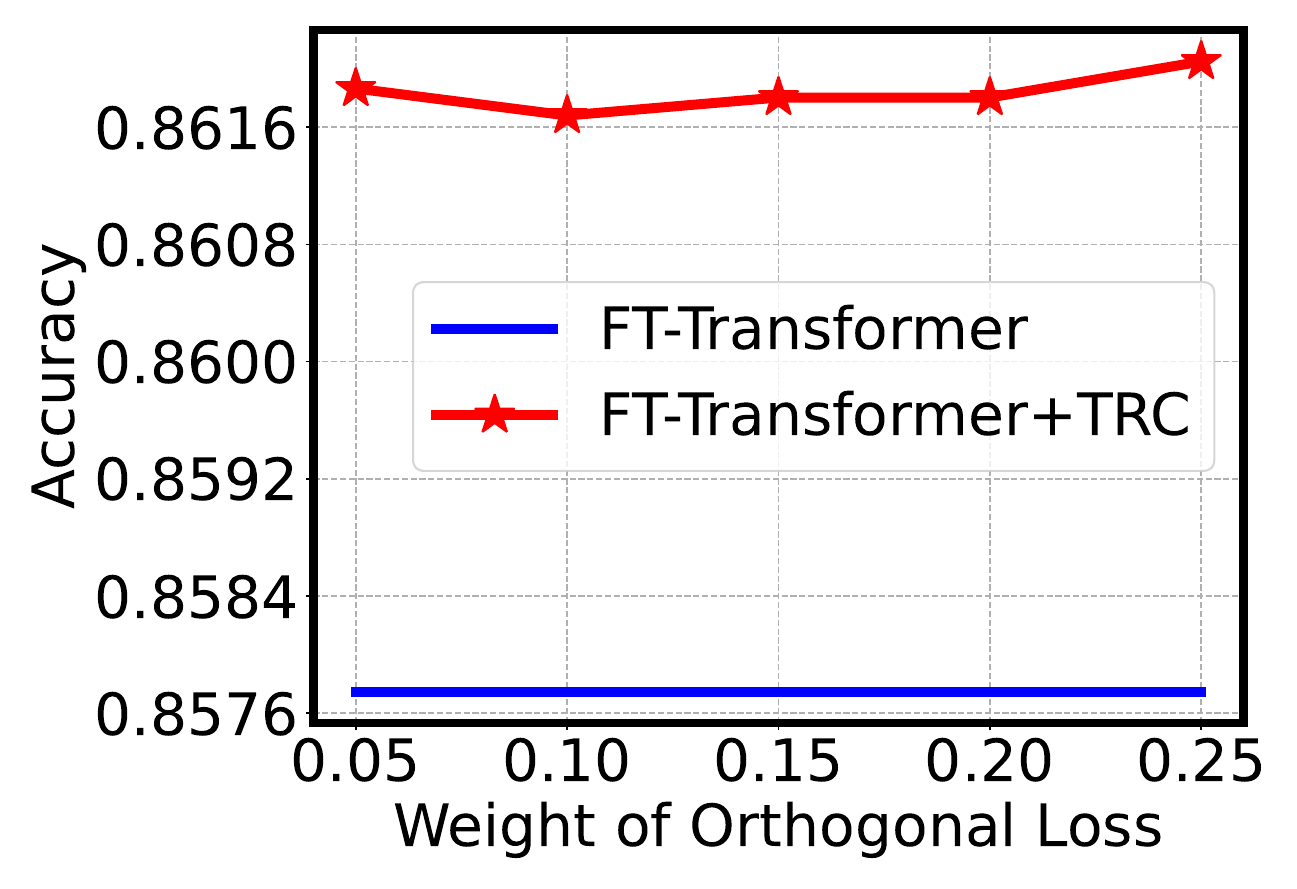}
        \caption{AD dataset $\uparrow$}
    \end{subfigure}
    \hfill
    \begin{subfigure}[b]{0.24\linewidth}
        \centering
        \includegraphics[width=\linewidth]{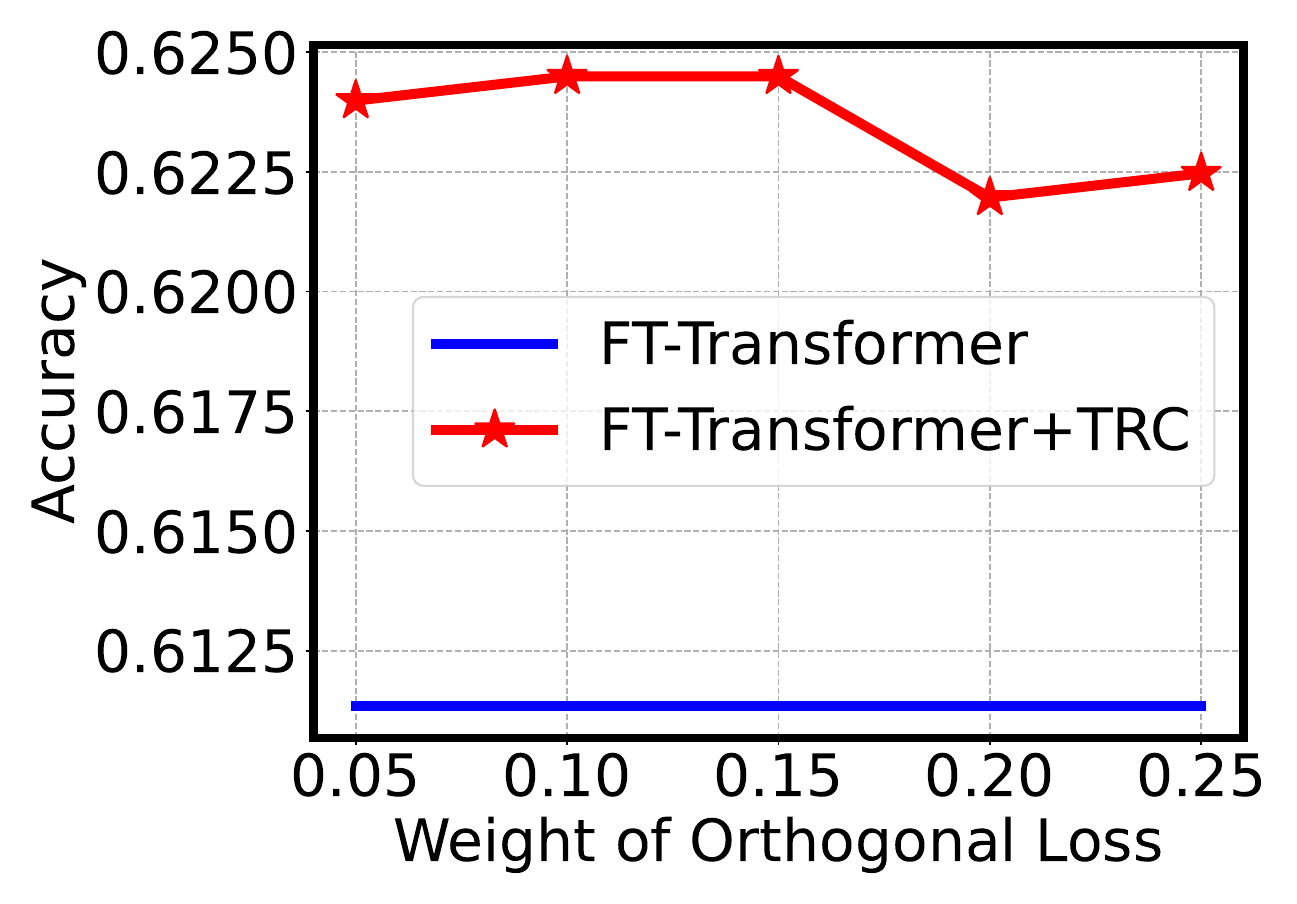}
        \caption{GE dataset $\uparrow$}
    \end{subfigure}
    \hfill
    \caption{The performance of \method with varying weights of the orthogonality loss.}
    \label{appendix:s1}
\end{figure*}

\begin{figure*}[h]
    \centering
    
    \hfill
    \begin{subfigure}[b]{0.24\linewidth}
        \centering
        \includegraphics[width=\linewidth]{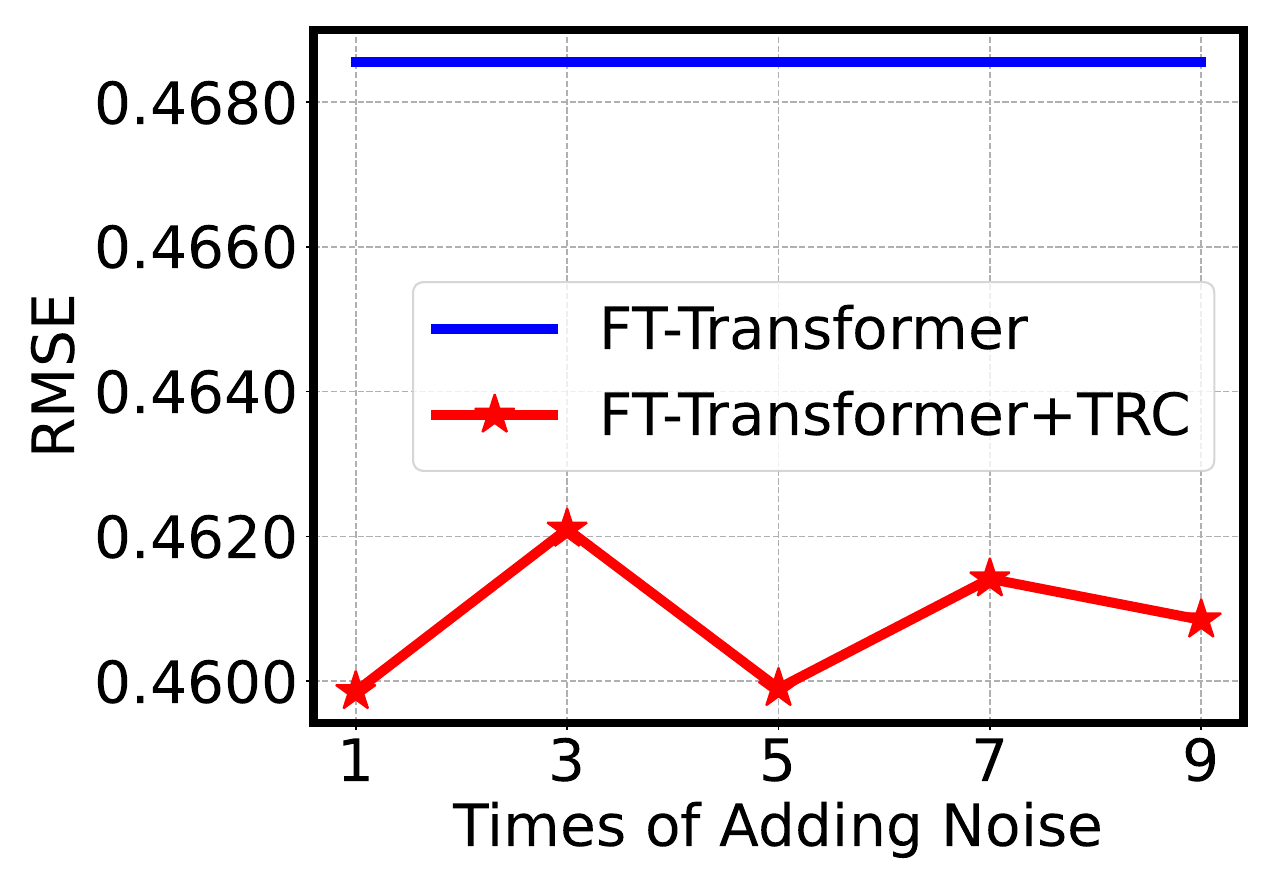}
        \caption{CA dataset $\downarrow$}
    \end{subfigure}
    \hfill
    \begin{subfigure}[b]{0.24\linewidth}
        \centering
        \includegraphics[width=\linewidth]{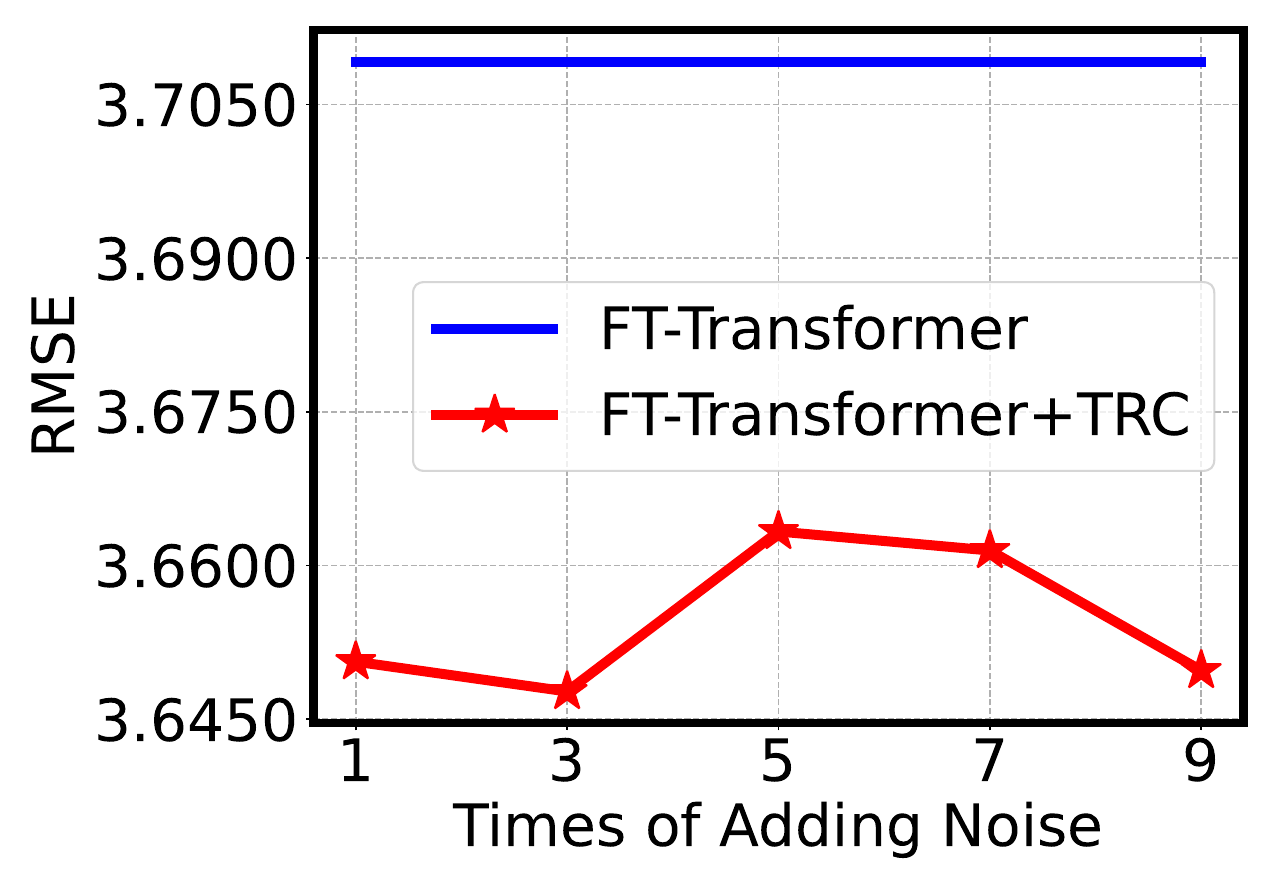}
        \caption{CO dataset $\downarrow$}
    \end{subfigure}
    \hfill
    \begin{subfigure}[b]{0.24\linewidth}
        \centering
        \includegraphics[width=\linewidth]{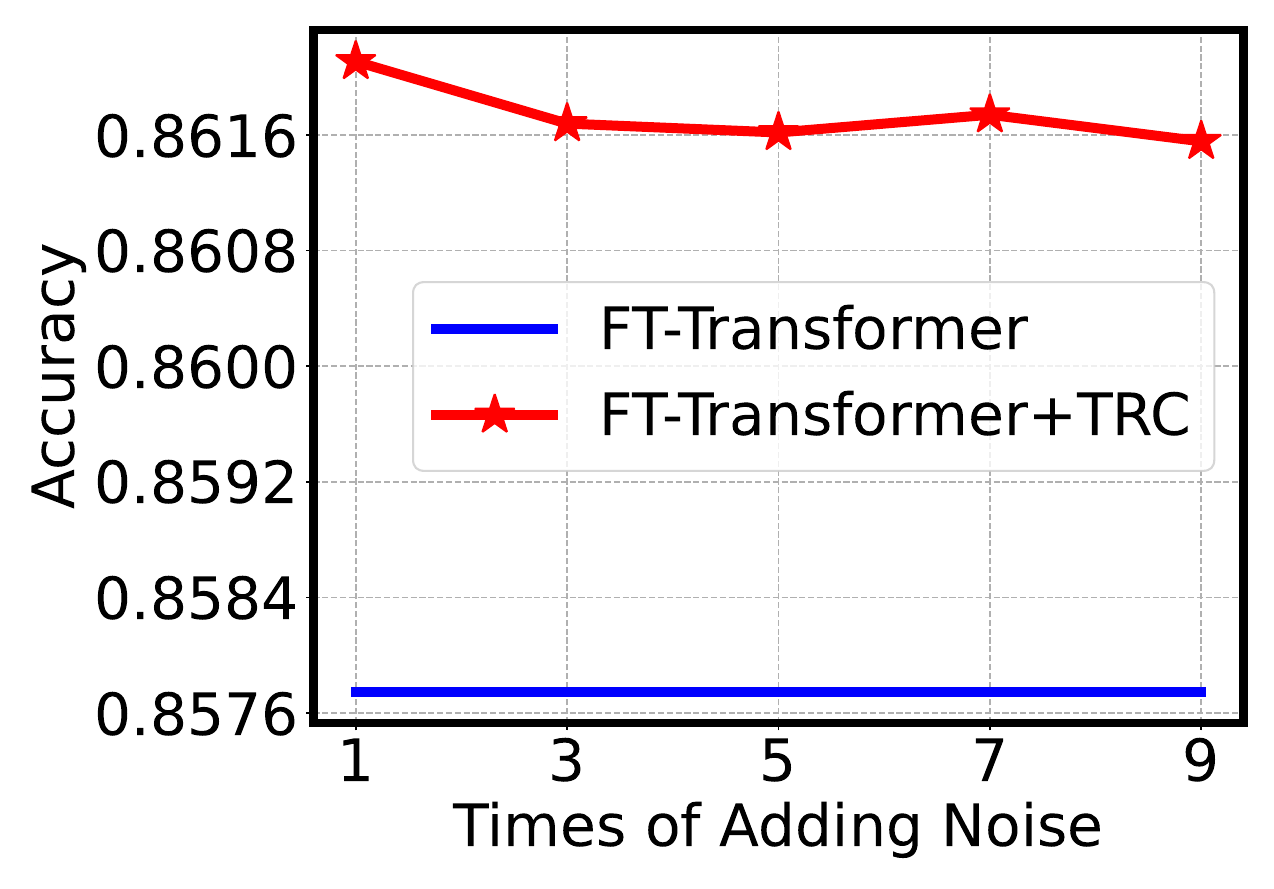}
        \caption{AD dataset $\uparrow$}
    \end{subfigure}
    \hfill
    \begin{subfigure}[b]{0.24\linewidth}
        \centering
        \includegraphics[width=\linewidth]{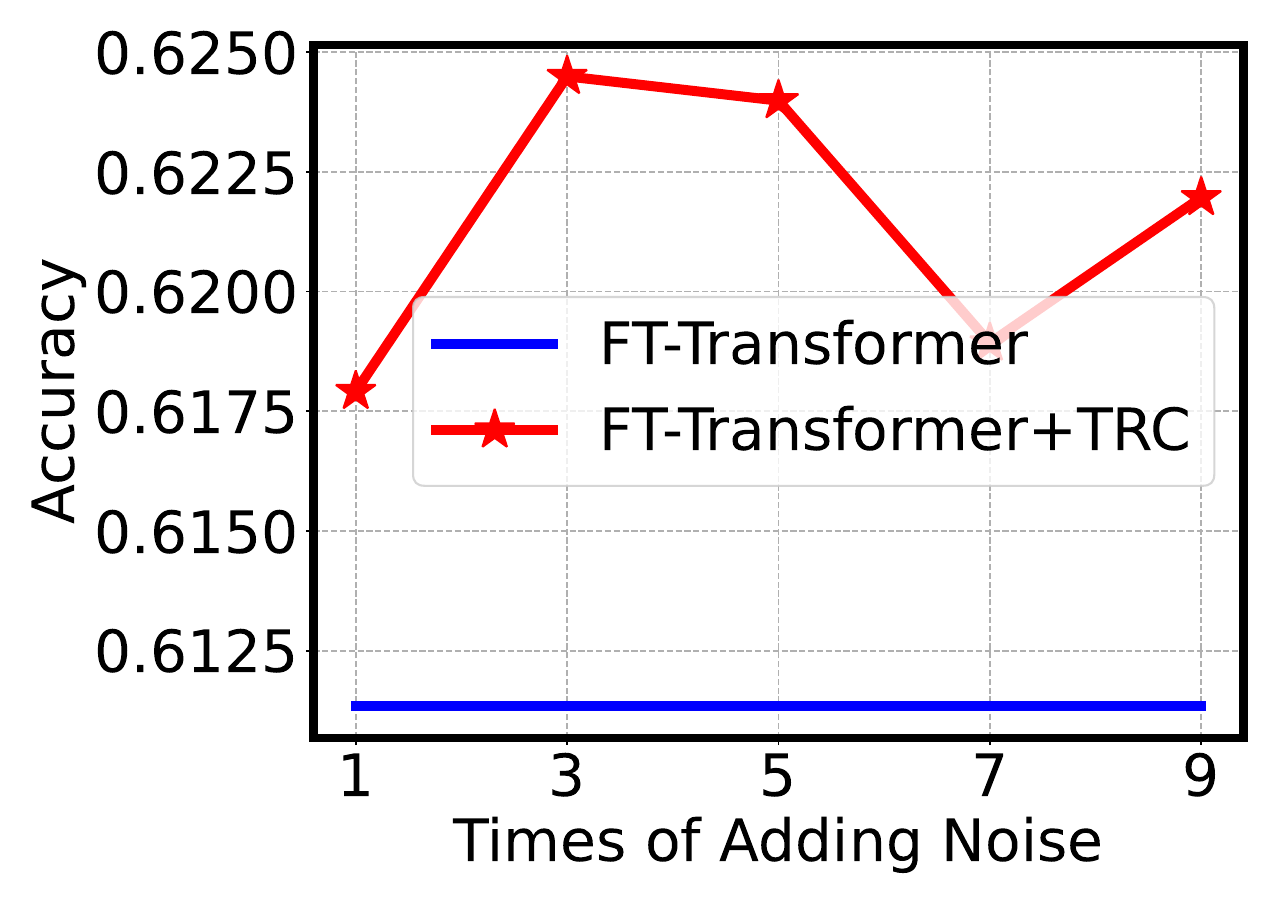}
        \caption{GE dataset $\uparrow$}
    \end{subfigure}
    \hfill
    \caption{The performance of \method with varying times of perturbing the observation in Inherent Shift Learning.}
    \label{appendix:s2}
\end{figure*}

\begin{figure*}[!h]
    \centering
    \hfill
    \begin{subfigure}[b]{0.42\linewidth}
        \centering
        \includegraphics[width=\linewidth]{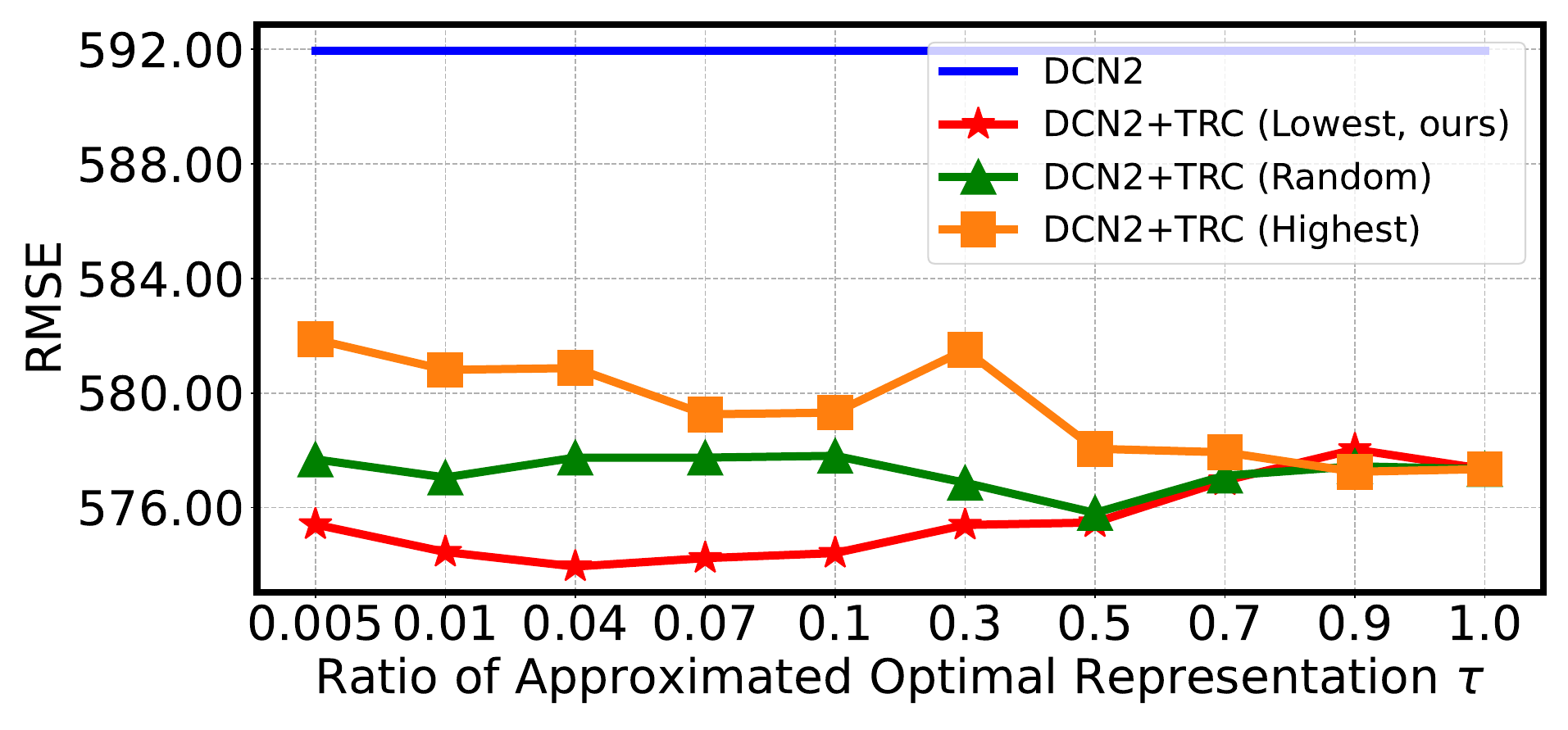}
        \caption{DI dataset $\downarrow$}
    \end{subfigure}
    \hfill
    \begin{subfigure}[b]{0.42\linewidth}
        \centering
        \includegraphics[width=\linewidth]{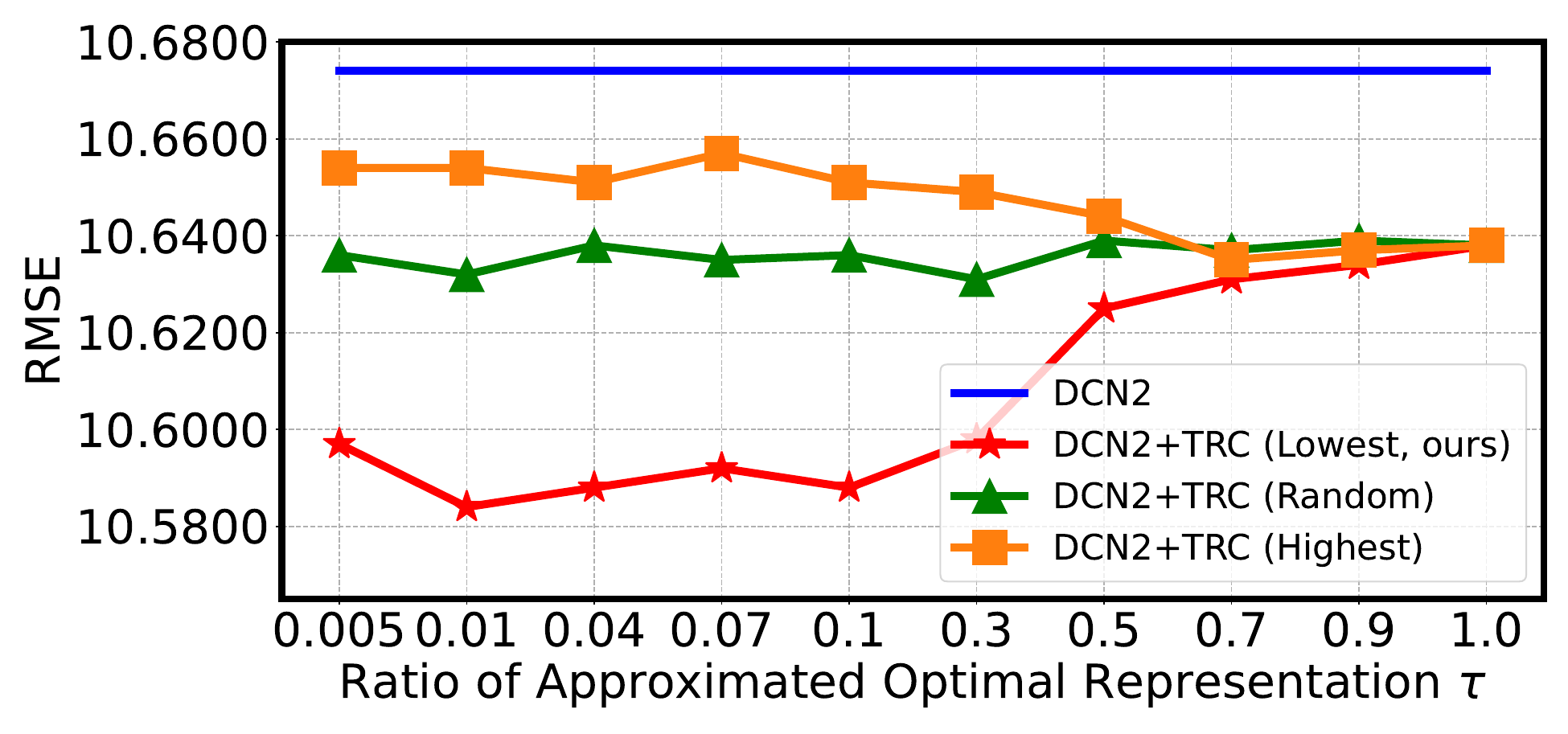}
        \caption{SU dataset $\downarrow$}
    \end{subfigure}
    \hfill
    
    \hfill
    \begin{subfigure}[b]{0.42\linewidth}
        \centering
        \includegraphics[width=\linewidth]{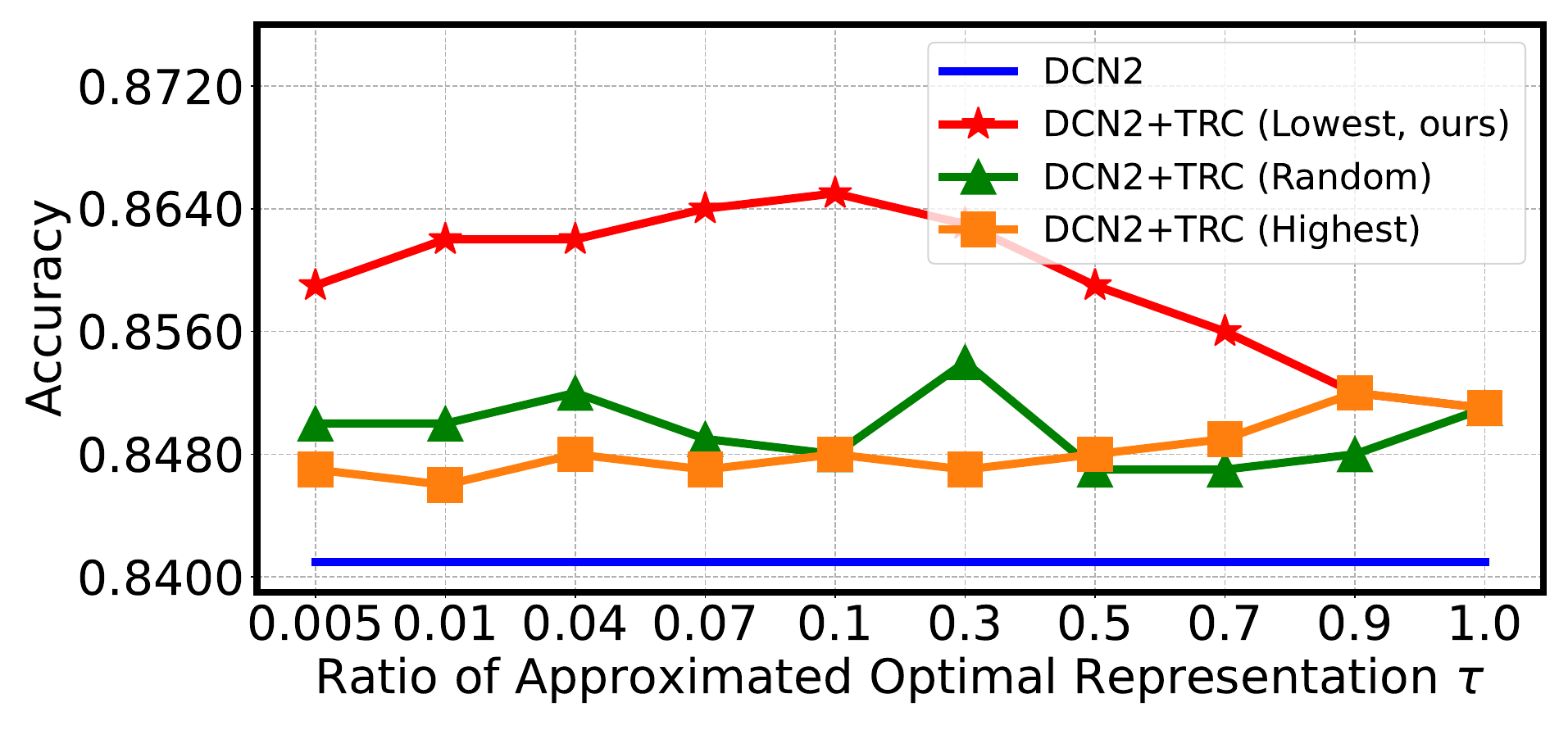}
        \caption{AU dataset $\uparrow$}
    \end{subfigure}
    \hfill
    \begin{subfigure}[b]{0.42\linewidth}
        \centering
        \includegraphics[width=\linewidth]{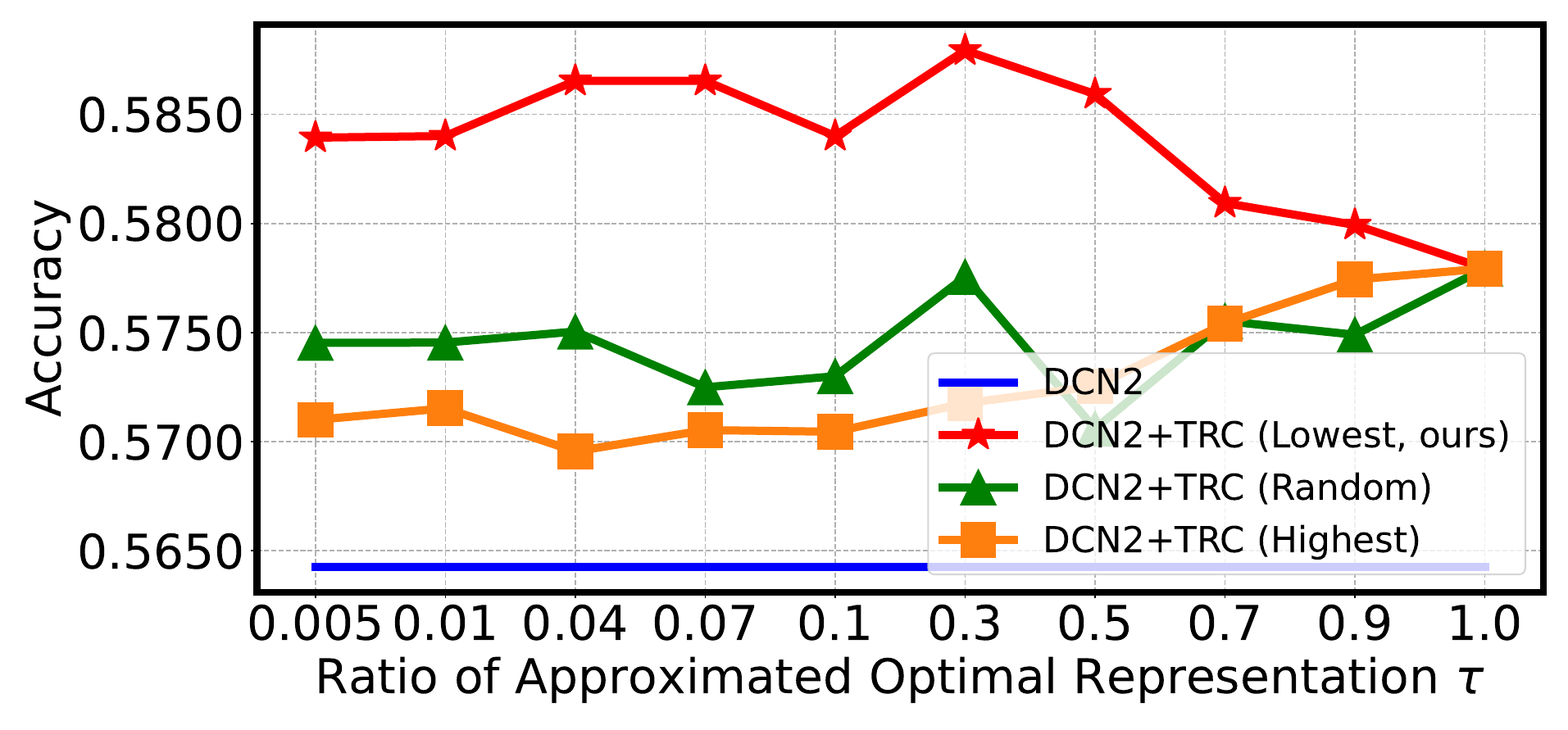}
        \caption{GE dataset $\uparrow$}
    \end{subfigure}
    \hfill
    \caption{\textcolor{blue}{The performance of \method with the varying proportion of the selected optimal representation over all of the validation set samples, i.e., $\tau$. This
figure serves as an extension of Fig. 10 in manuscript.}}
%     \caption{\textcolor{blue}{The performance of \method with the varying proportion of the selected optimal representation over all of the validation set samples, i.e., $\tau$. This
% figure serves as an extension of Fig.~\ref{fig:tau} in manuscript.}}
    \label{appendix:s3}
\end{figure*}

\begin{figure*}[h]
    \centering
    \hfill
    \begin{subfigure}[b]{0.42\linewidth}
        \centering
        \includegraphics[width=\linewidth]{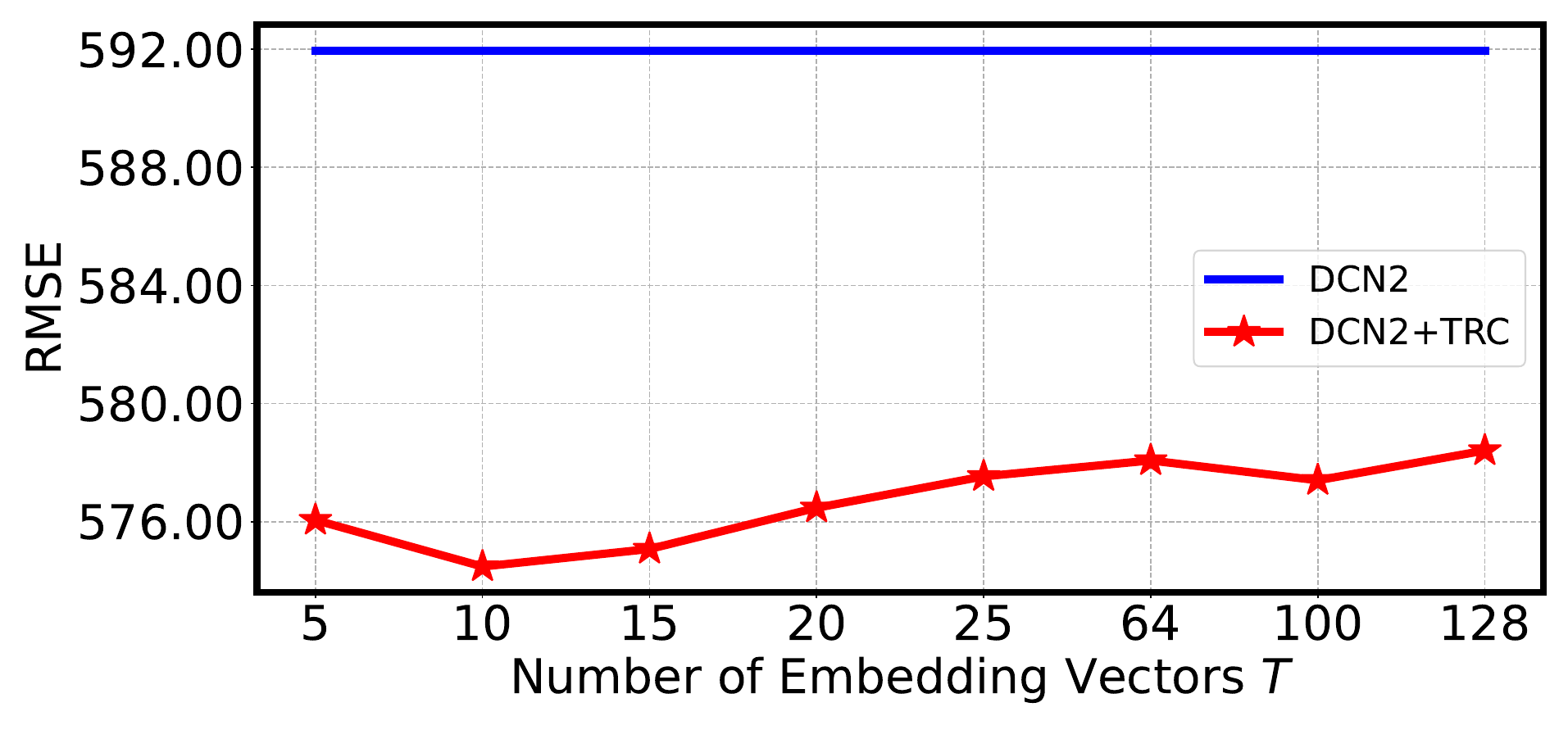}
        \caption{DI dataset $\downarrow$}
    \end{subfigure}
    \hfill
    \begin{subfigure}[b]{0.425\linewidth}
        \centering
        \includegraphics[width=\linewidth]{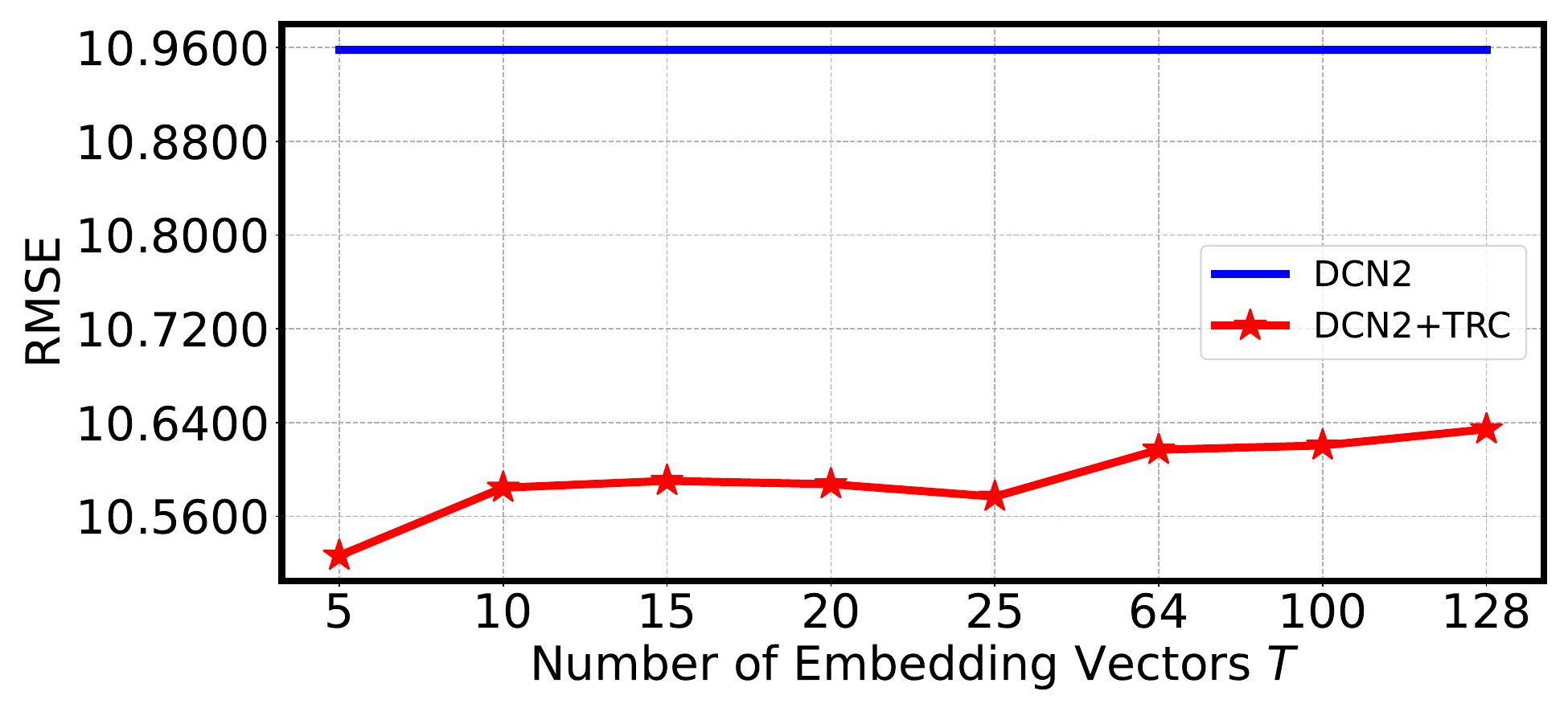}
        \caption{SU dataset $\downarrow$}
    \end{subfigure}
    \hfill
    
    \hfill
    \begin{subfigure}[b]{0.42\linewidth}
        \centering
        \includegraphics[width=\linewidth]{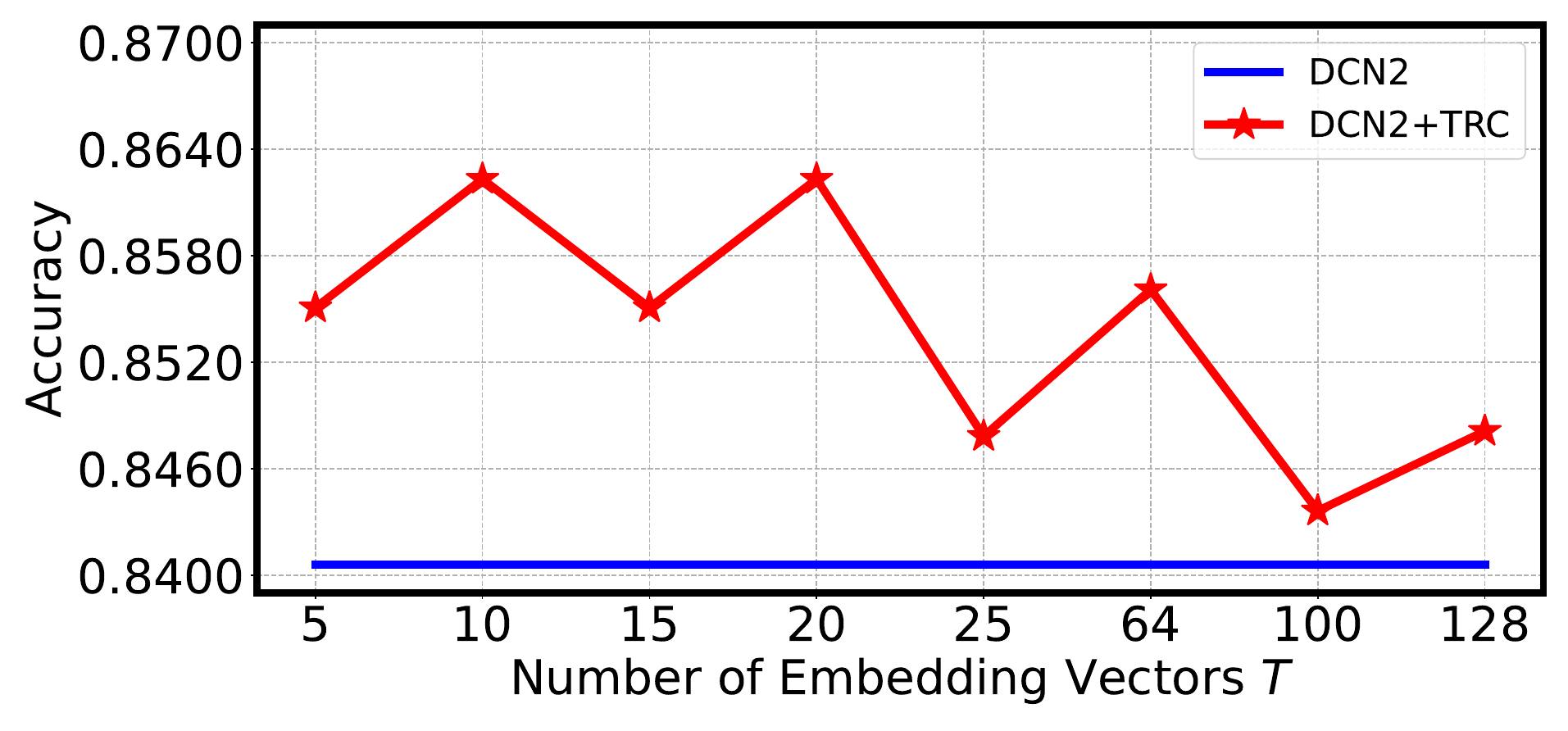}
        \caption{AU dataset $\uparrow$}
    \end{subfigure}
    \hfill
    \begin{subfigure}[b]{0.42\linewidth}
        \centering
        \includegraphics[width=\linewidth]{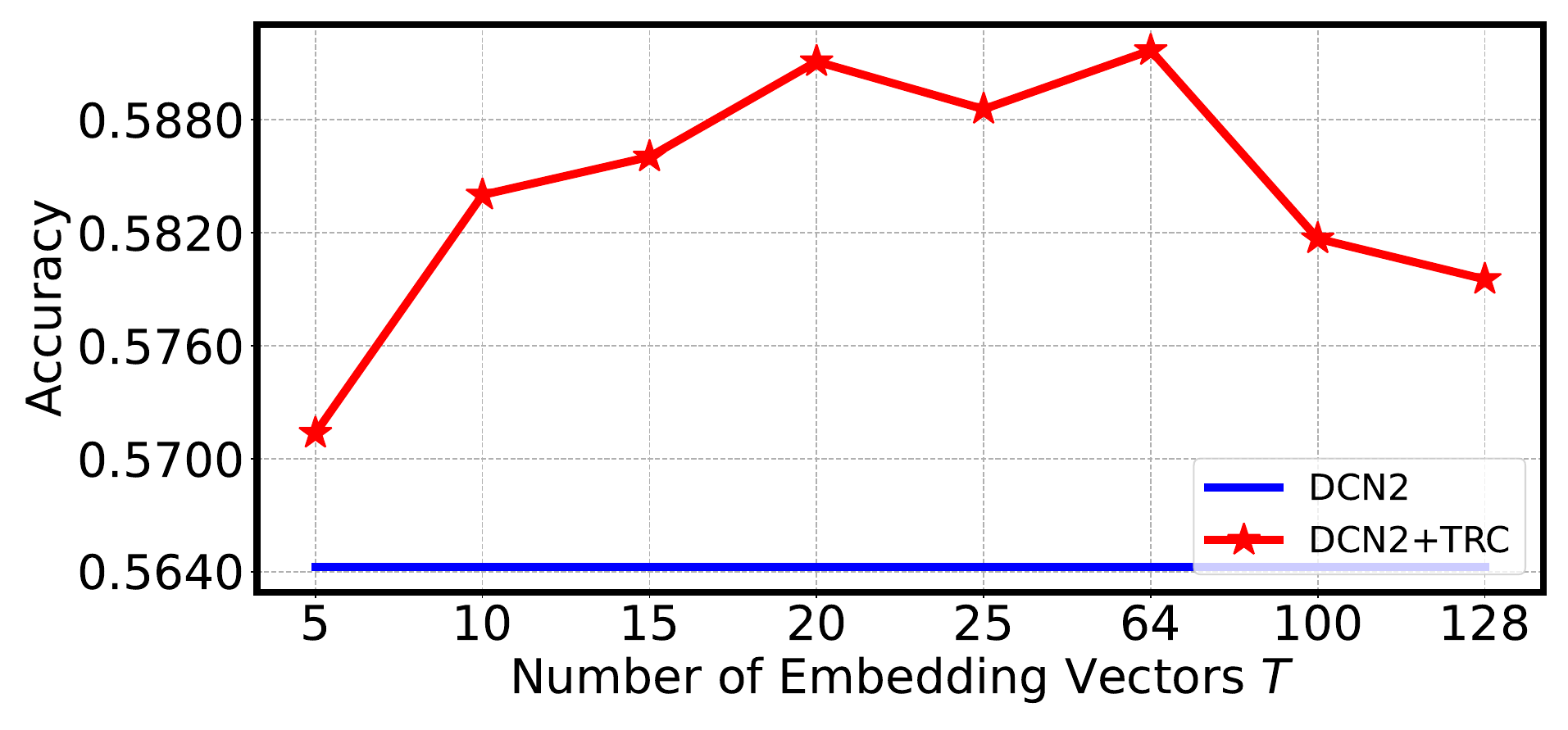}
        \caption{GE dataset $\uparrow$}
    \end{subfigure}
    \hfill
    \caption{\textcolor{blue}{The performance of \method with varying number of embedding vectors. This
figure serves as an extension of Fig. 11 in manuscript.}}
%     \caption{\textcolor{blue}{The performance of \method with varying number of embedding vectors. This
% figure serves as an extension of Fig.~\ref{fig:T} in manuscript.}}
    \label{appendix:s4}
\end{figure*}

We incorporate the sensitivity analysis for the weight of loss function $\mathcal{L}_{orth}$, the perturbing times for constructing simulated sub-optimal representations in Inherent Shift Learning, the ratio of selected optimal representations, and the number of embedding vectors in Fig.~\ref{appendix:s1}, Fig.~\ref{appendix:s2}, Fig.~\ref{appendix:s3}, and Fig.~\ref{appendix:s4}.

\newpage

% \textbf{Sensitivity analysis of the weight of orthogonal loss.}

% \textbf{Sensitivity analysis of the times of adding noise to the observation of optimal representation.}

\clearpage

% \textbf{Sensitivity analysis of the ratio of optimal representation.}
% \begin{figure*}[h]
%     \centering
%     \hfill
%    \begin{subfigure}[b]{0.24\linewidth}
%         \centering
%         \includegraphics[width=\linewidth]{pictures/robustness/pure_ratio/california_housing_FTTransformer.pdf}
%         \caption{CA dataset}
%     \end{subfigure}
%     \hfill
%     \begin{subfigure}[b]{0.24\linewidth}
%         \centering
%         \includegraphics[width=\linewidth]{pictures/robustness/pure_ratio/combined_cycle_power_plant_FTTransformer.pdf}
%         \caption{CO dataset}
%     \end{subfigure}
%     \hfill
%     \begin{subfigure}[b]{0.24\linewidth}
%         \centering
%         \includegraphics[width=\linewidth]{pictures/robustness/pure_ratio/adult_FTTransformer.pdf}
%         \caption{AD dataset}
%     \end{subfigure}
%     \hfill
%     \begin{subfigure}[b]{0.24\linewidth}
%         \centering
%         \includegraphics[width=\linewidth]{pictures/robustness/pure_ratio/GesturePhaseSegmentationProcessed_FTTransformer.pdf}
%         \caption{GE dataset}
%     \end{subfigure}
%     \hfill
%     \caption{The performance of \method with the varying proportion of the selected optimal representation over all of the validation set samples, i.e., $\tau$.}
%     \label{appendix:s3}
% \end{figure*}

\clearpage
\subsection{\textcolor{blue}{Theoretical Analysis and Discussion}}
\label{appendix:theoretical analysis and discussion}
\textcolor{blue}{
\textbf{Definition 3 Effective Rank.}
\textit{The effective rank~\cite{roy2007effective} of the matrix $\mathcal{Z}$, denoted $\text{erank} (\mathcal{Z})$, is defined as:
\begin{equation}
\label{definition_eq1}
\begin{aligned}
    \text{erank} (\mathcal{Z}) = \exp (\text{SVE}),
\end{aligned}
\end{equation}
where \text{SVE} is the entropy of normalized singular values, as defined in \textbf{Definition 1.} 
The effective rank can be considered as a real-valued extension of the rank. 
While the usual rank counts the number of non-zero singular values, the effective rank measures how the energy is distributed across the singular directions.
}}

\textcolor{blue}{
To intuitively understand the difference between rank and effective rank, consider a two-dimensional Gaussian random vector with highly correlated components. Its covariance matrix is of rank two, but the corresponding Gaussian distribution exhibits most of its energy along the direction of one singular vector. In such a case, the spectral entropy (SVE) approaches zero, where the effective rank is slightly greater than one.}

\textcolor{blue}{
\textbf{Lemma 1}
\textit{Let $\mathcal{Z}=\{z_i\}_{i=1}^N\in \mathbb{R}^{N\times D}$ denote the representations, where $N$ is the number of samples and $D$ is the dimensionality. Then we have}
\begin{equation}
\label{lemma_eq1}
\begin{aligned}
    \text{SVE} (\mathcal{Z}) = \log \text{erank}(\mathcal{Z}) \leq \log \text{rank} (\mathcal{Z}),
\end{aligned}
\end{equation}
\textit{which is given by~\cite{roy2007effective}.
}}

\textcolor{blue}{
\textbf{Proposition 1}
\textit{Consider a classification task with $C$ classes. Let $\mathcal{Z}=\{z_i\}_{i=1}^N\in \mathbb{R}^{N\times D}$ and $Y \in \mathbb{R}^{N\times C}$ denote the representations and one-hot label matrix, where $N$ is the number of samples, $D$ is the dimensionality and $C \leq D$. When the model is trained to near-zero loss, the} \text{SVE} \textit{of} $\mathcal{Z}$ \textit{satisfies} $\text{SVE} (\mathcal{Z}) \leq \log C$.}

\textcolor{blue}{
\textit{Proof.} 
Assuming a balanced dataset, when the model is trained to near-zero loss, the phenomenon neural collapse typically occurs~\cite{papyan2020prevalence, yang2022inducing}. This implies that all representations within a class converge to their within-class means.
Specifically, for class $c$, all representations $\{z_i\}$ belongs to class $c$ satisfies $z_i \to w_c$, where $w_c \in \mathbb{R}^D$ is the class center. Thus representation $\mathcal{Z}$ could be written as 
\begin{equation}
\label{proposition_eq1}
\begin{aligned}
    \mathcal{Z} = YW,
\end{aligned}
\end{equation}
where $W = \{w_c\}_{c=1}^C \in \mathbb{R}^{C\times D}$ contains the class centers as rows.
From properties of matrix multiplication, the rank of $\mathcal{Z}$ is upper bounded by:
\begin{equation}
\label{proposition_eq2}
\begin{aligned}
    \text{rank}(\mathcal{Z}) \leq \min(\text{rank}(Y), \text{rank}(W)) \leq \min(C, D) = C.
\end{aligned}
\end{equation}
Then we have:
\begin{equation}
\label{proposition_eq3}
\begin{aligned}
    \text{SVE} (\mathcal{Z}) \leq \log \text{rank}(\mathcal{Z}) \leq \log C, 
\end{aligned}
\end{equation}
which is given by \textbf{Lemma 1}.}

\textcolor{blue}{
\textbf{Discussion 1: SVE interpretation in TRC.}
TRC focuses on deep learning for supervised tabular prediction tasks, i.e., classification and regression. The purpose is to model the conditional distribution $P(y|x)$, where $x$ and $y$ are features and corresponding label. As we discussed in Section III of manuscript, the typical architecture used for such tasks consists of a backbone network that extracts latent representations, followed by a linear prediction head. 
Let us take classification as an example to illustrate.
Ideally, the learned representations should be linearly separable: that is, representations for the same class should cluster together, and different classes should be well-separated. 
}

\textcolor{blue}{
Let $\mathcal{Z}=\{z_i\}_{i=1}^N\in \mathbb{R}^{N\times D}$ denote the representations, where $N$ is the number of samples, $D$ is the dimensionality and $D \leq N$.
As shown in \textbf{Lemma 1} in \textcolor{blue}{Appendix} J, we have $\text{SVE} (\mathcal{Z}) \leq \log \text{Rank} (\mathcal{Z}) \leq \log D$, where SVE denotes singular value entropy.
\textbf{Proposition 1} in \textcolor{blue}{Appendix} J establishes that for well-learned representations that are informative for the prediction task, the energy tends to concentrate in a few dominant singular directions.
As a result, the SVE is upper bounded by $\log C$, where $C$ is the number of classes and $C\leq D$.
We assume that the trained deep tabular backbones has learned redundant information in addition to useful information for prediction.
This is supported by our empirical results in Fig. 2 and Fig. 6 of manuscript, where higher SVE often exhibits worse performance. 
% This is supported by our empirical results in Fig.~\ref{fig:motivation2} and Fig.~\ref{fig:sve and performance} of manuscript, where higher SVE often exhibits worse performance. 
In addition, we could observe from Fig. 8 of manuscript that with the redundant information (higher SVE), the representations tend to be more entangled and less linearly separable.
% In addition, we could observe from Fig.~\ref{fig:visualization with embeddings} of manuscript that with the redundant information (higher SVE), the representations tend to be more entangled and less linearly separable.
}

\textcolor{blue}{
Our method, TRC, leverages this insight to improve learned representations in a post-hoc, model-agnostic fashion. It reduces SVE while maintaining predictive performance, by jointly minimizing the supervised loss (Eq. 12) and the entropy of the singular value spectrum. This avoids modifying the original backbone and retains its capacity to learn complex feature interactions. Importantly, we do not advocate for extreme compression: excessively low SVE can discard task-relevant information and cause underfitting. 
Our results show that moderate SVE reduction strikes a good balance between compactness and predictive utility (Fig. 6 and Fig. 8 of manuscript).
% Our results show that moderate SVE reduction strikes a good balance between compactness and predictive utility (Fig.~\ref{fig:sve and performance} and Fig.~\ref{fig:visualization with embeddings} of manuscript).
}

\textcolor{blue}{
\textbf{Proposition 2}
\textit{Let $\mathcal{Z^*}\in \mathbb{R}^{N\times D}$ denote the representations obtained by Tabular Space Mapping in TRC, where $N$ is the number of samples, $D$ is the dimensionality and $T$ is the number of embedding vectors with $T\leq D$. Then we have} $\text{SVE} (\mathcal{Z^*}) \leq \log T$.} 

\textcolor{blue}{
\textit{Proof.} Given that any representation computed by Eq. 9 in Tabular Space Mapping of manuscript could be viewed as a weighted linear combination of the embedding vectors $r\mathcal{B}$, $\mathcal{Z^*}$ could be written as:
\begin{equation}
\label{proposition2_eq1}
\begin{aligned}
    \mathcal{Z^*} = R\mathcal{B},
\end{aligned}
\end{equation}
where $R = \{r_i\}_{i=1}^N\in \mathbb{R}^{N\times T}$ denotes the weights across all samples, and $\mathcal{B}\in \mathbb{R}^{T\times D}$ denotes the embedding vectors. 
From properties of matrix multiplication, the rank of $\mathcal{Z^*}$ is upper bounded by:
\begin{equation}
\label{proposition2_eq2}
\begin{aligned}
    \text{rank}(\mathcal{Z^*}) \leq \min(\text{rank}(R), \text{rank}(\mathcal{B})) \leq \min(T, T) = T.
\end{aligned}
\end{equation}
Then we have:
\begin{equation}
\label{proposition2_eq3}
\begin{aligned}
    \text{SVE} (\mathcal{Z^*}) \leq \log \text{rank}(\mathcal{Z^*}) \leq \log T,
\end{aligned}
\end{equation}
which is given by \textbf{Lemma 1}.}

\textcolor{blue}{
\textbf{Discussion 2: The choice of number of embedding vectors $T$ in TRC.}
Let $\mathcal{Z}=\{z_i\}_{i=1}^N\in \mathbb{R}^{N\times D}$ and $\mathcal{Z^*}\in \mathbb{R}^{N\times D}$ denote the representations obtained by backbone and Tabular Space Mapping in TRC respectively, where $N$ is the number of samples and $D$ is the dimensionality. 
As discussed in \textbf{Discussion 1}, the deep tabular backbones often exhibits high SVE, indicating that the learned representations $\mathcal{Z}$ contain redundancy in addition to task-relevant information.
We propose to jointly reduce the SVE of $\mathcal{Z^*}$ and minimize the supervised loss in Eq. 12 of manuscript to remain the critical information for prediction.
This encourages the energy of representations to concentrate along dominant singular directions that are most informative for the task.}

\textcolor{blue}{
As illustrated in \textbf{Proposition 2}, SVE of $\mathcal{Z^*}$ is upper bounded by $\log T$, where $T$ is the number of embedding vectors in LE-Space. Therefore the reduction of SVE could be achieved by decreasing the number of embedding vectors $T$. 
To make the SVE of $\mathcal{Z^*}$ in TRC effectively less than that of $\mathcal{Z}$ from backbone models, one could calculate the SVE of $\mathcal{Z}$ on train set, and set $T < \exp (\text{SVE}(\mathcal{Z}))$.
Notably, overly low SVE values may lead to underfitting by discarding useful task-relevant information.
Hence, we recommend choosing $T$ in a moderate range (e.g., $[5, \exp (\text{SVE}(\mathcal{Z}))]$) to maintain a good balance between compression and information preservation.}

\textcolor{blue}{
The sensitivity analysis w.r.t. $T$ are presented in Fig. 11 of manuscript and Fig.~\ref{appendix:s4} of Appendix~\ref{appendix:sensitivity analysis}. 
Overall, the TRC is robust across a wide range of $T$, and TRC consistently improve the performance of backbone models across all values of $T$.
As we suggested, when $T$ is within the range of $[5, \exp (\text{SVE}(\mathcal{Z}))]$ (e.g., $\exp (\text{SVE}(\mathcal{Z})) = 22$ on CO dataset, $\exp (\text{SVE}(\mathcal{Z})) = 29$ on DI dataset, $\exp (\text{SVE}(\mathcal{Z})) = 94$ on SU dataset, $\exp (\text{SVE}(\mathcal{Z})) = 81$ on AU dataset, $\exp (\text{SVE}(\mathcal{Z})) = 101$ on GE dataset), the performance of TRC is stable.
% When the $T$ is relatively low (e.g., less than 64), the performance of TRC is stable. 
As $T$ increases further, we observe a slight degradation in performance, yet TRC could still enhance the performance of backbone model. 
This is aligned with our claim that higher SVE may introduce redundant information hindering prediction.}

\subsection{\textcolor{blue}{How Well the Simulated Shift is Expected to Model the ``Inherent'' Shift}}
\label{appendix:simulated shift and inherent shift}
\textcolor{blue}{
We include experiment to evaluate how well the simulated shift, computed by shift estimator, can model the real inherent shift. The results are provided in Table~\ref{appendix:representation distance}.
Let us denote the representations extracted by the trained backbone $G_f(\cdot;\theta_f)$ as $\mathcal{Z}=\{z_i\}_{i=1}^N\in \mathbb{R}^{N\times D}$, where $z_i = G_f(x_i;\theta_f) \in \mathbb{R}^D$ corresponds to sample $x_i$, $N$ is the number of samples.
The true inherent shift is defined by the discrepancy between sub-optimal representation and the corresponding optimal representation. However, since the true optimal representations are not accessible in practice, directly measuring the inherent shift is challenging.}

\textcolor{blue}{
To approximate this, we propose the following experimental setup: we train the same backbone architecture $G_f(\cdot;\theta_f)$ in two different ways, (i) heavy training with the full number of epochs, producing representations $\mathcal{Z}^\text{heavy}=\{z_i^\text{heavy}\}_{i=1}^N$, and (ii) light training with only half the number of epochs, yielding $\mathcal{Z}^\text{light}=\{z_i^\text{light}\}_{i=1}^N$. We make an assumption that compared to the representations $\mathcal{Z}^{\text{light}}$ from light training, the representations $\mathcal{Z}^{\text{heavy}}$ from heavy training are closer to the optimal ones. Hence, we treat $\mathcal{Z}^{\text{heavy}}$ as optimal representations, and $\mathcal{Z}^{\text{light}}$ as sub-optimal representations. Notably, this assumption is adopted solely for the purpose of this evaluation, in the absence of ground truth for the inherent shift.}

\textcolor{blue}{
Under this setup, the true inherent shift for each $z_i^{\text{light}}$ is defined as $z_i^{\text{light}} - z_i^{\text{heavy}}$.
We apply the proposed TRC to $\mathcal{Z}^{\text{light}}$ and obtain the learned shift estimator $\phi(\cdot;\theta_\phi)$.
Next, we apply our shift estimator $\phi(\cdot;\theta_\phi)$ to estimate the inherent shift on each $z_i^{\text{light}}$, and obtain the re-estimated representation $\hat{z}_i$ by $\hat{z}_i = z_i^{\text{light}} - \phi(z_i^{\text{light}};\theta_\phi)$.
Intuitively, if the estimated shift $\phi(\cdot;\theta_\phi)$ could well model the true inherent shift $z_i^{\text{light}} - z_i^{\text{heavy}}$, $\hat{z}_i$ should be closer to $z_i^{\text{heavy}}$.
We then compare the average distance between $\hat{z}_i$ and $z_i^{\text{heavy}}$, against the average distance between $z_i^{\text{light}}$ and $z_i^{\text{heavy}}$.}

\textcolor{blue}{
Our results show that the re-estimated representation $\hat{z}_i$ are significantly closer to the optimal representation $z_i^{\text{heavy}}$ than the original suboptimal ones $z_i^{\text{light}}$. This demonstrates that the shift estimator learned from artificial corruption has the potential ability to model the inherent shift in the representation space.}
\begin{table}
\centering
\footnotesize
\caption{\textcolor{blue}{The L2 distance between sub-optimal representations and optimal representations w/o and w/ shift estimator. ``w/o Shift Estimator'' corresponds to the distance between $z_i^{\text{light}}$ and $z_i^{\text{heavy}}$, ``w/ Shift Estimator'' corresponds to the distance between $\hat{z}_i$ and $z_i^{\text{heavy}}$.}}
\label{appendix:representation distance}
\setlength{\tabcolsep}{1.8mm}{
\begin{tabular}{c|c|cccc}
\toprule
Backbone       & Discrepancy ($\downarrow$)                 & AD             & CA             & CO             & GE              \\ \midrule
MLP            & w/o Shift Estimator                    & 0.647          & 0.487          & 0.450          & 0.464           \\
MLP            & \textbf{w/ Shift Estimator} & \textbf{0.578} & \textbf{0.395} & \textbf{0.246} & \textbf{0.434}  \\ \midrule
FT-Transformer & w/o Shift Estimator                    & 2.217          & 2.460          & 2.352          & 3.870           \\
FT-Transformer & \textbf{w/ Shift Estimator} & \textbf{1.068} & \textbf{0.887} & \textbf{0.820} & \textbf{1.443}  \\
\bottomrule
\end{tabular}}
\end{table}

\subsection{\textcolor{blue}{Limitations of Assumption 1}}
\label{appendix:limitations of assumption 1}
\textcolor{blue}{
We acknowledge that the assumption may not always hold. In particular, as shown in Table~\ref{appendix:few training epochs}, when the backbone model is severely underfitting and far from convergence (e.g., trained for only 5\% of the total epochs), the performance of TRC based on Assumption 1 (lowest-gradient selection) becomes similar to that of random or highest-gradient selection. This suggests that, under severe backbone underfitting settings, low gradient norms may not indicate representations with the least shift.
However, when the number of training epochs increases, e.g., trained for 50\% of the total epochs, TRC based on Assumption 1 could enhance the backbone performance better than that of random or highest-gradient selection, which illustrates the effectiveness of Assumption 1. 
}

\begin{table*}[!t]
\centering
\footnotesize
\caption{\textcolor{blue}{The performance of TRC on underfitted backbone.}}
\label{appendix:few training epochs}
\setlength{\tabcolsep}{1.9mm}{
\begin{tabular}{c|c|cccc}
\toprule
Datasets & Backbone training epochs               & DCN2  & +\textbf{TRC (Lowest, ours)} & +TRC (Random) & +TRC (Highest)  \\ \midrule
CO $\downarrow$      & 5\%                                    & 4.57  & \textbf{4.498}              & 4.512        & 4.503          \\
CO $\downarrow$      & 50\%                                   & 4.384 & \textbf{4.165}              & 4.285        & 4.325          \\ \midrule
GE $\uparrow$      & 5\%                                    & 0.508 & \textbf{0.527}              & 0.525        & 0.524          \\ 
GE $\uparrow$      & 50\%                                   & 0.543 & \textbf{0.561}              & 0.55         & 0.548          \\
\bottomrule
\end{tabular}}
\end{table*}

% \vspace{-1em}
\subsection{Broader Impacts}
\label{appendix:broader impacts}
Tabular data is extensively utilized across various domains including healthcare, finance, engineering and psychology. Despite its wide-ranging application, much of the research in deep learning, has predominantly focused on other data modalities such as images, text, and time series. Our paper aims to bridge this gap by introducing a novel deep Tabular Representation Corrector, TRC, to enhance trained deep tabular models through two tasks. 
Progress in this direction has the potential to facilitate the construction of multi-modal pipelines for problems where only one part of the input is tabular, while other components involve images, text, and other Deep Learning-friendly data types. These pipelines can then be trained end-to-end through gradient optimization across all modalities. Such integration enables the fusion of insights from tabular data, such as demographics and genomics, with information from images, text, and time series. 
% However, it's crucial to recognize the limitations associated with such data integration, particularly concerning biases and privacy concerns that may arise.

\end{document}